\documentclass[a4paper,11pt,twoside]{ThesisStyle}

\usepackage{amsmath,amssymb}
\usepackage[T1]{fontenc}
\usepackage[utf8]{inputenc}
\usepackage[english]{babel}

\usepackage[left=1.5in,right=1.3in,top=1.1in,bottom=1.1in,includefoot,includehead,headheight=13.6pt]{geometry}

\usepackage{silence}

\usepackage{aecompl}
\usepackage{url}

\usepackage[printonlyused,withpage]{acronym}

\usepackage{ifpdf}

\ifpdf
  \usepackage[pdftex]{graphicx}
\else
  \usepackage{graphicx}
\fi

\graphicspath{{.}{images/}}

\usepackage{color}
\definecolor{linkcol}{rgb}{0,0,0.4}
\definecolor{citecol}{rgb}{0.5,0,0}

\usepackage[nottoc, notlof, notlot]{tocbibind}
\usepackage{minitoc}
\def\abs{\operatorname{abs}}
\def\argmax{\operatornamewithlimits{arg\,max}}
\def\argmin{\operatornamewithlimits{arg\,min}}

\usepackage{rotating}                    
\usepackage{fancyhdr}                    

\let\headruleORIG\headrule
\renewcommand{\headrule}{\color{black} \headruleORIG}

\usepackage{colortbl}
\arrayrulecolor{black}

\fancypagestyle{plain}{
  \fancyhead{}
  \fancyfoot{}
  
}

\usepackage{algorithm}
\usepackage[noend]{algorithmic}
\usepackage{scrextend}

\makeatletter

\def\cleardoublepage{\clearpage\if@twoside \ifodd\c@page\else%
  \hbox{}%
  \thispagestyle{empty}
  \newpage%
  \if@twocolumn\hbox{}\newpage\fi\fi\fi}

\makeatother

{%

\hrulefill
\vspace*{0.5cm}%
\end{minipage}
}

\let\minitocORIG\minitoc
\renewcommand{\minitoc}{\minitocORIG \vspace{1.5em}}

\usepackage{subfigure}
\usepackage{multirow}

\newenvironment{bulletList}%
{ \begin{list}%
	{$\bullet$}%
	{\setlength{\labelwidth}{25pt}%
	 \setlength{\leftmargin}{30pt}%
	 \setlength{\itemsep}{\parsep}}}%
{ \end{list} }

\renewcommand{\epsilon}{\varepsilon}

\ifpdf
  \usepackage[pagebackref,hyperindex=true]{hyperref}
\else
  \usepackage[dvipdfm,pagebackref,hyperindex=true]{hyperref}
\fi

\renewcommand*{\backref}[1]{}
\renewcommand*{\backrefalt}[4]{%
\ifcase #1 %
(Not cited.)%
\or
(Cited on page~#2.)%
\else
(Cited on pages~#2.)%
\fi}

\hypersetup
{
bookmarksopen=true,
pdftitle="Arthur Ouaknine PhD Thesis",
pdfauthor="Arthur Ouaknine",
pdfsubject="Deep learning for radar data exploitation of autonomous vehicle", 
pdfmenubar=true, 
pdfhighlight=/O, 
colorlinks=true, 
pdfpagemode=UseNone, 
pdfpagelayout=SinglePage, 
pdffitwindow=true, 
linkcolor=linkcol, 
citecolor=citecol, 
urlcolor=linkcol 
}

\usepackage{booktabs}
\usepackage{times}
\usepackage{epsfig}
\usepackage{graphicx}
\usepackage{amsmath}
\usepackage{amssymb}
\usepackage{stmaryrd}
\usepackage{tabularx,booktabs}
\usepackage{multirow}
\usepackage{lipsum}
\usepackage{array}
\usepackage{color, colortbl}
\usepackage{pifont}
\usepackage{flushend}
\usepackage{dsfont}
\usepackage{pdfpages}
\usepackage{lipsum}
\usepackage{makecell}
\usepackage[export]{adjustbox}
\usepackage{bm}
\usepackage{cleveref}
\usepackage{xfrac}
\usepackage{algorithm} 
\usepackage{algorithmic}
\usepackage{setspace}
\usepackage{eqparbox}
\usepackage{bbm}
\usepackage{xurl}

\newcommand{\mcl}[1]{\ensuremath{\mathcal{#1}}}

\newcommand{\vect}[1]{\ensuremath{\textbf{#1}}}

\newcommand{\etal}{\textit{et al.} }
\renewcommand{\algorithmiccomment}[1]{\bgroup\hfill\#~#1\egroup}

\definecolor{better}{rgb}{0.19, 0.55, 0.91}
\definecolor{worse}{rgb}{0.82, 0.1, 0.26}
\newcommand{\cmark}{\textcolor{better}{\ding{51}}}%
\newcommand{\xmark}{\textcolor{worse}{\ding{55}}}%

\newcolumntype{C}{>{\centering\arraybackslash}X}
\newcolumntype{L}{>{\centering\arraybackslash}p{0.2\textwidth}}
\newcommand{\NA}{---}

\newcommand{\NTx}{N_{\text{Tx}}}
\newcommand{\NRx}{N_{\text{Rx}}}
\newcommand{\BinA}{B_{\text{A}}}
\newcommand{\BinR}{B_{\text{R}}}
\newcommand{\BinD}{B_{\text{D}}}
\newcommand{\BinE}{B_{\text{E}}}
\newcommand{\BinM}{B_{\text{M}}}
\newcommand{\BinN}{B_{\text{N}}}

\DeclareUnicodeCharacter{2212}{-}

\begin{document}

\includepdf[pages=-]{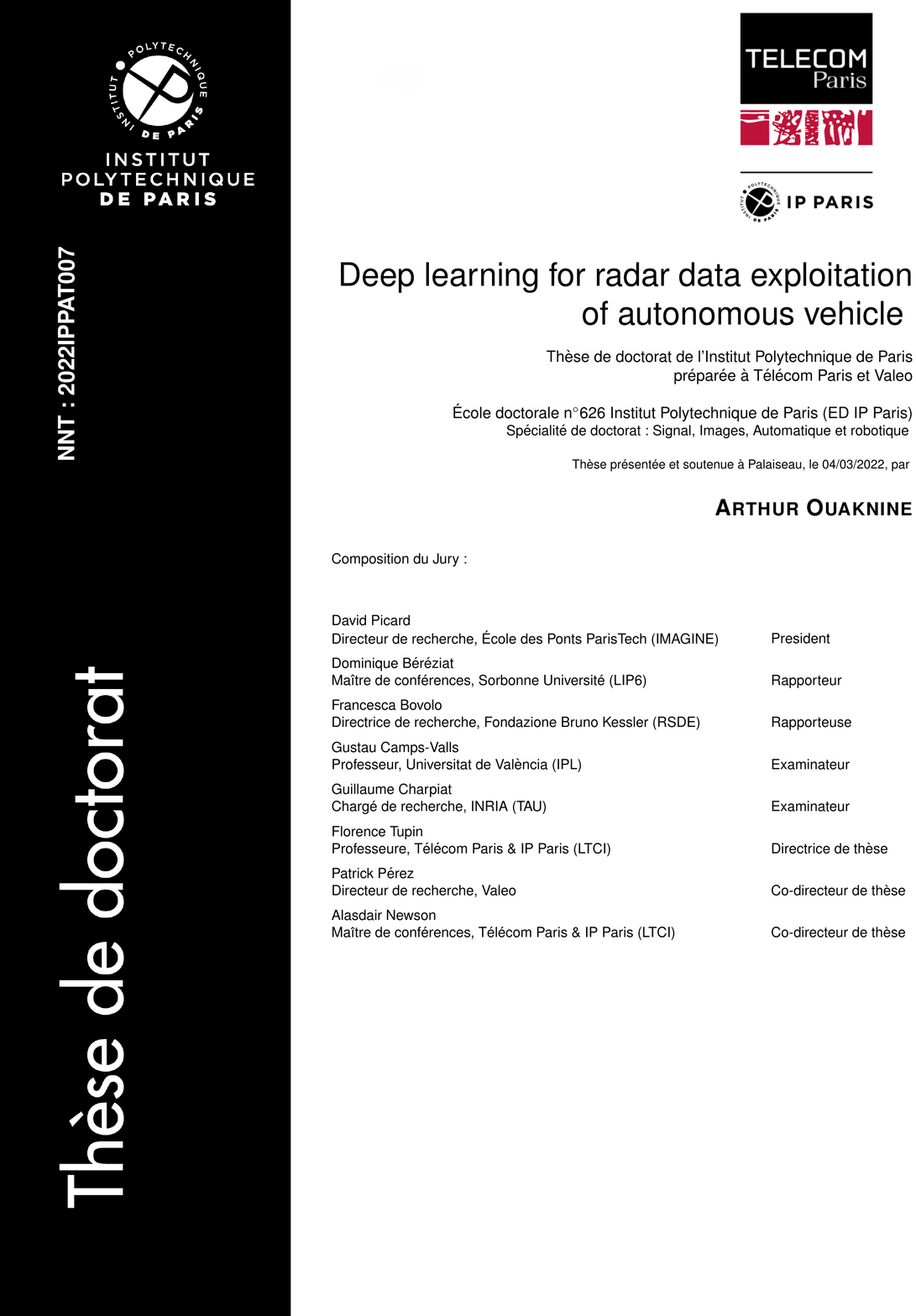}







\pagenumbering{roman}

\setcounter{page}{0}
\cleardoublepage

\section*{Acknowledgments}

Je souhaite tout d’abord remercier Alasdair, Florence et Patrick, mes encadrants de thèse, pour m’avoir fait confiance sur ce projet et pour avoir dirigé mes recherches avec rigueur pendant plus de trois années. Alasdair, merci de m’avoir soutenu et aidé à tout moment, aussi bien personnellement que professionnellement ; et merci pour ton implication rigoureuse dans tous mes projets. Tu as été un acteur majeur de cette thèse sans qui elle n’aurait pas pu aller aussi loin : ne change rien pour tes futurs doctorants. Florence, merci pour ta disponibilité, pour nos discussions et nos encadrements de projets communs qui m’ont beaucoup apporté humainement et scientifiquement. Patrick, merci pour nos discussions et tes idées qui ont inspiré nos travaux ainsi que pour ta confiance en m’intégrant à l’équipe naissante de valeo.ai. 

Je remercie également Julien pour avoir contribué à cette thèse de façon significative et sans qui ces travaux n’auraient pas pu voir le jour. Merci pour tes idées, ton expertise et ta disponibilité ayant permis des avancées importantes sur ces sujets que nous abordions tous les deux.

I would like to thank the jury of my thesis for your time and your relevant feedbacks which truly improved the quality of this thesis. I would like to especially thank Domique Béréziat and Francesca Bovolo for reviewing my manuscript. Je remercie aussi mon jury de mi-parcours pour leurs retours et conseils avisés concernant ma thèse.

Je remercie Télécom Paris pour son accueil et tout particulièrement l’équipe IMAGES du département IDS. J’ai eu la chance de faire de nombreuses rencontres entre Paris 13ème et le plateau de Saclay. Merci à mes co-bureaux des deux sites : Nicolas Go., Vincent, Raphaël et Matthis. Je remercie toute l’équipe IMAGES, les enseignants-chercheurs, les post-docs et les actuels et anciens doctorants pour cette vie animée à Télécom Paris. Merci à Emanuele, Mateus, Giammarco, Nicolas Ga., Clément, Antoine, Inès, Nicolas C., Robin, Xu, Xiangli, Corentin, Sylvain, Christophe, Alban, Zoé, Pietro, Saïd, Yann, Isabelle et Jean-Marc. Je remercie particulièrement Jean-Marie pour la patience dont il a fait preuve pour m’enseigner la physique élémentaire au début de ma thèse.  Et merci aux co-organisateurs et aux participants du Deep Learning Working Group, qui a été une source de discussions riches et de rencontre entre les doctorants.

Je remercie également toute l’équipe de valeo.ai qui a tant évolué depuis mon arrivée. Je remercie tout particulièrement les Bezos, Charles, Maxime et Simon, pour nos débats et nos blagues. Sans eux, l’aventure n’aurait pas été la même. Merci à Gabriel de m’avoir sorti de nombreuses situations difficiles. Merci à Alexandre pour tous nos échanges très enrichissants et pour ce projet avec beaucoup d’avenir. Je remercie tous les actuels et anciens permanents, doctorants et stagiaires pour ces échanges scientifiques d’une qualité impressionnante. Merci à Andrei, Hedi, Eloi, Tuan-Hung, Himalaya, Spyros, Florent, Huy, Antoine, Oriane, Laura, Gilles, Matthieu, Renaud, Mickael, Tristan, Bjoern, Corentin, Leon et Antonin.

Je remercie tous mes anciens collègues de Zyl, a.k.a. BIM, a.k.a. Comète, a.k.a. Crossroad, pour tous ces moments incroyables, avant, pendant et après mon stage de fin d’étude. Merci à Mathieu, Aurel, Anthony, Ophély, Samy, Gauthier, Camille, Thomas, Martin, Alexandre, Flo et Valentin. Je remercie tout particulièrement Long pour m’avoir transmis sa passion pour la recherche, pour son encadrement irréprochable et pour avoir poussé mon projet de thèse sans même le vouloir.

Je remercie tous mes amis pour faire partie de ma vie et pour m’avoir soutenu pendant toutes ces années d’études et de recherche. Merci à Gauthier, Garance et Félix, mes amis d’enfance qui ont toujours été et seront toujours là, de Paris au Guatemala. Merci aux Chocs, Toli, Matthieu, Raphaël, Simon, Mattia, Loris, Maxou et Laloum, mes frères qui ont partagé ma vie et la partageront pendant encore longtemps sdv. Merci à Lina et Anna pour m’avoir soutenu en toute condition ces dernières années, je vous en suis extrêmement reconnaissant ; notre amitié a encore de grandes choses à vivre.

Bien évidemment, je remercie Sofia, Zalie, Hélène et Fanny, les reines de Montalivet, pour votre indéfectible soutien, pour tout ce qu’on a vécu et tout ce qu’il nous reste à vivre ensemble.

Je remercie les puissant.e.s, a.k.a. grand est, pour tous ces moments incroyables de vie, de fêtes, de débats d’experts, entre Paris, Moliets et Bordeaux. Merci à Eugénie, Manon, Joachim, Romain, Solène, Hicham et Roxane, avec qui j’ai hâte de revivre ces joies.

Je remercie également Lou, Maud, Juliette, Violette, Nico, Vincent et Djav sans qui ces vacances, ces dîners, ces fêtes et autres moments n’auraient pas créé de si beaux souvenirs chers à mes yeux. 
Je remercie JB pour toutes ces discussions, ces apéros et ces soirées où nous avons refait le monde depuis les bancs de la Sorbonne.

Je remercie Antoine et Matthieu pour nos débats interminables, nos apéros au BM et à la Butte-aux-Cailles, depuis Télécom à aujourd’hui, vous avez réussi à me motiver pour cette thèse.

Je remercie toute ma famille, Suzanne, Christiane, Henri, Antoine, Isabelle, Didier, Michel, Mirjanna, Maud, Wahid et Jonathan pour toutes ces années de partage, qui ne cesseront jamais je l’espère.

Je remercie mes parents et ma sœur, Lydia, Robert et Lucie, pour leur amour, leur soutien, leur compréhension et leurs encouragements depuis tant d’années ; sans qui je n’aurais jamais pu arriver où je suis aujourd’hui. Cette thèse leur est dédiée.

Je remercie finalement celles et ceux que je n’ai pas cité et qui m’ont énormément apporté au cours de ma vie ; grâce à qui j’ai pu être heureux, me construire et apprendre à dépasser mes limites. Enfin, un immense merci aux lecteurs, qui auront la patience de se plonger dans mes travaux ayant occupé mes trois dernières années.

\dominitoc
\tableofcontents


\chapter*{List of Acronyms}
\mtcaddchapter[List of Acronyms]


\begin{acronym}
\renewcommand{\\}{}

\acro{3D-FFT}{Three Dimensional Fast Fourier Transform}

\acro{AD}{Angle-Doppler}
\acro{ADAS}{Advanced Driving Assistance Systems}
\acro{ADC}{Analog-to-Digital Converter}
\acro{AE}{AutoEncoder}
\acro{AN}{Artificial Neuron}
\acro{ANN}{Artificial Neural Network}
\acro{AoA}{Angle-of-Arrival}
\acro{AP}{Average Precision}
\acro{AR}{Average Recall}
\acro{ASPP}{Atrous Spatial Pyramidal Pooling}

\acro{BCE}{Binary Cross-Entropy}
\acro{BEV}{Bird's Eye View}

\acro{CAN bus}{Controller Area Network}
\acro{CE}{Cross-Entropy}
\acro{CFAR}{Constant False Alarm Rate}
\acro{cGAN}{conditional Generative Adversarial Network}
\acro{CNN}{Convolutional Neural Network}
\acro{CoL}{Coherence Loss}

\acro{DFT}{Discrete Fourier Transform}
\acro{DoA}{Direction-of-Arrival}

\acro{FCN}{Fully Convolutional Network}
\acro{FFT}{Fast Fourier Transform}
\acro{FMCW}{Frequency-Modulated Continuous Wave}
\acro{FoV}{Field-of-View}
\acro{FPN}{Feature Pyramid Network}
\acro{FPS}{Frames-per-Second}

\acro{GAN}{Generative Adversarial Network}
\acro{GFLOPS}{Giga FLoating-point Operations Per Second}
\acro{GPS}{Global Positioning System}
\acro{GPU}{Graphics Processing Unit}

\acro{HD}{High-Definition}
\acro{HNM}{Hard Negative Mining}

\acro{IF}{Intermediate Frequency}
\acro{iid}{independent and identically distributed}
\acro{ILSVRC}{ImageNet Large Scale Visual Recognition Challenge}
\acro{IoU}{Intersection over Union}

\acro{JS}{Jensen-Shannon}

\acro{KL}{Kullback-Leibler}

\acro{LD}{Low-Definition}
\acro{LeakyReLU}{Leaky Rectified Linear Unit}
\acro{LiDAR}{Light Detection And Ranging}
\acro{LSTM}{Long Short-Term Memory}

\acro{MAE}{Mean Absolute Error}
\acro{mAP}{mean Average Precision}
\acro{mAR}{mean Average Recall}
\acro{mDice}{mean Dice}
\acro{MIMO}{Multiple Input Multiple Output}
\acro{mIoU}{mean Intersection over Union}
\acro{MLP}{Multilayer Perceptron}
\acro{mPP}{mean Pixel Precision}
\acro{mPR}{mean Pixel Recall}
\acro{MSE}{Mean Squared Error}

\acro{NMS}{Non-Maximum Suppression}

\acro{PP}{Pixel Precision}
\acro{PR}{Pixel Recall}

\acro{RA}{Range-Angle}
\acro{RAD}{Range-Angle-Doppler}
\acro{RAED}{Range-Azimuth-Elevation-Doppler}
\acro{RCS}{Radar Cross Section}
\acro{RD}{Range-Doppler}
\acro{RADAR}{Radio Detection And Ranging}
\acro{ReLU}{Rectified Linear Unit}
\acro{RGB}{Red, Green and Blue}
\acro{RNN}{Recurrent Neural Network}
\acro{RoI}{Region of Interest}
\acro{RPN}{Region Proposal Network}
\acro{Rx}{Receiver antenna}

\acro{SAR}{Synthetic Aperture RADAR}
\acro{SDice}{Soft Dice}
\acro{SGD}{Stochastic Gradient Descent}
\acro{SORT}{Simple and Online Real time Tracking}

\acro{Tx}{Transmitter antenna}

\acro{wMAE}{weighted Mean Absolute Error}
\acro{wCE}{weighted Cross-Entropy}

\end{acronym}

\mtcaddchapter[List of Figures]

\listoffigures

\mtcaddchapter[List of Tables]

\listoftables

\chapter*{Abstract}
\mtcaddchapter[Abstract]

Autonomous driving requires a detailed understanding of complex driving scenes. The redundancy and complementarity of the vehicle’s sensors provide an accurate and robust comprehension of the environment, thereby increasing the level of performance and safety. This thesis focuses the on automotive \acs{RADAR}, which is a low-cost active sensor measuring properties of surrounding objects, including their relative speed, and has the key advantage of not being impacted by adverse weather conditions. 

With the rapid progress of deep learning and the availability of public driving datasets, the perception ability of vision-based driving systems (e.g., detection of objects or trajectory prediction) has considerably improved.  The \acs{RADAR} sensor is seldom used for scene understanding due to its poor angular resolution, the size, noise, and complexity of raw \acs{RADAR} data as well as the lack of available datasets.  This thesis proposes an extensive study of \acs{RADAR} scene understanding, from the construction of an annotated dataset to the conception of adapted deep learning architectures.

First, this thesis details approaches to tackle the current lack of data. A simple simulation as well as generative methods for creating annotated data are presented. It also describes the CARRADA dataset, composed of synchronised camera and \acs{RADAR} data with a semi-automatic method generating annotations on the \acs{RADAR} representations. Today, the CARRADA dataset is the only one to provide annotated raw \acs{RADAR} data for object detection and semantic segmentation tasks. The CARRADA dataset and its annotation method form our first major contribution presented at the International Conference on Pattern Recognition (ICPR).

This thesis then presents a proposed set of deep learning architectures for \acs{RADAR} semantic segmentation. A combination of specialized loss functions is also presented, including a coherence function that reinforces the spatial harmony of the predictions made by the model on the different views of the \acs{RADAR} tensor. The proposed architecture with the best results outperforms alternative models, derived either from semantic segmentation of natural images or from \acs{RADAR} scene understanding, while requiring much less parameters. 
The presented method for semantic segmentation of raw \acs{RADAR} data is our second major contribution presented at the International Conference on Computer Vision (ICCV).
This thesis also introduces a method to open up research into the fusion of \acs{LiDAR} and \acs{RADAR} sensors for scene understanding. The proposed method is an early fusion module propagating the \acs{RADAR} information in the \acs{LiDAR} point cloud by considering the resolution and accuracy of the sensor as a quantification of its uncertainty. In this way, our module is able to create a point cloud that benefits from the advantages of both sensors while compensating for their disadvantages.

Finally, this thesis exposes a collaborative contribution, the RADIal dataset with synchronised \acf{HD} \acs{RADAR}, \acs{LiDAR} and camera. A deep learning architecture is also proposed to estimate the \acs{RADAR} signal processing pipeline while performing multitask learning for object detection and free driving space segmentation simultaneously. This collaborative contribution is under review at an international conference.

\chapter*{French Summary}
\mtcaddchapter[French Summary]



La conduite autonome vise à révolutionner notre mobilité en comprenant et en prévoyant l'environnement de conduite tout en palliant les faiblesses humaines de manière sûre et efficace. Pour se faire, elle exige une compréhension détaillée de scènes de conduite complexes.
L'environnement du véhicule est observé par des caméras enregistrant des images facilement compréhensibles par l'œil humain. Au fil du temps, des capteurs actifs supplémentaires sont apparus, venant compléter les caméras et permettant une meilleure compréhension des scènes environnantes : le \acf{LiDAR}, le \acf{RADAR} et les capteurs à ultrasons.
La redondance et la complémentarité des capteurs du véhicule permettent ainsi une compréhension précise et robuste de l'environnement : ils augmentant ainsi le niveau de performance et de sécurité. Cette thèse se concentre sur le \acs{RADAR} automobile, qui est un capteur actif à faible coût mesurant les propriétés des objets environnants, y compris leur vitesse relative, tout en ayant l'avantage de ne pas être affecté par des conditions météorologiques défavorables. 

Avec les progrès rapides de l'apprentissage profond et la disponibilité d'ensembles de données publiques sur la conduite, la capacité de perception des systèmes de conduite basés sur la vision (par exemple, la détection d'objets ou la prédiction de trajectoire) s'est considérablement améliorée.  Le capteur \acs{RADAR} est rarement utilisé pour la compréhension de scène en raison de sa faible résolution angulaire, de la taille, du bruit et de la complexité des données brutes \acs{RADAR} ainsi que du manque de données disponibles.  Cette thèse propose une étude approfondie de la compréhension de scènes \acs{RADAR} : de la construction d'un jeu de données annotées à la conception d'architectures d'apprentissage profond adaptées.

Tout d’abord, le chapitre \ref{chap:introduction} de cette thèse introduit les motivations ayant conduit à nos travaux. En particulier, il aborde les grandes avancées de la conduite autonome ainsi que l’avènement des algorithmes d’apprentissage.

Le chapitre \ref{chap:background} présente ensuite la théorie de fonctionnement du capteur \acs{RADAR} et les méthodes de traitement du signal appliquées aux données enregistrées. Il présente également la théorie des réseaux de neurones artificiels ainsi que les architectures d'apprentissage profond pour l’analyse d’images qui sont, pour la plupart, exploitées dans nos travaux.

Le chapitre \ref{chap:related_work} présente les ensembles de données \acs{RADAR} existants et les travaux antérieurs sur la compréhension de scènes \acs{RADAR} pour la détection d'objets et la segmentation sémantique basées sur les algorithmes d'apprentissage profond. Il traite aussi des approches antérieures de fusion de capteurs combinant le \acs{RADAR} avec des capteurs de type caméra ou \acs{LiDAR}, ou les deux, pour la compréhension de scènes automobiles.

Le chapitre \ref{chap:datasets} de cette thèse détaille ensuite des approches permettant de remédier au manque de données. Une simulation simple ainsi que des méthodes génératives pour créer des données annotées sont présentées. Ce chapitre décrit également le jeu de données CARRADA, composé de données synchronisées de caméra et de \acs{RADAR} avec une méthode semi-automatique générant des annotations sur les représentations \acs{RADAR}. Aujourd'hui, le jeu de données CARRADA est le seul à fournir des données \acs{RADAR} brutes annotées pour des tâches de détection d'objets et de segmentation sémantique. Le jeu de données CARRADA et sa méthode d’annotation forment notre première contribution majeure présentée à la conférence internationale sur la reconnaissance de forme (International Conference on Pattern Recognition).

Le chapitre \ref{chap:radar_scene_understanding} présente ensuite un ensemble d'architectures d'apprentissage profond pour la segmentation sémantique de données \acs{RADAR} brutes. Une combinaison de fonctions de perte spécialisées est également proposée dont une fonction de cohérence renforçant l’harmonie spatiale des prédictions réalisées par le modèle sur les différentes vues du tenseur \acs{RADAR}.  L'architecture proposée la plus performante présente les meilleures performances comparées aux modèles alternatifs, dérivés soit de la segmentation sémantique d'images naturelles, soit de la compréhension de scènes \acs{RADAR}, tout en nécessitant beaucoup moins de paramètres. La méthode de segmentation sémantique de données \acs{RADAR} brutes présentée est notre seconde contribution majeure présentée à la conférence internationale de vision par ordinateur (International Conference on Computer Vision).
Ce chapitre décrit également une méthode permettant d'ouvrir la recherche sur la fusion des capteurs \acs{RADAR} et \acs{LiDAR} pour la compréhension de scènes. La méthode proposée est composée d’un module de fusion précoce propageant l'information \acs{RADAR} dans le nuage de points \acs{LiDAR} en considérant la résolution et la précision du capteur \acs{RADAR} comme une quantification de son incertitude. De cette manière, notre méthode est capable de créer un nuage de points bénéficiant des atouts des deux capteurs tout en palliant leurs inconvénients.

Le chapitre \ref{chap:hd_radar} expose une contribution collaborative, le jeu de données RADIal enregistré avec un \acs{RADAR} haute définition (HD), un \acs{LiDAR} et une caméra synchronisés. Une architecture d'apprentissage profond est également proposée pour estimer la chaîne de traitement du signal \acs{RADAR} tout en effectuant simultanément un apprentissage multitâche pour la détection d'objets et la segmentation de l'espace libre de conduite. Cette contribution collaborative est en cours de révision dans une conférence internationale. 

Enfin, le chapitre \ref{chap:conclusion} de cette thèse conclue sur nos contributions et les limites de nos travaux. Il présente aussi nos futurs travaux sur la compréhension de scènes \acs{RADAR}. En particulier, les pistes d’amélioration de notre méthode de segmentation sémantique de données \acs{RADAR} brutes sont abordées. Ce chapitre aborde aussi nos perspectives d'exploitation de nuages de points \acs{RADAR} et \acs{LiDAR} propagés et fusionnés dans le but d’obtenir une meilleure compréhension des scènes urbaines complexes.

\mainmatter

\chapter{Introduction}
\label{chap:introduction}
\minitoc

\section{Context}

Autonomous driving aims to revolutionize our mobility by understanding and predicting the driving environment while alleviating for human weaknesses in a safe and efficient manner.
Since the European project Eureka Prometheus (1987-1995) and the American DRAPA Grand Challenge (2004-2007), the spectacular progresses in visual scene understanding using learning algorithms have brought back the old dream of driverless cars \cite{janai_computer_2020}.
The automotive industry is getting closer and closer to this goal as the major players have massively invested in advanced technologies: multi-sensor perception, three-dimensional reconstruction, cartography, high-precision location, planning and commands.

Scene understanding is a prerequisite, but it goes hand in hand with decision making for car automation. Both must be mastered to deliver safe vehicles with automatic features to the public. The automotive industry distinguishes five levels of driving automation. Levels one and two consist of integrating driving assistance or partially automatic features, while a human continues to monitor all tasks, \textit{e.g.} automatic cruise control or lane centering. These levels are already integrated in the vehicles currently marketed to the public. 
At level three, the vehicle is capable of performing most driving tasks in certain conditions, \textit{e.g.} highway driving up to 60km/h, 
but human control is still required for these features. The automotive industry estimates that level 3 vehicles will be sold to the public in 2024 in European countries.
Levels four and five no longer require a driver and perform fully automatic driving in different circumstances. Depending on the progress of research and development, these levels could become publicly available in the coming decades.

\ac{ADAS} and autonomous driving require a detailed understanding of complex driving scenes. 
The environment of the vehicle is observed by cameras recording images that are easily understood by the human eye.
Additional active sensors have emerged complementing cameras and allowing a better understanding of the surrounding scenes: \acf{LiDAR}, \acf{RADAR} and ultrasound sensors.
With the transition from assisted to automated driving, the redundancy and complementary of these sensors aims to provide an accurate and robust comprehension of the environment, thereby increasing the level of performance and safety. 
Redundancy mechanisms are required at all levels of the system: from the sensing parts to the final decision modules. At the sensor level, they can be reached using sensors of different types, where each of them offers a vision of the world with its own properties and physical aspects.

\begin{figure}[!t]
  \begin{center}
    \includegraphics[width=0.8\textwidth]{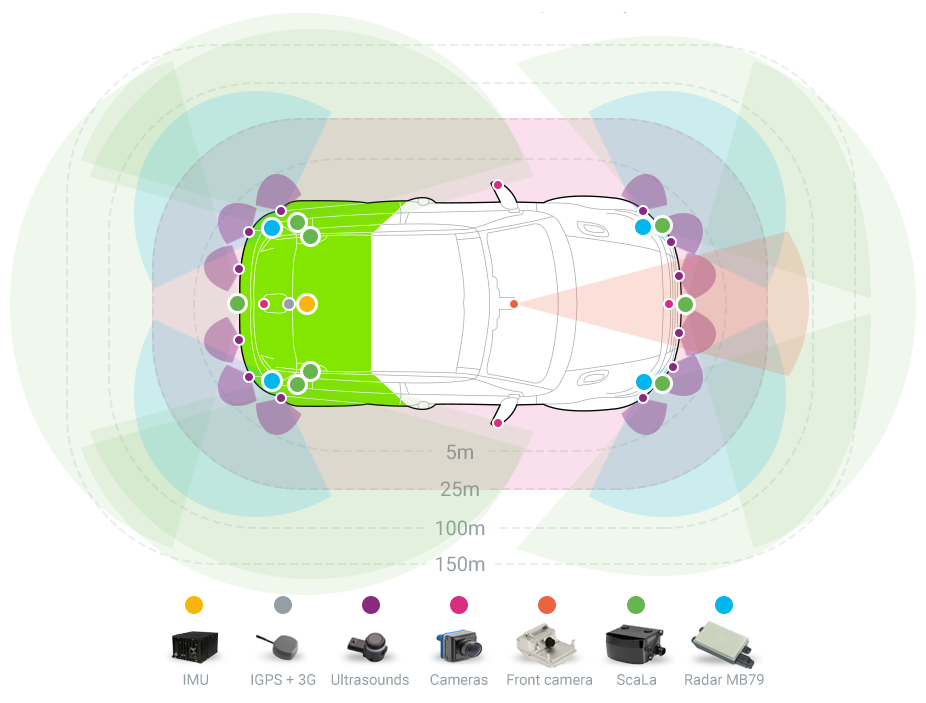}
  \end{center}
  \caption[Sensor setup of the Valeo Drive4U prototype]{\textbf{Sensor setup of the Valeo Drive4U prototype}. Source: Valeo.}
  \label{fig:intro_valeo_sensors}
\end{figure}

The automotive supplier Valeo is a world leader in the design and manufacturing of sensors for assistance driving. The company has been actively involved in autonomous driving research and development for almost ten years, designing sensor prototypes and real-world experiments (see Figure \ref{fig:intro_valeo_sensors}).
In this context, prototype cars are equipped with various sensors to record real-life scenes and to understand the complementarity and limitations between the sensors. 
The recorded and released data from the automotive industry in collaboration with researchers aim developing research on recent learning algorithms to better understand the car's environment. 

This thesis focuses on automotive \acp{RADAR} which are low-cost active sensors measuring properties of surrounding objects, including their relative speed, and have the key advantage of not being impacted by adverse weather conditions, \textit{e.g.} rain, snow or fog.
They have been used in the automotive industry for the last two decades, \textit{e.g.}, for automatic cruise control or blind spot detection. 
\ac{RADAR} has become the sensor of choice for applications requiring time to collision as it provides, besides localization, the radial velocity thanks to the Doppler information.
However, it is seldom used for scene understanding due to its poor angular resolution, the size, noise, and complexity of \ac{RADAR} raw data as well as the lack of available datasets.  
Representations provided by a \ac{RADAR} sensor are moreover difficult to understand for non-specialist humans.
For these reasons, \ac{RADAR} sensors were left out of scene understanding using learning algorithms.
This thesis aims to exploit \ac{RADAR} data for object recognition in complex driving scenes using modern machine learning approaches.

Machine learning, as part of artificial intelligence, is the study of computer algorithms that automatically improve their performance by experience. 
These algorithms use a dataset, or training data, to learn to recognize patterns and make predictions.
Machine learning algorithms have rapidly developed since the advent of the digital economy and the abundance of available data. 
They are applied in many fields, \textit{e.g.} to predict consumer behaviour, estimate financial time series, detect cancer cells in medical images, forecast weather, rank user preferences in video platforms or understand scenes and take decisions in autonomous driving.

This thesis is centered on a specific category of machine learning algorithms called \acfp{ANN}.
In the human brain, a neuron receives electrical signals via \textit{dendrites}. The signals are then processed in the \textit{soma}, transmitted by the \textit{axon} if the signal is above a threshold, and transferred to the next neuron with the \textit{synapses}.
Inspired by this phenomenon, \cite{mcculloch_logical_1943} proposed the first mathematical formulation of an artificial neuron processing a signal with fixed weights (similarly to \textit{dendrites}) and applying a threshold on the output.
An extension proposed by \cite{rosenblatt_perceptron_1958} updates the weights regarding the prediction error of the artificial neuron. This algorithm called the Perceptron is the first biologically inspired machine learning algorithm from the human brain.
In the late 1960's, \cite{minsky_perceptrons_1969} showed the limitations of a single artificial neuron which is a linear classifier and therefore cannot solve non-linearly separable classification problems (\textit{e.g.} XOR). They proposed to overcome this problem by stacking artificial neurons in layers and by connecting the neurons from a layer to another creating a neural network, also called \acf{MLP}.

\begin{figure}[!t]
  \begin{center}
    \includegraphics[width=1\textwidth]{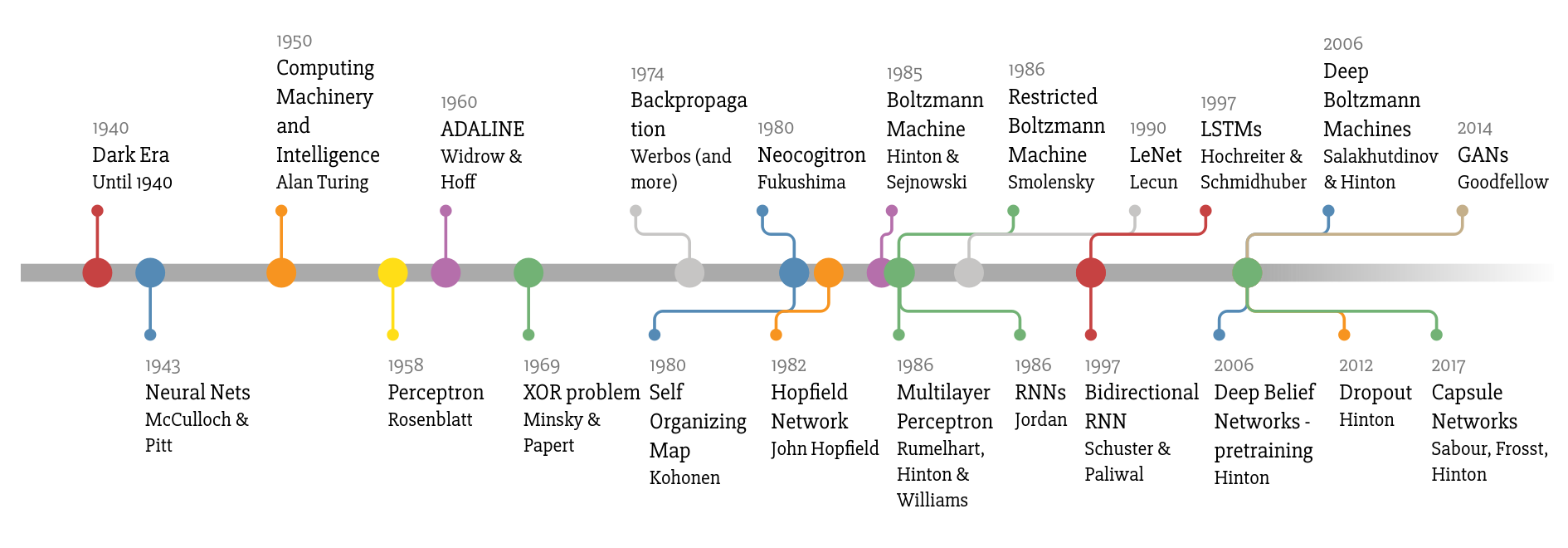}
  \end{center}
  \caption[Timeline of the evolution of Artificial Neural Networks and deep learning]{\textbf{Timeline of the evolution of Artificial Neural Networks and deep learning}\footnotemark.}
  \label{fig:intro_history_deep}
\end{figure}

\footnotetext{\sloppy\url{https://towardsdatascience.com/a-weird-introduction-to-deep-learning-7828803693b0}}

Over the years, several improvements have been published, as illustrated in Figure \ref{fig:intro_history_deep}, proposing neural network structures or innovating methods to train these algorithms.
An important step forward is the \acf{CNN}, introduced by \cite{lecun_backpropagation_1989} associated with a new learning method, which is particularly adapted to image analysis. However, it was hampered by the computer limitations of the time.
In the 2010's, the increase of computing power, in particular with \acfp{GPU}, and the availability of large-scale databases motivated the emergence of deep learning.
Deep learning is a subset of machine learning that involves stacking many layers in a neural networks.
These algorithms were highlighted with a major milestone reached in 2012 with the \acf{ILSVRC}, consisting of classifying images into 1,000 categories (\textit{e.g.} cats, dogs, or food).
The first deep learning model proposed by \cite{krizhevsky_imagenet_2012} outperformed all the existing methods classifying image in a category with a large margin.
In 2015, another deep learning model introduced by \cite{he_deep_2016} first beat human performance on the ImageNet challenge making deep learning methods the front page of the computer vision research community.

Since the 2010's, deep neural networks have outperformed humans in face recognition \cite{taigman_deepface_2014} or in the game of Go with AlphaGo \cite{silver_mastering_2016}. The \ac{ANN} have been adapted to many types of data: images, videos, point clouds, texts, sounds, time series or DNA sequences.
These impressive performances have attracted the automotive industry, which sees these algorithms as an answer to the fundamental requirements for autonomous driving.
They are particularly well suited for scene understanding to detect and classify objects in a car's environment using camera or \ac{LiDAR} data. 
This thesis aims to investigate deep learning methods for \ac{RADAR} data, which have barely been explored in recent years.

\section{Motivations}

\begin{figure}[!t]
  \begin{center}
    \includegraphics[width=1\textwidth]{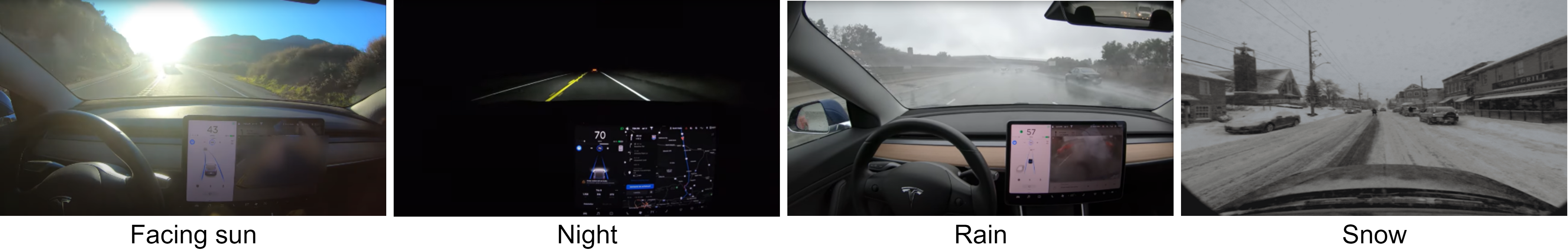}
  \end{center}
  \caption[Camera image examples of scenes with different lighting and weather conditions]{\textbf{Camera image examples of scenes with different lighting and weather conditions}.}
  \label{fig:intro_img_ex}
\end{figure}

\sloppy
Scene understanding requires a high level of perception around the car. For this purpose, complementary sensors are used to compensate their respective weaknesses.
Cameras record images that provide a comprehensive understanding of a scene, but they are impacted by lighting conditions, \textit{e.g.} during night or when facing the sun, or by adverse weather conditions that reduce their visibility (see Figure \ref{fig:intro_img_ex}).
\ac{LiDAR} is an active sensor transmitting laser beams in the environment and is not affected by lighting conditions.
This sensor measures the time it takes to receive reflected light and records the distance and the intensity of a reflection. 
The \ac{LiDAR} sensor provides a 3D point cloud measuring the geometry of a scene with a resolution below the degree for both azimuth and elevation angles.
However, the laser beams are reflected by droplets or snowflakes due to their wavelength. Difficult weather conditions create artefacts in the recorded 3D point clouds \textit{e.g.} with rain \cite{karlsson_probabilistic_2021}, snow \cite{kurup_dsor_2021} or fog \cite{bijelic_benchmark_2018, guan_through_2020}.
The \ac{RADAR} sensor emits electromagnetic waves, which are not impacted by adverse weather conditions, and records the location, Doppler and reflectivity of objects in the scenes through the received signals.
This sensor is an interesting candidate for scene understanding because its properties compensate for other sensors in specific scenarios. 

With the rapid progress of deep learning and the availability of public driving datasets, \textit{e.g.}, \cite{geiger_vision_2013, cordts_cityscapes_2016, huang_apolloscape_2019, yu_bdd100k_2020, sun_scalability_2020}, the perception ability of vision-based driving systems (detection of objects, structures, markings and signs, estimation of depth, forecasting of other road users’ movements) has considerably improved. This progress quickly extended to \ac{LiDAR}, with the help of specific architectures to deal with 3D point clouds \cite{qi_pointnet_2017}.
To improve further the performance of autonomous driving systems, extending the size and scope of open annotated datasets is a key challenge.
The \ac{RADAR} sensor provides cumbersome and noisy representations which are difficult to understand. For these reasons, there was no open source \ac{RADAR} dataset for automotive application before 2019, which has hampered research on deep learning applied to \ac{RADAR} data.
Motivated by the relevance of this sensor, this thesis fills the gap in existing works by proposing unique datasets and methods to generate annotations avoiding costly manual annotations.

While deep learning has brought major progresses in the automotive use of cameras and \acp{LiDAR} -- for object detection and segmentation in particular -- it is only recently that it has also embraced \ac{RADAR} signals. In fact, even though the \ac{RADAR} technology has greatly improved, the signal processing pipeline has remained the same for years. 
This sensor is now a source of interest since public datasets have been released as detailed in Section \ref{sec:related_datasets}.
Motivated by the recent advances in deep learning algorithms, this thesis proposes to adapt neural network architectures for \ac{RADAR} scene understanding.

\section{Contributions and outlines}

This thesis proposes an extensive analysis of \ac{RADAR} scene understanding, from the construction of an annotated dataset to the conception of adapted deep learning architectures.

Chapter \ref{chap:background} provides the background on \ac{RADAR} theory and signal processing methods applied to the recorded data. It also briefly introduces the theory behind \ac{ANN} as well as present deep learning architectures based on images which are exploited in our work.

Chapter \ref{chap:related_work} relates the existing \ac{RADAR} datasets and previous works on \ac{RADAR} scene understanding for object detection and semantic segmentation using deep learning algorithms. It also presents previous sensor fusion approaches combining \ac{RADAR} with either or both camera and \ac{LiDAR} sensors for automotive scene understanding.

Chapter \ref{chap:datasets} details approaches to tackle the lack of data. A simple simulation as well as generative methods for creating annotated data are presented.
This chapter also introduces one major contribution, the CARRADA dataset, composed of synchronised camera and \ac{RADAR} data with a semi-automatic method generating annotations on the \ac{RADAR} representations. Today, the CARRADA dataset is the only dataset providing annotated raw \ac{RADAR} data for object detection and semantic segmentation tasks.

Chapter \ref{chap:radar_scene_understanding} presents our second major contribution, a set of deep learning architectures with their associated loss functions for \ac{RADAR} semantic segmentation. Today, the proposed architecture performing the best outperforms alternative models, derived either from the semantic segmentation of natural images or from \ac{RADAR} scene understanding, while requiring significantly fewer parameters.
This chapter also introduces a work in progress to open up research into the fusion of \ac{LiDAR} and \ac{RADAR} sensors for scene understanding.

Chapter \ref{chap:hd_radar} exposes a collaborative contribution, the RADIal dataset with synchronised \acf{HD} \ac{RADAR}, \ac{LiDAR} and camera. A deep learning architecture is also proposed to estimate the \ac{RADAR} signal processing pipeline while performing multi-task learning for object detection and free driving space segmentation\footnote{The free driving space segmentation task consists to locate pixel-wise the available space that can be driven.} simultaneously. Today, RADIal is the only dataset providing raw data from a \ac{HD} \ac{RADAR} with annotations for both object detection and free space segmentation. Moreover, the proposed method is the only one to successfully learn a part of the \ac{RADAR} pre-processing pipeline in a multi-task framework using \ac{HD} \ac{RADAR} data. 

Finally, Chapter \ref{chap:conclusion} highlights the contributions of this thesis and discuss the perspectives of exploiting \ac{RADAR} data for scene understanding in the context of autonomous driving.
\\

\noindent
The contributions presented in this thesis have led to the following publications: 
\begin{tabular}{p{0.85\textwidth} c}
\toprule
Publication & Chapter \\
\midrule
\textbf{Arthur Ouaknine}, Alasdair Newson, Julien Rebut, Florence Tupin and Patrick Pérez. ``CARRADA Dataset: Camera and Automotive Radar with Range-Angle-Doppler Annotations'', in \textit{International Conference on Pattern Recognition (ICPR)}, 2020. & \ref{chap:datasets} \\
\midrule
\textbf{Arthur Ouaknine}, Alasdair Newson, Patrick Pérez, Florence Tupin and Julien Rebut. ``Multi-View Radar Semantic Segmentation'', in \textit{International Conference on Computer Vision (ICCV)}, 2021. & \ref{chap:radar_scene_understanding} \\
\midrule
Julien Rebut, \textbf{Arthur Ouaknine}, Waqas Malik and Patrick Pérez. ``Raw High-Definition Radar for Multi-Task Learning'', in \textit{Conference on Computer Vision and Pattern Recognition (CVPR)}, 2022. & \ref{chap:hd_radar} \\
\bottomrule
\end{tabular}

\chapter{Background}
\label{chap:background}
\minitoc

This chapter provides the background to understand the automotive \ac{RADAR} sensor and the neural network machine learning algorithms. Section \ref{sec:background_radar_theory} presents the theory of the automotive \ac{RADAR} sensor. Section \ref{sec:background_signal_process} details the pipeline commonly used to process the recorded \ac{RADAR} signals. Sections \ref{sec:background_artificial_nn}, \ref{sec:background_cnn} and \ref{sec:background_rnn} introduce the \ac{ANN}, \acf{RNN} and \acf{CNN} respectively. Finally, Section \ref{sec:background_deep} relates deep learning methods and architectures which are exploited in the following chapters.

\section{RADAR theory}
\label{sec:background_radar_theory}
A \acf{RADAR} sensor is an active sensor using its own source of signal. 
It generates electromagnetic waves which are emitted via one or several \acp{Tx}. 
The wavelengths used are significantly larger than the visible spectrum and therefore not affected by lighting or adverse weather conditions.
The waves are reflected by an object depending on the composition and the geometry of its surface. The signals are then received by the \ac{RADAR} via one or several \acp{Rx}. 
The comparison between the transmitted and the received waveforms allows inferring the distance, the relative velocity, the azimuth angle and the elevation of the reflector regarding the \ac{RADAR} position \cite{ghaleb_micro-doppler_2009} and the positioning of its antennas. 
Most of the automotive \acp{RADAR} use \ac{MIMO} systems \cite{donnet_mimo_2006}: each couple of (\ac{Tx}, \ac{Rx}) receives the reflected signal assigned to a specific \ac{Tx} transmitting a waveform. 

\begin{figure}[!t]
  \begin{center}
    \includegraphics[width=0.5\textwidth]{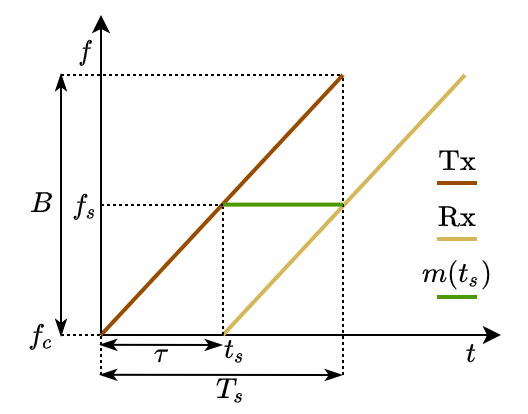}
  \end{center}
  \caption[Example of a RADAR chirp]{\textbf{Example of a RADAR chirp}. The chirp is generated considering a linearly modulated frequency as described in Equation \ref{eq:transmit_freq}. The signal is transmitted by a \ac{Tx} antenna and received by a \ac{Rx} antenna for each reflector in the environment. The IF signal is deduced from a mixer by comparing the transmitted and received signals (see Equation \ref{eq:mixed_signals}).
  A signal reflected by a single object is illustrated here. Multiple signals are separated regarding the time delay of the signal round-trips and the phase shift of the waves. It would be illustrated by distinct IFs, one per object.}
  \label{fig:chirp_schema}
\end{figure}

A \ac{FMCW} \ac{RADAR} transmits a signal, called a chirp \cite{brooker_understanding_2005}, whose frequency 
is linearly modulated over the sweeping period $T_s$: at time $t_s \in \{0, \cdots, T_s\}$, the emitted sinusoidal signal has the frequency:
\begin{equation} \label{eq:transmit_freq}
    f_s  = f_c + \frac{B}{T_s}t_s ,
\end{equation}
where $f_c$ is the carrier frequency, $B$ the bandwidth, $B/T_s$ is the linear slope of the frequency variation,
and its phase reads
\begin{equation}
    \phi_E(t) = 2\pi f_s t.
\end{equation}

After reflection on an object at distance $r(t)$ from the emitter, the received signal has phase:
\begin{equation}
\label{eq:phase_shift}
    \phi_R(t) = 2\pi f_s (t-\tau)  = \phi_E(t) - \phi (t),
\end{equation}
where $\tau=\frac{2 r(t)}{c}$ is the time delay of the signal round trip, with $c$ the velocity of the wave through the air considered as constant, and $\phi(t)$ is the phase shift:
\begin{equation}
    \phi (t) = 2 \pi f_s \tau = 2 \pi f_s \frac{2r(t)}{c}.
\end{equation}
Measuring this phase shift (or equivalently the time delay between the transmitted and the reflected signal) grants access to the distance between the sensor and the reflecting object.

Its relative velocity is accessed through the frequency shift between the two signals, a.k.a. the Doppler effect.
Indeed, the phase shift varies when the target is moving:
\begin{equation}
    f_d = \frac{1}{2\pi} \frac{d \phi}{d t} = \frac{2 v_R}{c}f_s ,
\end{equation}
where $v_R = d r / d t$ is the radial velocity of the target object w.r.t. the \ac{RADAR}.
This yields the frequency Doppler effect whereby frequency change rate between transmitted and received signals, $\frac{f_d}{f_s} = \frac{2 v_R}{c}$, depends linearly on the relative speed of the reflector. Measuring this Doppler effect hence amounts to recovering the radial speed:
\begin{equation}
    v_R = \frac{c f_d}{2 f_s}.
\end{equation}

Radar's capability to discriminate between two targets with same range and velocity but different angles is called its angular resolution. 
It is directly proportional to the antenna aperture, that is, the distance between the first and last receiving antennas. 
The time delay between the received signals of each Rx transmitted by a given \ac{Tx} carries the orientation information of the object.

The \ac{MIMO} approach \cite{donnet_mimo_2006} is commonly used to improve the angular resolution without increasing the physical aperture: angular resolution increases by a factor of 2 for each added emitting antenna (\ac{Tx}). 
Denoting $\NTx$ and $\NRx$ the number of its Tx and Rx channels respectively, a \ac{MIMO} system builds a virtual array of $\NTx \cdot \NRx$ antennas. 
The downside of this approach is that the reflected signal of an object appears $\NTx$ times, making the data interleaved. The \ac{RADAR} representation is easily de-interleaved considering a \ac{LD} \ac{RADAR} since it has at most 2 Tx antennas. This processing step applied to a \ac{HD} \ac{RADAR} is a greedy task; additional details are provided in Chapter \ref{chap:hd_radar}.

Depending on the positioning of the antennas, the azimuth angle and the elevation of the object are respectively deduced from the horizontal and vertical pairs of (\ac{Tx}, \ac{Rx}).
The \ac{AoA} is deduced from the variation between the phase shift of adjacent pairs of \ac{Rx}.
In particular, the azimuth angle\footnote{A geometric illustration is depicted in Figure \ref{fig:relative_velocity}} $\alpha$ is obtained with the phase shift variation of horizontal adjacent \ac{Rx} noted $\Delta \phi_\alpha = 2 \pi f_s \frac{2 h \sin \alpha}{c}$, where $h$ is the distance separating the adjacent receivers.

\section{Recordings and signal processing}
\label{sec:background_signal_process}

\begin{figure}[!t]
  \begin{center}
    \includegraphics[width=1\textwidth]{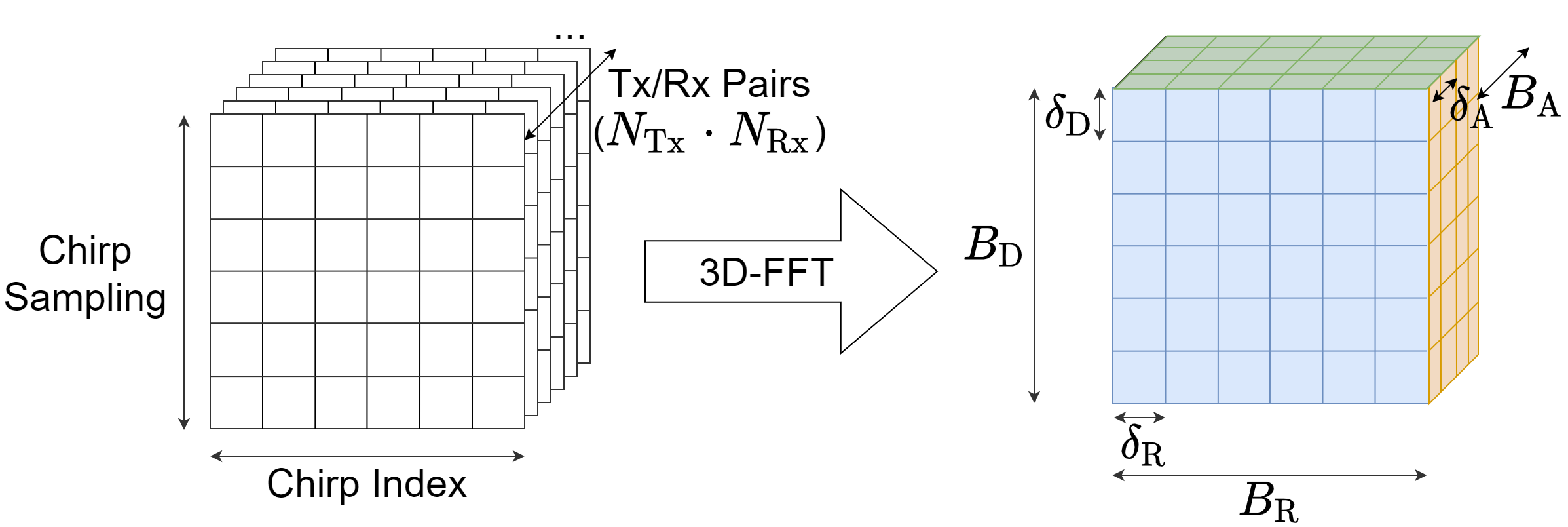}
  \end{center}
  \caption[Range-Angle-Doppler (RAD) tensor generation]{\textbf{Range-Angle-Doppler (RAD) tensor generation}. The signals received by all pairs of (\ac{Tx}, \ac{Rx}) antennas are transformed and stored in a 3D tensor in the frequency domain. Inverse FFT operations are respectively applied on the chirp index axis to deduce the range, on the chirp sampling axis (\textit{i.e.} the frequency sampling for each chirp) for the Doppler and on the (\ac{Tx}, \ac{Rx}) pairs axis for the angle. The generated RAD tensor in the temporal domain has dimensions $\BinR {\times} \BinA {\times} \BinD$. Each of its axes has a resolution of $\delta_\text{R}$, $\delta_\text{A}$ and $\delta_\text{D}$ respectively depending on the \ac{RADAR} sensor specificities. 
  The geometry and the number of \ac{RADAR} antennas define the recording resolutions in azimuth ($\delta_\alpha$) and elevation ($\delta_\text{E}$) angles.
  Considering low-definition \ac{RADAR}, the positioning of the antennas is only able to obtain the azimuth angle and $\delta_\text{A} = \delta_\alpha$.
  }
  \label{fig:schema_fft}
\end{figure}

\subsection{Transformations in the temporal and frequency domains}

A \ac{RADAR} sensor transmits and receives signals, respectively noted $s_E$ and $s_R$ are written:
\begin{align}
    s_E(t) &= A_E e^{j2 \pi f_s t} = A_E e^{j \phi_E(t)}, \\
    s_R(t) &= A_R e^{j2 \pi f_s (t - \tau)} = A_R e^{j \phi_R(t)},
\end{align}
where $A_E$ and $A_R$ denote their amplitude. 
The phase shift induced by the time delay between the transmitted and received signal is computed by correlating the two signals.
The convolution between two signals in the temporal domain is equivalent to a point-wise product of their \ac{DFT}. Therefore, the transmitted and received signals are compared in the frequency domain.

The \ac{FFT} algorithm applies a \ac{DFT} to the data from the temporal domain to the frequency domain. 
Let $x: \{ 0, \cdots, N_{T}-1 \} \rightarrow \mathbb{C}$ be a sequence of $N_T$ complex signals. For $\nu \in \{ 0, \cdots, N_{f}-1 \}$ a finite sequence of $N_f$ frequencies, the \ac{DFT} of $x$ is defined as:
\begin{equation}
    \hat{x}(\nu) = \sum_{t=0}^{N_{T}-1}x(t) e^{-j2 \pi \nu \frac{t}{N_{T}}}.
\end{equation}
A \ac{FFT} is applied on $N_T$ complex signals for each transmitted and received signals in the temporal domain, their \ac{DFT} is written:
\begin{align}
    \hat{s}_E(\nu) &= \sum_{t=0}^{N_T-1} s_E(t) e^{-j2\pi \nu \frac{t}{N_T}}, \\
    \hat{s}_R(\nu) &= \sum_{t=0}^{N_T-1} s_R(t) e^{-j2\pi \nu \frac{t}{N_T}}.
\end{align}
The signals are compared in the frequency domain with a mixer that generates the \ac{IF} signal:
\begin{align}
\label{eq:mixed_signals}
    \hat{m}(\nu) = \hat{s}_E(\nu) \cdot \hat{s}_R(\nu).
\end{align}
%
The \ac{IF} signal, illustrated in Figure \ref{fig:chirp_schema}, is deduced from the signal transmitted by \ac{Tx} and received by \ac{Rx}.
The mixed signal is filtered using a low-pass filter and digitized by an analog-to-digital converter (\acs{ADC}). This way, the recorded signal carries the Doppler frequencies, ranges and angles of all reflectors.

Consecutive filtered \ac{IF} signals are stored in a frame buffer which is a frequency-domain 3D tensor:
the first dimension corresponds to the chirp index; the second one is the chirp sampling defined by the linearly modulated frequency range; the third tensor dimension indexes (\ac{Tx}, \ac{Rx}) antenna pairs.

The recorded data are finally transformed in the temporal domain as illustrated in Figure \ref{fig:schema_fft}. 
\textit{E.g.} the inverse \ac{FFT} applied on the chirp index axis processes an inverse \ac{DFT} with $N_f$ frequencies on the \ac{IF} signals (Equation \ref{eq:mixed_signals}) as:
\begin{equation}
    m(t) = \sum_{\nu=0}^{N_{f}-1} \hat{m}(\nu) e^{j 2\pi t \frac{\nu}{N_{f}}}. 
    \label{eq:range_fft}
\end{equation}
This method estimates the peak reached in the \ac{IF} signal in the temporal domain due to time delay between the transmission and reception of the signals \cite{ghaleb_micro-doppler_2009, molchanov_radar_2014}. Considering the chirp index, the distance of the reflector is then deduced from the estimated time delay.

The 3D tensor in the frequency domain (see Figure \ref{fig:schema_fft}) is processed using an inverse three dimensional fast Fourier transform (3D-FFT): a range-\ac{FFT} along the rows deducing the object range, a Doppler-\ac{FFT} along the columns deducing the objects' relative velocity and an angle-\ac{FFT} along the depth deducing the angle between two objects (see the range-\ac{FFT} example in Equation \ref{eq:range_fft}). 
This sequence of inverse \ac{FFT}s results in the \ac{RAD} tensor of dimensions $\BinR {\times} \BinA {\times} \BinD$, a 3D data cube of complex numbers where each axis amounts to discretised values of the corresponding physical measurement as illustrated in Figure \ref{fig:schema_fft}. 

The range, velocity and angle bins in the output tensor correspond to discretized values defined by the resolution of the \ac{RADAR}. 
The range resolution is defined as  $\delta R = \frac{c}{2B}$. The relative velocity, or Doppler resolution $\delta D = \frac{c}{2 f_c T}$ is inversely proportional to the frame duration time. The angle resolution $\delta \alpha = \frac{c}{f_c N_{\text{Rx}} h \cos(\alpha)}$ is the minimum angle separation between two objects to be distinguished \cite{iovescu_fundamentals_2017}, with $N_{\text{Rx}}$ the number of \ac{Rx} antennas and $\alpha$ the horizontal azimuth angle between the \ac{RADAR} and an object reflecting the signal.

\begin{figure}[!t]
  \begin{center}
    \includegraphics[width=0.7\textwidth]{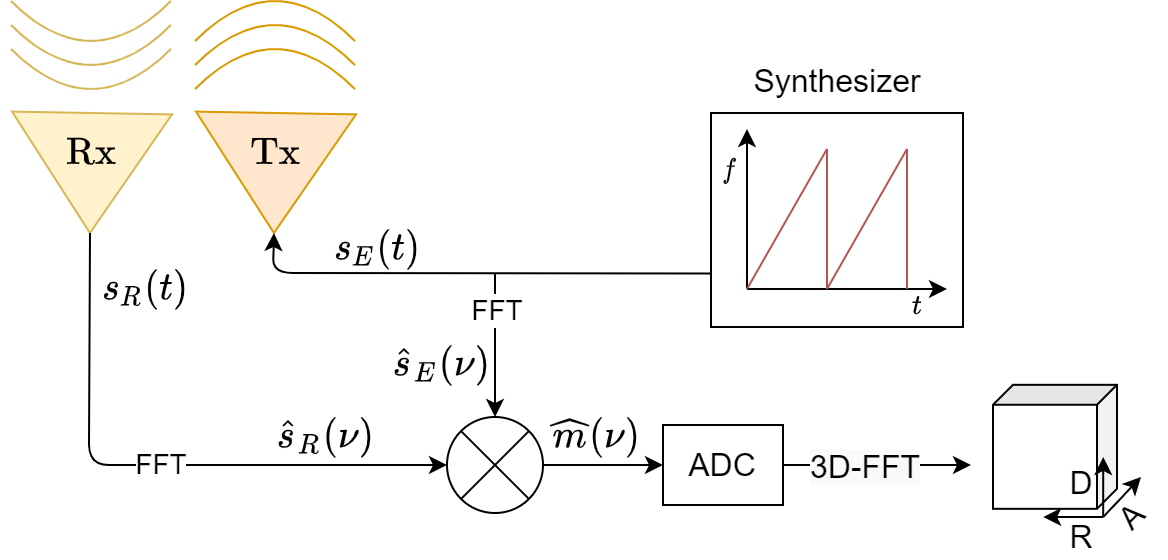}
  \end{center}
  \caption[Outline of the RADAR pipeline]{\textbf{Outline of the RADAR pipeline}. An electromagnetic wave is synthetized and transmitted in the environment with a \ac{Tx} antenna. The received signal is mixed with the original one and stored in a tensor in the frequency domain. An inverse \acl{FFT} is applied on each axis of the recorded data to generate the \acl{RAD} tensor in the temporal domain.}
  \label{fig:signal_processing_pipeline}
\end{figure}

The outline of the \ac{RADAR} pipeline, which is from the synthesis of the electromagnetic wave to the generation of the \ac{RAD} tensor, is depicted in Figure \ref{fig:signal_processing_pipeline}. The following section will describe the speckle noise strongly represented in \ac{RADAR} data.

\subsection{Speckle noise}
\label{sec:background_speckle}

\sloppy
The intensities (squared modulus) in the \ac{RAD} tensor provide information on the power backscattered by surrounding objects. A visualisation of the tensor by 2D pairs of axes is proposed in Appendix \ref{sec_app:mvrss_rad_tensor_vis}. It can be noticed that information is redundant regarding each slice of the tensor considering its pairs of axes.

Moreover, the backscattered signal presents strong fluctuations also known as speckle phenomenon and which have been well-studied by Goodman \cite{goodman_fundamental_1976}. The fluctuations are due to the addition of complex backscattered signals of elementary scatterers in the same resolution cell leading to a random intensity record in the absence of a significant reflector.
The model proposed by Goodman assumes that a large number of echoes are produced by individual scatterers in a resolution cell with \ac{iid} complex amplitudes. This phenomenon can be modeled as a multiplicative noise and can be written:
\begin{equation}
    w = u \times s,
\end{equation}
where $w$ is the measured intensity in a resolution cell, $u$ the underlying reflectivity and $s$ the speckle \cite{dalsasso_sar2sar_2021}.
According to the work of \cite{goodman_fundamental_1976} and considering multi-looked observations\footnote{The multi-looking method, used with Synthetic Aperture RADAR (SAR), consists in aggregating intensity values of different looks which can be obtained in different ways.},
%
%
the speckle can be modeled as a random variable following a gamma distribution defined as:
\begin{equation}
    p(s) = \frac{L^L}{\Gamma(L)} s^{L-1} e^{-Ls},
\end{equation}
where $L \geq 1$ is the number of looks and $\Gamma(.)$ the gamma function. As described by \cite{goodman_fundamental_1976}, $\mathbb{E}[s] = 1$ and $\mathbb{V}[s] = \frac{1}{L}$, the higher the number of looks, the lower the variance of the speckle. 

The speckle becomes additive after a logarithmic transform applied to the recorded intensity while stabilizing its variance \cite{deledalle_mulog_2017}. The log-transformed intensity $\tilde{w}$ is thus written:
\begin{equation}
    \tilde{w} = \tilde{u} + \tilde{s},
\end{equation}
where $\tilde{s}$ is the log-speckle following a Fisher-Tippett distribution given by:
\begin{equation}
    p(\tilde{s}) = \frac{L^L}{\Gamma(L)} e^{L\tilde{s}} e^{-Le^{\tilde{s}}}.
\end{equation}
The expectation and variance of the log-speckle no longer depend on the underlined reflectivity. They are respectively noted  $\mathbb{E}[\tilde{s}] = \psi(L) - \log(L)$ and $\mathbb{V}[s] = \psi(1, L)$, where $\psi(.)$ is the diagamma function and $\psi(., .)$ is the polygamma function of order L \cite{abramowitz_handbook_1965}. 

The multi-looking operation, consisting in averaging a few samples, reduces the noise strength, whereas the logarithmic transform induces a variance stabilization of the resulting signal \cite{deledalle_mulog_2017, dalsasso_sar2sar_2021}.
The next section will describe the transformations applied to the previously presented \ac{RAD} tensor to obtain interpretable \ac{RADAR} representations while reducing the speckle noise.

\subsection{RADAR representations}

\begin{figure}[!t]
  \begin{center}
    \includegraphics[width=1\textwidth]{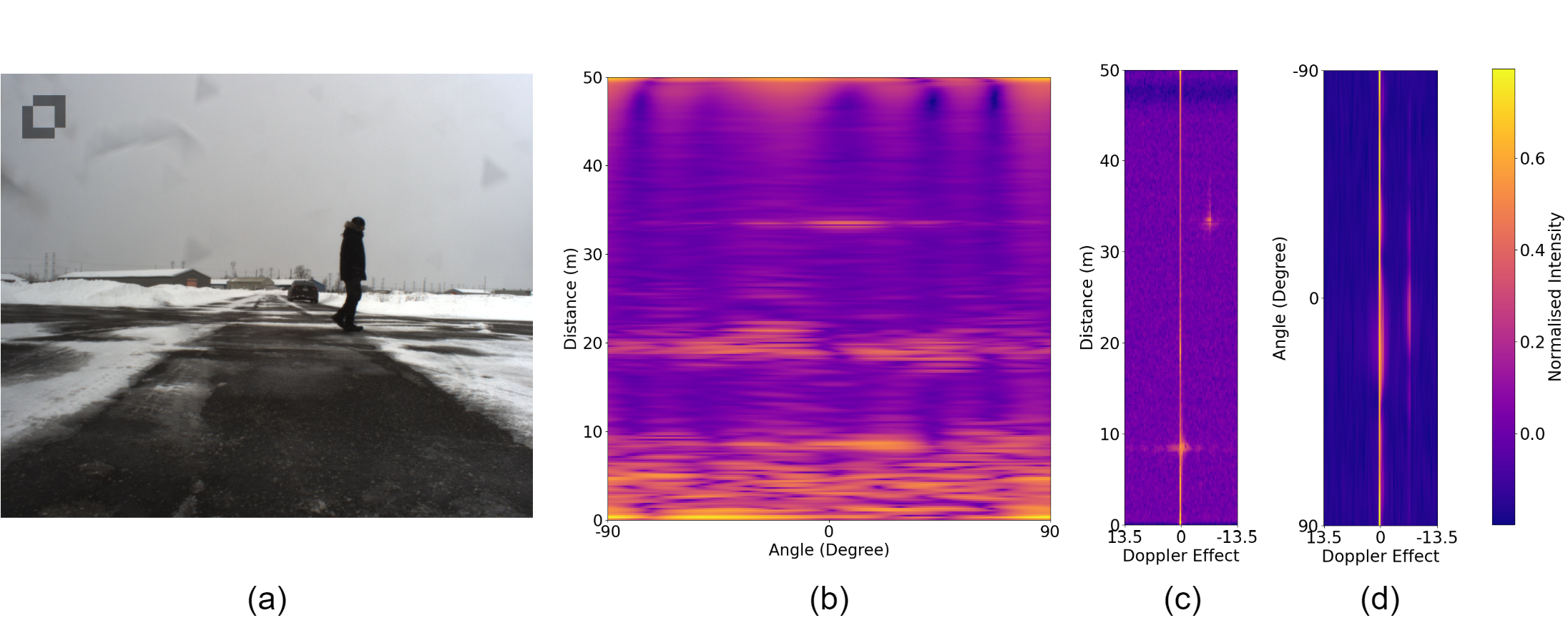}
  \end{center}
  \caption[An example of RADAR representations]{\textbf{An example of RADAR representations}. The \acf{RAD} tensor corresponding to the scene, illustrated with the camera image (a), is aggregated using Equation \ref{eq:agg_method_average} to create the 
  (b) Range-Angle, (c) Range-Doppler and (c) Angle-Doppler views corresponding to each 2D pair of axis of the tensor. This scene is a sample of the CARRADA dataset presented in Section \ref{sec:carrada}.
  }
  \label{fig:background_radar_data}
\end{figure}

Let $\vect{X}^{\text{RAD}}$ be the complex \ac{RAD} tensor.
Aggregating its intensities across one of its dimensions leads to three possible 2D views of the tensor: \ac{RA}, \ac{RD} and \ac{AD} illustrated in Figure \ref{fig:background_radar_data}.
For instance, we define the \ac{RA} view, expressed in decibels, as:
\begin{equation}
    \vect{x}^{\text{RA}}[r,a] = \Phi_{\text{D}}(\vect{X}^{\text{RAD}}),
\label{eq:general_agg_method}
\end{equation}
where $\Phi_{\text{D}}$ is an arbitrary aggregation function compressing the \ac{RAD} tensor through its Doppler dimension. 
As detailed in Section \ref{sec:background_speckle}, an averaging followed by logarithm transform are commonly used in practice.
In this work, we average the intensity of the \ac{RAD} tensor on its third axis to reduce the speckle noise.
Note that the hypothesis of independent and identically distributed samples is no longer verified along the third axis, which will have the consequence of attenuating the reflections\footnote{This limitation is discussed in Section \ref{sec:mvrss_conclusions} and Chapter \ref{chap:conclusion}.}.
With this method, the \ac{RA} view is then defined as:
\begin{equation}
    \vect{x}^{\text{RA}}[r,a] = 10 \log_{10}
    \Big(
    \frac{1}{\BinD} \sum_{d=1}^{\BinD} 
    \big |\vect{X}^\text{RAD}[r,a,d] \big|^2
    \Big),
\label{eq:agg_method_average}   
\end{equation}
with $|.|$ the modulus and $\BinD$ the number of Doppler bins. 
Turning \ac{RAD} tensors into views aims both to compress substantially the data representation and to reduce its noise while preserving the objects' signature. As a consequence, the view representations are interpretable, offering a viable way to exploit the temporal dimension in tensor sequences. They also meet practical computational and memory constraints imposed by in-car embedding.
Considering an example of application, it reduces the size of the data by a factor of 50\footnote{As an example, the CARRADA dataset \cite{ouaknine_carrada_2020}, detailed in Section \ref{sec:carrada_dataset}, provides \ac{RAD} tensors of 17MB each. Their corresponding aggregated views are 0.1MB for \ac{RD}/\ac{AD} and 0.5MB for \ac{RA}.}. An example of the \ac{RADAR} views is illustrated in Figure \ref{fig:background_radar_data}.

With conventional \ac{FMCW} \acp{RADAR}, the \ac{RAD} tensor is usually not available as it is too computationally intensive to estimate. A \ac{CFAR} algorithm \cite{rohling_radar_1983} is typically applied to filter \ac{RD} views. It keeps the highest intensity values while taking into account the local relation between points. The \ac{AoA} is then computed for each high reflection to obtain a sparse \ac{RADAR} point cloud, also called \ac{DoA} point cloud, in Cartesian coordinates while each point has its Doppler and \ac{RCS}.
To translate the \ac{AoA} into an effective angle, one needs to calibrate the sensor. 
An alternative to the third \ac{FFT} is to correlate in the complex domain the \ac{RD} spectrum with a calibration matrix, to estimate the angles (azimuth and elevation). 
The complexity of this operation for a single point of the \ac{RD} tensor is $\mathcal{O}(\NTx\NRx\BinA\BinE)$, where $\BinA$ and $\BinE$ are respectively the number of bins over which azimuth and elevation angles are discretized in the calibration matrix. 
A \ac{LD} \ac{RADAR} considers a single elevation ($\BinE=1$) and the azimuth resolution is low. This process is therefore computationally affordable to generate sparse \ac{RADAR} point clouds. However it meets a bottleneck with \ac{HD} \acp{RADAR}.

Multiple representations of \ac{RADAR} data are available depending on the signal processing pipeline applied on the recordings: \ac{ADC} data, \ac{RAD} tensor, \ac{RD}/\ac{RA}/\ac{AD} views or sparse point clouds.
The fewer the pre-processing of the data, the more precise the signal corresponding to the objects is, the more cumbersome the representation is and the higher the level of noise.
The point cloud is the lighter representation but the CFAR filtering has drawbacks: it drastically shrinks the shape of the objects' signature and loses small, low-reflection signatures, especially non-metallic objects such as pedestrians.
Selecting a \ac{RADAR} data representation depends on the application (classification, detection, semantic segmentation) and on a trade-off between volume of the data and the applied method.
The following section introduces the background on \ac{ANN}.

\section{Artificial neural networks}
\label{sec:background_artificial_nn}

\subsection{Introduction}

Machine learning is the process optimizing parameters of a statistical model to solve a task \cite{hastie_elements_2001}. 
The parameters of the model are adjusted to separate or combine the input data samples regarding redundant patterns.
The supervised learning scheme assumes that each data sample has a label corresponding to the task to solve (\textit{e.g.} categorizing the image of a cat, detecting cancer cells in medical images, predicting the values of a financial time series and so on). The optimization process consists in minimizing the error between the prediction of the model and the corresponding label. In other words, a machine learning algorithm is able to solve a task by learning to transform data.

Datasets are diverse and many different tasks can be learnt by a machine learning algorithm. The most common are classification (\textit{e.g.} categorizing an image), detection (\textit{e.g.} localizing an object in an image), regression (\textit{e.g.} predicting the future price of a product), denoising (\textit{e.g.} cleaning a corrupted sample) and machine translation (\textit{e.g.} translating a sentence from a language to another).

Let consider a supervised learning setting, where $\vect{X} = \left\{ \vect{x}_0, \cdots, \vect{x}_{N-1} \right\}$ is a set of data with $N$ samples and $\vect{Y} = \left\{ \vect{y}_0, \cdots, \vect{y}_{N-1} \right\}$ their corresponding ground-truth labels. 
A machine learning algorithm is a function noted $f_\theta(.)$ parameterized by a vector $\theta = ( \theta_0, \cdots, \theta_{d-1} )$, where $d$ is the number of parameters in the model. It tries to predict a label for each sample of the dataset as
\begin{equation}
    f_\theta (\vect{x}_i) = \hat{\vect{y}_i},
\end{equation}
where $\vect{x}_i \in \vect{X}$ and $\hat{\vect{y}_i} \in \hat{\vect{Y}} = \left\{ \hat{\vect{y}}_0, \cdots, \hat{\vect{y}}_{N-1} \right\}$ is the prediction of the model.

In practice, the dataset is randomly divided into a training set and a testing set, respectively $\left\{\vect{X}^{\text{Train}}, \vect{Y}^{\text{Train}} \right\}$ and $\left\{\vect{X}^{\text{Test}}, \vect{Y}^{\text{Test}} \right\}$.
It aims to estimate \ac{iid} training and testing sets with a random separation of the entire dataset.
Therefore, the algorithm will be evaluated on data that it has never seen during its training period.

Let $\mathcal{L}(., .)$ be a function quantifying the error between the ground truth and the predictions of the mode. The supervised learning optimization process tries to find a set of optimal parameters $\theta^\ast$ minimizing the error between the ground-truth labels and the predictions. During training, it is sought as:
\begin{equation}
    \theta^\ast \in \argmin_\theta \frac{1}{N^{\text{Train}}} \sum_{i=0}^{N^{\text{Train}}-1} \mathcal{L}(\vect{y}_i^{\text{Train}}, f_\theta(\vect{x}_i^{\text{Train}})),
\end{equation}
where $N^{\text{Train}}$ is the number of elements in $\vect{X}^{\text{Train}}$.
Once the optimization process has converged, \textit{i.e.} the training loss oscillates under a threshold, the generalization error of the model is evaluated as $\frac{1}{N^{\text{Test}}} \sum_{i=0}^{N^{\text{Test}}-1} \mathcal{L}(\vect{y}_i^{\text{Test}}, f_{\theta^\ast}(\vect{x}_i^{\text{Test}}))$, where $\frac{1}{N^{\text{Test}}}$ is the number of element in $\vect{X}^{\text{Test}}$, or with a defined evaluation metric related to the task to solve.

\subsection{At the neuron level}

\begin{figure}[!t]
  \begin{center}
    \includegraphics[width=1\textwidth]{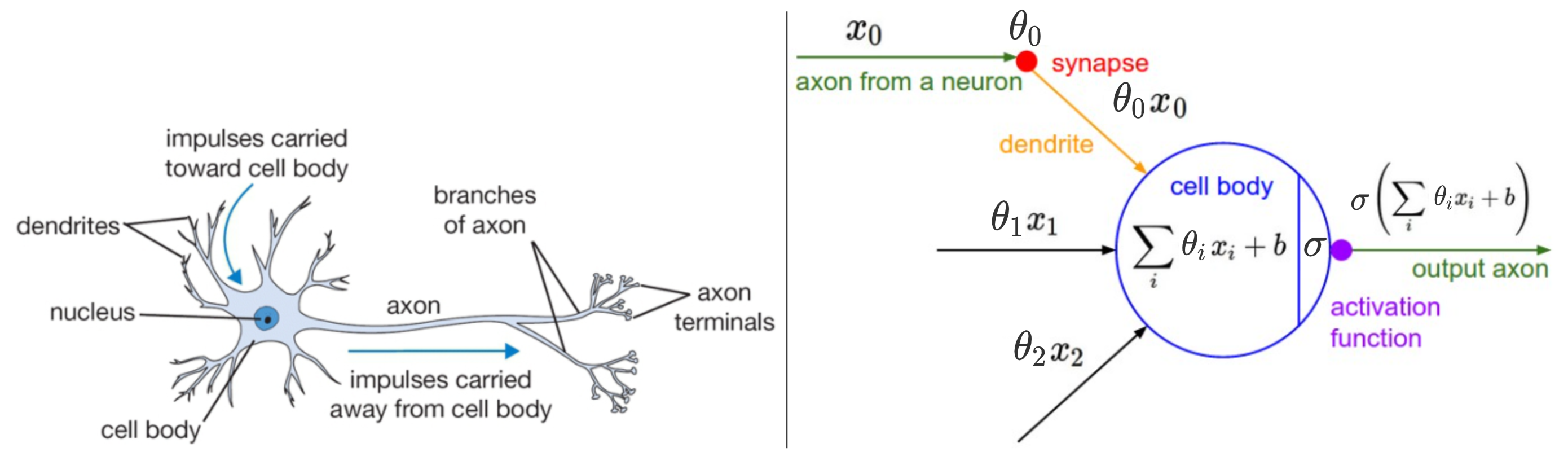}
  \end{center}
  \caption[Illustration of a biological neuron and its mathematical model]{\textbf{Illustration of a biological neuron (left) and its mathematical model (right)}. \cite{karpathy_cs231n_2021}.}
  \label{fig:background_neuron}
\end{figure}

In the human brain, a neuron receives electrical signals via \textit{dendrites} which are processed in the \textit{soma} (the cell body) and transmitted by the \textit{axons} if the signal is above a threshold. The \textit{synapses} finally transfer the signals to another neuron using the \textit{axon} terminals. A representation is depicted in Figure \ref{fig:background_neuron} (left).
A simple mathematical formulation of a biological neuron as illustrated in Figure \ref{fig:background_neuron} (right), we will name it \ac{AN}.

The Perceptron \cite{rosenblatt_perceptron_1958} algorithm is the simplest \ac{AN} optimizing its weights via a machine learning process.
It is a supervised linear classifier predicting a binary outputs as
\begin{equation}
    f_\theta (\vect{x}) = 
    \begin{cases}
    1 \quad \text{if} \quad \theta \cdot \vect{x} + b > 0, \\
    0 \quad \text{otherwise},
    \end{cases}
\end{equation}
where $b$ a bias scalar value. 
The optimization process will converge only if the data are linearly separable, which is usually not the case.
An activation function, noted $\sigma(.)$, is introduced to mimic the \textit{axon} behavior thresholding the signals in a biological neuron. This function aims to introduce a non-linearity and extend its capacity to separate the data.
The mathematical formulation of an \ac{AN} is written 
\begin{equation}
    f_\theta (\vect{x}) = \sigma (\theta \cdot \vect{x} + b),
\end{equation}
which is equivalent to the Perceptron when $\sigma$ is a Heaviside step function.
The most common activation functions are:
\begin{bulletList}
    \item The sigmoid function ``squashing'' the values into a range of $[0, 1]$: $\sigma(a) = \frac{1}{1 + \exp(a)}$;
    \item The hyperbolic tangent (tanh) ``squashing'' the values into a range of $[-1, 1]$: $\sigma(a) = \frac{2}{1+\exp(-2a)}-1$;
    \item The \ac{ReLU} function thresholding the value to zero: $\sigma(a)= \max (0, a)$;
    \item The \ac{LeakyReLU} function allowing a small negative slope of the input value: $\sigma(a)= \max (\epsilon a, a)$, where $\epsilon$ is the slope.
\end{bulletList}
Note that the activation functions have linear, or approximately linear, regions according to their limits. An activation is saturated when it reaches a linear region and has a constant, or almost constant value.

When the sigmoid function is used as the activation function, the neuron behaves like a binary linear classifier, a.k.a. logistic regression, providing a probability to belong to a class. The activation function is chosen according to the problem to solve and the expected distribution of the data.

\subsection{At the layer level}

\begin{figure}[!t]
  \begin{center}
    \includegraphics[width=0.5\textwidth]{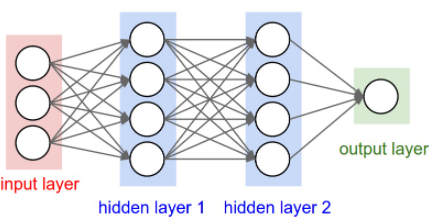}
  \end{center}
  \caption[Illustration of a Multilayer Perceptron]{\textbf{Illustration of a \acl{MLP}}. The illustrated algorithm \cite{karpathy_cs231n_2021} is a 3-layer neural network with three inputs, two hidden layers of four neurons each and a single output layer. There is no connection between the neurons of the same layer.}
  \label{fig:background_mlp}
\end{figure}

The Perceptron algorithm is not able to solve exactly the XOR problem which is not linearly separable \cite{minsky_perceptrons_1969}. The \ac{MLP} algorithm leverages this limitation by considering multiple Perceptrons stacked together to form a layer. Multiple layers of neurons are considered in the \ac{MLP} to learn specific patterns and non-linear dependencies between the input data as illustrated in Figure \ref{fig:background_mlp}.
The \ac{MLP} is a special case of the \ac{ANN} where each neuron of its hidden layer is connected to each neuron of the next layer with a unique weight. 
It is also named fully connected layer.
Note that there is no connection between the neurons of the same layer.
The \ac{MLP} algorithm consists in three types of layers:
\begin{bulletList}
    \item The input layer receiving the data as input of the model;
    \item The hidden layer which is a fully connected layer transforming and transmitting the input signal to the next layer. Note that several hidden layers can follow each others;
    \item The output layer returning the predictions of the model with one or multiple
neurons depending on the task to solve (\textit{i.e.} binary classification, multi-class classification, multi-value regression and so on).
\end{bulletList}

\subsection{Training a neural network}

Training an \ac{ANN} means adapting the weights associated to each neuron in order to minimize a loss function.
The optimization process iteratively updates the weights of the network by quantifying their gradient w.r.t. the loss function.
This function must be differentiable and has to be chosen carefully according to the problem to solve.

\paragraph{Loss functions}
Considering a regression task, the \ac{MSE} loss function is commonly used to minimize the gap between the true and the predicted values. The objective is formalized as
\begin{equation}
    \min_\theta \frac{1}{N} \sum_{i=0}^{N-1} (\vect{y}_i - f_\theta (\vect{x}_i))^2,
\end{equation}
 where $N$ is the number of elements in $\vect{X}$.

The binary classification task is tackled using a \ac{BCE} loss function. It quantifies the likelihood to predict the true labels. In practice and under the assumption of an activation function defined in $[0, 1]$, the optimization process minimizes the negative log-likelihood;
\begin{equation}
    \min_\theta - \frac{1}{N} \sum_{i=0}^{N-1} \vect{y}_i \log(f_\theta (\vect{x}_i)) + (1-\vect{y}_i) \log(1 - f_\theta (\vect{x}_i)).
\end{equation}

The multi-class classification consists in predicting for each $K$ class a probability that an input $\vect{x}_i$ belongs to each one of them. The ground-truth $\vect{y}_{i}$ is a one-hot vector of $K$ elements, with 1 at the corresponding index of the class, 0 otherwise. 
The general optimization problem for multi-class classification using a categorial \ac{CE} is written:
\begin{equation}
     \min_\theta - \frac{1}{K \cdot N} \sum_{k=0}^{K-1} \sum_{i=0}^{N-1} \vect{y}_{k,i} \log(f_\theta (\vect{x}_{i})_k),
\end{equation}
where $\sum_{k=0}^{K-1} \log(f_\theta (\vect{x}_{i})_k) = 1$.

\paragraph{Gradient descent optimization}
The weights of a neural network are initialized before starting the optimization process. 
The initialized set of weights $\theta \in \mathbb{R}^d$ denotes the starting point of the process in a $d$ dimensional space.
Initialization is an important choice as it leads to different local minima, and thus to performance variability.

Considering a neural network with a randomly initialized vector of weights,
their values are usually randomly drawn from a Gaussian or uniform distribution to avoid extreme values in long tail distributions. There is no better choice between these two distributions \cite{goodfellow_deep_2016}. However, the initialisation has a significant impact on both the optimization process and the generalisation capacities of the network.
Practical examples of uniform distributions used for the weight initialisation are $\mathcal{U}(-0.3, 0.3)$, $\mathcal{U}(0, 1)$ or $\mathcal{U}(-1, 1)$.

The Xavier initialization method \cite{glorot_understanding_2010} is commonly used and consists in randomly drawing the initial weight values in $\mathcal{U}(-\frac{1}{\sqrt{n}}, \frac{1}{\sqrt{n}})$, where $n$ is the number of inputs of the initialized layer.
An extension named normalized Xavier initialisation considers the uniform distribution defined as $\mathcal{U}(-\sqrt{\frac{6}{n+m}}, \sqrt{\frac{6}{n+m}})$, where $m$ is the number of neurons in the initialized layer. This initialization method is generally used to train neural network architectures because the magnitude of the values is well defined w.r.t. the input and current layers.

A common practice in deep learning, called pre-training or transfer learning, consists in initializing the model with a weight vector that has already been learnt on a pretext task with a large annotated dataset. The pre-trained parameters are then fine-tuned on a down-stream task, usually with fewer annotations available, benefiting from the knowledge acquired in the previous optimisation process.

The optimization process considers an arbitrary differentiable loss function $\mathcal{L}(., .)$. The error between the ground truth labels and the predictions of the model $f_\theta(.)$ is quantified as $\frac{1}{N} \sum_{i=0}^{N-1} \mathcal{L}(\vect{y}_i, f_\theta(\vect{x}_i))$ as described in the previous paragraphs. The optimization method commonly used during the training process is the \ac{SGD} \cite{robbins_stochastic_1951, kiefer_stochastic_1952}. It updates the weights iteratively in a direction opposite to the gradient of the loss function. Computing the gradients w.r.t. the loss is not scalable considering the entire training dataset. Let's consider a ``batch'' of $B$ samples, written $\vect{X}_B = \left \{ \vect{x}_0, \cdots, \vect{x}_{B-1} \right \}$, the \ac{SGD} method is defined as:
\begin{equation}
    \theta_{t} := \theta_{t-1} - \frac{\eta}{B} \nabla_\theta \sum^{B-1}_{i=0}  \mathcal{L}(\vect{y}_i, f_{\theta}(\vect{x}_i)),
\end{equation}
where $\eta > 0$ is the learning rate, $\theta_t$ the parameters of the model at iteration $t$ and $\vect{x}_i, \vect{y}_i$ the $i$-th sample of the batch with its associated label.

The learning rate is a key hyperparameter defining the amplitude of the weight update. It is chosen carefully because if it is small, the optimization will reach a local minima with numerous number of steps. And if it is high, the process may diverge.

The optimization landscape of a neural network training algorithm is complex and non-convex. Thus it converges to a local minima which is difficult to characterize. Exploring gradient descent algorithms and the optimization landscape of neural networks is an active field of research. 
In the presented experiments, the ADAM \cite{kingma_adam_2015} method is used for optimization. It is an efficient process considering momentum of the gradients to remember their past trajectories. The parameter update depends on $m_t$ the estimation of the bias-corrected first moment and $v_t$ the estimation of the bias-corrected second moment. Considering a batch of $B$ samples, the iterative gradient descent method proposed by \cite{kingma_adam_2015} is defined as:
\begin{align}
    m_t &:= \frac{\beta_1 m_{t-1} + \frac{(1- \beta_1)}{B} \nabla_\theta \sum^{B-1}_{i=0}  \mathcal{L}(\vect{y}_i, f_{\theta}(\vect{x}_i))}{(1 - \beta_1^t)}, \\
    v_t &:= \frac{\beta_2 v_{t-1} + \frac{(1-\beta_2)}{B} \nabla^2_\theta \sum^{B-1}_{i=0}  \mathcal{L}(\vect{y}_i, f_{\theta}(\vect{x}_i))}{(1-\beta_2^t)}, \\
    \theta_t &:= \theta_{t-1} - \eta \frac{m_t}{\sqrt{v_t} + \epsilon},
\end{align}
where $\beta_1, \beta_2 \in [ 0, 1 )$ the exponential decay rates of the momentum estimations, $\epsilon=10^{-8}$ and $\beta_1^t, \beta_2^t$ denoting respectively $\beta_1, \beta_2$ to the power $t$.

\paragraph{Forward and backward propagations}
During the training process, a feedforward pass through the network is performed to infer the predictions regarding an input. The loss function is evaluated by comparing the predictions and the ground-truth labels.
Then the back-propagation step aims to evaluate the gradients of the weights regarding the loss function thanks to the chain rule. 

Let $\theta = (\theta_0, \cdots, \theta_{d-1})$ be the vector of $d$ weights of a neural network. The network can be represented as an oriented graph where each neuron is a vertex. 
The neurons are connected to the previous and next layers via edges corresponding to the weights of the network.

According to the chain rule, the gradient of a weight will integrate all other weights on which it depends to calculate the loss, \textit{i.e.} the path on the graph that leads to the loss. 
Considering that $\theta_n$ is the last edge leading to the loss computed at the last neuron, 
the gradient of $\theta_i$ w.r.t. the loss depends on $(\theta_i, \cdots, \theta_n)$. It is computed as:
%
%
%
\begin{equation}
    \nabla_{\theta_i} \mathcal{L}(\vect{y}, f_{\theta}(\vect{x})) = \frac{\delta \mathcal{L}(\vect{y}, f_{\theta}(\vect{x}))}{\delta \theta_i} = \sum_j \frac{\delta \mathcal{L}(\vect{y}, f_{\theta}(\vect{x}))}{\delta f_{\theta_j}(\vect{x})} \frac{\delta f_{\theta_j}(\vect{x})}{\delta \theta_i},
\end{equation}
where $f_{\theta_j}(\vect{x})$ is the sub-function, \textit{i.e.} the sub-graph, of $f_{\theta}(\vect{x})$ parameterized by $ \left \{ \theta_j, \cdots, \theta_n \right \}$ weights.
We suggest that the reader refers to the work of  \cite{goodfellow_deep_2016} for in-depth details on graph computing for back-propagation.

\paragraph{Regularization}
A machine learning model is over-parameterized if its number of parameters is much larger than the number of training samples.
As a consequence, it has the capacity to learn the data itself, in other words to over-fit the training set, leading to a lack of generalization and poor performances on the test set. This problem is tackled using regularization methods to constraint the parameters of the model during the optimization process. 

The L1-norm is useful to reduce the dimension of the parameter space by forcing their value toward zero. The optimization is written
\begin{equation}
    \min_\theta \sum_{i=0}^{N-1} \mathcal{L}(\vect{y}_i, f_{\theta}(\vect{x}_i)) + \frac{\lambda}{2} \left\lVert \theta \right\rVert_1, 
\end{equation}
where $\lambda$ is an hyperparameter to scale the regularization strength.
The L2-norm used as a regularization method aims to penalize extreme values of the weights. It reduces over-fitting on outlier data. The optimization is written
\begin{equation}
    \min_\theta \sum_{i=0}^{N-1} \mathcal{L}(\vect{y}_i, f_{\theta}(\vect{x}_i)) + \frac{\lambda}{2} \left\lVert \theta \right\rVert_2^2.
\end{equation}

\begin{figure}[!t]
  \begin{center}
    \includegraphics[width=0.5\textwidth]{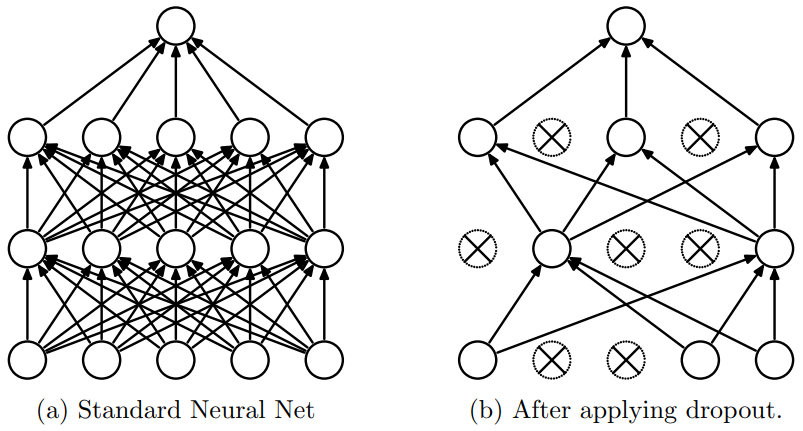}
  \end{center}
  \caption[Example of Dropout]{\textbf{Example of Dropout}. Example of the Dropout method \cite{srivastava_dropout_2014} with a probability of 0.5.}
  \label{fig:background_dropout}
\end{figure}

The dropout method \cite{hinton_improving_2012, srivastava_dropout_2014} regularizes neural networks by keeping a few active neurons with a propability $p$ at each iteration of the training. The targeted neurons are randomly chosen at each optimization step. Figure \ref{fig:background_dropout} illustrates this method with $p=0.5$.
During the forward propagation, the neurons of the $i$-th layer transmit the signal $f^i_{\theta_i}(\vect{x})$ to the $(i+1)$-th layer. Let $\vect{z}_{i+1}$ be a binary vector taking 1 with probability $p$ and 0 with probability $1-p$. The neurons at the layer $i+1$ are selected as:
\begin{equation}
    f^{i+1}_{\theta_{i+1}}(\vect{x}) = \vect{z}_{i+1} \odot (f^{i}_{\theta_{i}}(\vect{x}) \cdot \theta_i + b_i).
\end{equation}

At training time, each $\vect{z}_i$ is randomly drawn during the forward pass, they are fixed during the backward pass. A test time, the dropout method is not applied but a scaling factor of $\frac{1}{1-p}$ over the set of weights compensates the high number of neurons alive. The dropout method prevents over-fitting by reducing artificially the number of parameters to train, it is commonly used in fully connected layers.

\section{Convolutional neural network}
\label{sec:background_cnn}

\subsection{Convolutional layer}
Natural images recorded from cameras are presented as 3D matrices. Their pixels are in the range $[0, 255]$ for each \ac{RGB} channels. Using images as the input of a machine learning model, in particular considering a fully connected \ac{ANN}, is computationally expensive. 
As an example, considering an image of dimension $200{\times}200{\times}3$, a single fully connected layer will be represented by 120,000 parameters to process each pixel. The previous \ac{ANN} cannot scale to high dimensional input representations. Moreover, it is not able to learn spatial features directly from the input data. 

In recent years, \acp{CNN} using convolution operations has been explored to tackle these issues.
Let $f \in \mathbb{R}^{H \times W \times C}$ an image or feature map with $C$ channels and $g \in \mathbb{R}^{k_{1} \times k_{2} \times C}$ a set of $C$ 2D kernels. The 2D convolution operator is formally written
\begin{equation}
    (f*g)(h, w) = \sum_{i=-\infty}^{+\infty} \sum_{j=-\infty}^{+\infty} \sum_{c=-\infty}^{+\infty} f(i, j, c)g(h - i, w - j, c),
\end{equation}
where $h, w \in \mathbb{N}$ defines the coordinates in the output feature map.
The equivariance to translation property implies that the convolution of a shifted input contains the same information as the original one. As a consequence, the operation is able to detect patterns anywhere in the input.

A convolutional layer convolves a 2D kernel, over the entire input. 
At each iteration, it performs a dot product between the group of local values and the kernel, producing a scalar value. 
The scalar values are successively stored to create a feature map.
This type of layer aims to learn local and spatial features, while reducing the number of parameters, making convergence easier than in the case of fully-connected layers.
It is particularly well suited to high dimensional data, composed of redundant geometric and textural patterns, \textit{e.g.} images.
Moerover, the weights of a convolutional layer kernel are shared on the entire image meaning that the number of parameters regarding a fully connected \ac{ANN} is dramatically reduced.

\begin{figure}[!t]
  \begin{center}
    \includegraphics[width=0.6\textwidth]{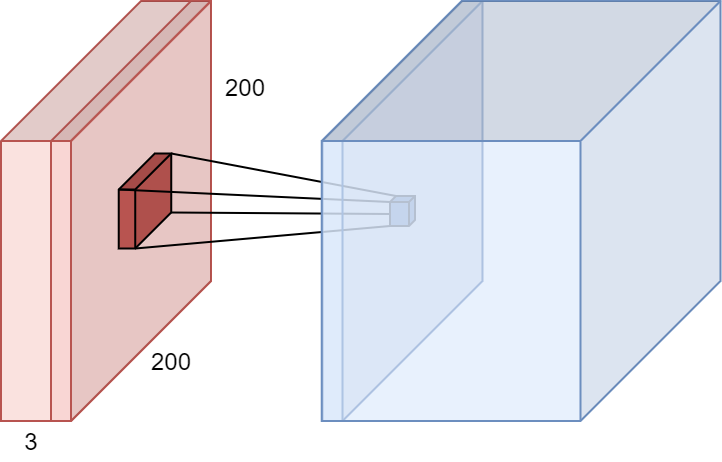}
  \end{center}
  \caption[Example of a convolutional layer]{\textbf{Example of a convolutional layer}. An $200 {\times} 200 {\times} 3$ input volume (light red) is convolved by a kernel (dark red) producing a scalar value (small blue cube) at each convolution step. The kernel creates a 2D feature maps once it has sliced the entire input. 
  The convolutional layer produces as many feature maps as kernels to learn; the output is a 3D tensor of feature maps.}
  \label{fig:background_conv_ex}
\end{figure}

As the convolution operation is differentiable, its kernels of weights can be learnt using the back-propagation method \cite{lecun_backpropagation_1989} with gradient descent regarding a loss function.

\subsection{Complementary methods and layers}
Practical methods have been introduced to better adapt to the dimensions of the image and to reduce the computational cost of the convolution operation. 
The ``padding'' method consists in adding default values at the border of the input to simple handle boundary conditions due to the convolution. 
Regarding the size of the kernel, additional values are required to create a feature map with the same size than the input. 
The default values are commonly set to zero (zero-padding) but many other methods are employed as duplicating the border values or replicating values as a mirror.
The ``stride'' method aims to slide the convolutional kernel on the input with a pixel gap at each iteration similarly as a down-sampling method.
A kernel usually convolves an image with a stride of 1 pixel. This parameter can be increased to reduce the computational cost of the operation, \textit{e.g.} a stride of 2 will slide the kernel with a gap of 1 pixel between each matrix product and thus, produces an output down-sampled by a factor of 2. However, small patterns may be missed. The stride is usually increased when a convolution is applied to high resolution images.
Let $D$ be the spatial dimension of input (either height or width), $S$ the stride of the convolution, $P$ the additional padding dimension and $k$ the size of the kernel. The dimension of the output after applying the convolution operation is $\frac{D - k - 2 S}{P} + 1$.

\begin{figure}[!t]
  \begin{center}
    \includegraphics[width=0.8\textwidth]{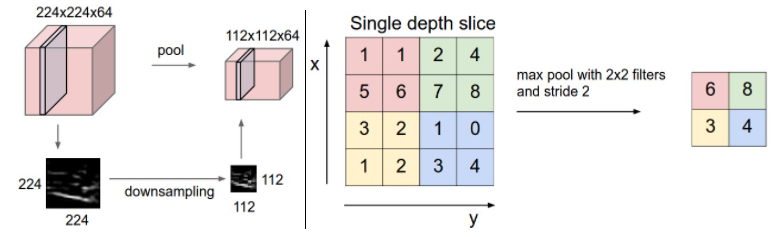}
  \end{center}
  \caption[Example of a max-pooling layer]{\textbf{Example of a max-pooling layer}. Left: example of the application of a pooling layer on a $200{\times}200{\times}32$ input with a kernel of size $2{\times}2$ and a stride of 2. Right: detail of the application on local regions of the input \cite{karpathy_cs231n_2021}.}
  \label{fig:background_maxpool}
\end{figure}

The architecture of a neural network dictates the composition of its successive layers. A \ac{CNN} extracts information from the input data with successive convolutional layers \cite{krizhevsky_imagenet_2012}. Each one of these layers takes as input the output feature maps of the previous layer.
To reduce the computational cost of such models and learn features at different scales, the size of the representations is progressively reduced.
A pooling layer is nested between two convolutions applying a $\max(.)$ operation on the intermediate feature map as illustrated in Figure \ref{fig:background_maxpool}. As the convolution, it slides the input with a kernel (usually $2{\times}2$) and stores the maximum value in the outputted feature map. The padding and stride methods are also applicable in the max-pooling layer.

Recent advances in the understanding of neural networks optimization schemes lead to a better control on the gradient descent algorithms used during training. The distribution of the inputs of each layer changes during training due to the initialization of the weights, the intrinsic randomness of the data and the changes in the distribution of the weights in the previous layer. 
This phenomena, named internal covariate shift \cite{ioffe_batch_2015}, slows down the training (requires low learning rates) and may lead to divergence due to saturated non-linearities. 
The batch normalization layer introduced by  \cite{ioffe_batch_2015} refining the distribution of the inputs of each layer during training to reach a Gaussian distribution $\mathcal{N}(0, 1)$.
The batch normalization layer has a significant impact on convolutional networks by smoothing the optimization landscape to propagate relevant information with the gradients \cite{goodfellow_deep_2016}.
Let $\vect{x}$ be the input of a layer over a mini-batch $\mathcal{B}= \left \{ \vect{x}_0, \cdots, \vect{x}_{m-1} \right \}$. The layer first normalizes $\vect{x}$ w.r.t. to the statistics of the mini-batch as
\begin{align}
    \mu_\mathcal{B} &= \frac{1}{m} \sum^{m-1}_{i=0} \vect{x}_i, \\
    \sigma_\mathcal{B}^2 &= \frac{1}{m} \sum^{m-1}_{i=0} (\vect{x}_i - \mu_\mathcal{B})^2, \\
    \hat{\vect{x}}_i &= \frac{\vect{x}_i - \mu_\mathcal{B}}{\sqrt{\sigma_\mathcal{B}^2} + \epsilon},
\end{align}
where $\hat{\vect{x}}_i$ is the normalized sample of the mini-batch input and $\epsilon>0$ a negligible value.
The normalized sample is then scaled and shifted as
\begin{equation}
    \vect{o}_i = \gamma \hat{\vect{x}}_i + \beta,
\end{equation}
where $\vect{o}_i$ is the output sample of the batch normalization layer and $\gamma$, $\beta$ two learnt parameters.
This layer is commonly applied after a convolutional layer to normalize the distribution of the transformed data. It is usually followed by an activation layer avoiding the saturation of the non-linearity.
The following section will briefly introduce recurrent neural network structures.

\section{Recurrent neural network}
\label{sec:background_rnn}

\begin{figure}[!t]
  \begin{center}
    \includegraphics[width=1\textwidth]{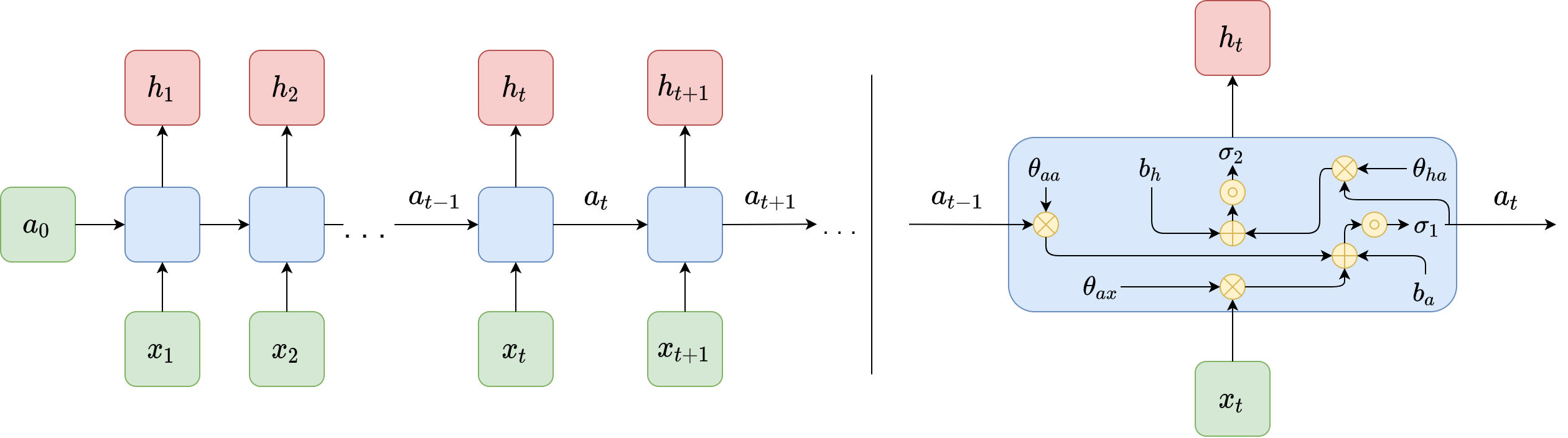}
  \end{center}
  \caption[Example of a recurrent neural network]{\textbf{Example of a recurrent neural network}. A traditional recurrent neural network processes a sequence iteratively and produces an hidden state ($h_t$) and an activation vector ($a_t$) (left). The simplest cell consists in applying three fully connected layers at different level of the data processing (right). }
  \label{fig:traditional_rnn}
\end{figure}

A \ac{RNN} is specialized in processing sequence of data. It is particularly used in temporal series analysis (\textit{e.g.} for music, video or stock market) and Natural Language Processing (NLP) (\textit{e.g.} textual analysis or translation).
Its structure is based on a cell taking as input an element of the sequence and an activation vector. It outputs an hidden state and an activation vector. 
At each iteration, the cell of the network processes the next element of a sequence with the activation vector produced at the previous step as illustrated in Figure \ref{fig:traditional_rnn} (left). Considering an application where a sequence is used to predict another sequence, the hidden states produced at each step correspond to the predictions of the network.

This type of network can be formalized differently depending on the task to solve. The \ac{RNN} presented and illustrated in \ref{fig:traditional_rnn} (left) is a specific case of a many-to-many application because it takes a sequence as input and outputs a sequence, both having the same length. It is generally used for named entity recognition.
A many-to-many application with different length of sequences is used for text translation, the input sequence of words is processed completely and the predicted sequence, in another language, is generated afterward.
The many-to-one processes an entire sequence to generate a single output, \textit{e.g.} for sentiment classification in texts. Finally, the one-to-many consists in generating a sequence by considering a single observation as input.

In the many-to-many application presented in Figure \ref{fig:traditional_rnn}, considering same length sequences, the cell of the \ac{RNN} defines the operations applied to the element of the sequence and the activation vector. Let $\theta_{aa}, \theta_{ax}, \theta_{ha}, b_a, b_h$ be trainable coefficients shared temporally and $\sigma_1, \sigma_2$ activation functions. The traditional \ac{RNN} cell, illustrated in Figure \ref{fig:traditional_rnn} (right), is defined as:

\begin{align}
    \vect{a}_t &= \sigma_1(\theta_{aa} \vect{a}_{t-1} + \theta_{ax}\vect{x}_t + b_a), \\
    \vect{h}_t &= \sigma_2(\theta_{ha} \vect{a}_{t-1} + b_h),
\end{align}
where $\vect{x}_t$ is the element of the input sequence at time $t$, $\vect{a}_t$ and $\vect{h}^t$ are respectively the activation and hidden state vectors produced at time $t$.

The \ac{RNN} models have multiple benefits: they process sequences of any length, the number of parameters does not increase with the input size, they consider the historical information and the parameters are shared at each timestamp. However, they are really slow to train and suffer from vanishing gradients which are computed at the end of the sequence. Long-term dependencies are also difficult to learn because they are lost in the processing of sequences.

The \acf{LSTM} cell \cite{hochreiter_long_1997} aims to alleviate these limitations with a particular structure learning long and short-term dependencies while easily back-propagating the gradients during time. The details of \ac{RNN} cell structures are beyond the scope of this thesis. We suggest that the reader refers to the work of  \cite{goodfellow_deep_2016} for in-depth information.
Many best practices and modules have been developed in the conception of neural network architecture for the past few years. 
The next section aims to provide intuitions on the widely used methods in deep learning for scene understanding, in particular for classification, object detection and semantic segmentation.

\section{Deep learning}
\label{sec:background_deep}

Deep learning models are composed of multiple layers stacked one after the other forming an end-to-end differentiable model trained via back-propagation. Layers and neural network architectures have been widely explored in the past few years improving the feature representations of the data. Recent large scale annotated datasets helped to train models with increasing the number of parameters while improving the performances on public benchmarks and challenges.

\begin{figure}[!t]
  \begin{center}
    \includegraphics[width=1\textwidth]{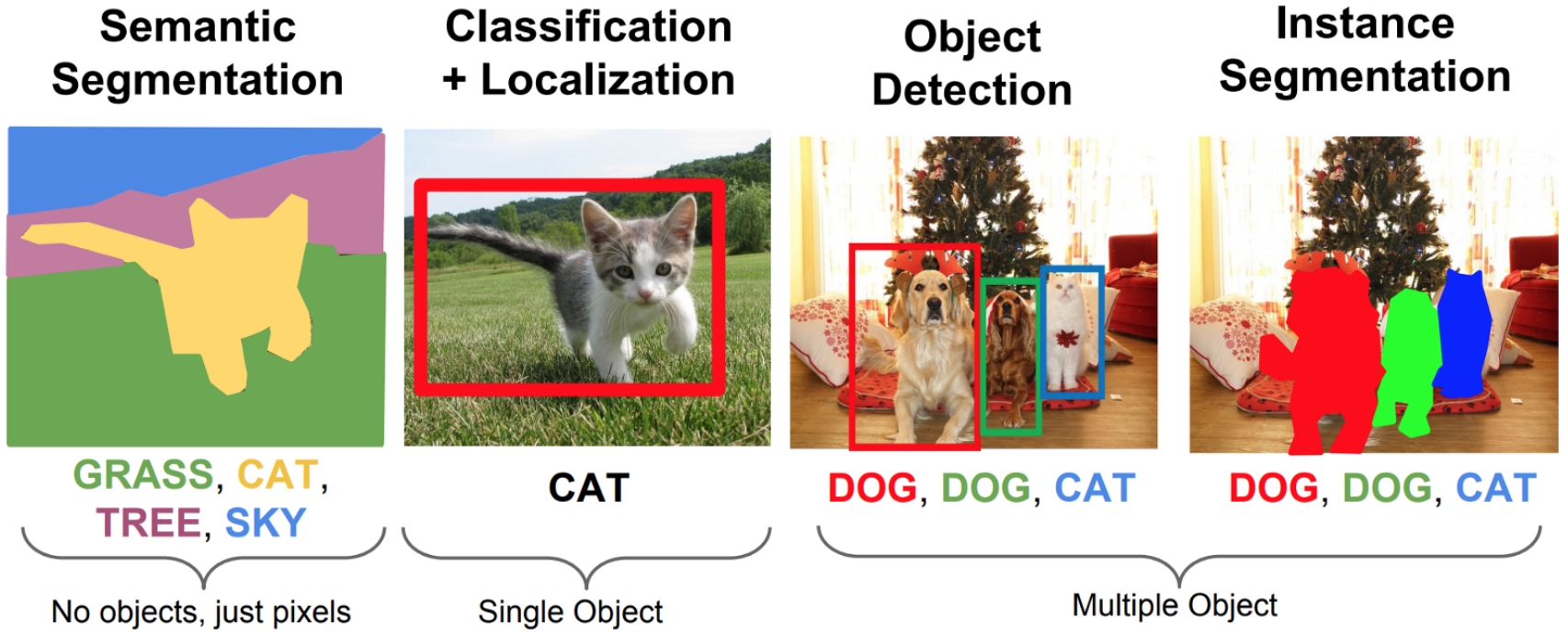}
  \end{center}
  \caption[Comparison between different computer vision applications]{\textbf{Comparison between different computer vision applications}. Examples of well-known computer vision tasks for scene understanding with one of multiple objects to recognize.}
  \label{fig:comparison_tasks}
\end{figure}

This section will focus on deep learning applications for computer vision, in particular on well-known tasks in scene understanding namely classification, object detection and semantic segmentation. Examples of these applications are illustrated in Figure \ref{fig:comparison_tasks}. This section aims to provide a background on existing layers and deep learning architectures which will be useful for the following chapters. The reported performances in Sections \ref{sec:background_classif}, \ref{sec:background_detection} and \ref{sec:background_segmentation} are evaluated on different tasks and datasets, thus they can not be directly compared \textit{per se}.

\subsection{Classification}
\label{sec:background_classif}
One of the most popular task in computer vision is classification, \textit{i.e.} associating a category to each image of a dataset.
Classification has been widely explored using the ImageNet dataset \cite{deng_imagenet_2009}. 
A collaboration between Stanford University and Princeton University led to this large scale dataset of fourteen millions images annotated in one thousand categories. They created the annotations using set of synonym rings (or synsets)\footnote{A synonym ring or synset, is a group of data elements that are considered semantically equivalent.}) of the WordNet\footnote{\url{https://wordnet.princeton.edu/}} lexicon tree.
The original challenge consisted in a simple classification task, each image belonging to a single category among one thousand, from specific breed of dog to precise type of food. 
Due to its large scale, the \acf{ILSVRC} is one of the most popular in computer vision. For clarity, we will call it the ImageNet challenge.
This section will detail advances in deep learning architectures and methods which have continuously improved the performances on the ImageNet challenges from 2012 to 2018.
In the following paragraphs, we note the top-$k$ error rate as the inverse of the top-$k$ accuracy, also written $1 - \frac{\text{TP} + \text{TN}}{\text{TP} + \text{FP} + \text{TN} + \text{FN}}$, where a prediction is considered as positive if it is ranked in the top-$k$ highest probabilities considering the predictions for all classes; and TP the true positive, TN the true negative, FP the false positive and TN the false negative.

\paragraph{The advent of deep learning (AlexNet).}
The ImageNet challenge has been traditionally tackled with image analysis algorithms such as SIFT \cite{lowe_distinctive_2004} with mitigated results until the late 90's. However, a leap in performances has been brought by using neural networks.
Inspired by \cite{lecun_efficient_2012}, the first deep learning model proposed by \cite{krizhevsky_imagenet_2012} drew attention to the public by beating all the previous computer vision methods with a top-5 error rate of 15.3\%. 
The proposed AlexNet model can be considered today as a simple architecture with two consecutive convolutional, max-pool layers and three fully-connected layers.

\paragraph{Going deeper (VGG).}
In 2014, \cite{simonyan_very_2015} proposed the VGG16 architecture, composed of sixteen convolutional layers, four nested max-pool layers and three final fully connected layers. One of its specificities is to chain multiple convolutional layers with ReLU activation functions creating non-linear transformations. The authors also introduced kernels of weights of size $3 {\times} 3$ for each convolution (as opposed to $11 {\times} 11$ filters in the AlexNet model).
They noticed that similar patterns can be learnt with smaller kernels while decreasing the number of parameters to learn.
Using smaller kernels allows more convolutional layers to be stacked. As a consequence, deep layers have a larger receptive field. They thus have the capacity to learn fine-grained patterns at different scales.
With these methods, the authors reduced by a factor of two the error rate of the AlexNet model reaching a top-5 error rate of 7.3\% on the 2012 ImageNet challenge.

\paragraph{Inception modules (GoogLeNet and Inception V2).}
The ``inception module'' has been inspired by the work of \cite{lin_network_2014}, consisting in training successive convolutional layers while introducing \ac{MLP} between two layers. 
This idea have been exploited by \cite{szegedy_going_2015} who proposed GoogLeNet (a.k.a. Inception V1), a deep neural network with 22 layers of inception modules for a total of over 50 convolutional layers.
Their proposed module is composed of parallel convolutions with $1 {\times} 1$, $3 {\times} 3$, $5 {\times} 5$ kernels of weights and a $3 {\times} 3$ max-pool layer to increase the sparsity in the model. The produced feature maps are then concatenated and analyzed by the next inception module.
The error rate on the 2012 ImageNet challenge has decreased to 6.7\% while requiring significantly less memory than the VGG16 architecture (55MB \textit{v.s.} 490MB). This gap is due to the three more fully-connected layers in the VGG.

In 2015, \cite{szegedy_rethinking_2016} developed the Inception V2 model, mostly
inspired by the first version. 
The authors have changed the $5 {\times} 5$ kernel in the inception modules by two $3 {\times} 3$ kernels. It reduces the computational cost of the model which reached a top-5 error rate of 5.6\% on the ImageNet challenge.
The authors also proposed to factorize the convolution kernels of an inception module. It consists in replacing a convolution with a $3 {\times} 3$ kernel with two convolutions of kernels $3 {\times} 1$ and $1 {\times} 3$ respectively. This method reduces the number of parameters and the computational cost of the model. The Inception V3 architecture is composed of inception modules with factorized kernels. The authors also changed the first layers to process higher resolution inputs. They finally reached a top-5 error rate of 3.58\% on the 2012 ImageNet challenge.

\paragraph{Residual learning (ResNet).}
In their work, \cite{he_deep_2016} noticed that extremely deep models are difficult to train, leading to decreasing performances. They introduced the ``residual learning'' method creating a connection between the output of one or multiple convolutional layers and their original input with an identity mapping. In other words, the model tries to learn a residual function keeping most of the information.
These connections also help to better propagate the gradients through deep networks by maintaining their magnitude.
Residual learning neither requires any additional parameters, nor increases the computational complexity of the model. 
The authors proposed several architectures, named ResNet-X, where X is the number of convolutional layers with $3 {\times} 3$ kernels using residual learning by blocks of two layers. The ResNet-152 performed a top-5 error rate of 4.49 \% on the 2012 ImageNet challenge (less that the inception V3) and it won the 2015 challenge with a top-5 error rate of 3.57\%.

\paragraph{Residual learning and inception modules (Inception-ResNet).}
\cite{szegedy_inception-v4_2017} have combined inception modules increasing the sparsity and residual blocks to learn deeper networks. The residual inception blocks are stacked with a similar architecture to the Inception V3 model. 
It results in the Inception V4\footnote{\cite{szegedy_inception-v4_2017} developed a pure (\textit{i.e.} without residual block) Inception V4 and an Inception-ResNet V2 model which uses inception modules and residual blocks. The aforementioned Inception V4 is the Inception-ResNet V2 providing the best performances.}, or Inception-ResNet, which can be trained faster and outperforms other methods with a top-5 error rate of 3.08\% on the 2012 ImageNet challenge.

\paragraph{Squeeze and excitation module.}
The ``Squeeze-and-Excitation'' module has been introduced by \cite{hu_squeeze-and-excitation_2018}. It performs a global pooling on the input feature maps, followed by a fully connected layer with ReLU activation, and a second fully connected layer with Sigmoid activation. The output is scaled by to the input resolution and a residual connection is performed. Its main advantage is the low number of parameters due to the down-sampling applied to the input maps. It won the 2017 ImageNet challenge with a top-5 error rate of 2.25\%.

\paragraph{Conclusion}
This section described the milestones reached in deep learning for the image classification task, in particular applied to the ImageNet challenge. Well-known modules and architectures have be detailed and they are still commonly used for feature extraction nowadays. However, it is not an exhaustive list of all the existing models between 2012 and 2018. The advances from 2018 to date, in particular on Transformers architectures \cite{dosovitskiy_image_2021, liu_swin_2021, touvron_training_2021}, are beyond the scope of this thesis.

\subsection{Object detection}
\label{sec:background_detection}
Classification methods detailed in Section \ref{sec:background_classif} categorize images into a single class, usually corresponding to the most salient object\footnote{Misleading examples in ImageNet have a single label although several objects are visible.}. Images, \textit{e.g.} recorded in urban scenes, are usually complex and contain multiple objects. In this case, such models are uncertain about which label to assign. 
The object detection task consists in localising and classifying all the objects in an input by predicting a bounding box around each one of them. It is therefore more appropriate for complex scene understanding.
In this section, well-known datasets and evaluation metrics are briefly introduced. Then, each paragraph will detail a commonly used deep neural network architecture for object detection. This list is not exhaustive and additional methods are detailed in Appendix \ref{sec_app:background_detection}.

\paragraph{Datasets.}
The PASCAL Visual Object Classification (PASCAL VOC)\footnote{\url{http://host.robots.ox.ac.uk/pascal/VOC/}} dataset \cite{everingham_pascal_2015} is a well-known dataset for classification, object detection and segmentation of objects. There are eight challenges spanning from 2005 to 2012, each of them having its own specificities. Considering the object detection task, there are around 10,000 images for training and validation with bounding boxes covering 20 categories.

Since 2013, ImageNet \cite{russakovsky_imagenet_2015} has released an object detection challenge with bounding boxes. The training dataset is composed of around 500,000 images only for training and 200 categories. It is rarely used because the size of the dataset requires a large computational power for training. Also, the high number of classes is difficult to tackle considering the object recognition task. A comparison between the 2014 ImageNet dataset and the 2012 PASCAL VOC dataset is available online \footnote{\url{http://image-net.org/challenges/LSVRC/2014/}}.

The Common Objects in COntext (COCO)\footnote{\url{http://cocodataset.org}} dataset \cite{lin_microsoft_2014}, developed by Microsoft, proposes four challenges: caption generation, object detection, key point detection and object segmentation.
This section focuses on the object detection task of the COCO dataset consisting in localizing the objects in an image with bounding boxes and categorizing each one of them between 80 categories.
The dataset changes each year but it is usually composed of more than 120,000 images for training and validation, and more than 40,000 images for testing.

\paragraph{Evaluation metrics.}
\label{sec:background_detection_metrics}
The object detection challenge contains a regression and a classification task.
First of all, the bounding boxes with low confidence (the model usually outputs
many more boxes than actual objects) are removed to assess the spatial precision.
Then, the \ac{IoU} is defined as the percentage $\frac{|A \cap B|}{|A \cup B|}$, where $A$ is the area of the predicted bounding box and $B$ is the ground-truth box. 
The higher the IoU, the better the predicted location of the box for a given object. 
A threshold is usually applied on the \ac{IoU} values to select the bounding box candidates.

The \ac{mAP} metric consists in computing the \ac{AP} over all the categories of the dataset. Let $\text{AP} = \frac{\text{TP}}{\text{TP}+\text{FP}}$ be the \acf{AP} with TP the true positive and FP the false positive predictions, we note $\text{mAP} = \frac{1}{K}\sum_{k=0}^{K-1} \text{AP}_k$ where $K$ is the number of classes in the dataset and $\text{AP}_k$ is the \ac{AP} for the $k$-th class. 
The \ac{mAP} metric avoids extreme specialization in a few classes and thus weak performances in the others. 
This metric considers only the predicted bounding boxes with a sufficient overlap with the ground truth, \textit{i.e.} with a threshold on the \ac{IoU}. The \ac{IoU} threshold is usually fixed but a high number of bounding boxes increases the number of candidate boxes. The COCO challenge has developed an official metric avoiding an over-generation of boxes. It computes a mean of the \ac{mAP} scores for a set of \ac{IoU} threshold values in order to penalize a high number of bounding boxes with wrong classifications. This set of thresholds is defined between 0.5 and 1 with a 0.05 step.

\paragraph{Region-based Convolutional Network (R-CNN).}
This method starts with a selective search \cite{uijlings_selective_2013} initializing small regions in an image and merging them with a hierarchical grouping. The detected regions
are merged according to a variety of color spaces and similarity metrics. It outputs a small number of region proposals which could contain an object.
The R-CNN model \cite{girshick_region-based_2016} combines the selective search method detecting region proposals and deep learning to classify the objects. Each region proposal is resized to match the input of a \ac{CNN} outputting a 4096-dimension vector of features. This vector is then used as input of binary SVM \cite{hearst_support_1998} classifiers, one for each class. It is also used as input of a linear regressor adapting the shapes of the corresponding bounding box to reduce the location error.
The \ac{CNN} is trained on the 2012 ImageNet dataset for classification. It is then fine-tuned using the region proposals corresponding to an \ac{IoU} greater than 0.5 with the ground-truth boxes. Two versions are produced, one version is using the 2012 PASCAL VOC dataset and the other the 2013 ImageNet dataset with bounding boxes. The SVM classifiers are also trained for each class of each dataset.
The best R-CNN models have achieved a 62.4\% \ac{mAP} score on the 2012 PASCAL VOC challenge (22.0 points increase w.r.t. the second best result on the leader board) and a 31.4\% \ac{mAP} score over the 2013 ImageNet dataset (7.1 points increase w.r.t. the second best result on the leader board).

\paragraph{Fast Region-based Convolutional Network (Fast R-CNN).}
The objective of Fast R-CNN \cite{girshick_fast_2015} is to reduce the computational cost due to the diversity of models required to analyse all region proposals.
A \ac{CNN} takes the entire image as input instead of a specialised one for each region proposal (R-CNN). \acp{RoI} are detected with the selective search method applied on the produced feature maps. 
Formally, the feature maps size is reduced using a \ac{RoI} pooling layer to get valid \aclp{RoI} with fixed height and width as hyperparameters. Each \ac{RoI} layer feeds fully connected layers\footnote{The entire architecture is inspired from the VGG16 model, thus it has 13 convolutional layers and 3 fully connected layers.} creating a vector of features. The vector is used to predict the observed object with a softmax classifier and to adapt the bounding box location with a linear regressor.
The best Fast R-CNN models have reached \ac{mAP} scores of 70.0\% for the 2007 PASCAL VOC challenge, 68.8\% for the 2010 PASCAL VOC challenge and 68.4\% for the 2012 PASCAL VOC challenge.

\paragraph{Faster Region-based Convolutional Network (Faster R-CNN).}
The aim of the Faster R-CNN \cite{ren_faster_2015} is to replace the selective search method with a \ac{RPN} to generate region proposals, predict bounding boxes and detect objects only with \acp{CNN}. The Faster R-CNN combines an \ac{RPN} and a Fast R-CNN model.

The \ac{RPN} takes as input the entire image and produces feature maps.
A window of size $3{\times}3$ slides over all the feature maps and outputs a 256-dimension vector of features linked to two fully connected layers, one for box regression and one for box classification. Multiple region proposals are predicted by the fully connected layers. A maximum of $k$ regions is fixed, thus the output of the box regression layer has a size of $4k$ (for each box, coordinates of a corner of the boxes and its height and width) and the
output of the box classification layer a size of $2k$ (``objectness'' scores to detect an object or not in the box). The $k$ region proposals detected by the sliding window are called anchors. When the anchor boxes are detected, they are selected with a threshold applied on the ``objectness'' score keeping only the relevant boxes. These anchor boxes and the feature maps computed by the initial \ac{CNN} model are used as input of a Fast R-CNN model.

Faster R-CNN uses \ac{RPN} to avoid the selective search method, it accelerates the training and testing processes while improving the performances. The \ac{RPN} is a pre-trained model using the ImageNet dataset for classification, it is then fine-tuned on the PASCAL VOC dataset. 
The best Faster R-CNN models have obtained \ac{mAP} scores of 78.8\% on the 2007 PASCAL VOC challenge and 75.9\% on the 2012 PASCAL VOC challenge. The models have been trained with PASCAL VOC and COCO datasets. One of these models\footnote{The fastest Faster R-CNN has an architecture inspired by the ZFNet model introduced by \cite{zeiler_visualizing_2014}. The commonly used Faster R-CNN has an architecture similar to the VGG16 model and it is only 10 times faster than the Fast R-CNN.} is 34 times faster than the Fast R-CNN using the selective search method.

\paragraph{You Only Look Once (YOLO).}
The YOLO model \cite{redmon_you_2016} is a single stage approach predicting bounding box coordinates and class probabilities with a single network. Its simplicity permits real time predictions during inference.
The model takes an image as input which is divided in an $S {\times} S$ grid. In each cell of this grid is predicted $B$ bounding boxes with a confidence
score. This confidence is defined as the probability to detect the object multiply
by the \ac{IoU} between the predicted and the ground-truth boxes.
The architecture is inspired by the GoogLeNet model \cite{szegedy_going_2015}
using inception modules. It has 24 convolutional layers followed by 2 fully-connected layers. The inception modules are replaced by $3 {\times} 3$ followed by $1{\times}1$ convolutions.
The final layer outputs a tensor of dimensions $S{\times} S {\times} (K + B {\times} 5)$ corresponding to the predictions in each cell of the grid, where $K$ is the number of classes, $B$ the fixed number of anchor boxes per cell, each anchor being characterized with 4 coordinates (coordinates of the center of the box, width and height) and a confidence value.
The YOLO model predicts a high number of bounding boxes since it does not localize \acf{RoI}. 
Thus, there are a lot of bounding boxes without any object. \ac{NMS} it is applied at the end of the network. It consists in merging highly-overlapping bounding boxes of a same object into a single one. 
The YOLO model reached a 63.7\% \ac{mAP} score on the 2007 PASCAL VOC challenge and a 57.9\% \ac{mAP} score over the 2012 PASCAL VOC challenge. 
A few years later, the YOLO architecture has been extended in YOLOv2 and the authors proposed a new model, YOLO9000, capable of detecting more than 9000 categories while running in almost real time (around 10 \ac{FPS}). Details are provided in Appendix \ref{sec_app:background_detection}.

\paragraph{Single-Shot Detector (SSD).}
The SSD \cite{liu_ssd_2016} is a single stage model, similarly to YOLO, predicting the bounding box coordinates and the class probabilities simultaneously.
This end-to-end architecture is fully convolutional using different kernel sizes ($10 {\times} 10$, $5 {\times} 5$ or $3 {\times} 3$). Feature maps from convolutional layers at different levels of the network are used to predict the bounding boxes. The feature maps are processed by a specific convolutional layers with a $3 {\times} 3$ kernel called extra feature layers producing a set of bounding boxes similar to the anchor boxes of the Fast R-CNN.
Each box has 4 parameters: the coordinates of its center, its width and its height. At the same time, a branch produces a vector of probabilities corresponding to the softmax considering all the classes of object.
The \ac{NMS} method selecting the relevant bounding boxes is also used at the end of the SSD model.
\ac{HNM} selects relevant boxes among the negative samples during training:
the boxes with the highest confidence are selected depending on the ratio between the negative and the positive samples (evaluated to $\frac{1}{3}$ in this work).
In their work, \cite{liu_ssd_2016} distinguished the SSD300 and the SSD512, the latter corresponds to a SSD300 with an extra convolution on the prediction heads.
The best SSD models are trained with the 2007, 2012 PASCAL VOC datasets and the 2015 COCO dataset with data augmentation. They obtained \ac{mAP} scores of 83.2\% on the 2007 PASCAL VOC challenge and 82.2\% on the 2012 PASCAL VOC challenge. 
On the 2015 COCO challenge, they reached a score of 48.5\% for an \ac{IoU} threshold of 0.5, 30.3\% for an \ac{IoU} threshold of 0.75 and 31.5\% for the official \ac{mAP} metric.

\paragraph{Mask Region-based Convolutional Network (Mask R-CNN).}
An extension of the Faster R-CNN has been proposed by \cite{he_mask_2017} performing simultaneously object detection and semantic segmentation as a multi-task problem. 
It uses the Faster R-CNN pipeline with three output branches for each candidate object: a
class label, a bounding box offset and the object mask. It uses \ac{RPN} to generate bounding box proposals and produces the three outputs at the same time for each \ac{RoI}.
The initial RoIPool layer, used in the Faster R-CNN to select object proposals in the feature maps, is replaced by a RoIAlign layer. It removes the quantization of the coordinates of the original \ac{RoI} and computes the exact coordinates of the locations. The RoIAlign layer provides scale-equivariance and translation-equivariance with the region proposals.
The backbone extracting the features is a ResNeXt architecture \cite{xie_aggregated_2017} with 101 layers. Each residual block is slightly modified from the work of \cite{he_deep_2016} by considering multiple parallel convolutions producing feature maps which are stacked and followed by a residual connection.
The model detects \acp{RoI} which are processed with a RoIAlign layer. One branch of the network is linked to a fully connected layer adjusting the coordinates of the bounding boxes and predicting the class probabilities. The other branch is linked to two convolutional layers, the last one computes the mask of the detected object. Additional details about semantic segmentation are provided in the next section.
The multi-task training is performed by summing the loss function corresponding to each task into a global one. The gradients are propagated in the entire network. 
The Mask R-CNN outperformed the state of the art in the four COCO challenges: the instance segmentation, the bounding box detection, the object detection and the key point detection.
It reached \ac{mAP} scores of 62.3\% with an \ac{IoU} threshold of 0.5,
43.4\% for an \ac{IoU} threshold of 0.75 and 39.8\% for the official metric over the 2016 COCO
challenge.

\paragraph{Conclusion.}
Through the years, object detection models tend to infer localisation and classification all at once to have an entirely differentiable network. They can be trained from head to tail with back-propagation. However, a trade-off between high performance and real time prediction capability is made in the last presented models.
This review of object detection methods using deep learning, supplemented by Appendix \ref{sec_app:background_detection}, is not exhaustive and ranges from 2015 to 2017. Recent advances have largely extended these approaches but the previously presented ones are still commonly used.

\subsection{Semantic segmentation}
\label{sec:background_segmentation}

The semantic segmentation task applied to natural images consists in classifying each pixel in a category. An extension of this task, called instance segmentation, consisting in classify each pixel in a category with an identification number to distinguish different instances of the same category.

Most of the object detection pipelines presented in Section \ref{sec:background_detection} require anchor boxes or proposals to localize an object in a scene.
Unfortunately, just a few models take into account the entire context of an image. They are still limited on localizing objects with small part of the information. They cannot provide a full comprehension of a scene.

Scene understanding requires high visual perception of each entity while considering the spatial information. 
In the past few years, other challenges have emerged to better understand the actions in a image or a video: key point detection, action recognition, video captioning, visual question answering and so on. 
This section will focus on the semantic segmentation task. It will introduce datasets and evaluation metrics; and detail several well-known architectures and methods.
This list is not exhaustive and additional methods are detailed in Appendix \ref{sec_app:background_detection}.

\paragraph{Datasets.}
The PASCAL VOC dataset\footnote{\url{http://host.robots.ox.ac.uk/pascal/VOC/}} (2012) \cite{everingham_pascal_2015}, mentioned in the previous section, is well-known and commonly used for object detection and segmentation covering 20 categories. More than 11,000 images compose the train and validation datasets while 10,000 images are dedicated to the test dataset.

The PASCAL-Context dataset\footnote{\url{https://cs.stanford.edu/~roozbeh/pascal-context/}} (2014) \cite{mottaghi_role_2014} is an extension of the 2010 PASCAL VOC dataset. It contains around 10,000 images for training, 10,000 for validation and 10,000 for testing. The specificity of this release is that the entire scenes are segmented between more than 400 categories. Note that the images have been annotated during three months by six in-house annotators.

There are two COCO challenges\footnote{\url{http://cocodataset.org}} (in 2017 and 2018) for image semantic segmentation: ``object detection'' and ``stuff segmentation''. The object detection task consists in segmenting and categorizing objects into 80 categories. The stuff segmentation task
consists in segmenting almost the entire visual information of in an image, \textit{e.g.} including sky, wall, grass. 
This section will denote the COCO semantic segmentation challenge as the corresponding object detection task since it is the widely explored of the two.
The COCO dataset \cite{lin_microsoft_2014} for semantic segmentation is composed of more than 200,000 images with over 500,000 object instances segmented which are splitted in train, validation and test\footnote{As in many challenges, the test dataset is divided in test-dev (for research) and test-challenge (for the challenge). The annotations for both datasets are not available}. 
These datasets contain 80 categories and only the corresponding objects are segmented. 

The Cityscapes dataset\footnote{\url{https://www.cityscapes-dataset.com/}} \cite{cordts_cityscapes_2016} is composed of complex segmented urban scenes from 50 cities. There are around 23,500 images for training and validation (fine and coarse annotations) and 1,500 images for testing (only fine annotation).
The images are fully segmented as the PASCAL-Context dataset with 29 classes, within 8 super categories: flat, human, vehicle, construction, object, nature, sky, void. 
This dataset is well-known for its semantic segmentation task because of its complexity and its similarity with real urban scenes for autonomous driving applications.

\paragraph{Evaluation metrics.}

The semantic segmentation task is commonly evaluated using the \ac{mIoU} metric. As presented in Section \ref{sec:background_detection}, the \ac{IoU} quantifies the ratio between the overlapping area and the union area between a predicted shape and its ground truth. The \ac{mIoU} is the average of \ac{IoU} of all the predicted shapes considering their ground truth over all the classes.

The official evaluation metric of the Pascal-VOC, PASCAL-Context and Cityscapes challenges is the \ac{mIoU}. 
The COCO challenge also considers the \ac{mAP} and \ac{mAR} metrics for evaluation. Differently from the object detection task, these two metrics are computed pixel-wise without filtering the shapes with an \ac{IoU} threshold. The \ac{mAR} metric is computed similarly than the \ac{mAP} detailed in Section \ref{sec:background_detection}.
Let $\text{AR} = \frac{\text{TP}}{\text{TP}+\text{FN}}$ be the \ac{AR} with TP the true positive and FN the false negative predictions, we note $\text{mAR} = \frac{1}{K}\sum_{k=0}^{K-1} \text{AR}_k$ where $K$ is the number of classes in the dataset and $\text{AR}_k$ is the \ac{AR} for the $k$-th class. 
The works detailed in this section refer to their performances in term of \ac{AP} and \ac{AR} but these metrics are equivalent to the presented \ac{mAP} and \ac{mAR} respectively. We chose to keep the same notations than in Section \ref{sec:background_detection} for easier understanding.

\paragraph{Fully Convolutional Network (FCN).}
The \ac{FCN} proposed by \cite{long_fully_2015} is the first model composed only of convolutional layers trained end-to-end with back-propagation for image segmentation.
\acp{FCN} are used to learn features at different scales. It processes an input image with convolutions and down-samplings. 
Fully connected layers of well-known architectures (AlexNet, VGG16, GoogLeNet) are replaced by convolutions to allow non-fixed size inputs.
The convolutions output feature maps with lower and lower dimensions. Thus, they are up-sampled with up-convolutions (the stride factor is inferior to 1) to recover the original input size.
The \ac{FCN} uses skip connections from features learnt at different levels of the network to generate the final output. We denote with \ac{FCN}-32s a network where the output mask is generated only by up-sampling and processing feature maps with $1/32$ resolution of the input. Similarly, \ac{FCN}-16s is a network where $1/32$ and $1/16$ resolution features maps are used to generate the output mask. In the same manner, \ac{FCN}-8s fuses $1/32$, $1/16$ and $1/8$ resolution feature maps for output prediction.  
This way, the model classifies each pixel of the input into a category and it is trained using a pixel-wise loss.
The \ac{FCN}-8s reached a 62.2\% \ac{mIoU} score on the 2012 PASCAL VOC segmentation challenge using pre-trained models on the 2012 ImageNet dataset.

\paragraph{U-Net.}
\cite{ronneberger_u-net_2015} proposed an extension of the \ac{FCN} \cite{long_fully_2015} for biological microscopy images. The U-net architecture is composed of two pathways: a contracting pathway to compute features and an expanding pathway to spatially localise patterns in the image. 
The down-sampling or contracting pathway has a \ac{FCN} architecture extracting features with $3 {\times} 3$ convolutions. The up-sampling or expanding pathway uses up-convolution (or deconvolution as detailed in Appendix \ref{sec_app:background_segmentation}) reducing the number of feature maps while increasing their dimensions (height and width). Cropped feature maps from the down-sampling pathway of the network are copied and stacked within the up-sampling pathway to avoid loosing information. Finally, a $1 {\times} 1$ convolution processes the feature maps to generate a soft mask of probability to categorise each pixel of the input image. 
Since then, the U-net architecture has been widely extended in recent works, a few of them will be presented in the next paragraphs (FPN, PSPNet, DeepLabv3). This architecture does not use any fully connected layer. As a consequence, the number of parameters of the model is reduced and it can be trained with a small labelled dataset (using appropriate data augmentation). \textit{E.g.} the authors have used a public dataset with 30 images in their experiments.

\paragraph{Feature pyramid network (FPN).}
The Feature Pyramid Network (FPN) developped by \cite{lin_feature_2017}, inspired from U-Net \cite{ronneberger_u-net_2015}, is used in object detection or image segmentation frameworks. 
The architecture is composed of a bottom-up pathway, a top-down pathway and lateral connections in order to join low-resolution and high-resolution features. 
The bottom-up pathway takes an image with an arbitrary size as input. It is processed with convolutional layers and down-sampled by max-pooling layers. 
Note that each group of feature maps with the same size is called a stage. 
The output feature maps of the last layer at each stage are used for the lateral connections of the feature pyramid. 
The top-down pathway consists in up-sampling the last feature maps with un-pooling while enriching them with feature maps from the same stage of the bottom-up pathway using lateral connections. Each lateral connection processes the feature maps of the bottom-up pathway with a $1 {\times} 1$ convolution (reducing the number of feature maps) and stacks its output with the un-pooled feature maps of the top-down pathway.
The concatenated feature maps are then processed by a $3 {\times} 3$ convolution producing the output of the stage. 
Finally, each stage of the top-down pathway generates a prediction to detect an object. 
For semantic segmentation, the authors use two fully connected layers to generate two masks with different sizes over the objects. It works similarly to \acl{RPN} with anchor boxes (see R-CNN \cite{girshick_region-based_2016}, Fast R-CNN \cite{girshick_fast_2015}, Faster R-CNN \cite{ren_faster_2015}). 
According to the authors, the efficiency of this architecture is due to an improvement in the propagation of the information from the last layers of in the network. The FPN based on DeepMask (\cite{pinheiro_learning_2015}) and SharpMask (\cite{pinheiro_learning_2016}) frameworks achieved a 48.1\% \ac{mAR} score on the 2016 COCO segmentation challenge.

\paragraph{Mask R-CNN.}

As presented in \ref{sec:background_detection}, the Mask R-CNN \cite{he_mask_2017} consists in a Faster R-CNN with three output branches. First, it uses an \ac{RPN} extracting \ac{RoI} and features are learnt with the RoIPool layer. Then the output branches are specialized in computing the bounding box coordinates, predicting the associated class and the binary mask\footnote{The Mask R-CNN model computes a binary mask for each object for a predicted class (instance-first strategy) instead of classifying each pixel into a category (segmentation-first strategy).} to segment the object. 
The binary mask has a fixed size and it is generated by a \ac{FCN} for each \ac{RoI}. It also uses a RoIAlign layer instead of a RoIPool to avoid misalignments due to the quantization of the \ac{RoI} coordinates. The particularity of the Mask R-CNN model is its multi-task loss combining the losses of the bounding box coordinates, the predicted class and the segmentation mask. The model tries to solve complementary tasks leading to better performances on each individual task. The best Mask R-CNN uses a ResNeXt \cite{xie_aggregated_2017} to extract features and an FPN architecture. It has obtained a 37.1\% \ac{mAP} score on the 2016 COCO segmentation challenge and a 41.8\% \ac{mAP} score on the 2017 COCO semantic segmentation challenge.

\paragraph{Atrous convolutions (DeepLab and extensions).}

Inspired by the FPN model of \cite{lin_feature_2017}, \cite{chen_deeplab_2018} proposed the DeepLab architecture combining atrous convolution, spatial pyramid pooling and fully connected CRFs.
This model is also called the DeepLabv2, it is an adjustment of the original DeepLab model which will not be detailed here to avoid redundancy. 
The authors have introduced the atrous convolution which is equivalent to the dilated convolution of \cite{zhao_pyramid_2017}. It consists of convolutions with sparse kernels targeting spread pixels.
The number of neurons in the kernel does not change but they are separated by a fixed number of pixels, named dilation rate.
The atrous convolution helps to capture multiple scales of objects. When it is used without max-pooling, it increases the resolution of the final output without increasing the number of parameters.
The \ac{ASPP}, inspired by PSPNet \cite{zhao_pyramid_2017} (see Appendix \ref{sec_app:background_segmentation}), consists in applying parallel atrous convolutions using the same input with different dilation rates. The features maps are processed in separate branches and concatenated using bilinear interpolation to recover the original size of the input. The \ac{ASPP} module helps to learn patterns at different scales with different receptive fields due to the various dilation rates used.
The feature maps are then processed by a fully connected Conditional Random Field (CRF) \cite{krahenbuhl_efficient_2011} computing edges between the features and long term dependencies to produce the semantic segmentation.
The best DeepLab using a ResNet-101 \cite{he_deep_2016} as backbone has reached a 79.7\% \ac{mIoU} score on the 2012 PASCAL VOC challenge, a 45.7\% \ac{mIoU} score on the PASCAL-Context challenge and a 70.4\% \ac{mIoU} score on the Cityscapes challenge.

In their work, \cite{chen_rethinking_2017} have revisited the DeepLab framework to create DeepLabv3 combining cascaded and parallel modules of atrous convolutions. The authors have modified the ResNet architecture to keep high resolution feature maps in deep blocks using atrous convolutions.
An \ac{ASPP} module is used with an additional $1 {\times} 1$ layer rand batch normalization.
The contenated outputs are processed by a final $1 {\times} 1$ convolution to predict the segmentation masks.
The best DeepLabv3 model with a ResNet-101 pretrained on ImageNet and JFT-300M \cite{sun_revisiting_2017} datasets has reached 86.9\% \ac{mIoU} score in the 2012 PASCAL VOC challenge. It also achieved a 81.3\% \ac{mIoU} score on the Cityscapes challenge with a model only trained with the associated training dataset.

The final version called DeepLabv3+ \cite{chen_encoder-decoder_2018} uses an encoder-decoder structure. The author first introduce atrous separable convolution composed of a depth-wise convolution (spatial convolution for each channel of the input using a dilation rate $>1$) and point-wise convolution ($1 {\times} 1$ convolution with the depth-wise convolution as input).
The DeepLabv3 architecture has been used as encoder. 
The authors extracted features with a Xception \cite{chollet_xception_2017} architecture with modifications: they added convolutional layers, they replaced max-pooling layers by atrous depth-wise separable convolutions, and they added batch normalization and ReLU activation following each $3 {\times} 3$ convolutions. 
The feature maps of the backbone are processed by an \ac{ASPP} module with a final $1 {\times} 1$ convolution reducing the number of maps while up-sampling them with a factor of four.
The decoder processes the feature maps from the backbone and from the \ac{ASPP} module with convolutions. The final maps are up-sampled by a factor of four to recover the input dimension and produce the segmentation masks.
The best DeepLabv3+ pre-trained on the COCO and the JFT-300M \cite{sun_revisiting_2017} datasets obtained a 89.0\% \ac{mIoU} score on the 2012 PASCAL VOC challenge. The model trained on the Cityscapes dataset reached a 82.1\% \ac{mIoU} score for the corresponding challenge.

\paragraph{Conclusion.}
The section, supplemented by Appendix \ref{sec_app:background_segmentation}, described a non-exhaustive list of neural network architectures for semantic segmentation published between 2015 and 2018.
A recurrent problem is these approaches is the lack of global visual context in the features learnt by the network, partially explored in the EncNet architecture (see Appendix \ref{sec_app:background_segmentation}).
The state of the art methods used multiple pathways in the network to better propagate the information and to learn relations between the objects.

Considering an entire image, pixel-wise predictions allow a more accurate understanding of a scene and its environment. It is especially true in the context of autonomous driving in complex urban scenes.
Multi-task learning has also shown that neural networks are able to learn more relevant features by solving complementary tasks at the same time.
As an improvement, a multi-head architecture could be considered to solve the semantic segmentation task with other tasks proposed in the COCO challenge (\textit{e.g.} key point detection, action recognition, video captioning or visual question answering).
The following section will briefly introduce deep learning methods and architectures for 3D point cloud scene understanding.

\subsection{Methods for 3D point clouds}
\label{sec:background_3d}

Scenes can also be represented in three dimensions, including 3D point clouds, in order to obtain accurate geometric information. These are points located in 3D Cartesian coordinates. The point cloud representation is sparse, which means that it does not completely fill the Cartesian space, unlike an image composed of dense adjacent pixels.
Methods presented in Sections \ref{sec:background_classif}, \ref{sec:background_detection} and \ref{sec:background_segmentation} cannot be directly applied to a point cloud representation as convolutional layers process dense inputs.

The tasks presented in the previous sections for scene understanding have their own equivalence when applied to point clouds: classification of an entire point cloud, prediction of bounding boxes and point-wise semantic segmentation.
In the automotive industry, the \ac{LiDAR} sensor is commonly used for theses tasks since it provides 3D point clouds recordings.
Processed \ac{RADAR} data can be represented as a point cloud in a Cartesian space (see Section \ref{sec:background_signal_process}). The low resolution of the elevation angle of a \ac{LD} \ac{RADAR} leads to process the \ac{RADAR} point cloud as a \ac{BEV} representation, but both representations are generally processed with similar methods.

This section introduces well-known deep neural network architectures specialised in point cloud processing. It aims to provide insights for a better comprehension of the related work on \ac{RADAR} point clouds that will be presented in Chapter \ref{chap:related_work}.

\paragraph{PointNet.}
The PointNet model proposed by \cite{qi_pointnet_2017} is the first end-to-end neural network architecture that processes unordered sets of 3D points while being adapted for point cloud classification, point-wise semantic segmentation and scene semantic parsing.
Its first module processes the point cloud with a mini-network called T-Net, based on successive fully connected layers extracting point-wise features. The final layer of the T-Net is a \ac{MLP}, shared between all the points, predicting an affine transformation matrix. This matrix is directly applied to the coordinates of the input point cloud to adjust its geometric representation.
Then, a shared \ac{MLP} learns point-wise features from the transformed point cloud and forwards them to the next stage. 
Successive T-Net transformation and shared \ac{MLP} process the representations before applying a max-pooling on the point dimension producing a global feature vector. This vector is then used either as input of an \ac{MLP} of point cloud classification, or in a segmentation network for semantic segmentation.

The segmentation network combines local features from the middle stage \ac{MLP} to the global feature vector. The concatenated representations are then processed by two successive \acp{MLP} shared between all the points: the first combines the local and global concatenated features, the second predicts a class for each point to perform semantic segmentation.

The PointNet architecture uses only shared \acp{MLP} and fully connected layers to learn local and global features from an unordered point cloud. Moreover, it can easily be adapted to the application. This method reached the best performances for point cloud classification, segmentation and scene parsing with less computational complexity than competing methods. To this day, PointNet is still used as a backbone for point cloud feature extraction for both \ac{LiDAR} and \ac{RADAR} point clouds.

\paragraph{PointNet++.}
An extension named PointNet++ \cite{qi_pointnet_2017-1} has been proposed to capture local patterns with increasing contextual scales in order to improve performances in complex scenes. The authors proposed a hierarchical neural network applying PointNet recursively to increase the receptive field of each layer similarly to stacked convolutions for image processing. This network learns fine-grained structures and either uses the last down-scaled point cloud as input of a \ac{MLP} for classification, or merges feature to obtain the global context of the scene for segmentation.

Additionally to the recursive PointNet, the authors proposed the Multi Scale Grouping (MSG) and the Multi Resolution Grouping (MRG) approaches. The MSG operates before the PointNet to concatenate features from different scale levels and obtain a multi-scale feature. 
The MRG initially aims to apply a local PointNet on each point provided by the previous layer. This method is computationally expensive, so the authors approximated it by using a single PointNet applied to each local group of features which have been aggregated at a higher level.
For the segmentation task, they proposed a Point Feature Propagation to segment the original point cloud from a down-sampled point cloud. Each classified point is interpolated at the upper level using an inverse weighting average of its local neighborhood while the skip connections of the network transmit the information to each level.

The PointNet++ succeeds to learn hierarchical features with a recursive PointNet while fusing multi-scale features with grouping approaches. The proposed architecture is more computationally expensive than PointNet, but it provides better performance and can also be adapted to various applications. It is also commonly used for 3D scene understanding depending on the trade-off to reach between the performances and the inference time.

\paragraph{Pixor.}

The PIXOR architecture \cite{yang_pixor_2018} has been introduced as a single stage method for 3D object detection from \ac{LiDAR} point cloud claiming real time inference useful for autonomous driving.
The authors transformed a 3D point cloud in dense 2D \ac{BEV} representations\footnote{One of the advantages of the \ac{BEV} representation is that objects do not overlap w.r.t. the front-views.}, each one corresponding to an elevation angle discretization.
The 2D \acp{BEV} are transformed in occupancy grids with fixed resolutions and stacked in the channel dimension.

The PIXOR architecture is composed of an FPN backbone with successive 2D convolutions and down-sampling layers while respecting that the last feature map has sufficient resolution per pixel to contain an object.
The head part of the network processes the backbone feature maps and performs object recognition and location simultaneously. The location task consists of classifying each pixel of the feature map as being an object or not. The recognition task regresses 6 parameters per pixel corresponding to the offset of the object's position from the pixel location, its size and orientation.

The proposed architecture uses 2D convolutions on 3D points clouds transformed in dense representations while being a single stage approach. It reached state of the art performances in 3D object detection while operating in real time and is therefore a good trade-off for autonomous driving applications.

\paragraph{Conclusion.}
Point clouds are sparse representations that need to be either processed point-wise, or transformed in dense representations to use conventional methods.
This section provided details on the most mainstream deep learning architectures for point cloud processing, namely PointNet, PointNet++ and PIXOR.
Many recent methods and specific sparse convolution operations have been designed for 3D point cloud processing but they are out of the scope of this thesis. We suggest that the reader refers to the work of \cite{guo_deep_2020} for in-depth information.

The following chapter will review the related work on \ac{RADAR} datasets and deep learning methods applied to \ac{RADAR} data for scene understanding, in particular for classification, object detection and semantic segmentation.

\chapter{Related work}
\label{chap:related_work}
\minitoc

Over the years, deep learning methods have evolved to suit many research areas.
This chapter reviews the related work of deep learning algorithms applied to \ac{RADAR} data. 
Section \ref{sec:related_diverse} briefly introduces diverse applications using \ac{RADAR} data. In Section \ref{sec:related_datasets}, we focuses on open source \ac{RADAR} datasets, which are a prerequisite for improvements in deep learning algorithms.
Sections \ref{sec:related_detection} and \ref{sec:related_segmentation} respectively review the related works on object detection and semantic segmentation applied to automotive \ac{RADAR}.
Finally, Section \ref{sec:related_fusion} reports on the applications of sensor fusion including a \ac{RADAR} sensor.

\section{Diverse applications}
\label{sec:related_diverse}

As detailed in Section \ref{sec:background_radar_theory}, a \ac{RADAR} sensor provides the position and Doppler of the surrounding reflectors, which is interesting to explore in many research areas. 
Unlike with a camera, it is difficult to identify and distinguish people in \ac{RADAR} data. Therefore this sensor helps to understand scenes while respecting the privacy of the users. 
In particular, hand gesture recognition using \ac{RADAR} data has been widely explored because of its application to human control on a mobile device, an electronic watch or inside a car.

Classification of hand gestures has been explored with end-to-end \ac{CNN} applied to Doppler spectrograms\footnote{A Doppler spectrogram representation is a time-Doppler representation corresponding to a single \ac{FFT} applied on the recorded \ac{RADAR} data considering the chirp sampling axis (see Section \ref{sec:background_signal_process})} \cite{kim_hand_2016, dekker_gesture_2017}.
Simulated Doppler spectrograms have also been used to train an \ac{MLP} for gesture classification \cite{ishak_human_2018}.
In their work, \cite{wang_interacting_2016} used \acp{CNN} to generate embeddings of \ac{RD} sequences used as input of an \ac{RNN} with an \ac{LSTM} cell. The presented results have highlighted that the temporal information is relevant for a classification task using \ac{RADAR} data.
In the same way, \cite{zhang_latern_2018} classified sequences of range spectrograms stacked in the temporal dimension with 3D convolutions and an \ac{LSTM} cell trained with a CTC loss\footnote{The Connectionist Temporal Classification (CTC) function \cite{graves_connectionist_2006} is specialized in calculating a loss between a continuous time series and a target sequence.} function \cite{graves_connectionist_2006}.
Unsupervised learning has also been explored to learn representations of range spectrograms with VGG \aclp{AE} \cite{zhang_u-deephand_2019}, which are then fine-tuned for hand gesture classification.
\cite{lei_continuous_2020} suggested an architecture with different branches to process \acl{RD} and \acl{RA} views with an \ac{RNN} exploiting the temporal axis for hand gesture classification. An additional spatio-Doppler attention has also been proposed by \cite{hazra_radar_2019}.
Since objects' signatures evolve through time, 
\cite{scherer_tinyradarnn_2021} proposed an architecture for hand gesture classification with a \ac{CNN} backbone and a Temporal Convolutional Network (TCN) using dilated 1D convolutions extracting temporal features from \ac{RD} and achieving better results than \ac{RNN} with \ac{LSTM}.
The \ac{RAD} tensor has also been considered for hand gesture classification by being aggregated in views which are processed by separate branches and finally merged \cite{wang_human_2021}.
\cite{sun_automatic_2019} detected multiple gestures with a Faster R-CNN including 3D convolutions on Doppler spectrograms recorded inside a car. 
Segmentation and detection of several gestures have also been explored by improving Mask R-CNN with additional modules of max and average pooling processed by \acp{MLP} \cite{wang_rammar_2019}.
Indoor car driver's hand gesture recognition has been tackled with multi-class classification using sensor fusion with camera, depth estimation and Doppler spectrograms processed with 3D convolutions \cite{molchanov_multi-sensor_2015}.

Deep learning models have also shown interesting results in differentiating human activities. Doppler spectrograms have been used for human detection and indoor activity classification \cite{kim_human_2016}. A more advanced architecture has been proposed by \cite{zhu_classification_2020} using fully 1D convolutions and \ac{RNN} with \ac{LSTM} cells to solve the same task.
Doppler spectrograms have been explored for human gait classification including the temporal information with \ac{RNN} and \ac{LSTM} cells \cite{klarenbeek_multi-target_2017} or more advanced architectures with dual stream of the Doppler representation \cite{chen_attention-based_2021} based on Vision Transformers (ViT) \cite{dosovitskiy_image_2021}.
Deformable convolutions \cite{dai_deformable_2017} have been used to adapt convolutional kernels to objects' signature on Doppler spectrograms for fall motion classification.

\ac{RADAR} sensors have also been exploited for outdoor application, \textit{e.g.} to classify Unmanned Aircraft Vehicles (UAV) using a temporal \ac{FCN} on Doppler spectrograms \cite{brooks_temporal_2018}.
\cite{wang_deep_2020} explored ice layer segmentation with a ground-penetrating \ac{RADAR}.
Counting people with \ac{RADAR} data has been addressed by adapting a network trained on natural images to \ac{RA} \ac{RADAR} views and Doppler spectrograms using knowledge distillation \cite{aydogdu_multi-modal_2020}.
Since \ac{RADAR} representations contain speckle noise, denoising \aclp{AE} have been explored on \ac{RD} views \cite{de_oliveira_deep_2020} using filtered representations as ground truth.
Micro-motions of raw \ac{RADAR} data in the temporal domain have been used as input of an \acl{AE} to estimate contactless electrocardiograms while considering domain transformation \cite{chen_contactless_2021}. 
Unsupervised learning has also been explored from \ac{RA} map of a scanning \ac{RADAR} to learn feature embeddings using a deep learning architecture for place recognition \cite{gadd_unsupervised_2021}.

\sloppy
The \ac{SAR} is used as an embedded sensor on orbital satellites, planes and aerial platforms. It exploits the motion of the \ac{RADAR}'s antennas to create an image and uses the combination of multiple pulses to create a synthetic aperture and improve its spatial resolution. \ac{SAR} data are similar to 2D \ac{BEV} representations, they have been used in many applications, \textit{e.g.} for pixel-wise land segmentation \cite{zhang_complex-valued_2017, zhang_polsar_2020, pham_very_2021} 
or sensor fusion for panoptic semantic segmentation \cite{garnot_multi-modal_2021}. Deep learning applications have been developed for \ac{SAR} data, in particular for earth observation, but are beyond the scope of this thesis.

The lack of open source \ac{RADAR} data (with or without annotation) has impeded deep learning research in many fields. Generating data as realistic Doppler spectrograms of simple objects using \acp{GAN} \cite{goodfellow_generative_2014} has been explored \cite{truong_generative_2019}. Synthetic \ac{RADAR} data have been simulated using CycleGAN \cite{zhu_unpaired_2017} for ice layer segmentation \cite{rahnemoonfar_radar_2020}. Doppler spectrograms have also been generated using \acp{GAN} for human gait classification.
The automotive industry also suffers from the lack of open source \ac{RADAR} dataset. Generative approaches could address this issue but they have shown limitations to reproduce the properties of \ac{RADAR} data (see Section \ref{sec:dataset_generation}). The following section will detail existing automotive \ac{RADAR} datasets specialized in scene understanding. 

\section{Automotive RADAR datasets}
\label{sec:related_datasets}

\begin{table*}[!ht]
    \centering
    \scriptsize
    \begin{tabular}{l|c|c|c|c|c|c|c|c|c|c|c}
    \bottomrule
    \multirow{5}{*}{Dataset} & \multirow{5}{*}{Year} &  \multirow{5}{*}{Scale} & \multicolumn{5}{c|}{RADAR data} &  \multirow{5}{*}{\rotatebox{90}{RADAR type}} &  \multirow{5}{*}{\rotatebox{90}{Modalities}} & \multirow{5}{*}{\rotatebox{90}{Sequence}} & \multirow{5}{*}{\shortstack{Annotation \\ type}}  \\
    & & & \rotatebox{90}{ADC} & \rotatebox{90}{RAD} & \rotatebox{90}{RA or RD} & \rotatebox{90}{PC} & \rotatebox{90}{Doppler}  &  &  &  \\
    \midrule
    nuScenes  & 2019 & Large & \xmark & \xmark & \xmark & \cmark & \cmark & LD &   CLO & \cmark  & 3D Boxes \\
    Astyx & 2019 & Small & \xmark & \xmark & \xmark & \cmark & \cmark & HD & CL  & \xmark  & 3D Boxes \\
    RadarRobotCar & 2020 &  Large  & \xmark & \xmark & \cmark & \xmark & \xmark & S & CLO  &  \cmark & \xmark  \\
    RADIATE & 2020 & Medium & \xmark & \xmark & \cmark & \xmark & \xmark  &  S  & CLO & \cmark & 2D Boxes \\
    MulRan & 2020 &  Medium  & \xmark & \xmark & \cmark & \cmark & \xmark &  S  &  CLO & \cmark  & \xmark  \\
    Zendar & 2020 & Small  & \xmark & \xmark & \cmark & \cmark & \cmark & HD & CL & \cmark  & 2D Boxes  \\
    \midrule
    CARRADA (Sec.~\ref{sec:carrada}) & 2020 &  Small  & \xmark & \cmark & \cmark & \cmark & \cmark & LD & C  & \cmark  &  2D Boxes, Seg. \\
    \midrule
    CRUW & 2021 & Medium & \xmark & \xmark & \cmark & \xmark & \xmark  & LD & C & \cmark &  Point Location \\
    RadarScenes & 2021 &  Large & \xmark & \xmark & \xmark & \cmark & \cmark & HD & CO  & \cmark & Point-wise \\
    RADDet & 2021 &  Small  & \xmark & \cmark & \cmark & \xmark & \cmark & LD & C  &  \cmark & 2D Boxes \\
    \midrule
    RADIal (Chap.~\ref{chap:hd_radar}) & 2021 & Medium & \cmark & \cmark & \cmark & \cmark & \cmark & HD & CLO & \cmark  & 2D Boxes, Seg. \\
    \bottomrule
    \end{tabular}
    
    \caption[Publicly-available driving RADAR datasets]{\textbf{Publicly-available driving RADAR datasets}. 
    The dataset scale is Small ($<15$k frames), Large ($>130$k frames) or Medium (in between).
    The used \ac{RADAR} is \ac{LD}, \ac{HD} or Scanning (S). 
    \ac{RADAR} data are released in different representations, amounting to different signal processing pipelines: \ac{ADC} signal, \ac{RAD} tensor, \ac{RA} view, \ac{RD} view, Point Cloud (PC). Presence of Doppler information depends on the \ac{RADAR} sensor. Other sensor modalities are Camera (C), \ac{LiDAR} (L) and Odometry (O).
    CARRADA (see Section \ref{sec:carrada}) is the only dataset providing both \ac{RAD} and PC of a \ac{LD} \ac{RADAR} with bounding box and segmentation annotations.
    RADIal (see Chapter \ref{chap:hd_radar}) is the only dataset providing each representation of a \ac{HD} \ac{RADAR}, combined with camera, \ac{LiDAR} and odometry, while proposing detection and free space segmentation tasks.}
    \label{tab:related_public_dataset}
\end{table*}

Deep learning algorithms, trained with large amounts of annotated data, reached impressive performances for scene understanding tasks.
Annotated datasets are required to drive research in scene understanding using \ac{RADAR} representations.
This section will review the existing open source datasets for \ac{RADAR} scene understanding which have been released over the past three years.

\subsection{Traditional RADAR}

Traditional \acp{RADAR}, or \ac{LD} \acp{RADAR}, offer a good trade-off between cost and performance. 
Most of them are composed of two transmitter and four receiver antennas, leading to eight virtual antennas. 
They provide accurate range and velocity while being robust to adverse weather conditions.
The \textbf{nuScenes} dataset \cite{caesar_nuscenes_2020} is composed of 5.5 hours of recorded sequences in two countries including night and rain weather conditions. This dataset is composed of simultaneous recordings from five \acp{RADAR}. It has an additional 32-beam \ac{LiDAR} and 6 cameras. The \ac{RADAR} data are filtered and released as sparse point clouds with Doppler information. Data are annotated with 3D bounding boxes, classified between 23 classes, and provided in both Cartesian coordinates and camera domain.
The \textbf{CRUW} dataset \cite{wang_rethinking_2021} proposes 3.5 hours of recorded sequences in multiple scenarios (parking, campus road, city street, highway). The authors used a camera and a \ac{LD} \ac{RADAR} but they released only \ac{RA} views. The provided annotations are single point object locations in the \ac{RA} map with three classes.
The recently released RADDet dataset \cite{zhang_raddet_2021} contains around 10,158 frames recorded in urban scenes with a stationary car mounted with stereo cameras and an \ac{LD} \ac{RADAR}. The dataset contains \ac{RAD} tensors, \ac{RA} and \ac{RD} views; it does not contain \ac{RADAR} point clouds. The annotations are 2D bounding boxes, classified in 6 classes, for each \ac{RADAR} view. Additional details on the \textbf{RADDet} dataset and the annotation pipeline proposed by the authors are provided in Section \ref{sec:mvrss_raddet_dataset}.
Traditional \acp{RADAR} have a poor angular resolution and multiple sensors are required to have a perception of the scene all around the car. This second limitation is overcome by scanning \acp{RADAR}.

\begin{figure}[!t]
  \begin{center}
    \includegraphics[width=0.7\textwidth]{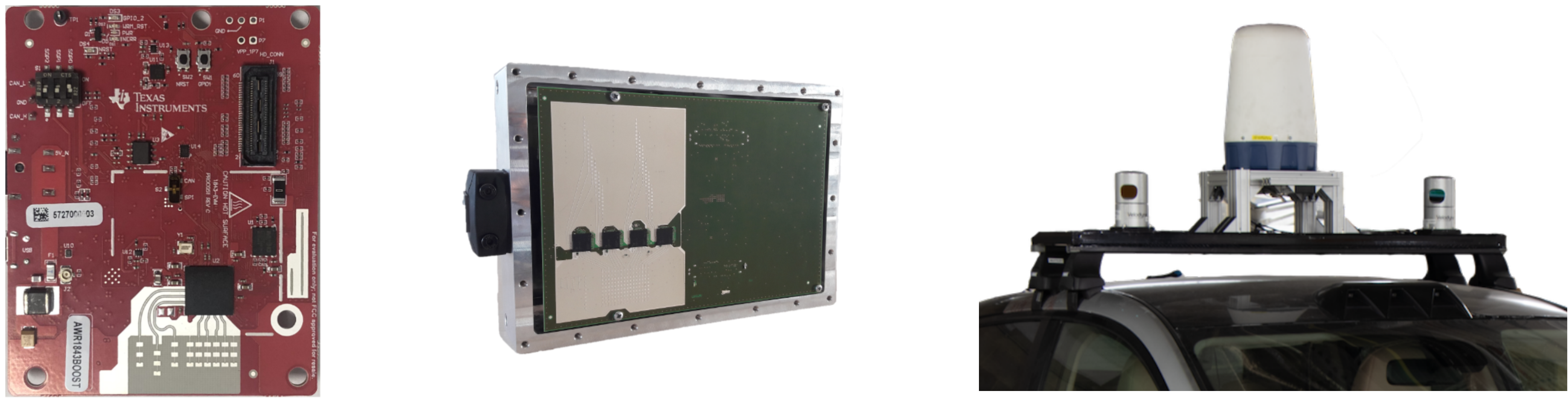}
  \end{center}
  \caption[Examples of Low-Definition, High-Definition and Scanning RADARs]{\textbf{Examples of Low-Definition, High-Definition and Scanning RADARs}. (Left) \acl{LD} \ac{RADAR} used in the CARRADA dataset (see Section \ref{sec:carrada}), (middle) \acl{HD} \ac{RADAR} used in the RADIal dataset (see Chapitre \ref{chap:hd_radar}) and (right) Scanning \ac{RADAR} on the roof of a car used in the Oxford RADAR RobotCar dataset \cite{barnes_oxford_2020}.
  }
  \label{fig:radar_sensors}
\end{figure}

\subsection{Scanning RADAR}

Transmitter antennas of scanning \acp{RADAR} generate signals successively to obtain a $360^{\circ}$ \ac{FoV} of the scene.
The \textbf{Oxford RADAR RobotCar} dataset \cite{barnes_oxford_2020} contains 280 km of recorded urban scenes in the city of Oxford at different periods of the year. The car is moving and mounted with three cameras, two 32-beam \acp{LiDAR} and a $360^{\circ}$ scanning \ac{RADAR}.
The provided \ac{RADAR} representations are \ac{RA} maps without annotation.
The \textbf{RADIATE} dataset \cite{sheeny_radiate_2021} is composed of 3 hours of \ac{RADAR} sequences in diverse types of scenes (parking, urban, motorway and suburban) with various weather conditions. The moving car is mounted with a camera, a 32-beam \ac{LiDAR} and a $360^{\circ}$ scanning \ac{RADAR}. The provided \ac{RA} maps are annotated with 2D bounding boxes classified between eight classes.
The \textbf{MulRan} dataset \cite{kim_mulran_2020} has a total of 41.2 km of recorded sequences in urban scenes within different cities at the month-level temporal gap. The authors used a 64-beam \ac{LiDAR} and a $360^{\circ}$ scanning \ac{RADAR} to record \ac{RA} representations without annotation on the \ac{RADAR} data.
The $360^{\circ}$ scanning \acp{RADAR} provide a high perception around the car with their large \ac{FoV}, however their angular resolution is limited as traditional \acp{RADAR} and they do not provide Doppler information. \ac{HD} \ac{RADAR} provides both improved angular resolution and Doppler information.

\subsection{High-definition RADAR}

Recent \ac{HD} \acp{RADAR} reach an azimuth angular resolution below the degree using large arrays of virtual antennas\footnote{\ac{HD} \acp{RADAR} contains hundreds of virtual antennas against tens for traditional or \ac{LD} \acp{RADAR}.}.
The \textbf{Astyx} dataset \cite{meyer_automotive_2019} is composed of only 546 frames without temporal information. The authors used a \ac{HD} \ac{RADAR} coupled with a camera and a 16-beam \ac{LiDAR}. The dataset provides a dense point cloud including the Doppler information with 3D bounding box annotations, classified in 7 classes, in both Cartesian coordinates and image domain.
The \textbf{Zendar} dataset \cite{mostajabi_high-resolution_2020} is composed of 7.2 minutes of recorded sequences in urban scenes with \ac{RD}, \ac{RA} representations and \ac{RADAR} point clouds. The scenes are recorded with a \ac{HD} \ac{RADAR}, a camera and a 16-beam \ac{LiDAR}. The annotations are provided with 2D bounding boxes considering a single category in Cartesian coordinates.
The \textbf{RadarScenes} dataset \cite{schumann_radarscenes_2021} provides four hours for urban scene sequences with diverse weather conditions, traffic density and road classes (highway, city). The provided data are point clouds recorded from an \ac{HD} \ac{RADAR} and camera images. The camera images are annotated with semantic segmentation masks and the \ac{RADAR} data with point-wise annotations. The objects are distinguished between 11 classes.
None of these datasets provides \ac{ADC} or \ac{RAD} tensor data from a \ac{HD} \ac{RADAR}. \ac{HD} \ac{RADAR} datasets usually propose \ac{RADAR} point clouds processed from the raw data. Unfortunately, the pre-processing pipeline looses information and requires a high computational cost.

\subsection{Our proposals}

In this thesis, we introduce two datasets to overcome the lack of annotated raw \ac{RADAR} data for both \ac{LD} and \ac{HD} \ac{RADAR} sensors.
In Section \ref{sec:carrada}, we propose the \textbf{CARRADA} dataset, the only dataset providing camera images with synchronised \ac{RAD} tensors and their corresponding views annotated with bounding boxes and semantic segmentation labels. A semi-automatic annotation tool generating the annotations in the \ac{RADAR} sequences is also presented.
In Chapter \ref{chap:hd_radar}, our collaborative work proposes the unique \textbf{RADIal} dataset composed of raw data recorded with \ac{HD} \ac{RADAR} together with camera and \ac{LiDAR} in various driving environments, filling a gap in existing automotive \ac{RADAR} datasets.
Table \ref{tab:related_public_dataset} summarizes the characteristics of the publicly-available driving datasets with \ac{RADAR}.
The presented open source \ac{RADAR} datasets have opened up research on scene understanding for automotive applications. The following sections will detail the related work on object detection, semantic segmentation and sensor fusion using \ac{RADAR} data.

\section{RADAR object detection}
\label{sec:related_detection}

Classification of \ac{RADAR} data for automotive applications has been little explored, \textit{e.g.} for vehicle classification \cite{capobianco_vehicle_2017}.
\ac{RADAR} point clouds have been exploited for person identification with the temporal dimension using PointNet \cite{qi_pointnet_2017} with an attention module \cite{cheng_person_2021} or with causal dilated temporal convolution for additional tracking \cite{pegoraro_real-time_2021}.
Classification is a restrictive task not suited to scene understanding, requiring to detect multiple objects in the environment of the car. 
The object detection task has been popular for the past few years to better exploit \ac{RADAR} representations and improve the comprehension of the environment of a car.

\subsection{Range-Angle-Doppler tensor}

The entire \acl{RAD} tensor has been explored to detect objects in the \ac{RA} view corresponding to polar coordinates. In their work, \cite{major_vehicle_2019} proposed to aggregate the views of the \ac{RAD} tensor by pair of axes, process them with individual \ac{CNN} branches and stack their features as the input of a single decoder exploiting the temporal dimension to detect object in \ac{RA}.
The \ac{RAD} tensor has also been aggregated in 2D views in the work of \cite{gao_ramp-cnn_2020}, each one being processed by a dedicated 3D autoencoder including the temporal information. The produced feature maps are fused and processed with an inception module for single point object location in the \ac{RA} view. The authors also proposed data augmentation methods specialized for \ac{RA} views. Additional details on the RAMP-CNN architecture proposed by \cite{gao_ramp-cnn_2020} are provided in Section \ref{sec:mvrss_radar_based_methods}.
In their work, \cite{palffy_cnn_2020} localized an object in \ac{RA} to crop the \ac{RAD} tensor and classify the sub-\ac{RAD} tensor with the Doppler information into a road user class.
Recent work proposed to extract features from the \ac{RAD} tensor with a ResNet backbone, considering the third dimension as channels \cite{zhang_raddet_2021}. Feature maps are then used to detect objects with two independent YOLO detection heads, one for 3D coordinates in the \ac{RAD} tensor, the other for 2D coordinates in a Cartesian \ac{RA} map.
The \ac{RAD} tensor is a cumbersome \ac{RADAR} representation usually processed with a sub-representation to benefit from both the location and the Doppler of the reflectors. However, the entire tensor is not always available, therefore most of the previous related works are single view approaches.

\subsection{Range-Angle or Range-Doppler view}

Single view approaches consider either the \acl{RA} or the \acl{RD} view to detect objects in one of them. Information related to the reflectors is reduced because it does not process the location and the Doppler information of the reflectors simultaneously.
Considering \ac{RA} view only helps to detect objects in a 2D space but it excludes the Doppler.
The \ac{RA} views have been used in a two-stage approach \cite{gao_experiments_2019}, a clustering localizes objects' signatures and a VGG architecture classifies them. Objects have been detected in \ac{RA} views with single-stage approaches comparing the YOLO and SSD architectures \cite{stroescu_object_2020}.
Objects have been detected and classified in \ac{RA} views of indoor scenes using a 300Ghz \ac{RADAR} with an end-to-end \ac{CNN} \cite{sheeny_300_2020}.
Objects have been detected in Cartesian \ac{RA} with an architecture similar to SSD while minimizing the aleatoric uncertainty of the model, \textit{i.e.} the variance of its predictions \cite{dong_probabilistic_2020}.

The recent CRUW dataset \cite{wang_rethinking_2021} has opened up research on deep learning architectures for object location (single point detection) in \ac{RA} view.
The RODNet architecture proposed by \cite{wang_rodnet_2021} processes \ac{RA} with 3D convolutions and inception modules to localize objects with a single point considering a confidence map.
Objects have been localised with a teacher-student method based on RODNet while improving spatio-temporal features with densely connected residual blocks based on atrous convolutions in both spatial and temporal dimensions \cite{hsu_efficient-rod_2021}.
Multiple variants of the RODNet architectures have been considered \cite{sun_squeeze-and-excitation_2021} including temporal inception modules and fused their prediction to localize objects in \ac{RA}.
A multi-scaled U-Net architecture has also been explored for this task \cite{ju_danet_2021} while proposing an inception module with 3D factorized convolutions.
In their work, \cite{zheng_scene-aware_2021} introduced a neural network with two branches, one recognizing urban scenes and the other localizing objects in \ac{RA}. The authors also introduced SceneMix, a set of data augmentation methods for \ac{RA} views specialized by type of scenes.
More recently, \cite{azam_channel_2021} proposed to detect objects in a \acl{BEV} of \ac{RA} by considering \ac{RADAR} representation in grey scale transformed in RGB, LAB and LUV as the input of a transformer network using multi-head attention layers.

A few works have performed object detection in \ac{RD} view providing the velocity information of the objects. However their position is ambiguous, only their distance can be deduced.
Objects have been detected in \ac{RD} using a YOLO architecture while defining anchor priors with the objects' signature in the grid of the representation \cite{perez_deep_2019}.
The U-Net architecture has also been considered by \cite{ng_range-doppler_2020} to process simulated \ac{RD} with complex values.
In their work, \cite{lee_wei-yu_spatio-temporal_2021} suggested a self-supervised method for feature alignment using self-weighted masks and a spatio-temporal consistency loss exploiting both annotated and non-annotated frames. Their method is applicable to both \ac{RA} and \ac{RD} views.

Other methods have proposed to process the \ac{RD} view to deduce information about the location of the objects in the \ac{RA}.
A two-stage approach has been proposed to detect an object in the \ac{RD} and then to estimate its angle \cite{brodeski_deep_2019}.
Complex \acp{RD} have been processed with U-Net with complex convolutions to predict bounding boxes in \ac{RA} \cite{stephan_human_2021}.
A recent work of \cite{meyer_graph_2021} exploited a tensor of \ac{RD} view, without the third \ac{FFT} on the antenna axis, to create a graph which is processed with graph convolutions and a ResNet \ac{FPN} backbone for 3D bounding box detection in \ac{RA}.
These methods try to localise objects without the entire \ac{RADAR} information. Advanced neural network architectures have been proposed but they do not exploit both the location and the Doppler information of the reflectors. The point cloud \ac{RADAR} is a lighter representation than the \ac{RAD} tensor with less information about each object but it carries both their location and Doppler.

\subsection{RADAR point cloud}

The \ac{RADAR} representation in Cartesian point cloud contains both the Doppler (excepting for scanning \ac{RADAR}) and the \ac{RCS} information. The point cloud is either sparse or dense depending on the \ac{RADAR} sensor (\ac{LD} or \ac{HD}).
In their work, \cite{danzer_2d_2019} processed a 4D \ac{RADAR} point cloud (Cartesian coordinates, Doppler and \ac{RCS}) with modified Frustum PointNets \cite{qi_frustum_2018} generating patch proposals, classifying each patch, segmenting the points in each patch and regressing the 2D bounding box coordinates.
The work of \cite{scheiner_seeing_2020} used \ac{RADAR} point clouds to detect, classify and track masked objects considering reflected waves received by the \ac{RADAR} with a specific \ac{AE} architecture.
The PointNet++ \cite{qi_pointnet_2017-1} architecture has been explored by \cite{svenningsson_radar-pointgnn_2021} to process \ac{RADAR} point clouds with a 3D convolutional \ac{FPN} and two output branches, one detecting objects and the other performing tracking.
The work of \cite{svenningsson_radar-pointgnn_2021} proposed to consider PointNet++ features as a graph and apply graph convolutions to generate contextual embeddings used to detect and classify objects in a \ac{BEV} representation of the sparse point cloud.
A dense point cloud of a \ac{HD} \ac{RADAR} has been processed for 3D object detection using a self-attention mechanism to learn global pillar\footnote{Point clouds are sometimes transformed into a pillar representation: each point with a 3D Cartesian position has its height axis extended with a fixed value.} features used as input of a 3D \ac{CNN} and an \ac{RPN} head \cite{xu_rpfa-net_2021}.
These methods considered point clouds from either \ac{LD} or \ac{HD} \ac{RADAR}. However, their raw signals are processed to create the point cloud and objects may be missed as explained in Section \ref{sec:background_signal_process}.
\\

\begin{figure}[!t]
  \begin{center}
    \includegraphics[width=0.7\textwidth]{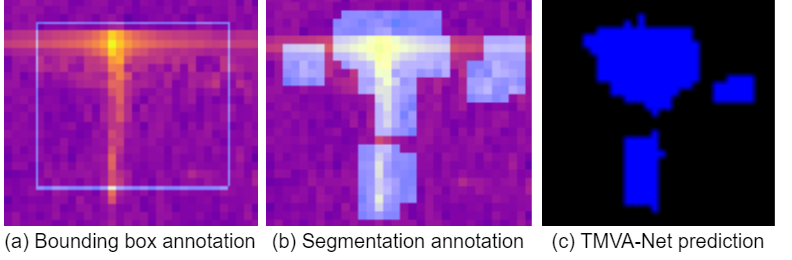}
  \end{center}
  \caption[Example of the annotation of a `Car' signature in Range-Doppler]{\textbf{Example of the annotation of a `Car' signature in \acl{RD}}. The annotations of a `Car' signature in a cropped \acl{RD} view are compared between (a) bounding box and (b) semantic segmentation. The bounding box annotation includes speckle noise while the segmentation mask is well suited to the object signature.
  Our proposed TMVA-Net neural network architecture detailed in Section \ref{sec:mvrss_archi_ours} successfully segments the object signature (c).} 
  \label{fig:background_annotations}
\end{figure}

The object bounding box detection task has been widely explored but it is not well suited to \ac{RADAR} data. Objects' signatures have non-compact shapes, they are extended on an axis of the representation and they have various sizes as illustrated in Figure \ref{fig:background_annotations}. Bounding boxes include noise and sometimes shapes from different categories. The next section will detail the related work on \ac{RADAR} semantic segmentation to overcome these issues.

\section{RADAR semantic segmentation}
\label{sec:related_segmentation}

The characteristics of the object signature in \ac{RADAR} representations make semantic segmentation well suited for scene understanding (see Figure \ref{fig:background_annotations}).
The shapes are non-compact due to the surface of the objects which reflecting the signals. Moreover, objects can be close to each other and partially mixed due to the sensor resolution.
The pixel-wise segmentation of the representations takes these problems into account with more accurate predictions.
\ac{RADAR} semantic segmentation has not been extensively explored due to the lack of annotated data.
The CARRADA dataset opened up research in this field thanks to the semi-automatic pipeline generating annotations, including semantic segmentation, proposed in Section \ref{sec:carrada_annot_pipeline}.
Nonetheless, to the best of our knowledge, there is no previous work on semantic segmentation using the entire \ac{RAD} tensor as our method proposed in Section \ref{sec:mvrss}. The following sections detail applications of \ac{RADAR} semantic segmentation on single \ac{RADAR} views or on \ac{RADAR} point clouds.

\subsection{Range-Angle view}

To the best of our knowledge, there is no previous work on \ac{RADAR} semantic segmentation on \ac{RD} views; this section thus focuses on \ac{RA} views only.
Occupancy grid segmentation is a popular area of research using \ac{RADAR} data.
The work of \cite{lombacher_semantic_2017} proposed a segmentation of Cartesian \ac{RA} maps in \ac{BEV} to distinguish occupied (cars or others) and free pixels using a 6-layer \ac{CNN} architecture.
The segmentation of Cartesian \acl{BEV} \ac{RA} map has been extended to classify the pixels between static (occupied or free) and dynamic road users, using an \acl{AE} architecture with up-convolutions while predicting the motion of dynamic objects \cite{hoermann_dynamic_2018}.
In their work, \cite{prophet_semantic_2019} generated occupancy grids from a dense \ac{HD} \ac{RADAR} point cloud and segmented them between four classes using FCN-8s \cite{long_fully_2015}, U-Net \cite{ronneberger_u-net_2015} and SegNet \cite{badrinarayanan_segnet_2017}.
The DeepLabv3+ \cite{chen_encoder-decoder_2018}, FCN and a lightweight version of FCN have been compared for binary occupancy segmentation of \ac{RA} views in both Cartesian and polar coordinates \cite{nowruzi_deep_2020}.
Occupancy grids have also been generated using \ac{LiDAR} and sparse \ac{RADAR} point clouds \cite{sless_road_2019} and segmented using a simple \ac{CNN} architecture similar to U-Net.
\ac{HD} \ac{RADAR} point clouds have been used to create 3D occupancy grid \cite{prophet_semantic_2020}, projected in 2D \ac{BEV} for occupancy segmentation using Deeplabv3+ and SegNet.
The work of \cite{kaul_rss-net_2020} proposed RSS-Net, the first architecture for \ac{RADAR} semantic segmentation trained on Cartesian \acl{BEV} \ac{RA} maps with seven categories. They also proposed a method to transfer the annotations from the \ac{LiDAR} sensor to the \ac{RADAR} representation. The RSS-Net architecture is similar to DeepLabv3+ with an \ac{ASPP} module, it is trained with a custom weighted cross entropy loss. Additional details on the RSS-Net architecture are provided in Section \ref{sec:mvrss_radar_based_methods}.


\subsection{RADAR point cloud}

The point cloud segmentation task consists in classifying each point in a category. Point clouds are usually represented in \ac{BEV} and Cartesian coordinates due to the low elevation resolution of the \ac{RADAR} sensor.
The PointNet++ architecture \cite{qi_pointnet_2017-1} has been used to segment dense \ac{HD} \ac{RADAR} point clouds \cite{feng_point_2019}.
In their work, \cite{schumann_semantic_2018} segmented 2D \ac{RADAR} point clouds using PointNet++ with additional multi-scale grouping and feature propagation modules.
The PointNet++ has also been extended to segment \ac{RADAR} point clouds with a mean shift \cite{comaniciu_mean_2002} feature extractor learning local dependencies and a feature fusion module based on attention mechanism \cite{cennamo_leveraging_2020}. 
The work of \cite{cheng_new_2021} proposed to use the coordinates of the points and the \ac{RD} view as inputs of a segmentation network similar to U-Net in order to classify each point as occupied or free.
\ac{RADAR} point clouds in \ac{BEV} have been segmented for anomaly detection \cite{griebel_anomaly_2021}, as ghost or multi-reflection, using PointNet++.
A 4D \ac{RADAR} point cloud from multiple \acp{RADAR} has been explored by \cite{schumann_radarscenes_2021}, combining Cartesian coordinates, Doppler and \ac{RCS}, and processed with three PointNet++ models and a recurrent point classifier for instance point segmentation.
Instance segmentation has also been explored using \ac{HD} \ac{RADAR} with PointNet++ and an additional clustering method processing the semantic information of the point clouds \cite{liu_deep_2021}.
Recently, the kernel point convolution (KPConv) layer and a variant of the LSTM cell for temporal dimension processing have been proposed by \cite{nobis_kernel_2021} for semantic segmentation of sparse \ac{RADAR} point clouds.
\\

\ac{RADAR} semantic segmentation is an interesting approach for scene understanding as detailed in Section \ref{sec:mvrss}. However, many situations are difficult to understand using \ac{RADAR} alone because of its low angular resolution and the difficulties of human interpretation. The following sections will review existing works on sensor fusion using \ac{RADAR} with camera and, or, \ac{LiDAR} providing improvements in many tasks.

\section{Sensor fusion}
\label{sec:related_fusion}

\subsection{RADAR and camera fusion}

The camera is a passive sensor that provides a dense and comprehensive representation. It has been usually preferred to be fused with the \ac{RADAR} sensor as they are complementary: they respectively provide dense semantic information and the location and Doppler of the objects. Camera and \ac{RADAR} fusion methods commonly use sparse \ac{RADAR} point clouds. The following related works are differentiated among three fusion approaches: early (at the data level, the input of the network), mid-level (at any feature level of the network) and late (at the prediction level, or last layer of the network). In the presented methods, the \ac{RADAR} data have been used to improve performances in image-based tasks.

\paragraph{Early fusion.}
Early fusion methods project \ac{RADAR} point clouds in the image using the calibration of the camera.
The precursor work of \cite{garcia_data_2012} was not a machine learning method, but the authors projected a sparse \ac{RADAR} point cloud in the camera image to perform block matching and track objects according to their Doppler information.
The work of \cite{kowol_yodar_2021} expended the \ac{RADAR} points on the entire height of the image, w.r.t. the poor \ac{RADAR} elevation resolution, and used it as input of a YOLO model to perform 2D object detection.
Image depth estimation has been explore by \cite{long_radar-camera_2021} with a two-stage approach. The authors trained a network to associate \ac{RADAR} points to the objects in the image with a confidence area and a second model predicts the depth map with the fused camera and \ac{RADAR} points joined with the confidence areas.
In their work, \cite{long_full-velocity_2021} explored full velocity estimation of sparse \ac{RADAR} points by mapping a \ac{RADAR} point to an object in an image, as in the previous method, and deducing the closed form of the velocity vector of the point using the optical flow of the object. 
The \ac{RADAR} point coordinates have been projected in the image plan and fused with the image to learn an association between feature vectors of the \ac{RADAR} points and the corresponding object bounding boxes \cite{dong_radar_2021}.
\ac{RADAR} point clouds have been aligned with detected objects using a two-path Faster R-CNN for 2D object detection \cite{liu_robust_2021}.

%
%
%
%

\paragraph{Mid-level fusion.}
Mid-level fusion methods process the image and the \ac{RADAR} data with independent neural network backbones and merge their feature maps in a common latent space.
In the work of \cite{chadwick_distant_2019}, \ac{RADAR} points have been projected in the image plan, two ResNet branches process the camera image and the \ac{RADAR} point cloud image producing feature maps which are fused and used as input of detection heads similarly to the SSD architecture improving long range detection.
In their work, \cite{nobis_deep_2019} used a similar point cloud processing but the two \ac{CNN} branches are linked with intermediate connections at each stage, a final \ac{FPN} detects 2D objects while an additional module focuses the weights on a certain sensor.
In the same manner, \cite{yadav_radarrgb_2020} fused the features of the two branches in a mid-level stage of an \ac{FPN} for 2D object detection.
Feature maps have been fused by considering an SSD architecture with \ac{RADAR} convolutional attention modules balancing the sensors and improving detection of small objects \cite{bai_vehicle_2020}.
The work of \cite{meyer_deep_2019} used \ac{HD} \ac{RADAR} point clouds with the camera image as the input of an AVOD architecture \cite{ku_joint_2018} producing representations of each modality with 3D anchors and 3D proposals from the \ac{RPN}. The authors showed better performances than the same pipeline using \ac{LiDAR} point cloud combined with the camera image.
In their work, \cite{nabati_radar-camera_2020} detected 3D objects by processing \ac{RADAR} point clouds in the image (early fusion) with convolutions to detect objects and an additional refinement block takes the \ac{RADAR} point cloud to refine proposal coordinates.
In the work of \cite{niesen_camera-radar_2020}, the entire \ac{RA} view has been processed with the image considering individual branches. Their feature maps have been fused and up-sampled with a decoder for 3D depth reconstruction.
The \ac{RADAR} point cloud has also been projected in the image plane with Gaussian kernels creating a range density map \cite{cheng_robust_2021} and processed as an image. The authors proposed a vision-\ac{RADAR} fusion based on self-attention and global-attention layers used as input of an \ac{FPN} for 2D object detection. This method showed improvements in small and long range object detection.
\ac{HD} \ac{RADAR} point clouds have been considered in a \ac{BEV} and front view representations, processed with individual backbones \cite{cui_3d_2021}. Their feature maps are fused for 3D object detection and tracking is performed using a Kalman filter.
In the work of \cite{zhang_rvdet_2021}, \ac{RADAR} point clouds have also been used to predict \ac{BEV} occupancy grid. The authors have projected images from multiple cameras in a \ac{BEV} representation with a specific neural network and its feature maps from both branches have been fused for 2D object detection.
Traffic flow has been estimated in the work of \cite{jin_traffic_2021}, \textit{i.e.} consisting in 3D bounding boxes, speed and traffic density estimation. The authors used a point pillar expansion on \ac{RADAR} point clouds and fused them with image features. They also used a CycleGAN \cite{zhu_unpaired_2017} generating image data with different lighting conditions showing the limitations of image-based methods.
Depth estimation has been explored using a self-supervised learning scheme by expanding \ac{RADAR} point cloud for weak supervision and image reconstruction \cite{gasperini_r4dyn_2021}.

\paragraph{Late fusion.}
These methods consist in fusing feature maps generated by individual backbones at the very end of the pipeline, before the predictions.
In their work, \cite{lekic_automotive_2019} used Cartesian \ac{RA} views as input of \acp{cGAN} \cite{isola_image--image_2017}. Their outputs are fused lately to generate free space segmentation in camera images from \ac{RADAR} occupancy grid representation and segmented images.
\ac{RADAR} point clouds have been used to generate 2D proposals for 2D object detection (mid-level fusion), each modality having its own network. Their features are finally fused before the bounding box predictions \cite{shuai_millieye_2021}.
The work of \cite{kim_grif_2020} used individual backbones to process images and \ac{RADAR} voxel grids creating features maps which are fused and used as input of a 3D \ac{RPN} (mid-level fusion). The feature maps are then transmitted to the detection head with a gated \ac{RoI} fusion module. They showed better performances in middle and long range detection than with \ac{LiDAR} point clouds.
Similarly, \cite{kim_low-level_2020} generated features maps from images and \ac{RA} with independent backbones as input of a 3D \ac{RPN} (mid-level fusion). The Cartesian \ac{RA} are also used at the detection head level.
In their work, \cite{kuang_multi-modality_2020} fused feature maps from each bounding box and the \ac{RADAR} points to create an affinity matrix with their features and proposed a dynamic coordinates alignment method to refined the detection predictions.
In the work of \cite{li_feature_2020}, the \ac{RADAR} points have been projected in the camera plan and expanded with pillars and circles. Then, the authors used a YOLOv3 \cite{redmon_yolov3_2018} with an \ac{FPN} attention to fuse the \ac{RADAR} image with the feature maps at different scales.
\ac{RADAR} point clouds have been projected in the image plan and processed with convolutions in the work of \cite{hussain_rvmde_2021}. They processed the image in parallel with an \ac{FPN} and fused their feature maps which are up-sampled for depth map estimation.
In their work, \cite{lo_depth_2021} proposed a similar approach projecting the \ac{RADAR} point clouds in the image plan to create a \ac{RADAR} depth map; then, separated backbones generate representations of the modalities which are fused before an ordinal regression \cite{fu_deep_2018} for depth map estimation.
A two-stage method for depth estimation has also been proposed by \cite{lin_depth_2020}. A first \ac{CNN} generates a coarse depth map to filter \ac{RADAR} outliers using images with project \ac{RADAR} points. In the second stage, the coarse map, the original image and the filtered image with \ac{RADAR} points are processed with a second network predicting the dense depth map.
In the work of \cite{nabati_centerfusion_2021}, expanded \ac{RADAR} points in pillars have been associated to object bounding boxes. Then, the modalities are processed by individual backbones and their feature maps are stacked for 3D bounding box and velocity estimation.
Fusing camera and \ac{RADAR} representations have increased performances in vision-based tasks. However the fusion is not easy since the modalities are not recorded in the same space.
The following section will detail methods fusing \ac{RADAR} and \ac{LiDAR} sensors.

\subsection{RADAR and LiDAR fusion}

The \ac{LiDAR} sensor provides dense 3D point clouds of a scene in polar coordinates. 
However, it does not provide the object velocity and its point cloud is dense only at short range. These limitations are overcome by using a \ac{RADAR} sensor recording Doppler information with a longer maximum range capability.
\ac{RADAR} and \ac{LiDAR} fusion has not been deeply explored due to the lack of open source datasets.
The related work is generally based on early fusion since both representation can be projected in a Cartesian space.

The work of \cite{weston_probably_2019} proposed an inverse sensor model (ISM) converting a Cartesian \ac{BEV} of the \acl{RA} \ac{RADAR} view into an occupancy probability grid based on its uncertainty. The model is trained with self-supervised learning based on an occupancy grid created from the \ac{LiDAR} point cloud.

In their work, \cite{shah_liranet_2020} suggested a pipeline processing a HD map, \ac{LiDAR} and \ac{RADAR} point clouds in \ac{BEV} representations with individual backbones 
producing feature maps which are fused for 2D object detection and trajectory prediction.
The \ac{RADAR} backbone is composed of a spatial network extracting sparse local information using parametric continuous convolutions, and a temporal network using \acs{MLP} to combine the spatial features in the temporal axis acting as an attention layer.

A voxel-based early fusion for better long range detection has also been explored by \cite{yang_radarnet_2020} and combined with an attention-based late fusion to perform 3D object detection and trajectory estimation using two detection heads from a PnPNet architecture \cite{liang_pnpnet_2020}.

\ac{RADAR} \acl{BEV} \ac{RA} has been fused with \ac{LiDAR} at the early stage of the network to predict \ac{RADAR} occupancy grid with a U-Net architecture using \ac{LiDAR} supervision \cite{kung_radar_2021}. The authors proposed a sliding window approach to train the network at short range and demonstrated generalization outside the \ac{LiDAR} sensing range using the \ac{RADAR} only to predict the occupancy grid.

The work of \cite{qian_robust_2021} processed \ac{RADAR} \ac{BEV} \ac{RA} and \ac{LiDAR} \ac{BEV} map, corrupted with simulated fog, with independent U-Net backbones. An \ac{RPN} extracts proposals from their fused feature maps (mid-level fusion). Each modality has an \ac{RoI} pooler producing a feature vector from each proposal which is fused with self-attention and cross-attention (late fusion) for 2D object detection.

Realistic Cartesian \ac{RADAR} scenes have been generated by improving the elevation of \ac{RADAR} points using an adversarial approach with simulated elevation and real \ac{LiDAR} elevation considered as ground truth \cite{weston_there_2021}.

Methods exploiting combined \ac{RADAR} and \ac{LiDAR} data without deep learning have recently emerged. 
In the work of \cite{farag_real-time_2021} and \cite{liu_surrounding_2021}, authors proposed to detect objects with clustering in \ac{LiDAR} and \ac{RADAR} representations independently. Their features are fused and the objects are tracked using a custom Kalman filter algorithm while estimating their velocity individually. \ac{RADAR} and \ac{LiDAR} data have been combined to detect surfaces in adverse weather conditions highlighting robustness against fog \cite{wallace_combining_2021}.

Fusion of \ac{RADAR} and \ac{LiDAR} data in Cartesian coordinates benefits from the density of the \ac{LiDAR} and the long range with Doppler of the \ac{RADAR}. It is still at the early stage, fusion methods and adapted neural network architectures will be explored in the future. In Section \ref{sec:sensor_fusion}, we will introduce a fusion method propagating \ac{RADAR} information through the \ac{LiDAR} point cloud. The following section will detail methods for \ac{RADAR}, camera and \ac{LiDAR} sensor fusion.

\subsection{RADAR, camera and LiDAR fusion}
Combining \ac{RADAR}, camera and \ac{LiDAR} requires to project the data or their representations in a common space leading to complicated pipelines to perform end-to-end learning. 

The work of \cite{bijelic_seeing_2020} projected \ac{LiDAR} and \ac{RADAR} point clouds in a range view, \textit{i.e.} the image view, and extended the \ac{RADAR} points on the entire height of the image. The entropy of each modality is estimated and used as input of the \ac{CNN} backbone corresponding to its modality. The entire pipeline is similar to SSD \cite{liu_ssd_2016} with additional connections linking each backbone and heads for 2D detection in the camera (details about the SSD architecture are provided in Section \ref{sec:background_detection}).

The architecture of \cite{shah_liranet_2020} introduced in the previous section has been modified by including camera images and feature gaits simulating sensor dropout to perform 3D object detection and trajectory estimation \cite{mohta_investigating_2020}. It uses a MultiXNet \cite{djuric_multixnet_2021} feature extractor specialized in multi-modality processing. The authors showed that their method reduces sensor over-fitting while improving generalization when a sensor is missing.

Partial optical flow has been explored by \cite{grimm_warping_2020} considering camera to localize objects, \ac{LiDAR} as a label for depth estimation and \ac{RD} \ac{RADAR} view to estimate velocity vectors pixel-wise. They proposed to learn a module warping the segmented \ac{RD} map into the image domain and to use the depth map from the \ac{LiDAR} point cloud to supervise the training.

In their work, \cite{nobis_radar_2021} projected image pixels in the \ac{LiDAR} point cloud and fused \ac{RADAR} and \ac{LiDAR} point clouds in voxels for 2D object detection. Features are extracted using a VoxelNet \cite{zhou_voxelnet_2018}, processed with 3D sparse sub-manifold convolutions \cite{graham_3d_2018} and projected in a dense \ac{BEV} representation to apply standard convolutions for the detection heads.

Similarly, \cite{wang_high_2020} associated an image pixel to each \ac{LiDAR} point with the temporal dimension creating a first 7D point cloud, and grouped \ac{RADAR} features to obtain a 8D \ac{RADAR} point cloud. Both point clouds are processed separately with the frustum PointNets framework \cite{qi_frustum_2018} and their features are lately used for 3D object detection and velocity estimation.

\ac{RADAR}, camera and \ac{LiDAR} fusion is still at its early stage, there is neither clear common latent representation to fuse all the representations nor relevant backbone extracting features from the three sensors simultaneously.

\section{Conclusions}
In recent years, deep learning algorithms applied to autonomous driving have left out the \ac{RADAR} sensor for scene understanding. The nuScenes dataset has been the first to propose sparse \ac{RADAR} point clouds without annotation. Since then, datasets were released including various scenes, annotations and sensors (see Table \ref{tab:related_public_dataset}). The Chapter \ref{chap:datasets} will discuss methods to tackle the lack of \ac{RADAR} data for scene understanding. In particular, Section \ref{sec:carrada} will detail our proposed CARRADA dataset and a semi-automatic pipeline to generate its annotations. Result of a recent collaboration, our RADIal dataset including raw \ac{HD} \ac{RADAR} data is presented in Chapter \ref{chap:hd_radar}. The two proposed datasets are unique in their kind; they are publicly available to the machine learning community to support research in scene understanding. 

Deep learning for \ac{RADAR} scene understanding has been opened up with the release of large scale datasets. \ac{RADAR} semantic segmentation has not been widely explored but it is the most suited task regarding the objects' signature in the \ac{RADAR} representation.
Moreover, exploiting raw \ac{RADAR} data is important since the pre-processing steps reduce information and miss small object signatures. The \acl{RAD} tensor is noisy and cumbersome and thus should be aggregated in views. 
To the best of our knowledge, there is no related work on \ac{RAD} tensor segmentation for scene understanding.
In Section \ref{sec:mvrss}, we propose the first approach segmenting multiple views of the \ac{RAD} tensor simultaneously outperforming competitive methods while requiring significantly less parameters.

To the best of our knowledge, there is no previous work on end-to-end object detection that is capable to scale with raw \ac{HD} \ac{RADAR} data. Moreover, there is no previous work either on free driving space segmentation\footnote{The free driving space segmentation task consists to locate pixel-wise the available space that can be driven. Further details are provided in Chapter \ref{chap:hd_radar}.} or semantic segmentation using only \ac{RD} views of \ac{HD} \ac{RADAR} signals. In addition, there is no existing multi-task model that performs both \ac{RADAR} object detection and semantic segmentation simultaneously.
In Section \ref{chap:hd_radar}, we propose a deep neural network architecture learning the costly pre-processing steps and performing multi-task learning using raw \ac{HD} \ac{RADAR} data: 2D object detection and free space segmentation simultaneously.

Sensor fusion including \ac{RADAR} has been recently explored. It has shown improvements in long range detection, small object detection and global performances in adverse weather conditions.
\ac{LiDAR} and \ac{RADAR} fusion has not been extensively investigated although the two type of sensors have complementary properties and they can be both represented in a Cartesian space.
In Section \ref{sec:sensor_fusion}, we propose a preliminary sensor fusion approach aiming to propagate \ac{RADAR} information through a dense \ac{LiDAR} point cloud.

\chapter{Proposed automotive RADAR datasets}
\label{chap:datasets}
\minitoc

\ac{RADAR} sensors generate electromagnetic wave signals that are not affected by weather conditions or darkness. These sensors inform not only about the 3D position of other objects, as \ac{LiDAR}, but also about their relative speed (radial velocity).   
However, in comparison to other sensory data, \ac{RADAR} signals are difficult to interpret, very noisy and cumbersome while having a low angular resolution. 
\ac{RADAR} has been left behind \acp{LiDAR} and cameras for these reasons and thus, has not been integrated in the data recording campaigns of large automotive datasets.

As detailed in Section \ref{sec:related_datasets}, tackling the lack of open source datasets has been a challenge in recent years. In this chapter, a simple \ac{RADAR} simulation is proposed in Section \ref{sec:rd_simulation}, a method to generate \ac{RADAR} data is detailed in Section \ref{sec:radar_generation}; and a novel dataset with synchronised camera and \ac{RADAR} data and a semi-automatic algorithm generating its annotations is described in Section \ref{sec:carrada}.

Section \ref{sec:carrada} presents a work mainly inspired from our article published at the International Conference on Pattern Recognition (ICPR) \cite{ouaknine_carrada_2020}.

\section{RADAR simulation}
\label{sec:rd_simulation}

At the beginning of 2019, there was no open source \ac{RADAR} dataset for automotive scene understanding. A costly and time consuming solution is to record and annotate a dataset. 
A second one is to consider a simple simulation of the behavior of a \ac{RADAR} sensor. The simulation has two major benefits: it is an unlimited source of data and fine-grained annotations are available without any cost.
In this first approach, we propose a simple simulation of the \ac{RADAR} sensor based on geometric considerations. Starting from a \ac{RADAR} position and a moving object and knowing its size and trajectory, we can deduce its distance to the sensor and its relative velocity and thus predict its approximated signature in the \ac{RD} data.
We have created a simulated dataset named RadarSim to train a classifier to distinguish \ac{RD} sequences between four categories of objects. Object classification experiments have been conducted using deep neural network architectures.

\subsection{Parameters and properties}

The simulation starts from a class of objects (among 4 categories) and a straight line trajectory in a delimited 2D space. Each object is represented by a square whose size depends on its category.
The real velocity and the angle of the object direction define a vector in polar coordinates which will determine the trajectory of the object.
The category also defines the range of values for the trajectory parameters (real velocity and angle).

The object is moving in front of a \ac{RADAR}, the sensor is considered as a single stationary point. The transmitted wave and its reflection on the object are not modeled. Instead, several points belonging to the reflector are selected and the simulation computes their relative velocity and distance w.r.t. the \ac{RADAR} position. 

Four categories are supported by the simulation: pedestrian, motorcycle / bicycle (same category), car and truck / bus (same category). An integer value is associated to each category, ordered in order of the size of the object. The category defines the size of the objects, the higher the larger. For each one of them, the size parameter is expressed in meters and is randomly drawn as $l \sim \mathcal{U}([l_{\text{min}}, l_{\text{max}}])$. The range of values depends on the category, see Table \ref{tab:simulation_parms}. Each object has a real velocity expressed in meters per second ($m \cdot s^{-1}$). Its value is randomly drawn as $v \sim \mathcal{U}([v_{\text{min}}, v_{\text{max}}])$, where the estimated range of real world velocities of the objects also reported in Table \ref{tab:simulation_parms}. The time interval between two frames is fixed to $\Delta t = 0.1$ second. 

While the object is moving, a vector (angle and length) defines its direction. The angle of direction of the object is randomly drawn as $\alpha \sim \mathcal{U}([0, 2 \pi])$ while its length depends on the velocity previously defined. Depending on this value, the object will start to move in one of the four regions of the space. These splits are made to ensure that the object will not cross the boundaries of the space.
Only a few points belonging to the object are estimated to create the signature. For each one of them, the intensity (in dB) of the reflection is estimated according to real observations and drawn as $I \sim \mathcal{U}([I_{\text{min}}, I_{\text{max}}])$. The ranges of intensity values are defined in Table \ref{tab:simulation_parms}.

\begin{table}
    \begin{center}
    \begin{tabular}{l | c c c}
        \toprule
         Category & Size $l$ (m) & Velocity $v$ ($m \cdot s^{-1}$) & Intensity $I$ (dB)\\
         \midrule
         0: Pedestrian &  $[0.1; 0.5]$ & $[1.0; 3.0]$ & $[50.0; 65.5]$\\
         1: Motorcycle / Bicycle &  $[0.5; 2.0]$ & $[3.0; 11.0]$ & $[50.0; 80.0]$\\
         2: Car &  $[2.0; 4.0]$ & $[4.0; 14.0]$ & $[65.5; 80.0]$\\
         3: Truck / Bus & $[4.0; 10.0]$ & $[4.0; 14.0]$ & $[65.5; 80.0]$\\
         \bottomrule
    \end{tabular}
    \end{center}
    \caption[Range of values for the simulated RADAR points properties]{\textbf{Range of values for the simulated RADAR points properties}. An object is defined by its category, size (in meter), its velocity (in meter per second) and the estimated intensity of its reflected signal (in decibel). }
    \label{tab:simulation_parms}
\end{table}

The \ac{RD} representation contains a high level of noise divided into three forms when considering averaged and log-transformed data (see Section \ref{sec:canada_rd}): speckle noise, uniform noise and zero-Doppler noise. 
The logarithmic transformation aims to transform the speckle from a multiplicative to an additive noise as $I = u {\times} s$, where $I$ is the intensity, $u$ the underlined reflectivity and $s$ the speckle as detailed in Section \ref{sec:background_speckle}.
The simulation considers log-intensity values and thus includes the speckle noise as a global additive noise. 
We have estimated the Fisher-Tippett distribution of the background (combining the reflectivity of the background and the speckle) from real log-transformed data \cite{goodman_speckle_2007}. It can be approximated by a Gaussian distribution as illustrated in Figure \ref{fig:canada_speckle} since the data have been multi-looked.
Regarding our estimations, the simulation will use a random variable $s \sim \mathcal{N}(35.473,\,\sqrt{0.244})$ to generate the background of the representation, including the speckle noise. 
%
We also introduce a uniform noise to mimic reflections of false alarms that can occur in recordings (ghost or multi-path reflections, ambiguity, and so on). 
It will activate each bin according to a random variable defined with a Bernoulli distribution $r \sim \operatorname{Bern} \left({0.01}\right)$. 
Finally, the zero-Doppler noise will contain all the reflections which have a zero Doppler value (stationary objects) at each range bin.
This phenomena is modeled on the zero Doppler bins with a random variable defined with a Bernoulli distribution defined as  $z = \operatorname{Bern} \left(p\right)$ at the $0$ range bin, where the probability parameter $p$ decreases from 1 by a factor $0.005$ for each range bin.
For a single bin of the representation, the simulated background including the noise is thus a random variable written $n = s + r + z \cdot \mathbbm{1}_{\text{Doppler}=0} $ which is added to the simulated log-intensity.
The following section will present the entire pipeline of simulation and the RadarSim dataset.

\subsection{RadarSim dataset}
\label{radarsim}

\begin{figure}[!t]
  \begin{center}
    \includegraphics[width=0.9\textwidth]{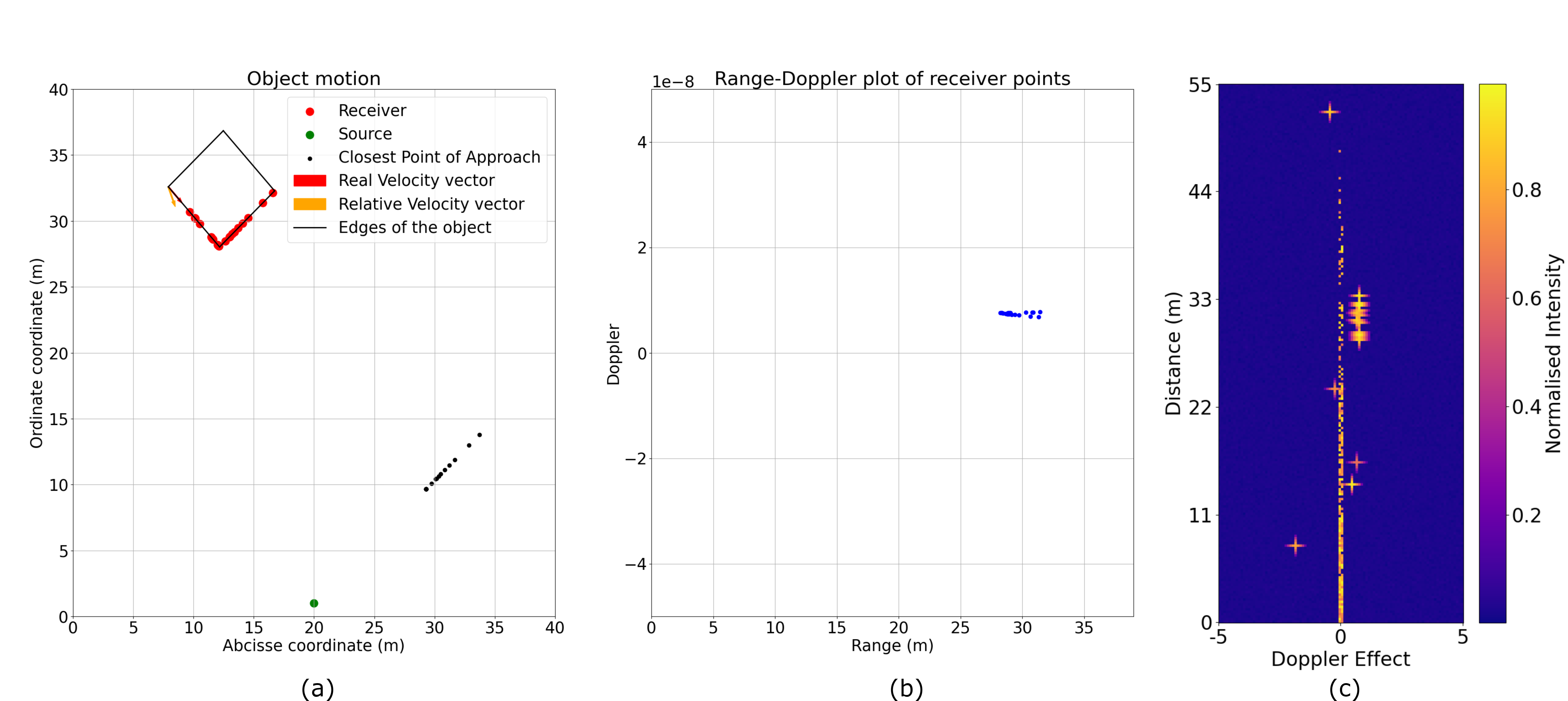}
  \end{center}
  \caption[Example of a Range-Doppler simulation for the category `Car']{\textbf{Example of a \acl{RD} simulation for the category `Car'}. (a) Simulation of the objects motion in a Cartesian space with a constant velocity vector and a fixed \ac{RADAR}. The closest position of each sampled point w.r.t. radar position and the velocity of the object is noted the closest point of approach. (b) Range-Doppler points of the simulated reflections on the object. (c) Range-Doppler representation with the simulated object signature and additional noise phenomena.}
  \label{fig:ex_rd_simulation}
\end{figure}

The dataset simulation starts by defining the object: randomly drawing its category, real velocity, direction and its starting point in the space. A sequence of twenty \ac{RD} frames is generated (equivalent to two seconds of recordings). Each one of them will be a matrix of size $220 {\times} 100$. An example of a frame from a generated scene is shown in Figure \ref{fig:ex_rd_simulation}. The following process is executed at each timestamp:
\begin{itemize}
    \item The range-Doppler matrix is created by discretizing the estimated values: the range is defined between $0$ and $55$ meters with a resolution of $0.25$ meters per bin; the radial velocity (Doppler) is defined between $-5$ and $5$ meters per second with a resolution of $0.1$ per bin. The matrix is initialized with the estimated background and the randomly generated noise.
    \item Random points are selected on the visible edges of the moving object that are visible to the \ac{RADAR}.
    \item The radial velocity component w.r.t. the \ac{RADAR} is deduced from the real velocity vector for each point as well as their distance.
    \item For each point, an intensity value is randomly drawn depending on the category of the object. An additional \textit{sinc} effect is considered to simulate the shape of the object reflection in the \ac{RADAR} representation. The cardinal sinus function is defined as: $\operatorname{sinc}(x) = \frac{\sin(\pi x)}{\pi x}$.
\end{itemize}
The simulation process generating a sequence of \acp{RD} is detailed in Algorithm \ref{alg:simulation}.

\begin{algorithm}[!t]
\setstretch{1.35}
\caption{Simulation of a sequence of Range-Doppler (RD)}
\label{alg:simulation}
\begin{algorithmic}[1]
\REQUIRE $T, e \in \mathbb{N}$, $(x_{\text{RADAR}}, y_{\text{RADAR}}) \in \mathbb{R}^2$, $d_\text{max}, r_\text{max} \in \mathbb{R}$
\\ \COMMENT{$T$ the number of frames in the sequence, $e$ the number of bins affected by the $sinc$ function, $(x_{\text{RADAR}}, y_{\text{RADAR}})$ the coordinates of the RADAR sensor, $d_\text{max}$ and $r_\text{max}$ the maximum Doppler and distance.}

\STATE $p = 1$ \COMMENT{Initialize the parameter of the random variable $z$.}
\STATE $c \in \left \{0, 1, 2, 3 \right \}$ \COMMENT{Set the class of the object.}
\STATE $x \sim \mathcal{U}([x_{\text{min}}, x_{\text{max}}])$ \COMMENT{Init. $x$ coord. of the object in a bounded Cartesian space.}
\STATE $y \sim \mathcal{U}([y_{\text{min}}, y_{\text{max}}])$ \COMMENT{Init. $y$ coord. of the object in a bounded Cartesian space.}
\STATE $l \sim \mathcal{U}([l_{\text{min}}, l_{\text{max}}])$ \COMMENT{Set $l$ the size of the object depending on $c$.}
\STATE $v \sim \mathcal{U(}[v_{\text{min}}, v_{\text{max}}])$ \COMMENT{Set $v$ the velocity of the object depending on $c$.}
\STATE $\alpha \sim \mathcal{U}([0, \pi]$ \COMMENT{Set $\alpha$ the angle of arrival of the object.}

\STATE $s \sim \mathcal{N}(35.473,\,\sqrt{0.244})$ \COMMENT{Random variable for the speckle noise.}
\STATE $r \sim \operatorname{Bern} \left(0.01\right)$ \COMMENT{Random variable for the false alarm reflection.}
\STATE $z \sim \operatorname{Bern} \left(p\right)$ \COMMENT{Random variable for the zero-Dopppler noise.}
\STATE $n = s + r + z \cdot \mathbbm{1}_{\text{Doppler}=0}$ \COMMENT{Global random variable for noise and background generation.}

\FOR[$T$ the length of the sequence to be generated.]{$t=0$ \TO $T-1$}

    \STATE $\vect{A}:$ $220 \times 100$ matrix filled with $n$ where $p$ decreases by $0.005$ for each range bin. 
    \STATE $K$: random points on the visible edges of the objects + number of false reflections.

    \FOR[$K$ the reflected points.]{$k=0$ \TO $K-1$}
        \STATE $I \sim \mathcal{U}([I_{\text{min}}, I_{\text{max}}])$ \COMMENT{Set $I$ the intensity of the reflection depending on $c$.}
        \STATE $d$: compute the radial velocity component of $k$
        \STATE $r = \sqrt{(x - x_{\text{RADAR}})^2 + (y - y_{\text{RADAR}})^2}$ \COMMENT{Compute the range of the point $k$}
        
        \STATE $\vect{A}_{\lceil \frac{d_\text{max}}{d} \rceil, \lceil \frac{r_\text{max}}{r} \rceil} \leftarrow \vect{A}_{\lceil \frac{d_\text{max}}{d}  \rceil, \lceil \frac{ r_\text{max}}{r} \rceil} + I$ \COMMENT{Add the reflected point $k$ to the RD.}
        
        \STATE $\vect{A}_{\lceil \frac{d_\text{max}}{d} \rceil \pm e, \lceil \frac{r_\text{max}}{r} \rceil \pm e} \leftarrow \operatorname{sinc} (\vect{A}_{\lceil \frac{d_\text{max}}{d} \rceil \pm e, \lceil \frac{r_\text{max}}{r} \rceil \pm e})$ \COMMENT{Apply $\operatorname{sinc}$ on $e$ bins around the reflection.}
        
        \STATE $x \leftarrow x + v_x \times \cos(\alpha)$ \COMMENT{Update $x$ w.r.t. the velocity and orientation of the object.}
        \STATE $y \leftarrow y + v_y \times \cos(\alpha)$ \COMMENT{Update $y$ w.r.t. the velocity and orientation of the object.}
    \ENDFOR
\ENDFOR

\end{algorithmic}
\end{algorithm}

The generated RadarSim dataset contains 1000 sequences with 20 frames for each scene. Each scene lasts 2 seconds, which means that an \ac{RD} is recorded every 0.1 second. 
The probability of randomly drawing a category is set to $0.25$ to obtain a well-balanced data set. 
The dataset is split into training ($60\%$), validation ($20\%$) and test ($20\%$). 
The classification task consists in categorizing sequences between four categories. Depending on the model used, it either classifies each frame of a sequence or directly classifies the entire sequence. Further details are provided in the following section.

\subsection{Experiments and results}
\label{radarsim_exp}

Each sequence of the RadarSim dataset represents the \ac{RD} signature of an object belonging to one of the four categories. The objective of the experiments is to classify each \ac{RD} matrix or each sequence into a category depending on the training approach. For this purpose, a classifier $f_{\theta}(.)$ parameterized by $\theta \in \mathbb{R}^d$ is trained. The complexity $d$ of the model depends on the chosen neural network architecture. The parameters are updated with the gradient based method Adam \cite{kingma_adam_2015} for a fixed learning rate value. The parameters $\theta$ of the function are initialized using the Xavier initialization method \cite{glorot_understanding_2010}. The \acl{CE} loss is optimized during training: 
\begin{equation}
    \displaystyle \min_{ \theta \in \mathbb{R}^d} \sum_{i=0}^{S-1} \vect{y}_i \log(f_\theta^*(\vect{x}_i)),
\end{equation}
where $S$ is the number of sequences, $\vect{y}_i$ the labels of the \ac{RD} sequence $\vect{x}_i$ and $f_\theta^{*}(.)$ the estimated model.
Performances are evaluated on the validation and the test datasets using the accuracy metric defined as: $\frac{\text{TP}}{\text{TP} + \text{FN}}$, where $\text{TP}$ is the number of True Positive predictions and $\text{FN}$ the number of False Negative predictions.

The performances are evaluated at the sequence level. A few architectures do not take into account the time dimension and thus predict a class for each frame of the sequence: this is the frame-based approach. During the training of the second approach, the loss is computed using the predictions and ground-truths of each frame. While during validation and testing, the accuracy is computed after a vote for each sequence regarding the predictions over its frames. For a given sequence $\vect{x}$, its attributed category is voted as following:
\begin{equation}
    \displaystyle \argmax_{k \in \{0, \hdots, 3\}} \sum_{i=0}^{N-1} \mathds{1}_{\{f_{\theta}^*(\vect{x}_i) = k\}}.
\end{equation}
The temporal information is incorporated by concatenating the frames of a sequence in the channel dimension to directly classify the sequence. The sequence-based approach consists in taking into account the time dimension and classifying the sequence with a single label.

Each architecture has been explored with several parameters including its number of layers. The highest performances for each architecture are reported in Table \ref{perf_radarsim} and discussed in the following section.

\begin{table}[t!]
\footnotesize
\begin{center}
\begin{tabular}{l c c c c}
    \toprule
    \multirow{2}{*}{Architecture} & \multicolumn{2}{c}{Approach} & \multicolumn{2}{c}{Accuracy}\\
    \cmidrule(lr{3pt}){2-3}
    \cmidrule(l{3pt}r{3pt}){4-5}
      & Frame & Sequence & Validation & Test \\
    ResNet-101 \cite{he_deep_2016} & \cmark & \xmark & 87.5 & 90.0 \\
    ResNet-101 \cite{he_deep_2016} & \xmark & \cmark & 83.0 & 85.0\\
    ResNet-34 \cite{he_deep_2016} + LSTM (3 cells) & \xmark & \cmark & 84.0 & 84.5\\
    3D Convolutions (2 layers + 1 FC) & \xmark & \cmark & 63.0 & 59.0\\
     \bottomrule
\end{tabular}
\end{center}
\caption[Performances on the RadarSim-Val and -Test datasets]{\textbf{Performances on the RadarSim-Val and -Test datasets}. Performances of the explored architectures on the validation and test splits of the RadarSim dataset.}
\label{perf_radarsim}
\end{table}

\subsection{Discussions}
\label{radarsim_discussions}

Results presented in Section \ref{radarsim_exp} show that the architectures used have succeeded in estimating the parameters of the simulation. The generated scenarios are not challenging enough and it is easy for the system to overfit the training dataset since the distributions between the splits are similar. 
We noticed that a light backbone with LSTM cells leads to similar results as those of extremely deep neural networks. Moreover, a model using 3D convolutions with only two layers has reached almost $60\%$ accuracy on the test dataset which is a promising result.

The wrong classifications show a common pattern regardless of their ground truth category: they cannot be classified by a human. Either because the object is not visible due to zero Doppler noise, or because only a small part of a large object is visible to the \ac{RADAR} and therefore poorly reflects the signal (a large object will be wrongly classified as a small one).

Real raw \ac{RADAR} data are necessary to take into account the entire complexity of the representation. They also should be annotated to train models in a supervised manner. To this end, the next section will describe methods to generate \ac{RD} with their entire complexity.

\section{RADAR data generation}
\label{sec:radar_generation}

This section presents methods to generate \acp{RD} from natural images (camera). A dataset of complex urban scenes with paired raw \ac{RADAR} data and natural images has been recorded and used for our experiments. However, the \ac{RADAR} data are not directly annotated due to the complexity of the representation. Because the objects in the raw \ac{RADAR} representations are difficult for a human to distinguish, manual annotation is expensive and of poor quality. In this work, we tried to approximate a transformation to map the natural image domain (source) to the \ac{RD} domain (target).
In other words, the model will learn to generate the signature of objects in the \ac{RADAR} representation by recognizing them in the camera image.
This transformation is learnt with deep neural networks by reconstructing the data of the target domain using the source domain as input. The motivation of this work is to find a common latent space between the two domains to transfer information from the source to the target. In this way, annotations in the natural images can be transferred to the raw \ac{RADAR} representations without human supervision.

\subsection{Dataset}
\label{sec:dataset_generation}

\begin{figure}[!t]
  \begin{center}
    \includegraphics[width=1\textwidth]{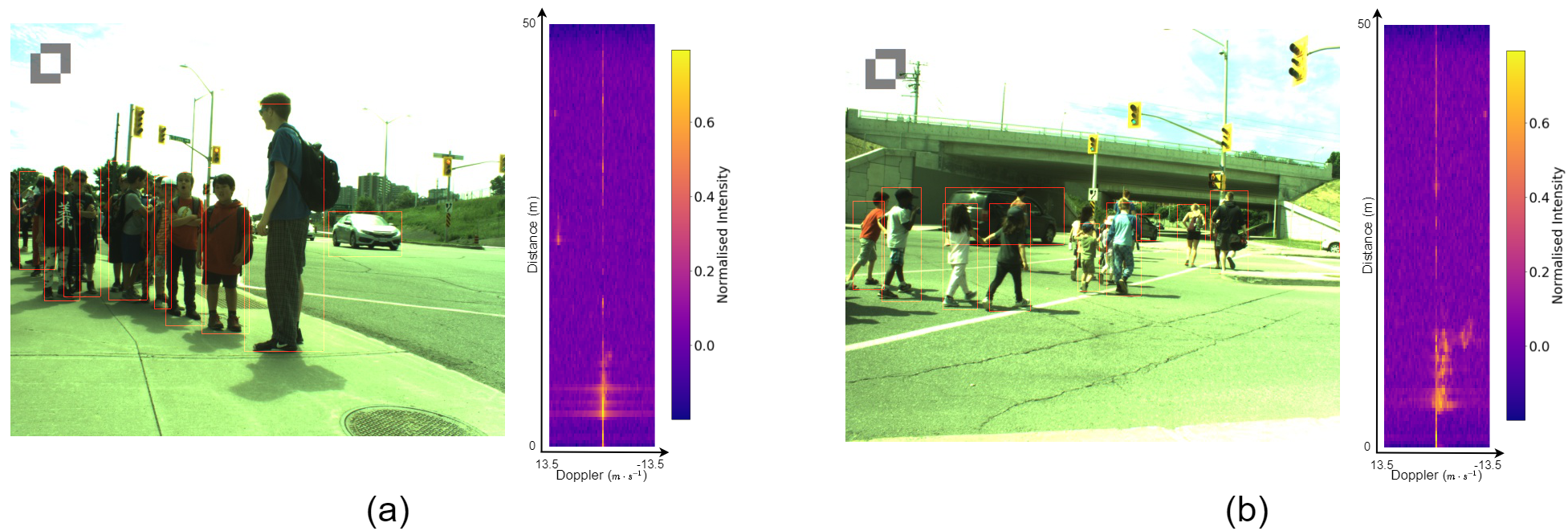}
  \end{center}
  \caption[Example of two scenes with pedestrians and moving cars]{\textbf{Example of two scenes with pedestrians and moving cars}. For both scenes (a) and (b); Left: natural image of the scene with bounding boxes around the pedestrian and the moving cars; Right: \acl{RD} representation of the scene without annotation.}
  \label{fig:canada_ex}
\end{figure}

The dataset was recorded in Canada in an urban environment. The acquisition setup consists of a 77GHz \ac{FMCW} \ac{RADAR} and a camera mounted on a stationary car. The \ac{RADAR} uses the \ac{MIMO} system configuration with 2 \ac{Tx} and 4 \ac{Rx} producing a total of $\NTx \cdot \NRx = 8$ virtual antennas. The dataset contains $26,085$ frames divided into 8 sequences for a total duration of 44 minutes. It consists of complex urban scenes with paired raw \ac{RADAR} and natural images recorded at 10 \ac{FPS}. 
In this work, we used the recorded natural images and \ac{RD} representations which have a resolution of $616 {\times} 514$ and $128 {\times} 64$ respectively. Bounding boxes annotations are provided for $\frac{1}{10}$ of the frames but only for the natural images. They are classified between two categories: pedestrian and vehicle. Figure \ref{fig:canada_ex} shows examples of scenes with pedestrians and cars at a crossroad.

The dataset is split between the training ($90\%$), the validation ($5\%$) and the test ($5\%$) datasets. The training split is deliberately large to be able to train deeper neural network architectures. The next section will provide details about the range-Doppler representation.

\subsection{Range-Doppler representation}
\label{sec:canada_rd}

The raw \ac{RADAR} data are recorded in a 3D tensor in the frequency domain which is processed by an inverse \ac{FFT} on each axis to obtain the \ac{RAD} tensor in the temporal domain (see Section \ref{sec:background_signal_process}). 
Let $\vect{X}^{\text{RAD}}$ be the complex \ac{RAD} tensor. The \ac{RD} is obtained by aggregating the tensor on the second axis following Equation \ref{eq:agg_method_average}. 
This method aims to reduce the global noise on the representation as presented in Section \ref{sec:background_signal_process}. 

The raw \ac{RADAR} data is prone to a high level of noise, partly due to the speckle noise as detailed in Section \ref{sec:background_speckle}.
It is a noise inherent to the \ac{RADAR} sensor covering the entire representation. While working on the reconstruction of the \ac{RD} representation, it is helpful to estimate the distribution of the noise that the model will need to overcome. For a given sequence, the speckle noise is extracted by selecting an image patch in the same spatial position, over $1,050$ frames. The patch has been chosen so that almost no signal appears in the area.

\begin{figure}[!htbp]
  \begin{center}
    \includegraphics[width=1\textwidth]{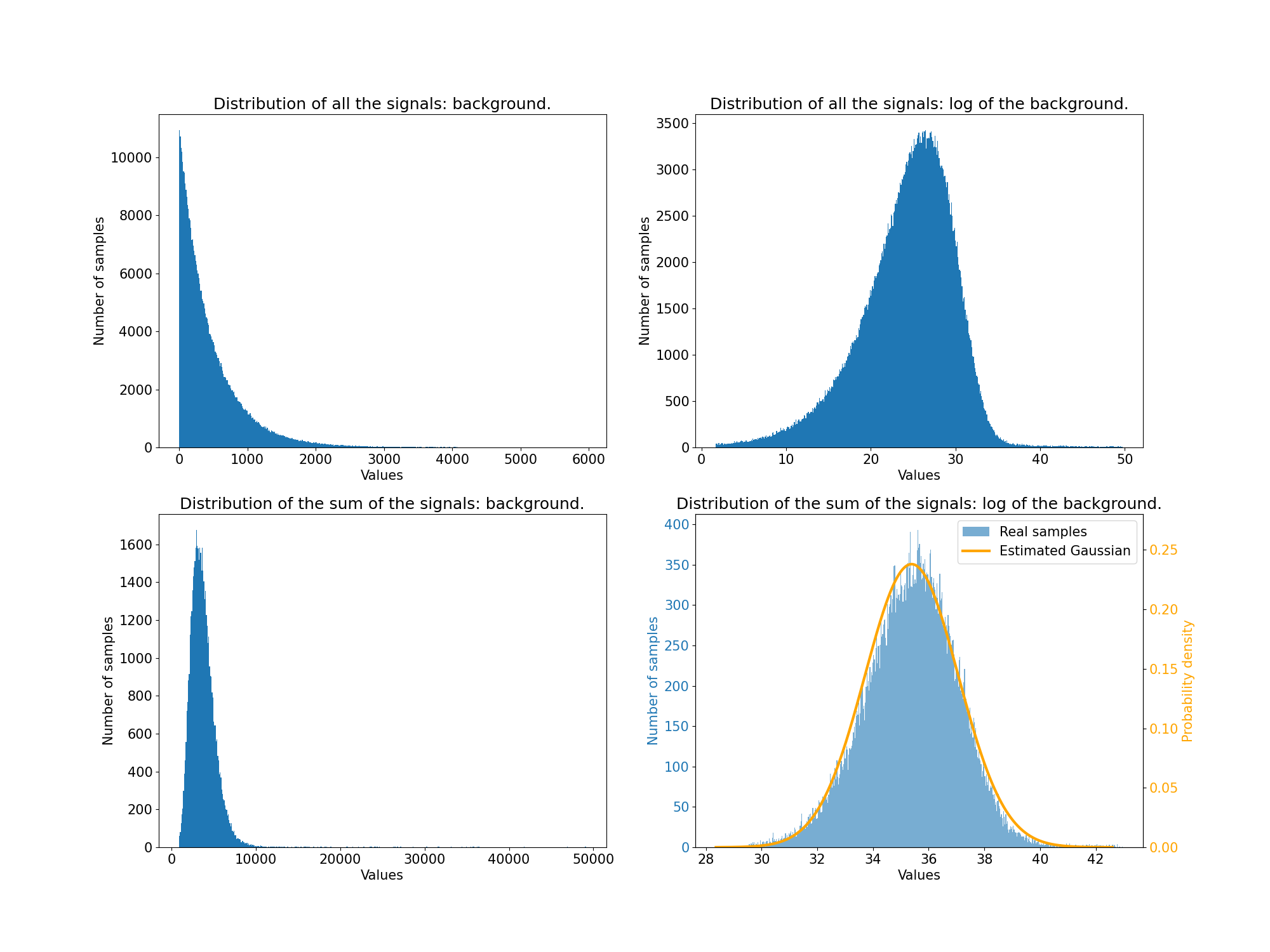}
  \end{center}
  \caption[Distributions of the background in RADAR data]{\textbf{Distributions of the background in RADAR data}. Distribution of the background for a given patch in a sequence before applying Equation \ref{eq:agg_method_average}. These sample distributions follow the model proposed by \cite{goodman_speckle_2007}; they are described in Section \ref{sec:background_speckle}. Top left: distribution of the signal values (decreasing exponential distribution). Top right: distribution of the logarithm of the signal values (Fisher-Tippett distribution). Bottom left: distribution of the signal values summed over the antennas axis (Gamma distribution). Bottom right: distribution of the signal values after applying Equation \ref{eq:agg_method_average} (blue) and distribution of random samples of an estimated Gaussian distribution (orange).}
  \label{fig:canada_speckle}
\end{figure}

Figure \ref{fig:canada_speckle} shows the distributions of the background with different methods to aggregate the data. 
The data follows the model proposed by \cite{goodman_speckle_2007}: the original signal has a decreasing exponential distribution and the log-signal has a Fisher-Tippett distribution.
The speckle aggregated in the third dimension of the tensor follows a Gamma distribution.
Finally, after applying Equation \ref{eq:agg_method_average} (averaging and log-transform), the speckled background follows a Fisher-Tippett distribution approximated by a Gaussian distribution estimated as $\mathcal{N}(35.38,\,\sqrt{2.82})$.
Details about the speckle and its distributions are provided in Section \ref{sec:background_speckle}.
The following section will explain the methods and experiments conducted on the presented dataset.

\subsection{Methods and Experiments}
\label{sec:canada_exp}

\begin{figure}[!t]
\centering
\includegraphics[width=1\textwidth]{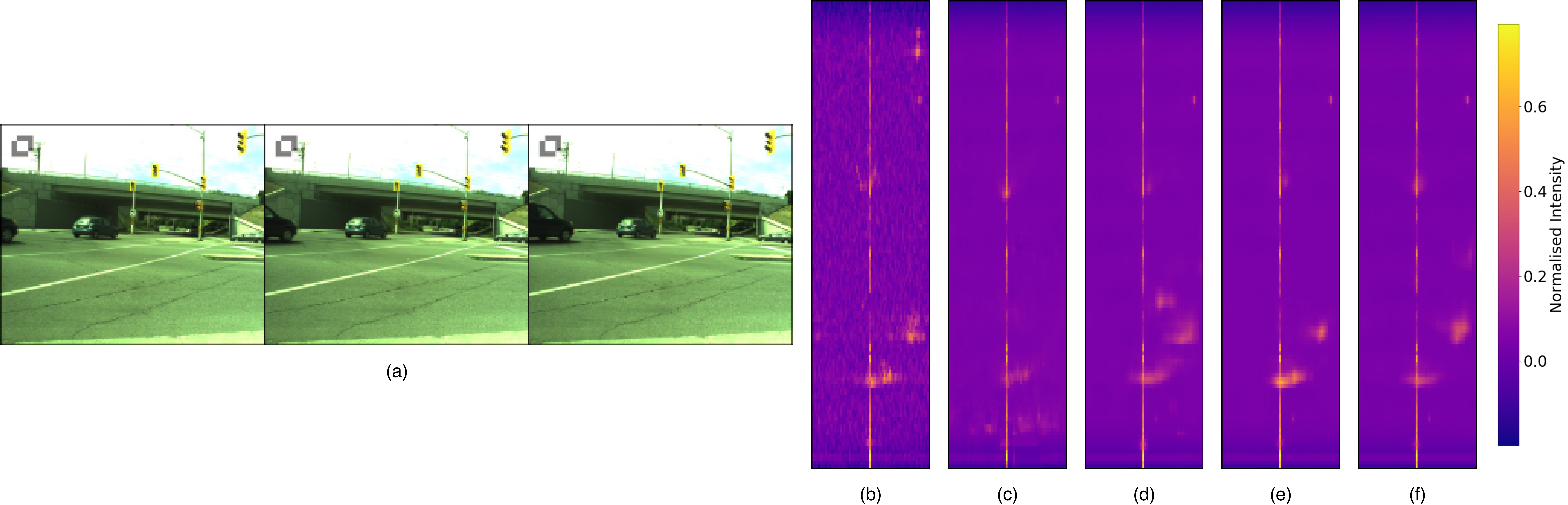}
\caption[Qualitative results of Range-Doppler generations]{\textbf{Qualitative results of \acl{RD} generations}. (a) Consecutive natural images images of a scene with (b) the \acl{RD} ground-truth associated to the image in the center. \acl{RD} generated from the sequence of natural images using: (c) VGG9, (d) \acl{AE}, (e) Light \acl{AE}, (f) \acl{AE} with noise map (ours).
}
\label{fig:rd_generation_quali}
\end{figure}

This section describes the methods we explored to generate \ac{RD} representations from natural images. 
We modified them to propose adapted architectures for \ac{RD} reconstruction from sequence of natural images.
The results of the experiments are presented in Table \ref{tab:perf_canada}. Qualitative results of \ac{RD} generation are compared in Figure \ref{fig:rd_generation_quali}.

The \ac{RD} representation contains both the distance and the relative velocity of the reflectors in a scene w.r.t. the \ac{RADAR}. At least two images are required to estimate the velocity of an object. The sequences are truncated to choose varying window sizes of natural images. Our experiments have shown that considering a short sequence of three consecutive images to generate an \ac{RD} leads to better performances without incurring a significant computational cost (especially when considering 3D convolutions). In the following experiments, the images of the short sequence are stacked in the depth dimension to train neural network architectures.

Let $\mathcal{S}$ be the input domain (any sequence of three consecutive natural images), and $\mathcal{R}$ the output domain of \ac{RD} representations. 
Our goal is to determine a function $f_\theta$, parameterized by $\theta$, which carries out this domain conversion, such as:
\begin{equation}\label{eq:im2rd}
    \begin{array}[t]{lrcl}
f_\theta : & \mathcal{S} & \longrightarrow & \mathcal{R} \\
    & \vect{x}_{s} & \longmapsto & \vect{x}_{r}. \end{array}
\end{equation}
Note that the dimension of $\vect{x}_s$ is much larger than $\vect{x}_r$ ($\mathcal{S} \gg \mathcal{R}$). In these experiments, $\vect{x}_r$ is reconstructed using $\vect{x}_s$.

During training, the \ac{MSE} is used as a loss function to reconstruct the \ac{RD} representation: $\text{MSE} = \frac{1}{\BinR \cdot \BinD} \left\lVert \vect{x}_{r} - \vect{x}_{r}^{*}\right\rVert_2 ^2$ where $\vect{x}_{r}^{*} = f_{\theta}^{*}(\vect{x}_s)$, $f_{\theta}^{*}$ is the approximated reconstruction function and $\BinR \cdot \BinD$ the number of elements in the \ac{RD} representation.
The \ac{MAE} has also been used as a loss function but its primary goal is to compare the obtained results as an evaluation metric.
It is written: $\text{MAE} = \frac{1}{\BinR \cdot \BinD} \left\lVert \vect{x}_{r} - \vect{x}_{r}^{*}\right\rVert_1$. 

Well-known deep neural network methods and architectures have been considered. We adapted them in our experiments for \ac{RD} reconstruction.
In particular, we explored stacked convolutions with 2D or 3D kernels, \acl{AE} architectures, radar-based prior information to train the network, specific loss functions and generative approaches. The following paragraphs describes these methods in details.

\paragraph{Stacked 2D convolutions.} This group of layers has been considered as a first approach while using down-samplings to reduce the size of the natural images and directly match the \ac{RD} dimensions, noted $(\BinR {\times} \BinD)$.
A final 1D convolution layer is used to reduce the number of feature maps and generate the \ac{RD} map. For this method, we adapted the VGG11 \cite{simonyan_very_2015} architecture in a VGG9 by replacing the last three fully-connected layers by a single 1D convolution layer to perform the reconstruction.

\paragraph{Stacked 3D convolutions.} 
The interest of using 3D convolutions introduced by \cite{ji_3d_2012} is to take into account the temporal patterns in the sequence of natural images used as input. Deep architectures are not feasible due to the high computational cost of 3D convolutions. We created a neural network architecture with 10 layers of 3D convolutions processing the sequence of images for \ac{RD} reconstruction. 

\paragraph{Convolutional \acl{AE}.} This method \cite{masci_stacked_2011} consists in processing the input data with 2D convolutions while deducing its dimensions with down-samplings.
The latent space of feature maps is then up-sampled to recover the output dimensions and reconstruct the \ac{RD} representation. A well-known extension of Convolutional \ac{AE} is U-Net \cite{ronneberger_u-net_2015}. Skip connections process the features learnt at different stages of the down-sampling pathway of the network and add the information to the paired stage of the up-sampling pathway. It helps to keep information even after reducing it into a low dimensional latent space. We modified the original U-Net architecture to match our input and output dimensions. 
The proposed \ac{AE} architecture has the same down-sampling and up-sampling architectures as U-Net with 15 convolution layers in total while the U-Net have additional skip connections. In total, the U-Net contains 19 convolutions with skip connections. We also proposed a lighter version of the \ac{AE} to reduce the training time (`Light-\ac{AE}'). Reducing the number of filters leads to small improvements with three times fewer parameters.

\paragraph{Additional Noise Map.} Our experiments have shown that the \ac{AE} always tries to estimate the speckle noise distribution to reconstruct the \ac{RD} representation. To leverage this observation, we proposed to stack the input data with a noise map randomly generated from the estimated background distribution detailed in Section \ref{sec:canada_rd}. 
The noise map has the same dimensions as the input images. The noise map and the images are stacked in the channel dimension to provide prior information to the network on the speckle distribution.
In the presented experiments, our proposed ``Noise Map'' approach leads to a significant gain of time during training to reach similar performances. It also improves results for a similar training time.

\paragraph{Custom Losses.} Reflectors in a scene are represented by high intensity values in the \ac{RD} representation. Previous methods have shown that a focus is made on the noise reconstruction while the shape of the object signature is difficult to recover. Considering the \ac{MSE} as a loss leads to learn a global average of the intensity values and thus prioritizes the speckle noise. The reconstruction of the object's signature is important because the final objective of this work is to transfer object annotations from a domain to another. To focus the reconstruction on high intensity values, we have used a \ac{wMAE}. It considers the ground-truth intensity value as a weight which is applied pixel-wise in the \ac{MAE} loss. The \ac{wMAE} loss is defined as:
\begin{equation}
    \text{\ac{wMAE}} = \frac{1}{\BinR \cdot \BinD}\sum_{n=0}^{\BinR \cdot \BinD-1} \vect{x}_{r, n} \left\lVert \vect{x}_{r, n} - \vect{x}_{r, n}^*\right\rVert_1,
\end{equation}
where $\vect{x}_{r, n}$ is the $n$-th element of the ground-truth \ac{RD} representation in the $\mathcal{R}$ domain and $\vect{x}_{r}^{*} = f_{\theta}^{*}(\vect{x}_s)$ with $f_{\theta}^{*}$ the approximated reconstruction function.

To both take into account the high intensity value penalization and the global noise estimation, the \ac{wMAE}\_\ac{MSE} loss has been experimented which combines both the \ac{wMAE} and the \ac{MSE} losses. 
The two terms are weighted by positive hyper-parameters $\lambda_1$ and $\lambda_2$. These values have not been explored yet, they are fixed to 1 in the experiments. The loss is defined as:

\begin{equation}
    \text{\ac{wMAE}\_\ac{MSE}} = \frac{1}{\BinR \cdot \BinD} \biggl[\lambda_1 \sum_{n=0}^{\BinR \cdot \BinD-1} \vect{x}_{r, n} \left\lVert \vect{x}_{r, n} - \vect{x}_{r, n}^*\right\rVert_1 + \lambda_2 \left\lVert \vect{x}_{r} - \vect{x}_{r}^*\right\rVert_2 ^2 \biggl].
\end{equation}

\paragraph{\acl{GAN}.} This approach introduced by \cite{goodfellow_generative_2014} is usually used for data synthesis, in particular for natural images. It combines a set of two neural networks: a generator and a discriminator. For a given sample of data $x$, the first one generates a fake data sample using a random variable $z$ as input (usually normally distributed). The second one is a binary classifier trying to detect if a sample is fake or real. Both networks are complementary, the generator tries to fool the discriminator with a high quality sample so it could not distinguish a real from a fake. Note that the generator takes only a random variable as input in the original approach, it can also take the initial sample $x$ as input.

Both networks are trained with a minmax game using a cost function $V(G, D)$:
\begin{equation}
\min_{G} \max_{D} V(D,G) = \mathbb{E}_{x \sim p_{\text{data}}(x)}[\log D(x)] + \mathbb{E}_{z \sim p_{\text{z}}(z)}[\log (1 - D(G(z)))],
\end{equation}
where $G$ is the generator and $D$ the discriminator. In their work, \cite{goodfellow_generative_2014} explain that minimizing $\log(1 - D(G(z)))$ leads to poor gradient at the beginning of the training since the generator is not efficient enough. In practice, $\log(D(G(z)))$ is maximized in the optimization process, it is called the Non-Saturating \ac{GAN} Loss. 

In the presented experiments, we proposed to learn a generator to create a fake \ac{RD} using the natural image as input instead of a random noise. The discriminator takes either the generated or real \ac{RD} and classifies it as fake or real. The loss function is composed of a \ac{MSE} reconstruction loss for the generator and a \ac{BCE} for the discriminator. The final objective is written:

\begin{equation}
    \min_{G} \max_{D} \mathcal{L}_{\text{\ac{GAN}}} (D, G) = \min_{G} \max_{D}  \frac{1}{\BinR \cdot \BinD}  \left\lVert \vect{x}_{r} - G(\vect{x}_{s}) \right\rVert_2 ^2 + V(D,G).
\end{equation}

Note that the implemented \ac{GAN} also uses patchGAN \cite{isola_image--image_2017} for the discriminator classification. This method is described in the following paragraph.

We also considered a second approach called \ac{cGAN} \cite{isola_image--image_2017}. It consists in providing additional information to the networks as a condition. 
This condition helps the generator to better estimate to distribution of the generated samples.
For a given sample $(x,y)$, the generator tries to reconstruct $y$ using the original data $x$ and a random noise $z$. 
In this case, the condition $x$ provides the distribution of the data in the source domain to help the generator in estimating the distribution of the data in the target domain.
The discriminator classifies either fake or real the generated or the real sample $y$ while observing $x$. Considering Equation \ref{eq:objective_cgan}, the objective function is defined as:

\begin{equation}
\label{eq:objective_cgan}
\min_{G} \max_{D} V_{x}(D,G) = \mathbb{E}_{x,y}[\log D(x, y)] + \mathbb{E}_{x, z}[\log (1 - D(x, G(x, z)))].
\end{equation}

In their work, \cite{isola_image--image_2017} described the patchGAN method to improve the discriminator consistency. Instead of directly classifying a sample as fake or real, each feature map of the last layer of the discriminator is classified in a category. The final discrimination result is obtained by aggregating the sub classifications on the feature maps. This way, the discriminator takes into account local pattern similarities. The \ac{MAE} loss is used for the generator and the \ac{BCE} loss (negative log likelihood) for the discriminator. The objective function is defined as:

\begin{equation}
    \min_{G} \max_{D} \mathcal{L}_{\text{\ac{cGAN}}} (D, G) = \frac{1}{\BinR \cdot \BinD} \left\lVert \vect{x}_{r} - G(\vect{x}_{s}) \right\rVert_1 + V_{x}(D,G).
\end{equation}

In our experiments, we note `Light-\ac{GAN}' and `Light-\ac{cGAN}' the corresponding generative methods using a Light-\ac{AE} architecture for the generator, and the original architectures proposed by \cite{goodfellow_generative_2014} and \cite{isola_image--image_2017} for the discriminator.

\begin{table}[t!]
\centering
\scriptsize
\begin{tabular}{l c c c c c c}
    \toprule
    \multirow{2}{*}{Method} & \multirow{2}{*}{\#Parameters} & \multirow{2}{*}{Loss} & \multicolumn{2}{c}{\ac{MSE} Scores} & \multicolumn{2}{c}{\ac{MAE} Scores}\\
    \cmidrule(lr{3pt}){4-5}
    \cmidrule(l{3pt}r{3pt}){6-7}
      & & & Validation & Test & Validation & Test \\
    VGG-9 & 9.23M & \ac{MSE} & 8.08 & 7.74 & 1.69 & 1.67 \\
    3D Conv. & 0.23M & \ac{MSE} & 9.10 & 8.74 & 1.83 & 1.81 \\
    \ac{AE} & 12.9M & \ac{MSE} & 4.76 & 4.84 & 1.50 & 1.50 \\
    U-Net \cite{ronneberger_u-net_2015} & 14.4M & \ac{MSE} & 5.22 & 5.16 & 1.52 & 1.52 \\
    AE + NoiseMap & 12.9M & \ac{MSE} & \textbf{4.52} & \textbf{4.51} & \textbf{1.48} & \textbf{1.48}\\
    Light-\ac{AE} & 3.24M & \ac{MSE} & 4.64 & 4.63 & 1.49 & 1.49\\
    Light-\ac{AE} & 3.24M & \ac{wMAE} & 5.16 & 5.16 & 1.50 & 1.49\\
    Light-\ac{AE} & 3.24M & \ac{wMAE}\_\ac{MSE} & 5.16 & 5.17 & 1.50 & 1.50\\
    Light-\ac{GAN} & 3.31M & $\mathcal{L}_{\text{\ac{GAN}}}$ & 4.92 & 4.87 & 1.50 & 1.50\\
    Light-\ac{cGAN} & 4.98M & $\mathcal{L}_{\text{\ac{cGAN}}}$ & 5.14 & 5.12 & 1.49 & 1.48\\
     \bottomrule
\end{tabular}
\caption[Quantitative performances of Range-Doppler reconstruction]{\textbf{Quantitative performances of Range-Doppler reconstruction}. Performances of the methods detailed in Section \ref{sec:canada_exp}, trained and tested on the splits of the dataset presented in Section \ref{sec:dataset_generation}, regarding the MSE and MAE evaluation metrics.}
\label{tab:perf_canada}
\end{table}

\paragraph{Results.}
Quantitative results presented in Table \ref{tab:perf_canada} show that our proposed Convolutional \ac{AE} an Light-\ac{AE} architectures, inspired from U-Net and detailed in the previous paragraph, reached the best performances regarding the \ac{MSE} and \ac{MAE} metrics. In particular, our proposed \ac{AE} processing an additional noise map generated from the speckle noise distribution obtained the higher reconstruction performances. Qualitative results on the test dataset presented in Section \ref{sec:dataset_generation} are illustrated in Figure \ref{fig:rd_generation_quali}. The generated \ac{RD} representations using \ac{AE} architectures succeed to recover several object signatures while reconstructing the zero-Doppler reflections. However, the speckle noise is estimated as a global average leading to a smooth background of intensities. 

\subsection{Discussions}
\label{sec:canada_discussions}
This section has described several methods to generate \acl{RD} representations using natural images. The presented experiments have shown that our Noise Map method integrating a map of generated speckle noise leads to increase the performances while reducing the training time. \ac{GAN} and \ac{cGAN} architectures have produced interesting results but further experiments should be conducted to reach better performances. The selected methods did not generate high quality \ac{RD} so the annotation transfer from the image domain to the \ac{RADAR} domain is not usable with this approach. In the next section, a dataset is introduced with non-urban scenes and a different pipeline to annotate the \ac{RADAR} representations in a semi-supervised manner.

\section{CARRADA dataset}
\label{sec:carrada}

\begin{figure}[!t]
\centering
\includegraphics[width=1\textwidth]{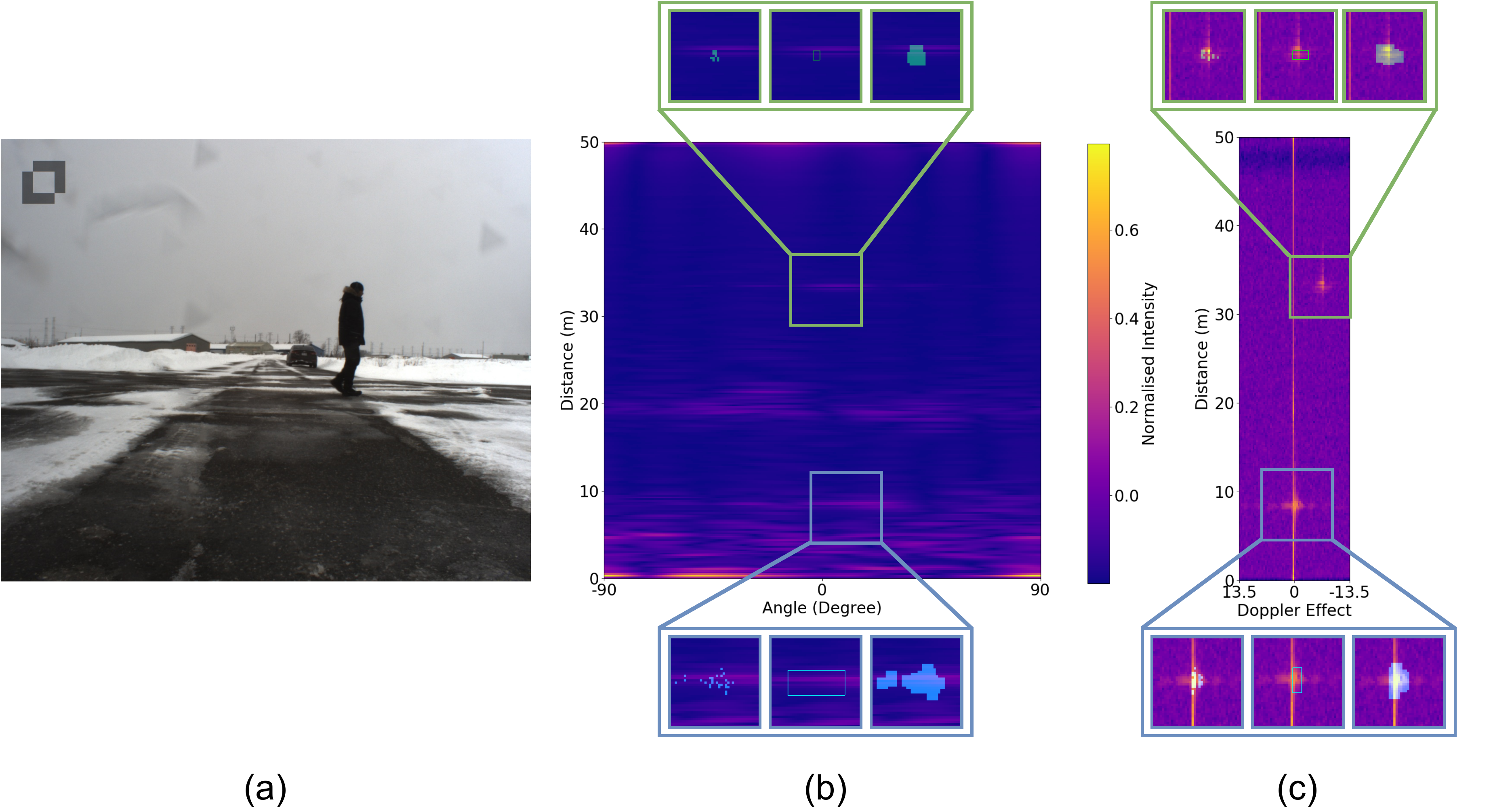}
\caption[A scene from CARRADA dataset, with a pedestrian and a car]{\textbf{A scene from CARRADA dataset, with a pedestrian and a car}. (a) Video frame provided by the frontal camera, showing a pedestrian at approximately 8m from the sensors and a car in the background at approximately 33m; (b-c) \ac{RADAR} signal at the same instant in \acl{RA} and \acl{RD} representation respectively. Three types of annotations are provided: sparse points, bounding boxes and dense masks. The blue squares correspond to the pedestrian and the green ones to the car. Note that the \acl{RA} view illustrated in (b) has been obtained using a maximum aggregation and a log-transform (Eq. \ref{eq:general_agg_method}) while the \acl{RD} view in (c) is computed using Equation \ref{eq:agg_method_average}.
}
\label{fig:carrada_teaser}
\end{figure}

High quality perception is essential for autonomous driving systems.
To reach the accuracy and robustness that are required by such systems, several types of sensors must be combined.  
Currently, mostly cameras and lidar are deployed to build a representation of the world around the vehicle. 
While \ac{RADAR} sensors have been used for a long time in the automotive industry, they are still under-used for autonomous driving despite their appealing characteristics (notably, their ability to measure the relative speed of obstacles and to operate even in adverse weather conditions).
To a large extent, this situation is due to the relative lack of automotive datasets with real \ac{RADAR}  signals that are both raw and annotated as detailed in Section \ref{sec:related_datasets}. 
In this section, CARRADA is introduced, a dataset of synchronized camera and \ac{RADAR} recordings with \acl{RAD} annotations.
A semi-automatic annotation approach is also presented, which was used to annotate the dataset, and a \ac{RADAR} semantic segmentation baseline evaluated on several metrics.
A sample of camera and \ac{RADAR} data with generated annotation is illustrated in Figure \ref{fig:carrada_teaser}.
Both our code and dataset are available online \footnote{\url{https://github.com/valeoai/carrada\_dataset}}.
This section presents a work mainly inspired from our article published at the International Conference on Pattern Recognition (ICPR) \cite{ouaknine_carrada_2020}.

\subsection{Dataset}
\label{sec:carrada_dataset}

The dataset has been recorded in Canada on a test track to reduce environmental noise. The acquisition setup consists of an AWR1843-BOOST\footnote{\url{https://www.ti.com/tool/AWR1843BOOST}} \ac{FMCW} \ac{RADAR} and a camera mounted on a stationary car. 
The \ac{RADAR} uses the \ac{MIMO} system configuration with 2 \ac{Tx} and 4 \ac{Rx} producing a total of 8 virtual antennas. The parameters and specifications of the sensor are provided in Table \ref{tab:carrada_radar_specs}. The image data recorded by the camera and the \ac{RADAR} data are synchronized to have the same frame rate in the dataset.
The sensors are also calibrated to have the same Cartesian coordinate system. The image resolution is $1238 {\times} 1028$ pixels. The \acl{RD} and \acl{RA} representations are respectively stored in 2D matrices of size $256 {\times} 64$ ($\BinR {\times} \BinD$) and $256 {\times} 256$ ($\BinR {\times} \BinA$).

One or two objects are moving in the scene at the same time with various trajectories to simulate urban driving scenarios. 
The distribution of these scenarios across the dataset is shown in Figure \ref{fig:carrada_dataset_distrib}.
The objects are moving in front of the sensors: approaching, moving away, going from right to left or from left to right (see examples in Figure \ref{fig:carrada_temporal_ex}). Each object is an \textit{instance} tracked in the sequence. The distribution of mean radial velocities for each object category is provided in Figure \ref{fig:carrada_velocity_distrib}, while other global statistics about the recordings can be found in Table \ref{tab:carrada_dataset}.

Object signatures are annotated in both \ac{RA} and \ac{RD} representations for each sequence. Each instance has an identification number, a category and a localization in the data. Three types of annotations for localization are provided: sparse points, boxes and dense masks. The next section will describe the pipeline used to generate them.

\begin{table}[!t]
\centering
\scriptsize
\begin{minipage}[b]{.4\linewidth}
\begin{tabular}{cc} \toprule
Parameter & Value\\ \midrule
Frequency & 77 Ghz\\
Sweep Bandwidth & 4 Ghz\\
Maximum Range & 50 m\\ 
\ac{FFT} Range Resolution & 0.20 m\\
Maximum Radial Velocity & 13.43 m/s\\
\ac{FFT} Radial Velocity Resolution & 0.42 m/s\\
Field of View & 180$^{\circ}$\\
\ac{FFT} Angle Resolution & 0.70$^{\circ}$\\ 
Number of Chirps per Frame & 64\\
Number of Samples per Chirp & 256\\ \bottomrule
\end{tabular}
\caption[Parameters and settings of the RADAR sensor]{\textbf{Parameters and settings of the RADAR sensor}.}
\label{tab:carrada_radar_specs}
\end{minipage}
\quad
\begin{minipage}[b]{.45\linewidth}
\renewcommand{\arraystretch}{1.0}
\centering
\scriptsize
\begin{tabular}{cc} \toprule
Parameter & Value\\ \midrule
Number of sequences & 30\\
Total number of instances & 78\\
Total number of frames & 12666 (21.1 min)\\ 
Maximum number of frames per seq. & 1017 (1.7 min)\\
Minimum number of frames per seq. & 157 (0.3 min)\\
Mean number of frames per seq. & 422 (0.7 min)\\
\multirow{2}{*}{\shortstack{Total number of annotated frames \\ with instance(s)}} & \multirow{2}{*}{7193 (12.0 min)}\\
\\
\bottomrule
\end{tabular}
\caption[Statistics of the CARRADA dataset]{\textbf{Statistics of the CARRADA dataset}.}
\label{tab:carrada_dataset}
\end{minipage}
\end{table}

\subsection{Pipeline for annotation generation}
\label{sec:carrada_annot_pipeline}

Automotive \ac{RADAR} representations are difficult to understand compared to natural images. Objects are represented by shapes with varying sizes carrying physical measures. It is not a trivial task to produce good quality annotations on this data.
This section details a semi-automatic pipeline based on video frames to provide annotations on \ac{RADAR} representations.

\subsubsection{From vision to physical measurements}
\label{sec:from_vision}

The camera and \ac{RADAR} recordings are synchronized. Visual information in the natural images is used to obtain physical prior knowledge about an instance as well as its category. The real-world coordinates of the instance and its radial velocity are estimated generating the annotation in the \ac{RADAR} representation. This first step instantiates a tracking pipeline propagating the annotation in the entire \ac{RADAR} sequence.

\begin{figure}[!t]
\centering
\begin{minipage}[b]{0.45\textwidth}
\centering
    \hspace*{-0.3cm}
    \includegraphics[width=7cm]{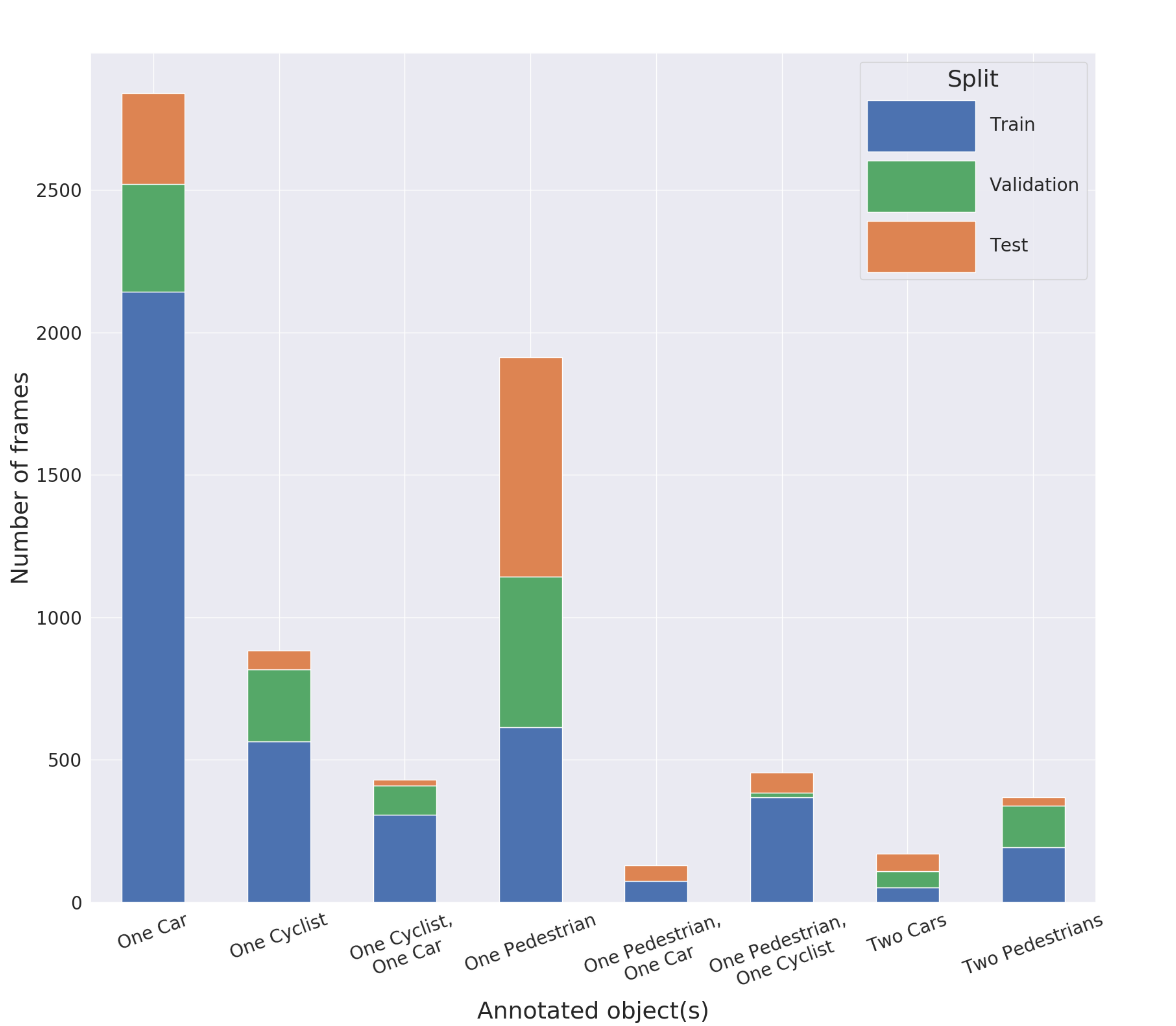}
    \caption[Object distribution across CARRADA]{\textbf{Object distribution across CARRADA}. Distribution of the eight object configurations present in the dataset, expressed as frame numbers across the three parts of the proposed split.}
    \label{fig:carrada_dataset_distrib}
\end{minipage}
\qquad
\quad
\begin{minipage}[b]{0.45\textwidth}
\centering
    \hspace*{-0.5cm}
    \includegraphics[width=7cm]{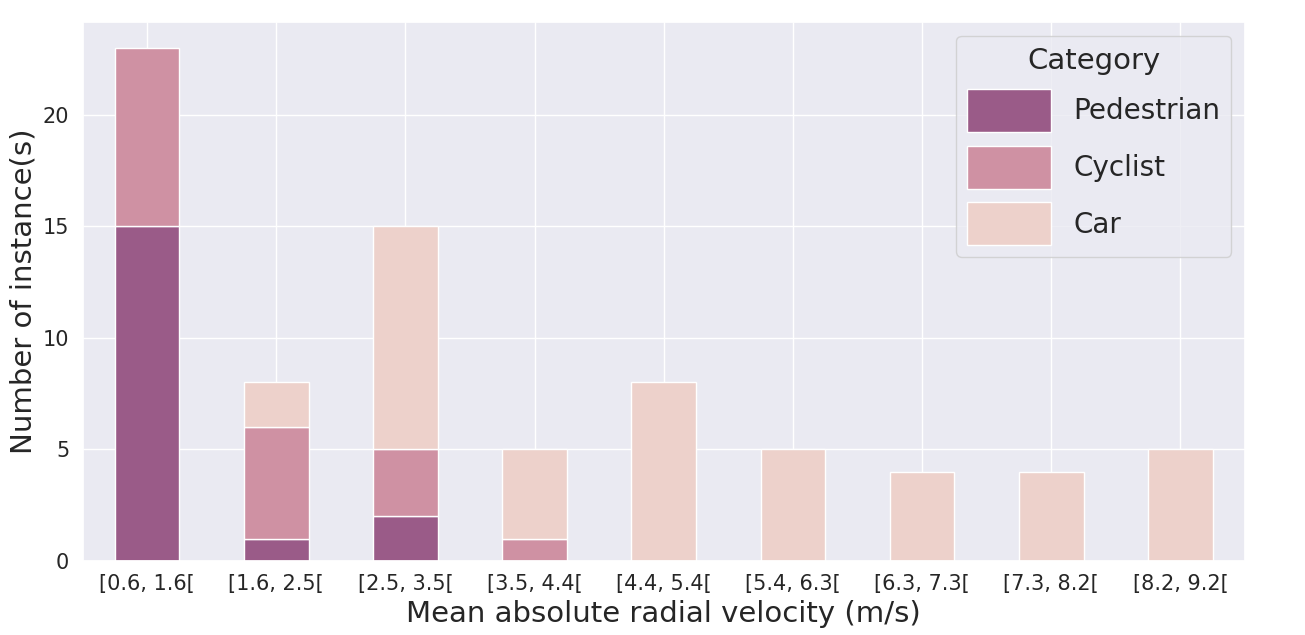}
    \caption[Distribution of radial velocities for all categories]{\textbf{Distribution of radial velocities for all categories}. For each annotated instance, its absolute radial velocity in m/s is averaged, over its sparse annotations in each frame and over time, and one histogram is built for each object class. Note that annotated velocities are actually signed (negative when the object if moving away and positive when it approaches the radar).}
    \label{fig:carrada_velocity_distrib}
\end{minipage}
\end{figure}

Each video sequence is processed by a Mask R-CNN \cite{he_mask_2017} model providing both semantic segmentation and bounding box predictions for each detected instance. Both are required for our pipeline to compute the center of mass of the object and to track it.
Instance tracking is performed with the \ac{SORT} algorithm \cite{bewley_simple_2016}. 
This light-weight tracker computes the overlap between the predicted boxes and the tracked boxes of each instance at the previous frame. The selected boxes are the most likely to contain the same instance, \textit{i.e.} the boxes with the highest overlap.

The center of mass of each segmented instance is projected on the bottom-most pixel coordinates of the segmentation mask. This projected pixel localized on the ground is considered as the reference point of the instance.
Using the intrinsic and extrinsic parameters of the camera, pixel coordinates of a point in the real-world space are expressed as:
\begin{equation}
s \; \vect{p} = A \; B \; \vect{c} ,
\end{equation}
where $\vect{p} = [p_x,p_y,1]^{\top}$ and $\vect{c}= [c_x,c_y,c_z,1]^{\top}$ are respectively the pixel coordinates in the image and the real-world point coordinates, $s$ a scaling factor, and $A$ and $B$ are the intrinsic and extrinsic parameters of the camera defined as:
\begin{equation}
\label{eq:camera_params}
A=
\begin{bmatrix}
f_x & 0 & a_x \\
0 & f_y & a_y \\
0 & 0 & 1
\end{bmatrix},\,
B=
\begin{bmatrix}
r_{11} & r_{12} & r_{13} & m_1 \\
r_{21} & r_{22} & r_{23} & m_2 \\
r_{31} & r_{32} & r_{33} & m_3
\end{bmatrix}.
\end{equation}

Using Equation \ref{eq:camera_params}, one can determine $\vect{c}$ knowing $\vect{p}$ with a fixed value of elevation.

Given the time interval $\delta t$ separating two frames $t-\delta t$ and $t$, the velocity vector $\mathbf{v}^t$ is defined as:
\begin{equation}
    \mathbf{v}^t = \vect{c}^t - \vect{c}^{t-\delta t},
\end{equation}
where $\vect{c}^t$ is the real-world coordinate in frame $t$.
The time interval chosen in practice is $\delta t = 1 $ second.

\begin{figure}[!t]
\centering
\includegraphics[width=3in]{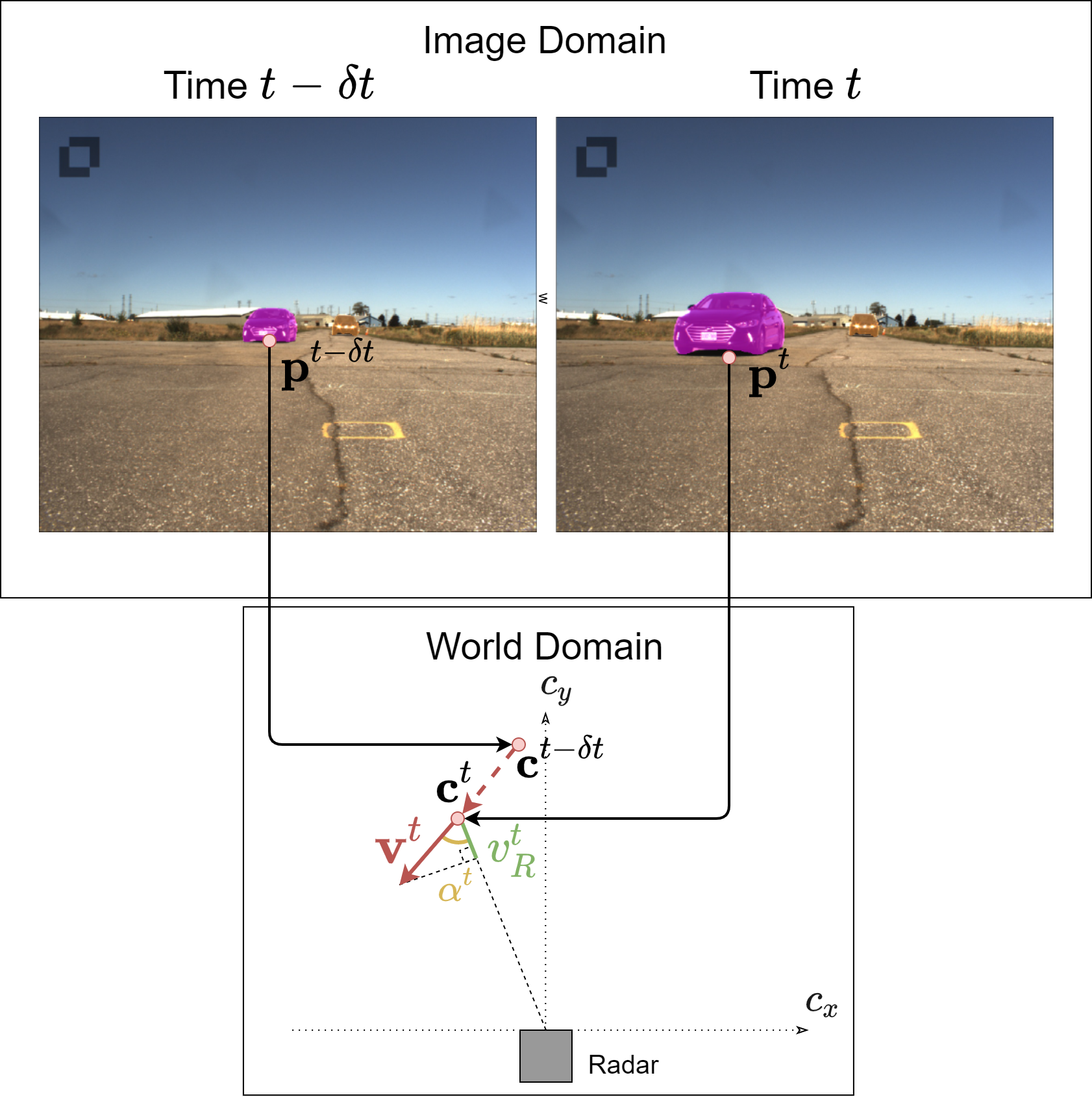}
\caption[Estimation of the radial velocity from natural images]{\textbf{Estimation of the radial velocity from natural images}.
The space $(c_x, c_y, c_z)$ defines real-world coordinates considering the \ac{RADAR} as origin, $c_z$ is fixed to zero. Points in the real-world domain $\vect{c}^{t-\delta t}$ and $\vect{c}^{t}$ (bottom) are estimated using the points in the pixel domain $\vect{p}^{t-\delta t}$ and $\vect{p}^{t}$ (top). The velocity vector $\vect{v}^t$ is estimated with the real-world points. The radial velocity $v^t_R$ of the object at time $t$ corresponds to the projection of its velocity vector on the straight line between the \ac{RADAR} of the object.}
\label{fig:relative_velocity}
\end{figure}

The Doppler effect recorded by the \ac{RADAR} is the radial velocity of the instance reflecting the signal. The radial velocity $v_R^t$ at a given frame $t$ is defined as:
\begin{equation}
    v_R^t = \cos{\alpha^t} \; \lVert \mathbf{v}^t \rVert ,
\end{equation}
where $\alpha^t$ is the angle formed by $\mathbf{v}^t$ and the straight line between the \ac{RADAR} and the instance. The quantization of the radial velocity is illustrated in Figure \ref{fig:relative_velocity}.
This way, each instance detected in the frame is characterized by a feature point $I^t = [\vect{c}^t, v_R^t]^{\top}$.
This point will be projected in a \ac{RADAR} representation to annotate the raw data and track it in this representation.

\subsubsection{DoA clustering and centroid tracking}
\label{sec:doa_tracking}

\begin{figure}[!t]
\centering
\includegraphics[width=2.5in]{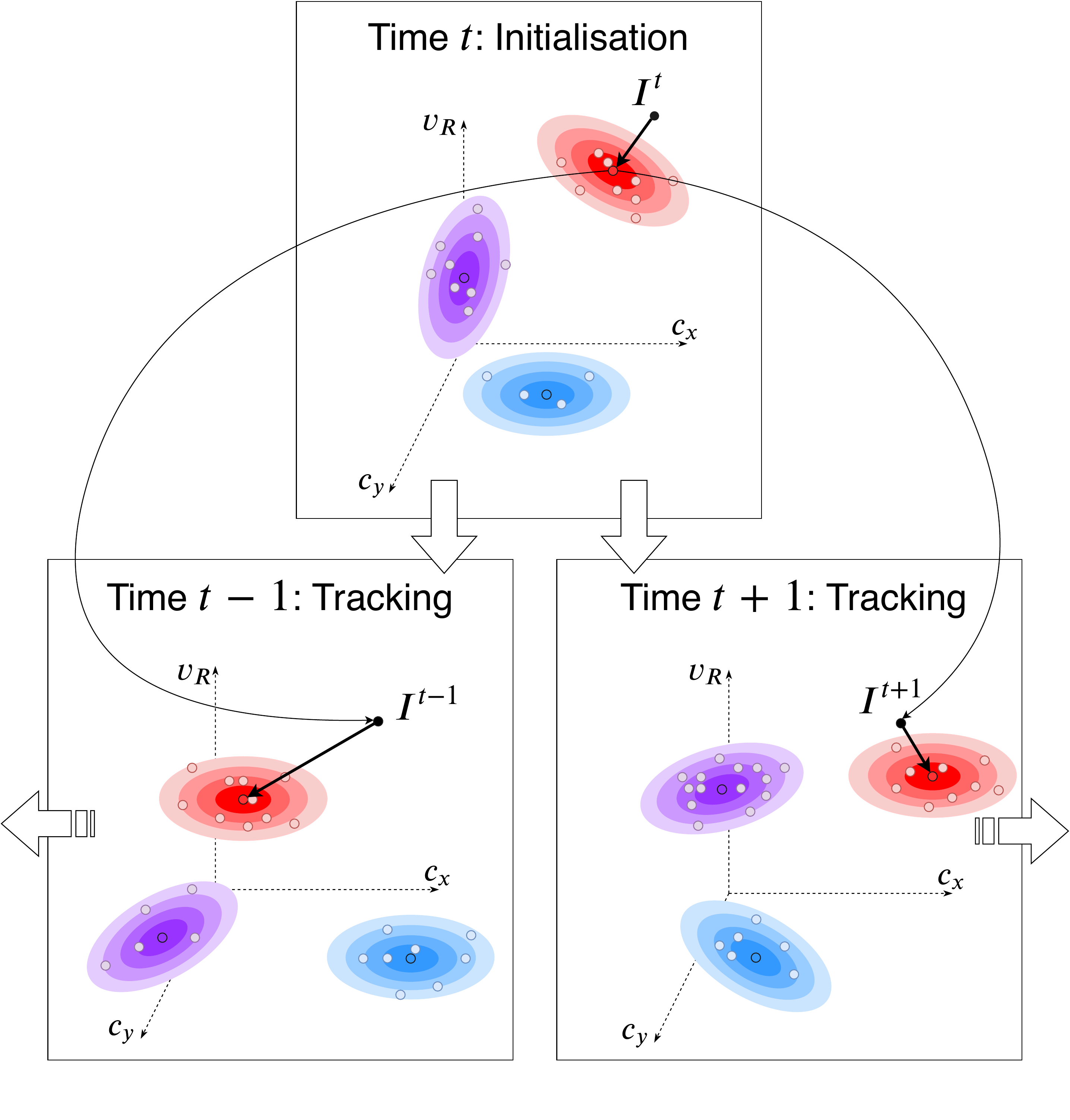}
\caption[Tracking of the Mean Shift cluster to propagate the annotation in the sequence]{\textbf{Tracking of the Mean Shift cluster to propagate the annotation in the sequence}.
The Mean Shift algorithm used with the bandwidth selection method is applied to the \ac{DoA}-Doppler representation at time $t$. The estimated point $I^t$, using the computer vision pipeline and corresponding to the tracked object, is associated to its closest cluster. The centroid of this cluster considered as $I^{t+1}$ and $I^{t-1}$ in the next and previous \ac{DoA}-Doppler frames is tracked iteratively.}
\label{fig:clustering_tracking}
\end{figure}

The \ac{RA} representation is a \ac{RADAR} scene in polar coordinates. Its transformation into Cartesian coordinates is called \ac{DoA}. Points are filtered by a \ac{CFAR} algorithm \cite{rohling_radar_1983} keeping the highest intensity values while taking into account the local relation between points. The \ac{DoA} is then a sparse point cloud in a 2D coordinate space similar to a \ac{BEV} representation.
The representation is enriched using the recorded Doppler for each point as the third axis of the point cloud. The 3D point cloud combines the Cartesian coordinates of the reflected point and its Doppler value. This helps to distinguish the signature boundaries of different objects. The feature point $I^t$ is projected in this space and assigned to a cluster of points considered as the reflection of the targeted instance. It is then tracked in the past and future using the following process, illustrated in Figure \ref{fig:clustering_tracking}.
This initialisation requires a human supervision to validate the matching between $I^t$ and the cluster of \ac{RADAR} points corresponding to the object before the tracking. This supervision could be avoided by considering an automatic validation between the camera and the \ac{RADAR} frames, it will be explored in a future work. 

At a given timestamp chosen by the user, a 3D \ac{DoA}-Doppler point cloud is clustered using the Mean Shift algorithm \cite{comaniciu_mean_2002}. Let $\{\vect{x}_0, \cdots, \vect{x}_{n-1} \}$ be a point cloud of $n$ points. For a given starting point, 
the algorithm iteratively computes a weighted mean of the current local neighborhood and updates the point until convergence. Each iteration reads:
\begin{equation}
    \vect{x} \leftarrow \frac{\sum_{i=0}^{n-1}\vect{x}_i K(\vect{x};\vect{x}_i, \sigma)}{\sum_{i=0}^{n-1}K(\vect{x};\vect{x}_i, \sigma)},
\end{equation}
where $K( \vect{x};\vect{x}_i,\sigma) = \mathcal{N}(\left\| \frac{\vect{x} - \vect{x}_i}{\sigma} \right\|;0,1)$ is the multivariate spherical Gaussian kernel with bandwidth $\sigma$, centered at $\vect{x}_i$.
All initial points leading to close final locations at convergence are considered as belonging to the same cluster. 

Mean-Shift clustering is sensitive to the bandwidth parameter $\sigma$. Its value should depend on the point cloud distribution and it is usually defined with prior knowledge about the data. 
In this application, it is not straightforward to group points belonging to the same object in the \ac{DoA}-Doppler point cloud representation. The number of points and their distribution depend on the distance and the surface of reflectivity of the target. Moreover, these characteristics change during a sequence while the instance is moving in front of the \ac{RADAR}. 
In the work of \cite{bugeau_bandwidth_2007}, the authors proposed to compute several times the Mean-Shift algorithm with ordered bandwidth values. They finally selected the optimal bandwidth by comparing the distribution of the clusters.
Our method applied the same procedure to point clouds with multiple instances. It automatically finds an optimal bandwidth for each instance contained in each point cloud. We proposed additional metrics to compare the cluster distributions, they are described in the following paragraphs.

For a given \ac{DoA}-Doppler point cloud, the closest cluster to the feature point $I^t$ is associated to an instance. Let $\sigma_b \in \{ \sigma_0, \cdots, \sigma_{B-1} \}$ be a bandwidth in a range of $B$ ordered values. A Mean-Shift algorithm noted $\text{MeanShift}(\sigma_b)$ selects the closest cluster $\mathcal{C}_b$ to $I^t$ containing $n_b$ points. After running the algorithm with all bandwidth values, $\{\mathcal{C}_0, \cdots, \mathcal{C}_{B-1} \}$ clusters are found with their corresponding optimal bandwidths. 

The following paragraphs will detail methods to select the optimal bandwidth $\sigma_{b^*}$ regarding the distribution and the stability of the clustered point clouds.
Once the optimal bandwidth is found, the closest cluster to $I^t$ using $\text{MeanShift}(\sigma_{b^*})$ is considered as belonging to the targeted instance.
The points $I^{t+1}$ and  $I^{t-1}$ are assigned with the centroid of this cluster. The process is then iterated in the previous and next frames to track the center of the initial cluster until the end of the sequence. 


\paragraph{Log-Ratio}
The further away the object, the smaller its signature in the \ac{RADAR} representation and the smaller its \ac{DoA} point cloud.
The proposed Log-Ratio method defines the bandwidth for the Mean Shift algorithm as a function of the distance of the centroid of the DoA cluster tracked from the previous or next \ac{RADAR} frame \footnote{It depends on the direction of the annotation propagation in the \ac{RADAR} sequence, either it is in the future or in the past depending on the initialization frame.}.
Let $d^{\text{max}}$ and $d_{t-1}$ be respectively the maximum distance of the \ac{RADAR} detection (50 meters in our case) and the distance to the centroid of the tracked \ac{DoA}-Doppler point cloud at $t-1$ (or $t+1$ depending on tracking direction). At time $t$, the optimal bandwidth $\sigma_{b^*}(d_{t-1})$ is defined as: 
\begin{equation}
    \sigma_{b^*}(d_{t-1}) = \frac{1 + \log(1 + d^{\text{max}})}{1 + \log(1 + d_{t-1})} = 
    \begin{cases}
    1 & \text{if} \quad  d_{t-1} = d^{\text{max}}\\
    1 + \log(1 + d^{\text{max}}) & \text{if} \quad d_{t-1} = 0 \\
    \frac{1 + \log(1 + d^{\text{max}})}{1 + \log(1 + d_{t-1})} & \text{otherwise}.
    \end{cases}
\end{equation}

\paragraph{Determinant}
In this method, the optimal bandwidth $\sigma_{b^*}$ is selected by comparing the stability of the determinant between the selected clusters.
For each $b \in \{0, \cdots, B-1\}$ ordered values and a cluster $\mathcal{C}_b =  \{\vect{x}_0, \cdots, \vect{x}_{n_b-1} \}$ of $n_b$ points, its Gaussian distribution $\mcl{N}(\widehat{\mu}_b, \widehat{\Sigma}_b)$ is estimated with expectation $\widehat{\mu}_b = \frac{1}{n_b} \sum_{i=0}^{n_b-1} \vect{x}_i$ and variance $\widehat{\Sigma}_b = \frac{1}{n_b-1} \sum_{i=0}^{n_b-1} (\mathbf{x}_{i \cdot} - \widehat{\mu}_b)(\mathbf{x}_{i \cdot} - \widehat{\mu}_b)^{\top}$.
For each bandwidth $b$, the determinant $|\widehat{\Sigma}_b| = \text{Det}(\widehat{\Sigma}_b)$ of the variance-covariance matrix of the corresponding Gaussian distribution is computed. Using these determinants from the fitted distributions as signatures, the bandwidth $\sigma_{b^*}$ is selected by choosing the one which is the most ``stable'' with respect to a varying bandwidth: 
\begin{equation}
    b^* = \underset{b \in \{1, \cdots, B-2\}}{\text{argmin}} (|\widehat{\Sigma}_b| - |\widehat{\Sigma}_{b-1}|) + (|\widehat{\Sigma}_{b+1}| - |\widehat{\Sigma}_b|).
\label{eq:detMin}
\end{equation}

\paragraph{Number of points}
The optimal bandwidth $\sigma_{b^*}$ is selected by comparing the stability of the number of points between the selected clusters.
For each $b \in \{0, \cdots, B-1\}$ ordered values, the cluster $\mathcal{C}_b \in \{\mathcal{C}_0, \cdots, \mathcal{C}_{B-1}\}$ corresponds to the closest cluster regarding the tracked centroid in a sequence using $\text{MeanShift}(\sigma_b)$.
The bandwidth $\sigma_{b^*}$ is selected by choosing the one which is the most ``stable'' with respect to a varying bandwidth:
\begin{equation}
b^* = \underset{b \in \{1, \cdots, B-2\}}{\text{argmin}} (\# \mathcal{C}_b - \# \mathcal{C}_{b-1}) + (\# \mathcal{C}_{b+1} - \# \mathcal{C}_b),
\label{eq:ptsMin}
\end{equation}
where $\# \mathcal{C}_b = \sum_{x \in \mathcal{C}_b} 1$ is the number of points in the cluster $\mathcal{C}_b$.

\paragraph{Jensen-Shannon divergence}
The optimal bandwidth $\sigma_{b^*}$ is selected by comparing the stability of the probability distribution of the points between the selected clusters.
For each $b \in \{0, \cdots, B-1\}$ ordered values, the probability distribution $p_b$ estimated with the $n_b$ points of the cluster $\mathcal{C}_b =  \{\vect{x}_0, \cdots, \vect{x}_{n_b-1} \}$ is the Gaussian distribution $\mcl{N}(\widehat{\mu}_b, \widehat{\Sigma}_b)$  with expectation $\widehat{\mu}_b = \frac{1}{n_b} \sum_{i=0}^{n_b-1} \vect{x}_i$ and variance $\widehat{\Sigma}_b = \frac{1}{n_b-1} \sum_{i=0}^{n_b-1} (\mathbf{x}_{i \cdot} - \widehat{\mu}_b)(\mathbf{x}_{i \cdot} - \widehat{\mu}_b)^{\top}$.
Using these fitted distributions, the bandwidth $\sigma_{b^*}$ is selected by choosing the one which is the most ``stable'' with respect to a varying bandwidth: 
\begin{equation}
\begin{split}
    b^* = \underset{b \in \{1, \cdots, B-2\}}{\text{argmin}} & \Big[\textit{JS} \big( p_b \|  p_{b-1} \big)
     + \textit{JS} \big(p_b \| p_{b+1} \big)\Big],
\end{split}
\label{eq:jsMin}
\end{equation}
where \acs{JS} is the \acl{JS} divergence \cite{endres_new_2003}. This is a proper metric derived from \ac{KL} 
divergence \cite{kullback_information_1951} as 
\begin{equation}
\textit{JS}(p\|q)^2 = \frac{\textit{KL}(p \| \frac{p + q}{2}) + \textit{KL}(q \| \frac{p + q}{2})}{2},    
\end{equation}
where $\textit{KL}(p \| \frac{p + q}{2}) = \sum_i p(i) \frac{2p(i)}{p(i) + q(i)}$, 
for two discrete probability distributions $p$ and $q$. The \ac{JS} divergence measure the similarity between $q$ and $p$ since it is symmetric and has a finite value, contrary to the \acl{KL} divergence.

\begin{figure}[!t]
\centering
\includegraphics[width=4.5in]{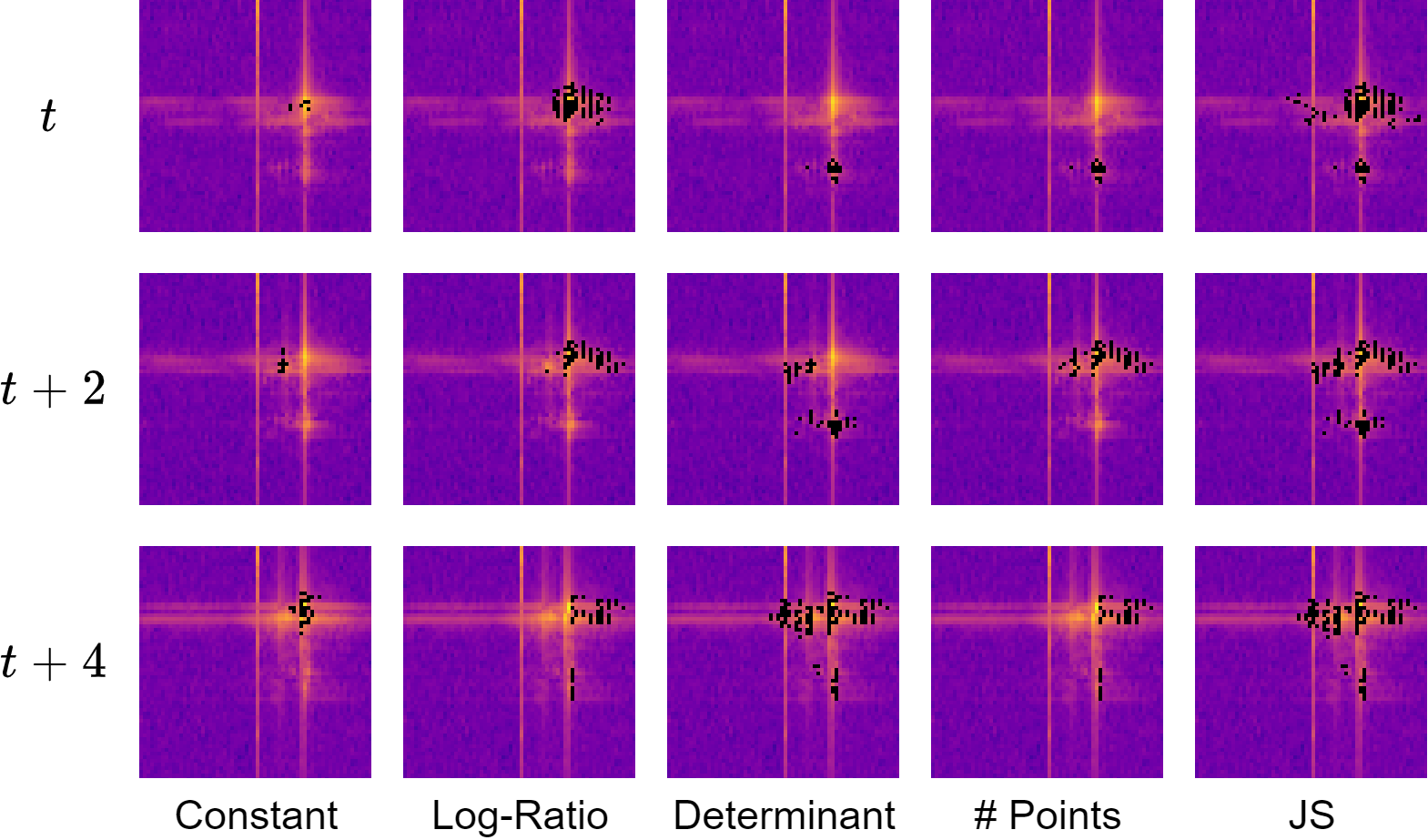}
\caption[Comparison of bandwidth selection methods]{\textbf{Comparison of bandwidth selection methods}. Qualitative results of the bandwidth selection methods applied to the Mean Shift algorithm presented in Section \ref{sec:from_vision}. The algorithm is applied at each timestamp of the \ac{RADAR} sequence in order to track the cluster corresponding to the object. Results are illustrated for a sequence between timestamps $t$ and $t+4$. The black points correspond to the tracked cluster of points in the \ac{DoA}-Doppler space. They are projected in the \acl{RD} representation cropped for visualisation purpose.}
\label{fig:quali_tracking}
\end{figure}

The presented methods have been tested and qualitatively compared on the CARRADA dataset. The \ac{JS} divergence used to quantify the stability between the probability distributions of compared clustered has reached the best results. 
It succeeds in including a large number of \ac{RADAR} points belonging to the object through time. 
An example comparing the proposed methods is illustrated in Figure \ref{fig:quali_tracking}. The temporal consistency of the bandwidth selection using the \ac{JS} divergence is also highlighted in Figure \ref{fig:carrada_temporal_ex}.

\subsubsection{Projections and annotations}
\label{sec:proj_and_annot}

We recall that $\mathcal{C}_{b^*}$ is the cluster associated to the point $I^t$ at time $t$ using $\text{MeanShift}(\sigma_{b^*})$, where $\sigma_{b^*}$ is the estimated optimal bandwidth. This cluster is considered as belonging to the tracked object. A category is associated to it by using the segmentation model on the image (Section \ref{sec:from_vision}).
The points are projected onto the \ac{RD} representation using the radial velocity and the distance is computed with the real-world coordinates. They are also projected onto the \ac{RA} representation by converting the Cartesian coordinates to polar coordinates. 

Let $f_{D}$ be the function which projects a point from the \ac{DoA}-Doppler representation into the \ac{RD} representation. Similarly, we denote with $f_{A}$ the projection into the \ac{RA} representation. The sets of points $\mathcal{M}_D = f_D(\mathcal{C}_b)$ and $\mathcal{M}_A = f_A(\mathcal{C}_b)$ correspond, respectively, to the \ac{RD} and \ac{RA} representations of $\mathcal{C}_b$. They are called the sparse-point annotations.

The bounding box of a set of points in $\mathbb{R}^2$ (either from $\mathcal{M}_D$ or $\mathcal{M}_A$) is defined as a rectangle parameterized by $\{ (x_{\min}, y_{\min}),  (x_{\max}, y_{\max})\}$ where $x_{\min}$ is the minimum $x$-coordinate of the set, $x_{\max}$ is the maximum, and similarly for the $y$-coordinates.

Mathematical morphology has been experimented to expand the sparse annotation creating a dense mask.
%
Let $\mathcal{E} \subset \mathbb{R}^2$ be a structuring element, the \textbf{dilation} of $\mathcal{M}$ by $\mathcal{E}$ is defined as $\mathcal{M} \oplus \mathcal{E} = \cup_{e \in \mathcal{E}} \mathcal{M}_e$, where $\mathcal{M}_e$ is the translation of $\mathcal{M}$ by $e$.
The \textbf{erosion} of $\mathcal{M}$ by $\mathcal{E}$, reducing the shape of $\mathcal{M}$ as opposed to the dilation, is written $\mathcal{M} \ominus \mathcal{E} = \cap_{e \in \mathcal{E}} \mathcal{M}_{-e}$, where $\mathcal{M}_{-e}$ is the translation of $\mathcal{M}$ by $-e$.
Finally, the \textbf{closing} of $\mathcal{M}$ by $\mathcal{E}$ is the erosion of the dilation of this set. It is noted $\mathcal{M} \bullet \mathcal{E} = (\mathcal{M} \oplus \mathcal{E}) \ominus \mathcal{E}$.

The dense mask annotation is obtained by dilating the sparse annotated set with a circular structuring element: given the sparse set of points $ \mathcal{M} = \{(x_0,y_0), \dots, (x_{N-1},y_{N-1})\}$, the associated dense mask is the set of discrete coordinates in $\cup_{i=0}^{N-1} \mcl{B}_r(x_i, y_i)$, where $\mcl{B}_r(x, y)$ is the disk of radius $r$ centered at $(x,y)$. It is called the ``\textbf{Dilation/Circle}'' method.
We compared it with three other combinations of operations: a dilation with a cross structuring element (Dilation/Cross), the Dilation/Circle followed by a closing with a square structuring element (Dilation/Circle-Closing/Square) and the Dilation/Circle followed by a closing and an erosion, both with a square structuring element (Dilation/Circle-Closing/Square-Erosion/Square). Note that the structuring elements are centered in a $3 {\times} 3$ binary patch. The cross structuring element has five 1 values and the square has nine 1 values.

\begin{figure}[!t]
\centering
\includegraphics[width=1\textwidth]{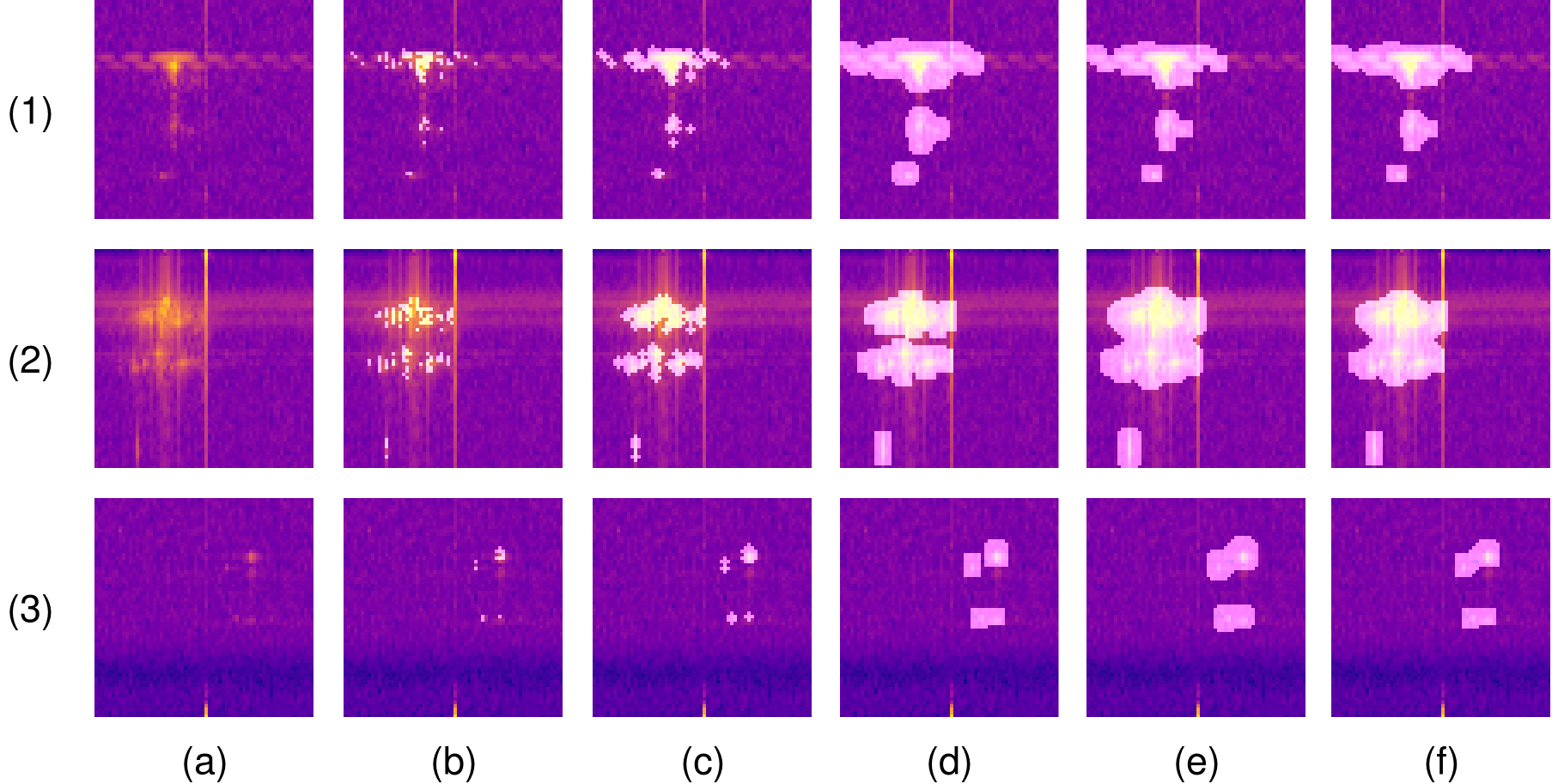}
\caption[Comparison of methods for dense mask annotation generation]{\textbf{Comparison of methods for dense mask annotation generation}. For each (1-3) independent example, the (a) source data is an object signature in a cropped \acl{RD}. Using the tracking pipeline and the projections presented in Sections \ref{sec:doa_tracking} and \ref{sec:proj_and_annot}, the cluster (b) is projected back in the representation, generating the sparse annotation. Qualitative results of the following methods are illustrated: (c) Dilation/Cross, (d) Dilation/Circle, (e) Dilation/Circle-Closing/Square, (f) Dilation/Circle-Closing/Square-Erosion/Square.
}
\label{fig:carrada_annot_methods}
\end{figure}

These combinations of mathematical morphology operations are compared to create the dense mask from the sparse annotation. Qualitative results are illustrated in Figure \ref{fig:carrada_annot_methods}. The Dilation/Circle method, used on its own, reached convincing results by covering homogeneously the shape of the object signature while including fewer noise bins than other methods for small objects.

In the following section, a baseline is proposed for \ac{RADAR} semantic segmentation trained and evaluated on the annotations detailed above.

\subsection{Semantic segmentation baseline}
\label{sec:carrada_baseline}

We proposed a baseline for semantic segmentation using \ac{RD} or \ac{RA} \ac{RADAR} representation to detect and classify annotated objects. \acp{FCN} \cite{long_fully_2015} are used here to learn features at different scales by processing the input data with convolutions and down-sampling. Feature maps from convolutional layers are up-sampled with up-convolutions to recover the original input size. Each bin of the output segmentation mask is then classified. The particularity of \ac{FCN} is that is uses skip connections from features learnt at different levels of the network to generate the final output. We denote with \ac{FCN}-32s a network where the output mask is generated only by up-sampling and processing feature maps with $1/32$ resolution of the input. Similarly, \ac{FCN}-16s is a network where $1/32$ and $1/16$ resolution features maps are used to generate the output mask. In the same manner, \ac{FCN}-8s fuses $1/32$, $1/16$ and $1/8$ resolution feature maps for output prediction.

The models are trained to recover dense mask annotations with four categories: \textit{background}, \textit{pedestrian}, \textit{cyclist} and \textit{car}. The background corresponds to speckle noise, sensor noise and artefacts which are covering most of the raw \ac{RADAR} data.
Parameters are optimized for 100 epochs using a categorical cross entropy loss function and the Adam optimizer \cite{kingma_adam_2015} with the recommended parameters ($\beta_1 = 0.9$, $\beta_2 = 0.999$ and $\epsilon = 1 \cdot 10^{-8}$). The batch size is fixed to 20 for the \ac{RD} representation and to 10 for the \ac{RA} representation to fill the memory capacities of the GPU (11GB). For both representations, the learning rate is initialized to $1 \cdot 10^{-4}$ for \ac{FCN}-8s and $5 \cdot 10^{-5}$ for \ac{FCN}-16s and \ac{FCN}-32s. The learning rate has an exponential decay of 0.9 each 10 epochs. Training has been completed using the PyTorch framework with a single \textit{GeForce RTX 2080 Ti} GPU. 

Performances are evaluated for each \ac{RADAR} representation using the \ac{IoU}, the \ac{PP} and the \ac{PR} for each category. Metrics by category are aggregated using arithmetic and harmonic means.
To ensure consistency of the results, all performances are averaged from three trained models initialized with different seeds.
Results are presented in Table \ref{tab:carrada_fcn_baseline}. 

\begin{table*}[t]
\def\arraystretch{1.2}
\setlength\tabcolsep{0.5pt}
\scriptsize
\begin{tabularx}{\textwidth}{c @{\hskip 0.1in} c @{} CCCCCC CCCCCC CCCCCC @{}}
\toprule
 \multicolumn{1}{c}{Data} & \multicolumn{1}{c}{Model} & \multicolumn{6}{c}{\ac{IoU}} &   \multicolumn{6}{c}{\ac{PP}} & \multicolumn{6}{c}{\ac{PR}} \\
 \cmidrule(lr{3pt}){3-8}
 \cmidrule(l{3pt}r{3pt}){9-14}
 \cmidrule(l{3pt}r){15-20}
 & & \rotatebox{90}{Background} & \rotatebox{90}{Pedestrian} & \rotatebox{90}{Cyclist} & \rotatebox{90}{Car} & \rotatebox{90}{mIoU} & \rotatebox{90}{hIoU}
 & \rotatebox{90}{Background} & \rotatebox{90}{Pedestrian} & \rotatebox{90}{Cyclist} & \rotatebox{90}{Car} & \rotatebox{90}{mPP} & \rotatebox{90}{hPP}
 & \rotatebox{90}{Background} & \rotatebox{90}{Pedestrian} & \rotatebox{90}{Cyclist} & \rotatebox{90}{Car} & \rotatebox{90}{mPR} & \rotatebox{90}{hPR} \\ 
\midrule
    \multirow{5}{*}{\textbf{\ac{RD}}} 
    
    & \multirow{2}{*}{\ac{FCN}-32s} & 99.6 (\NA)  &  16.8 (\NA)  &  3.2 (\NA)  &  27.5 (\NA)  &  36.8 (\NA)  &  8.1 (\NA)  &  99.6 (\NA)  &  69.4 \textbf{(17.3)}  &  5.4 (1.0)  &  64.2 \textbf{(17.2)}  &  59.7 (11.9)  &  13.8 (2.2)  &  99.9 (\NA)  &  20.3 (28.3)  &  6.7 (8.1)  &  32.3 (47.5)  &  39.8 (28.0)  &  13.3 (14.0)  \\
    
    & \multirow{2}{*}{\ac{FCN}-16s} &  99.6 (\NA)  &  28.9 (\NA)  &  7.2 (\NA)  &  42.1 (\NA)  &  44.5 (\NA)  &  17.2 (\NA)  &  99.7 (\NA)  &  64.7 (15.1)  &  18.0 (4.7)  &  67.6 (17.0)  &  62.5 (12.3)  &  40.3 (8.6)  &  99.9 (\NA)  &  39.2 (50.9)  &  10.7 (16.9)  &  54.1 (74.1)  &  51.0 (47.3)  &  23.4 (27.5)  \\
    
    & \multirow{2}{*}{\ac{FCN}-8s} &  99.7 (\NA)  &  \textbf{45.2} (\NA)  &  \textbf{15.5} (\NA)  &  \textbf{51.3} (\NA)  &  \textbf{52.9} (\NA)  &  \textbf{34.2} (\NA)  &  99.8 (\NA)  &  \textbf{72.3} (15.5)  &  \textbf{35.2} \textbf{(9.6)}  &  \textbf{69.8} (17.0)  &  \textbf{69.3} \textbf{(14.0)}  &  \textbf{59.8} \textbf{(13.1)}  &  99.9 (\NA)  &  \textbf{55.0} \textbf{(76.4)}  &  \textbf{22.1} \textbf{(35.7)}  &  \textbf{66.8} \textbf{(88.9)}  &  \textbf{60.9} \textbf{(67.0)}  &  \textbf{44.3} \textbf{(56.7)} \\

\midrule

\multirow{5}{*}{\textbf{\ac{RA}}} 
    
    & \multirow{2}{*}{\ac{FCN}-32s} &  99.8 (\NA) & 0.0 (\NA)  &  0.0 (\NA)  &  14.2 (\NA)  &  28.5 (\NA)  &  0.0 (\NA)  &  99.9 (\NA)  &  18.8 (6.1)  &  0.0 (0.0)  &  \textbf{69.3} \textbf{(13.7)}  &  47.0 (6.6)  &  0.0 (0.0)  &  100.0 (\NA)  &  0.0 (0.1)  &  0.0 (0.0)  &  15.5 (24.6)  &  28.9 (8.2)  &  0.0 (0.0)\\
    
    & \multirow{2}{*}{\ac{FCN}-16s} &  99.8 (\NA)  &  0.9 (\NA)  &  0.0 (\NA)  &  13.7 (\NA)  &  28.6 (\NA)  &  0.0 (\NA)  &  99.9 (\NA)  &  39.8 (5.2)  &  \textbf{0.2} (0.0)  &  68.0 (12.2)  &  \textbf{52.0} (5.8)  &  \textbf{0.9} (0.0)  &  100.0 (\NA)  &  0.9 (1.7)  &  0.0 (0.0)  &  15.4 (22.7)  &  29.1 (8.1)  &  0.0 (0.0)\\
    
    & \multirow{2}{*}{\ac{FCN}-8s} &  99.9 (\NA)  &  \textbf{5.5} (\NA)  &  0.0 (\NA)  &  \textbf{25.1} (\NA)  &  \textbf{32.6} (\NA)  &  \textbf{0.1} (\NA)  &  99.9 (\NA)  &  \textbf{42.2} \textbf{(10.0)}  &  0.1 (\textbf{0.1})  &  65.4 (12.4)  &  51.9 (\textbf{7.5})  &  0.6 (\textbf{0.2})  &  100.0 (\NA)  &  \textbf{6.3} \textbf{(11.3)}  &  0.0 (\textbf{0.1})  &  \textbf{30.0} \textbf{(45.5)}  &  \textbf{34.1} \textbf{(19.0)}  &  \textbf{0.1} \textbf{(0.3)} \\

\bottomrule
\end{tabularx}
 \caption[Semantic segmentation performances (\%) on the test dataset for Range-Doppler (RD) and Range-Angle (RA) representations]{\textbf{Semantic segmentation performances (\%) on the test dataset for Range-Doppler (RD) and Range-Angle (RA) representations}. Models are trained on dense mask annotations and evaluated on both dense mask (top values) and sparse points (bottom values in parentheses) annotations. Results are evaluated with \ac{IoU}, \ac{PP} and \ac{PR}. Metrics are computed by category and aggregated with both arithmetical (m) and harmonic (h) means. Lines (\NA) replacing values indicate non-applicable metrics, for example \ac{IoU} results on sparse annotations.}
\label{tab:carrada_fcn_baseline}
\end{table*}

Models are trained on dense mask annotations and evaluated on both dense mask (top values) and sparse points (bottom values in parentheses) annotations. 
Sparse points are more accurate than dense masks, therefore evaluation on this type of annotation provides information on the behaviour of predictions on key points. 
However, localization should not be evaluated for sparse points using a model trained on dense masks, therefore \ac{IoU} performances are not reported. 
The background category cannot be assessed for the sparse points because some of the points should belong to an object but are not annotated \textit{per se}.
Thus, arithmetic and harmonic means of sparse points evaluations are computed for only three classes against four for the dense masks. 

The baseline shows that meaningful representations are learnt by a popular 
visual semantic segmentation architecture. These models succeed in detecting and classifying shapes of moving objects in raw \ac{RADAR} representations even with sparse-point annotations. 
Performances on \ac{RA} are not as good as in \ac{RD} because the angular resolution of the sensor is low, resulting in less precise generated annotations. An extension to improve performances on this representation could be to transform it into Cartesian coordinates as done in \cite{major_vehicle_2019}.
For both representations, results are promising since the temporal dimension of the objects signatures has not yet been taken into account.

\subsection{Discussions}

The semi-automatic algorithm presented in Section \ref{sec:carrada_annot_pipeline} generates precise annotations on raw \ac{RADAR} data, but it has limitations. Occlusion phenomena are problematic for tracking, since they lead to a disappearance of the object point cloud in the \ac{DoA}-Doppler representation. An improvement could be to detect such occlusions in the video frames and include them in the tracking pipeline. The clustering in the \ac{DoA}-Doppler representation is also a difficult task in specific cases. When objects are close to each other with a similar radial velocity, point clouds are difficult to distinguish. Further work on the bandwidth selection and optimisation of this selection could be explored.

The CARRADA dataset provides precise annotations to explore a range of supervised learning tasks. 
A simple baseline is proposed for semantic segmentation trained on dense mask annotations. It could be extended by using temporal information or both dense mask and sparse points annotations at the same time during training. Current architectures and loss functions could also be optimized for semantic segmentation of sparse ground-truth points.
Object detection could be considered by using bounding boxes to detect and classify object signatures. 
As off-the-shelf object detection algorithms are not adapted to the \ac{RADAR} data representation and to the unusual size of the provided annotations, further work is required to redesign these methods.
By identifying and tracking specific instances of objects, other opportunities are opened.
Tracking of sparse points or bounding boxes could also be considered. 

\subsection{Conclusions}

\begin{figure}[!t]
\centering
\includegraphics[width=5.5in]{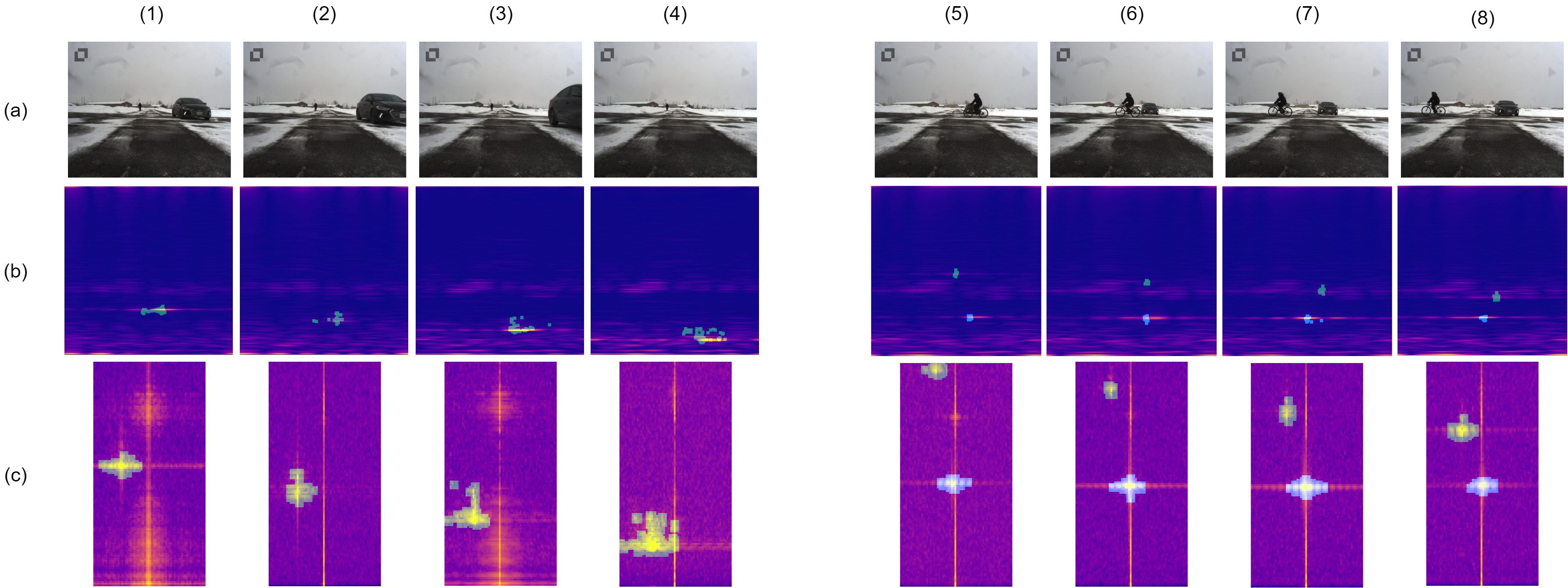}
\caption[Two scenes from CARRADA dataset, one with a car, the other on with a cyclist and a car]{\textbf{Two scenes from CARRADA dataset, one with a car, the other on with a cyclist and a car}.
(a) Video frames provided by the frontal camera showing moving objects in a fixed environment; (b-c) \ac{RADAR} signals at the same instants in \acl{RA} and cropped \acl{RD} representation  respectively. (1-4) First sequence; (5-8) Second sequence. Frames in both sequences are selected with an interval of 5 frames. In the first sequence, the segmentation mask corresponds to the annotation of the approaching car in the scene. Range-Doppler data (c)(1) and (c)(3) show that our method is robust to recording noise. In frames (4), the car is still in the radar's  field of view but it has disappeared from the camera. In the second sequence, the segmentation masks correspond respectively to the annotations of the moving cyclist (blue) and car (green). The cyclist is moving from right to left in front of the \ac{RADAR}, its radial velocity is progressively changing from positive to negative. Note that the \acl{RA} view illustrated in (b) has been obtained using a maximum aggregation and log-transform in Equation \ref{eq:general_agg_method} while the \acl{RD} view in (c) is computed using Equation \ref{eq:agg_method_average}.
}
\label{fig:carrada_temporal_ex}
\end{figure}

The CARRADA dataset contains synchronised video frames, \acl{RAD} tensor and thus, \acl{RA} and \acl{RD} raw \ac{RADAR} representations. \ac{RADAR} data are annotated with sparse points, bounding boxes and dense masks to localize and categorize the object signatures. A unique identification number is also provided for each instance. Annotations are generated using a semi-supervised algorithm based on visual and physical knowledge. 
The proposed semi-automatic pipeline generating the annotations is robust to \ac{RADAR} sensor noise and camera occlusion while being able to track an object in the \ac{RADAR} data which is not visible in the camera. These results are illustrated in Figure \ref{fig:carrada_temporal_ex}.
This pipeline could be used to annotate any camera-\ac{RADAR} recordings with similar settings. 
The dataset, code for the annotation algorithm and code for dataset visualisation are publicly available \footnote{\url{https://github.com/valeoai/carrada\_dataset}}.
Even if the recorded scenes contain only one or two objects moving in a controlled environment, the dataset is unique in its kind by providing raw \ac{RADAR} data with detection and semantic segmentation annotations.
This work also aims to motivate deep learning research applied to \ac{RADAR} sensor and multi-sensor fusion for scene understanding as proposed in the following chapter.

\section{Conclusions}

This chapter has outlined methods to address the lack of open and annotated \ac{RADAR} data.
We presented a simple simulation of objects' signature in \ac{RD} representations by considering the geometric properties of a moving object in a scene and incorporating different types of noise (Section \ref{sec:rd_simulation}). The proposed simulator is too simple to generate realistic samples. As our experiments show, deep neural network architectures are capable of easily over-fitting the simulation.

Secondly, we proposed methods to generate \ac{RD} representations from camera images using real data recorded in complex urban scenes (Section \ref{sec:radar_generation}). Our proposed \ac{AE} architecture combined with the ``Noise Map'' method, reached the best quantitative and qualitative results. However, the results were not convincing enough to further explore the transfer of annotations from the source domain (camera image) to the target domain (\ac{RD} representation).

Finally, we presented CARRADA, a unique dataset including synchronized camera images and raw \ac{RADAR} data with object detection and semantic segmentation annotations (Section \ref{sec:carrada}). We also proposed a semi-automatic pipeline based on camera images generating annotations on \ac{RADAR} representations while being robust to object occlusion, narrow camera \ac{FoV} and \ac{RADAR} sensor noise, as illustrated in Figure \ref{fig:carrada_temporal_ex}.
The CARRADA dataset and the presented annotation method were presented at the International Conference on Pattern Recognition (ICPR).

The proposed dataset contains simple scenes with only one or two moving objects at the same time. However, it is an interesting source of data for designing deep neural network architectures, while being realistic enough to be generalized to complex urban scenes, as we detail in the next chapter. The annotation pipeline is also limited in distinguishing two objects with similar Doppler while being close together. Furthermore, it requires a human intervention to instantiate the tracking method. This is a preliminary work on automatic \ac{RADAR} annotation generation, which has been extended in the recent work of \cite{zhang_raddet_2021}, as detailed in Section \ref{sec:mvrss_raddet_dataset}.

In the following chapter, we will present methods for \ac{RADAR} scene understanding. First, we will propose deep neural architectures for multi-view \ac{RADAR} semantic segmentation with their corresponding loss functions. Then, we will introduce a method for \ac{LiDAR} and \ac{RADAR} point cloud fusion to take advantage of both sensors for scene understanding applications.

\chapter{RADAR scene understanding}
\label{chap:radar_scene_understanding}
\minitoc




The \ac{RADAR} sensor is recently a source of interest since public datasets have been released as detailed in Section \ref{sec:related_datasets}.
\ac{RADAR} scene understanding is in its infancy, but it provides key information to compensate for weaknesses of the other sensors. The \ac{RADAR} data presented as a \ac{RAD} tensor contains signatures of the objects surrounding the car with enough details to distinguish them in the representation. 
Unlike object detection using bounding boxes, semantic segmentation is appropriate for this task, since the object signatures have extremely variable sizes and may be mixed up due to the sensor's resolution.

Multi-sensor fusion is essential to improve scene understanding by taking advantage of complementary sensor properties.
\ac{LiDAR} and \ac{RADAR} point clouds are easy to fuse in the Cartesian coordinates around the ego vehicle. Even if the \ac{RADAR} point cloud is degraded through the signal processing pipeline (see Section \ref{sec:background_signal_process}), it offers reflectivity and velocity information useful to localise objects in the dense \ac{LiDAR} dense representation. 

In this chapter, we propose a method for \ac{RADAR} scene understanding.
First, a multi-view approach is proposed for \ac{RADAR} semantic segmentation in Section \ref{sec:mvrss} details neural network architectures and their associated loss functions adapted to the \ac{RAD} tensor. 
This section is the second main contribution of this thesis, it is mainly inspired from our article published at the International Conference on Computer Vision (ICCV) \cite{ouaknine_multi-view_2021}.
Section \ref{sec:sensor_fusion} introduces a method to fuse and propagate \ac{RADAR} point cloud properties through the \ac{LiDAR} point cloud aiming to improve scene understanding tasks.


\section{Multi-view RADAR semantic segmentation}
\label{sec:mvrss}

\begin{figure}[t]
   \includegraphics[width=1.0\linewidth]{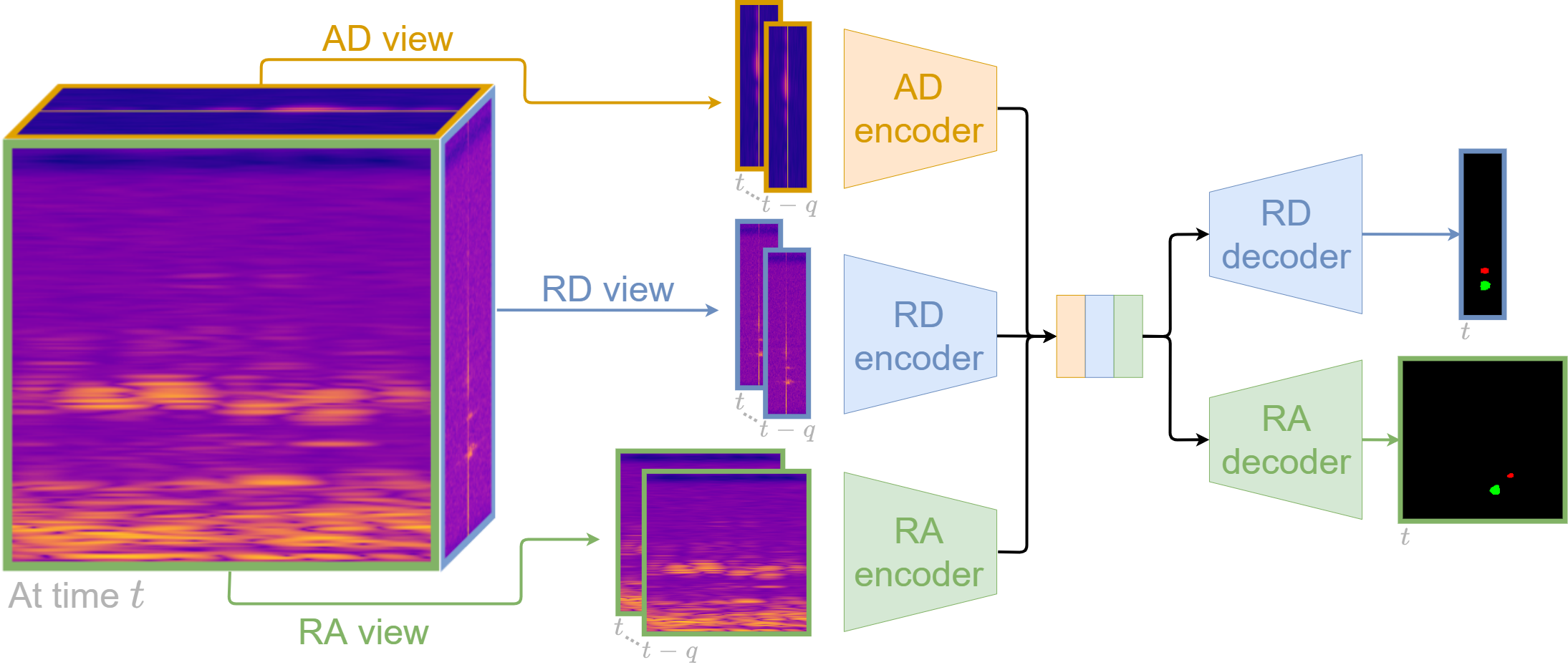}
   \caption[Overview of our multi-view approach to semantic segmentation of RADAR signal]{\textbf{Overview of our multi-view approach to semantic segmentation of RADAR signal}. At a given instant, \ac{RADAR} signals take the form of a \acl{RAD} tensor. Sequences of $q+1$ 2D views of this data cube are formed and mapped to a common latent space by our proposed multi-view architectures. Two heads with distinct decoders produce a semantic segmentation of the \acl{RA} and \acl{RD} views respectively (`background' in black, `pedestrian' in red and `cyclist' in green in this example). 
   }
\label{fig:mvrss_teaser}
\end{figure}

\subsection{Motivations}

In this section, an approach to multi-view \ac{RADAR} semantic segmentation is proposed, illustrated in Figure \ref{fig:mvrss_teaser}, that exploits the entire data  while addressing the challenges of its large volume and high level of noise (see Appendix \ref{sec_app:mvrss_rad_tensor_vis}). 
The segmentation is performed on the \ac{RD} and \ac{RA} views, which suffices to deduce the localisation and the relative speed of objects.
The first contribution is a set of lightweight neural network architectures designed for multi-view semantic segmentation of \ac{RADAR} signal (see Section \ref{sec:mvrss_archi_ours}. The second contribution is a set of loss terms to train models on these tasks while preserving coherence between the multi-view predictions (see Section \ref{sec:mvrss_losses}). 
Experiments in Sections \ref{sec:mvrss_expe_carrada} and \ref{sec:mvrss_expe_urban} demonstrate that our proposed best model outperforms alternative models, derived either from the semantic segmentation of natural images or from \ac{RADAR} scene understanding, while requiring significantly fewer parameters. Both our code and trained models are available online\footnote{\url{https://github.com/valeoai/MVRSS}}. Section \ref{sec:mvrss_conclusions} finally concludes and proposes perspectives.

\subsection{Methods and architectures}
\label{sec:mvrss_methods_and_archis}

The following Sections \ref{sec:mvrss_img_based_methods} and \ref{sec:mvrss_radar_based_methods} briefly present several methods for image segmentation and \ac{RADAR} scene understanding, to which this work is compared. They are chosen for their performance and their relevance to the \ac{RADAR} semantic segmentation task. Further details on the architectures are provided in Sections \ref{sec:background_segmentation} and \ref{sec:related_segmentation}. 
Except RSS-Net \cite{kaul_rss-net_2020}, 
these architectures were not originally designed for \ac{RADAR} semantic segmentation, nor to handle multiple views of a data volume.
Consequently, for each of them, two models have been trained independently for \ac{RD} and \ac{RA} segmentation respectively. More details are provided in Section \ref{sec:mvrss_method_modif}.

The three proposed architectures for multi-view \ac{RADAR} semantic segmentation are introduced in Section \ref{sec:mvrss_archi_ours}.

\subsubsection{Image-based competing methods}
\label{sec:mvrss_img_based_methods}

Long \etal \cite{long_fully_2015} propose \ac{FCN}, consisting of convolutional and down-sampling layers followed by transposed convolutions (``up-convolutions''). The final representations are processed by a 1D convolution with softmax to predict a category for each pixel. 
Several versions are proposed 
depending on the feature-map scales used to generate the output.
\ac{FCN} has been used for semantic segmentation of \ac{RADAR} data in Section \ref{sec:carrada_baseline}, where FCN-8s version achieves the best performance.

The U-Net architecture \cite{ronneberger_u-net_2015} is composed of equal-depth down-sampling and up-sampling pathways linked by skip connections.
Originally used for medical images, it is especially well suited for small-object segmentation.

The DeepLabv3+ \cite{chen_encoder-decoder_2018} is a popular encoder-decoder model for semantic segmentation of natural images. 
The encoder uses ``atrous'' separable convolutions which increase the receptive field of the network. The proposed \ac{ASPP} layer \cite{chen_deeplab_2018} combines atrous convolutions with various dilation rates to learn multi-scale features followed by a 1D convolution.

These methods are presented in detail in Section \ref{sec:background_segmentation}.

\subsubsection{Radar-based competing methods}
\label{sec:mvrss_radar_based_methods}

Kaul \etal \cite{kaul_rss-net_2020} propose RSS-Net, specialised in \ac{RADAR} semantic segmentation, in particular for \ac{RA} representations. Its architecture is similar to DeepLabv3+ with an encoder composed of 8 atrous convolutional layers, an \ac{ASPP} module and a convolutional decoder with up-sampling.
The architecture is designed to reduce the dimension of the feature maps in the encoder, leading to excellent performance for \acl{RA} \ac{BEV} semantic segmentation. 

Gao \etal \cite{gao_ramp-cnn_2020} propose the \ac{RADAR} Multiple-Perspective Neural Network (RAMP-CNN) for object detection in \ac{RA} representations. They aggregate the \ac{RAD} tensor into 2D \ac{RADAR} views which are processed by separate encoder-decoders with 3D convolutional layers.
The final \ac{RA} features are processed by a 3D inception module. RAMP-CNN achieves state of the art performance in localising and classifying multiple objects in \ac{RA} views.

These methods are presented in detail in Section \ref{sec:related_segmentation}.

\subsubsection{Proposed multi-view methods}
\label{sec:mvrss_archi_ours}

\begin{figure}[t!]
\centering
\includegraphics[width=5.5in]{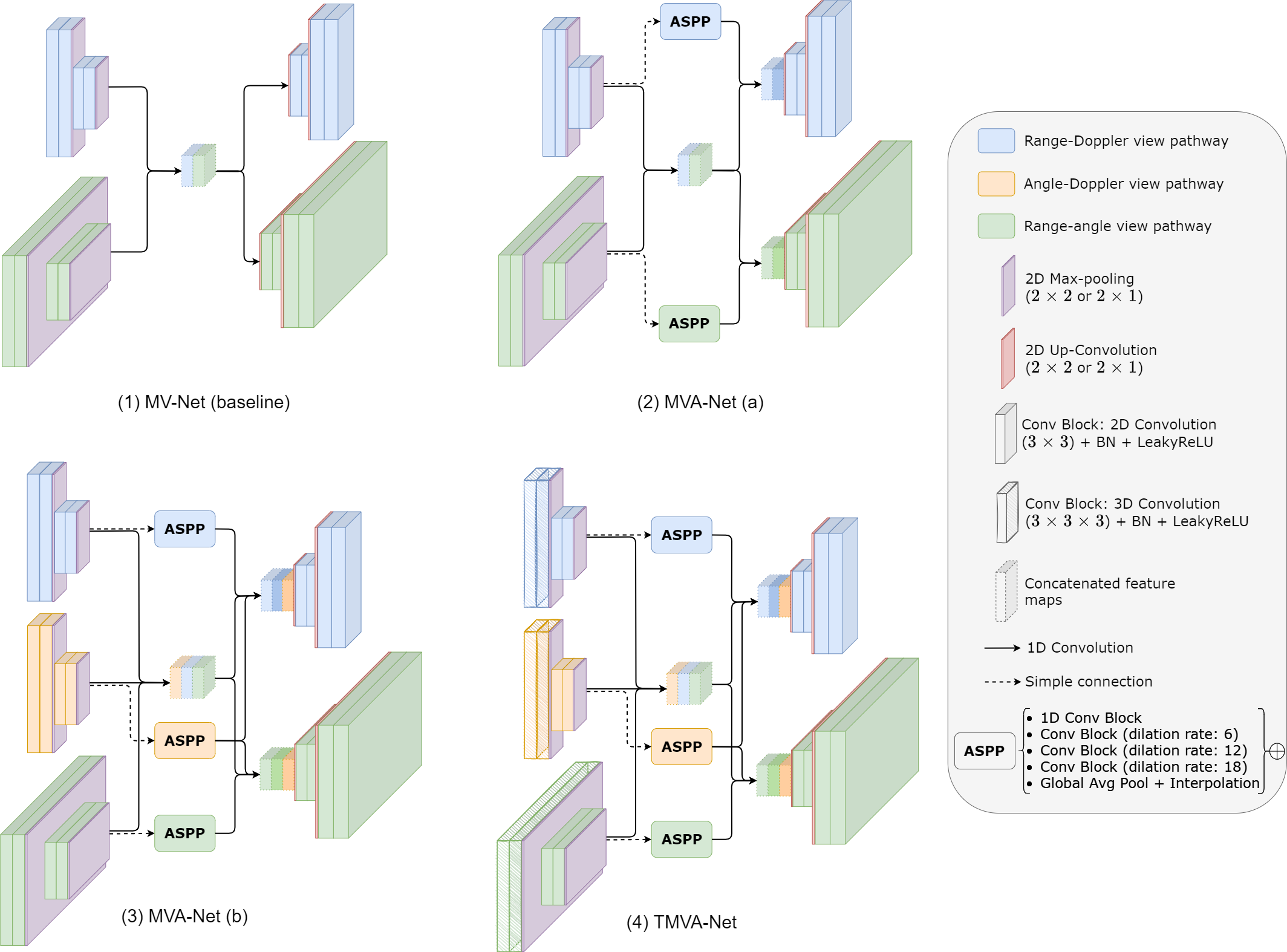}
\caption[Multi-view architectures for RADAR semantic segmentation]{\textbf{Multi-view architectures for RADAR semantic segmentation}. 
    (1) The multi-view network (MV-Net), considered as a baseline, is composed of two encoders, two decoders and a common latent space. 
    (2) The MVA-Net(a) has an additional \acl{ASPP} module for each view pathway;
    (3) The MVA-Net(b) has an additional \acl{AD} pathway;
    (4) The TMVA-Net architecture is similar to the MVA-Net(b) with 3D convolutions at the top of the encoders exploiting the temporal dimension.
    The detailed architectures are provided in Appendix \ref{sec_app:mvrss_archi_details}.}
\label{fig:mvrss_architectures}
\end{figure}

We proposed four lightweight neural network architectures. They are specialised in multi-view \ac{RADAR} semantic segmentation, whose general principle is illustrated in Figure \ref{fig:mvrss_architectures}. 
They take a temporal stack of \ac{RADAR} views as their input and process them with dedicated encoders.
The generated feature maps are fused in a shared latent space from which different decoders predict semantic segmentation maps for each output view.
Since neither encoders nor decoders share weights, they are kept simple to reduce the size of the total network. 
Thus, there are specialised parts of the network in each view, and a reasonable number of parameters altogether. The proposed architectures are detailed layer by layer in Appendix \ref{sec_app:mvrss_archi_details}.

\smallskip\noindent\textbf{Multi-view network (MV-Net).~}  Firstly, we proposed a baseline (see Figure \ref{fig:mvrss_architectures} (1)) in the form of a double encoder-decoder architecture that processes stacked \ac{RD} and \ac{RA} views and predicts simultaneously the \ac{RD} and \ac{RA} semantic segmentation maps. It is important to note that the third view (\ac{AD}) can be deduced from the \ac{RA} and \ac{RD} views, and it is therefore not necessary to segment it in the output. This is the case for all proposed architectures.
 
Each encoder is composed of two blocks, each one with two sequences of convolution, batch normalisation and activation layers. The two blocks are separated by a max-pooling operation to down-sample the feature maps along the range axis (the Doppler's resolution being lower, it is kept unchanged).
The feature maps from both encoders are down-sampled, processed by a 1D convolution and stacked into a common latent space. Since the input views are stacked according to the time dimension, this linear combination of the feature maps aims to learn temporal correlations.
The features in the shared latent space are then processed with 1D convolution layers and used as the input of each decoder.
There are two decoders predicting respectively the \ac{RD} and \ac{RA} semantic segmentation maps. Each one is composed of two blocks with two sequences of convolution, batch normalisation and activation layers. The up-sampling between the blocks is carried out by up-convolutions. A final 1D convolution performs a combination of the outputs of each decoder to generate $K$ feature maps, where $K$ is the number of classes. A softmax operation is then applied pixel-wise to the $K$ feature maps generating soft masks.

\smallskip\noindent\textbf{Multi-view network with \ac{ASPP} modules (MVA-Net).~} 
The \ac{ASPP} module \cite{chen_deeplab_2018} used in DeepLabv3 \cite{chen_encoder-decoder_2018} allows features to be jointly learned at different scales at a given depth of the network with no need for larger kernels or additional parameters. As shown in RSS-Net \cite{kaul_rss-net_2020}, it is well suited for \ac{RADAR} semantic segmentation since the objects' signatures can vary considerably.
The second architecture, MVA-Net, introduces the use of this \ac{ASPP} module at the end of each decoder of our MV-Net baseline.  
The generated multi-scale feature maps are concatenated, processed by a 1D convolution
and stacked to the input of each corresponding decoder.
Two variants of MVA-Net are proposed: MVA-Net(a), shown in Figure \ref{fig:mvrss_architectures} (2), consists in a double encoder-decoder architecture with \ac{ASPP} modules to process and segment \ac{RD} and \ac{RA} views; MVA-Net(b), shown in Figure \ref{fig:mvrss_architectures} (3), has an additional encoding branch learning features from the \ac{AD} view. 
Similarly to the other encoding branches, the \ac{AD} backbone generates features that are stacked in the common latent space. However, the outputs of its \ac{ASPP} module contain both the angle and Doppler information. Thus the multi-scale feature maps from the AD pathway are stacked to the inputs of both the \ac{RD} and \ac{RA} decoders.

\smallskip\noindent\textbf{Temporal multi-view network with \ac{ASPP} modules (TMVA-Net).~} 
The temporal dimension provides valuable information for \ac{RADAR} semantic segmentation. It helps in estimating the shape of an object's signature despite high noise levels, and distinguishing objects that are close together with similar velocities.
The third architecture, which will show to have the best performances, TMVA-Net in Figure \ref{fig:mvrss_architectures} (4), extends MVA-Net by explicitly leveraging the temporal dimension. 
The first block's 2D convolutions are replaced by 3D convolutions in each encoder branch, making it able to learn the spatio-temporal characteristics with limited increase in the number of parameters. Since 3D convolutions require a large number of parameters, full 3D-convolutional encoders, such as in \cite{gao_ramp-cnn_2020}, have not been retained.
Hence, TMVA-Net is composed of three encoders with 3D and 2D convolutions, one for each input view.
Each one of them has a dedicated \ac{ASPP} module. The feature maps generated from each encoding backbone are stacked into a shared latent space. From there, two decoders segment respectively the \ac{RD} and \ac{RA} views. They take as input the stacked features from the processed latent space and the multi-scale feature maps from the dedicated \ac{ASPP} modules.

\subsubsection{Losses}
\label{sec:mvrss_losses}

In what follows, the following generic notation are used: $f_\theta(\vect{x})\!=\!\vect{p}$ for a segmentation model with parameters $\theta$, input $\vect x$ and output $\vect p$. Training $f_{\theta}$ amounts to minimising w.r.t. $\theta$ a suitable loss function, given training examples $\vect x$ with ground truth $\vect y$.  
The architectures presented in Sections \ref{sec:mvrss_img_based_methods} and \ref{sec:mvrss_radar_based_methods} take single-view inputs stacked in the temporal dimension and predict, for each target view, a soft segmentation mask with class ``probabilities" for each bin. 
For instance, the output of a model processing only the \ac{RA} view is $f_\theta(\vect{x}^{\text{RA}})\!=\! \vect{p}^{\text{RA}}\!\in\![0,1]^{\BinR \times \BinA \times K}$, if $\BinR{\times}\BinA$ is the size of the view and $K$ the number of classes.
Our architectures, detailed in Section \ref{sec:mvrss_archi_ours}, take instead multi-view inputs: either $\vect{x}\!=\!(\vect{x}^{\text{RD}}, \vect{x}^{\text{RA}})$ or $\vect{x}\!=\!(\vect{x}^{\text{RD}}, \vect{x}^{\text{AD}}, \vect{x}^{\text{RA}})$. 
In both cases, their goal is to predict soft masks $\vect{p}\!=\!(\vect{p}^{\text{RD}}, \vect{p}^{\text{RA}})$ for both \ac{RD} and \ac{RA} views.

The following section details the loss functions applied to each segmented view to train our proposed architectures. A ``coherence'' loss is also introduced to enforce consistency between the predictions over the two views of the scene. 
Finally a combination of these loss terms is proposed.

\smallskip\noindent\textbf{Weighted Cross Entropy.~}
\label{sec:mvrss_wce}
Semantic segmentation models that predict a score for each class at each pixel are usually trained by minimising a \ac{CE} loss function.
This loss is not ideal for unbalanced segmentation tasks such as the \ac{RADAR} semantic segmentation, since the optimisation process tends to focus on the classes that are most represented. 
In the present case, background and speckle noise dominate, in comparison to the signatures of the objects we wish to detect.
In the same manner as RSS-Net \cite{kaul_rss-net_2020}, a \ac{wCE} loss is employed to tackle this issue.

Given a training example $\vect x$, let $\vect{y}\in \{0,1\}^{\BinM \times \BinN \times K}$ be its one-hot ground truth and $f_\theta(\vect{x}) = \vect{p} \in [0,1]^{\BinM \times \BinN \times K}$ the associated prediction where $\BinM{\times}\BinN$ is the size of the view $\vect x$. 
The \ac{wCE} loss function is defined as:
\begin{equation}
\mcl{L}_{\text{wCE}} (\vect{y}, \vect{p}) = - \frac{1}{K} \sum_{k=1}^{K} w_k \!\!\!\!\sum_{(m,n)\in\Omega}\!\!\!\!\vect{y}[m,n,k] \log \vect{p}[m,n,k],
\label{eq:mvrss_lossWCE}
\end{equation}
where $\Omega = \llbracket 1,\BinM\rrbracket \times \llbracket 1,\BinN\rrbracket$,
and $w_k$'s are normalized positive weights. Weight $w_k$ is inversely proportional to the frequency of class $k$ in the training set, that is $w_k\propto \big(\sum_{\vect y}\sum_{(m,n)\in\Omega}\vect{y}[m,n,k]\big)^{-1}$. The fewer the bins with ground-truth class $k$, the larger $w_k$ becomes. 

\smallskip\noindent\textbf{Soft Dice.~}
\label{sec:mvrss_sdice}
Object signatures in \ac{RADAR} representations often correspond to small regions. This is a well known issue in medical image segmentation, where the Dice metric (detailed in Section \ref{sec:mvrss_carrada_data_metrics}) is usually reformulated in a function called Dice loss, ranging between 0 and 1. 
In their work, \cite{milletari_v-net_2016} have proposed the \ac{SDice} loss defined as:
\begin{equation}
\mcl{L}_{\text{SDice}} = \frac{1}{K} \sum_{k=1}^{K} \Bigg[ 1 - \frac{2 \sum_{(m,n)} \vect{y}[m,n,k] \vect{p}[m,n,k]}{\sum_{(m,n)} \vect{y}^2[m,n,k] + \vect{p}^2[m,n,k]} \Bigg],
\end{equation}
where $(m,n) \in \Omega$, 
as in Equation \ref{eq:mvrss_lossWCE}. 
This formulation has proved useful for 2D and 3D medical image semantic segmentation, including for small objects. 

\begin{figure}[!t]
\begin{center}
\includegraphics[width=0.7\linewidth]{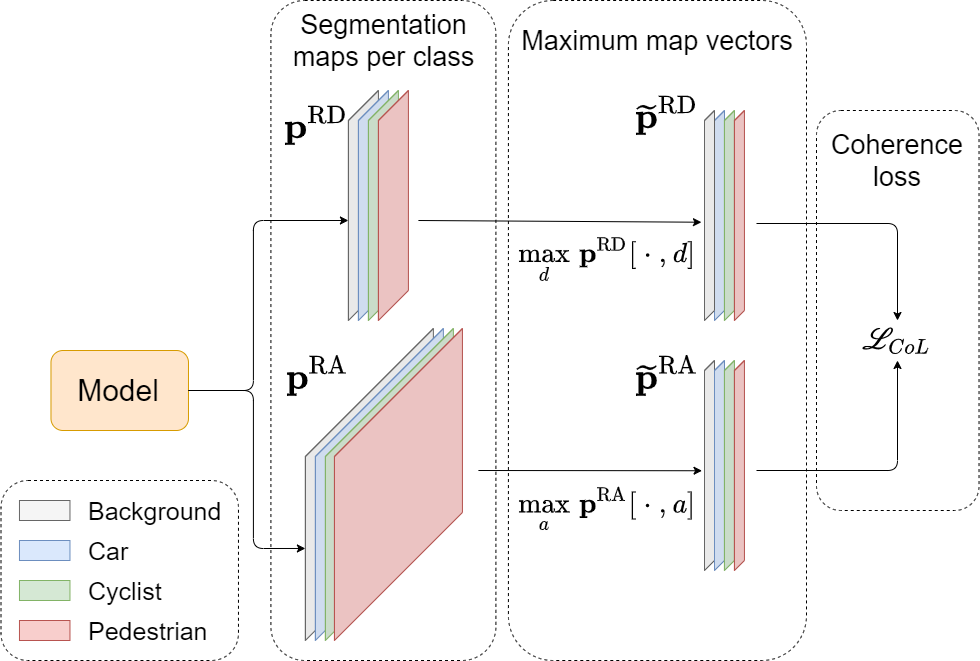}
\end{center}
   \caption[Computation of the coherence loss]{\textbf{Computation of the coherence loss}. The segmentation maps $\vect{p}^{\text{RD}}$ and $\vect{p}^{\text{RA}}$ of the two views are aggregated by max pooling along the axis
   that they do not share (either the Doppler or the angle). The coherence loss is the \acl{MSE} between the two resulting vectors $\tilde{\vect{p}}^{\text{RD}}$ and $\tilde{\vect{p}}^{\text{RA}}$.}
\label{fig:mvrss_coherence_loss}
\end{figure}

\smallskip\noindent\textbf{Coherence.~}
\label{sec:mvrss_col}
The objective of multi-view \ac{RADAR} semantic segmentation is to simultaneously segment several views of the aggregated \ac{RAD} tensor. 
The objects we wish to detect are observed in the different \ac{RADAR} views, thus it is clear that a certain coherence must be maintained between the segmented views. For example, one view should not represent a pedestrian, while another represents a cyclist. A \ac{CoL} is introduced to preserve a consistency between the predictions of the model. The procedure to calculate this loss is illustrated in Figure \ref{fig:mvrss_coherence_loss}. 

Let $(\vect{p}^{\text{RD}}, \vect{p}^{\text{RA}})$ be the segmentation maps predicted by the model $f_\theta$ after the softmax operation. These two maps are aggregated by applying a $\max(.)$ operator along the axis that they do not share (either the Doppler or the angle). The two resulting maps of same size, denoted  $\tilde{\vect{p}}^{\text{RD}}$ and $\tilde{\vect{p}}^{\text{RA}}$, contain the highest probability of each range bin for each class. In other words, they indicate if the model predicts a high probability to observe a category at a given distance. The coherence loss is the \ac{MSE} between these maximum range probability vectors. It encourages the network to predict high probability values at the same distance and in the same class for both views. The \ac{CoL}, in the interval $[0,1]$, is defined as:
\begin{equation}
\mcl{L}_{\text{CoL}}(\vect{p}^{\text{RD}}, \vect{p}^{\text{RA}}) = \frac{1}{B_\text{R} \cdot K} \big \lVert \tilde{\vect{p}}^{\text{RD}} -
\tilde{\vect{p}}^{\text{RA}}\big \rVert^2_{\mathrm{F}},
\end{equation}
where $\|\cdot\|_{\mathrm{F}}$ denotes the Frobenius norm, $\BinR$ the number of range bins and $K$ the number of classes.

\smallskip\noindent\textbf{Combination of losses.~}
\label{sec:mvrss_combi_loss}
The \ac{CE} loss is specialised in pixel-wise classification and does not consider spatial correlations between the predictions. The \ac{SDice} is particularly effective for shape segmentation, but it is difficult to optimise as a single loss function due to its gradient formulation.
Finally, the \ac{CoL} is useful where neither the \ac{CE} nor the \ac{SDice} is able to leverage a coherence between the prediction of the \ac{RD} and the \ac{RA} views. To combine the different strengths of these losses, the following final loss is proposed to train multi-view architectures:
\begin{equation}
\mcl{L} = \lambda_{\text{wCE}} (\mcl{L}_{\text{wCE}}^{\text{RD}} + \mcl{L}_{\text{wCE}}^{\text{RA}}) + \lambda_{\text{SDice}} (\mcl{L}_{\text{SDice}}^{\text{RD}} + \mcl{L}_{\text{SDice}}^{\text{RA}}) + \lambda_{\text{CoL}} \mcl{L}_{\text{CoL}},
\end{equation}
where $\lambda_{\text{wCE}}$, $\lambda_{\text{SDice}}$ and $\lambda_{\text{CoL}}$ are weighting factors set empirically.

\subsection{Experiments on the CARRADA dataset}
\label{sec:mvrss_expe_carrada}

This section presents the experimental evaluation of our proposed models on the CARRADA dataset described in Section \ref{sec:carrada}.
The dataset and the evaluation metrics are briefly described.
Modification made to the competing methods are then explained.
Finally, details are provided concerning the experiments and analyse their results quantitatively and qualitatively.

\subsubsection{Dataset and evaluation metrics}
\label{sec:mvrss_carrada_data_metrics}

\smallskip\noindent\textbf{Dataset.}
The CARRADA dataset contains synchronised camera and automotive \ac{RADAR} recordings with 30 sequences of various scenarios with one or two moving objects.
The \ac{RADAR} views are annotated using a semi-automatic pipeline (see Section \ref{sec:carrada_annot_pipeline}). This is the only publicly-available dataset providing \ac{RAD} tensors and dense semantic segmentation annotation for both \ac{RD} and \ac{RA} views. The objects are separated into four categories: \textit{pedestrian}, \textit{cyclist},  \textit{car} and \textit{background}. 
The provided \ac{RAD} tensors have dimensions $\BinR{\times}\BinA{\times}\BinD = 256{\times} 256{\times} 64$.
Additional details on the dataset are provided in Section \ref{sec:carrada_dataset}.
The experiments presented in Section \ref{sec:mvrss_carrada_train_and_res} use the proposed dataset splits (see Figure \ref{fig:carrada_dataset_distrib}, denoted CARRADA-Train, CARRADA-Val and CARRADA-Test.

\smallskip\noindent\textbf{Evaluation metrics.~} 
A classic performance metric in semantic segmentation is the \ac{IoU}
This metric has been introduced in Section \ref{sec:background_detection_metrics}.
From the perspective of a single object in a given scene, the \ac{IoU} measures how well and how completely it is segmented. 
Averaging this metric over all classes yields the \ac{mIoU} score.
Another related, yet slightly different metric, is provided by the Dice score: For a given class and with same notations as above, it is defined as $\frac{2|A \cap B|}{|A| + |B|}$. For global performance, it is averaged over all classes into the \ac{mDice}. Seeing segmentation as a local 1-\textit{vs}.-all classification problem for each class, the Dice amounts to the harmonic mean of the precision and recall (a.k.a. F1 score).
The \ac{IoU} and Dice metrics are considered as complementary; Both of them are reported in our experiments.

\subsubsection{Implementation of competing methods}
\label{sec:mvrss_method_modif}

This section describes the architectures used for comparisons. 
For each method, one model is trained specifically for single-view semantic segmentation of either \ac{RD} or \ac{RA}. Details concerning pre-processing procedures are provided in the Appendix \ref{sec_app:mvrss_preproc_procedures}.

\sloppy
The non-radar-based architectures have been used ``as is'':
The FCN-8s architecture is based on a VGG16 \cite{simonyan_very_2015} backbone; 
DeepLabv3+ uses a ResNet-101 \cite{he_deep_2016}; 
The U-Net architecture is identical to the one 
in \cite{ronneberger_u-net_2015}. 

In the experiments with RSS-Net, the number of down-sampling layers in the encoding part has been reduced to be trained with lower resolution inputs. 

The RAMP-CNN architecture dedicated to the \ac{RD} view has been adapted with two major changes. Firstly, the fusion module has been modified to aggregate and duplicate the feature maps to suit the \ac{RD} space. Secondly, the size of the output feature maps has been reduced on the Doppler axis using an additional convolutional layer with $3 {\times} 3$ filters and a stride factor of 4.
For both \ac{RD} and \ac{RA} segmentation tasks, an additional 1D convolutional layer with a softmax operation processes the last feature maps to predict segmentation maps.

\subsubsection{Training and results}
\label{sec:mvrss_carrada_train_and_res}

\begin{table}[t]
\def\arraystretch{1.1}
\setlength\tabcolsep{3pt}
\scriptsize
\begin{tabularx}{\textwidth}{c @{\hskip 0.1in} c @{} r @{\hskip 0.2in} CCCC|C| CCCC|C @{}}
\toprule
 \multirow{5}{*}{View} & \multirow{5}{*}{Method} & \multirow{5}{*}{\# Param. (M)} & \multicolumn{5}{c}{\ac{IoU} (\%)} &  \multicolumn{5}{c}{Dice (\%)} \\
 \cmidrule(lr{3pt}){4-8}
 \cmidrule(l{3pt}r{3pt}){9-13}
 & & & \rotatebox{90}{Bkg.} & \rotatebox{90}{Ped.} & \rotatebox{90}{Cyclist} & \rotatebox{90}{Car} & \rotatebox{90}{\textbf{\ac{mIoU}}} 
 & \rotatebox{90}{Back.} & \rotatebox{90}{Ped.} & \rotatebox{90}{Cyclist} & \rotatebox{90}{Car} & \rotatebox{90}{\textbf{\ac{mDice}}} \\ 
\midrule
    \multirow{9}{*}{\textbf{\ac{RD}}} 
    & FCN-8s \cite{long_fully_2015} &  134.3~~ & 99.7 & 47.7 & 18.7 & 52.9 &  54.7  &  99.8 & 24.8 & 16.5 & 26.9 &  66.3 \\
& U-Net \cite{ronneberger_u-net_2015} &  17.3~~  & 99.7 & \underline{\textcolor{blue}{51.0}} & \textbf{\textcolor{red}{33.4}} & 37.7   &  55.4   &   99.8 & \underline{\textcolor{blue}{67.5}} & \textbf{\textcolor{red}{50.0}} & 54.7  &  68.0 \\
    & DeepLabv3+ \cite{chen_encoder-decoder_2018} &  59.3~~  & 99.7 & 43.2 & 11.2 & 49.2   &  50.8   &   99.9 & 60.3 & 20.2 & 66.0  &  61.6 \\
    & RSS-Net &  10.1~~  &  99.3 & 0.1 & 4.1 & 25.0 &  32.1   &   99.7 & 0.2 & 7.9 & 40.0 &  36.9 \\
& RAMP-CNN &  106.4~~  &  99.7 & 48.8 & 23.2 & \underline{\textcolor{blue}{54.7}} &  \underline{\textcolor{blue}{56.6}}   &   99.9 & 65.6 & 37.7 & \underline{\textcolor{blue}{70.8}}  &  \underline{\textcolor{blue}{68.5}} \\
    & MV-Net (ours-baseline) &  2.4*  & 98.0 & 0.0 & 3.8 & 14.1  &  29.0  &   99.0 & 0.0 & 7.3 & 24.8  &  32.8 \\
    & MVA-Net (a) (ours) & 3.6* & 99.7 & 26.5 & 20.7 & 48.8 & 48.9 & 99.8 & 41.9 & 34.3 & 65.6 & 60.4 \\
    & MVA-Net (b) (ours) & 4.8* & 99.7 & 30.2 & 22.0 & \textbf{\textcolor{red}{59.6}} & 52.9 & 99.9 & 46.4 &  36.1 & \textbf{\textcolor{red}{74.7}} & 64.3 \\

& TMVA-Net (ours) &  5.6*  & 99.7 & \textbf{\textcolor{red}{52.6}} & \underline{\textcolor{blue}{29.0}} & 53.4 & \textbf{\textcolor{red}{58.7}} & 99.8 & \textbf{\textcolor{red}{68.9}} & \underline{\textcolor{blue}{45.0}} & 69.6 & \textbf{\textcolor{red}{70.9}} \\
    \midrule

\multirow{9}{*}{\textbf{\ac{RA}}} 
    & FCN-8s \cite{long_fully_2015} &  134.3~~  &  99.8 & 14.8 & 0.0 & 23.3 &  34.5  &  99.9 & 25.8 & 0.0 & 37.8  &  40.9 \\
    & U-Net \cite{ronneberger_u-net_2015} &  17.3~~ & 99.8 & \underline{\textcolor{blue}{22.4}} & \textbf{\textcolor{red}{8.8}} & 0.0  &  32.8   &   99.9 & \underline{\textcolor{blue}{36.6}} & \textbf{\textcolor{red}{16.1}} & 0.0  &  38.2 \\
    & DeepLabv3+ \cite{chen_encoder-decoder_2018} &  59.3~~  & 99.9 & 3.4 & 5.9 & 21.8  &  32.7  &   99.9 & 6.5 & 11.1 & 35.7  &  38.3 \\
& RSS-Net &  10.1~~  & 99.5 & 7.3 & 5.6 & 15.8 &  32.1   &   99.8 & 13.7 & 10.5 & 27.4  &  37.8 \\
    & RAMP-CNN &  106.4~~  &  99.8 & 1.7 & 2.6 & 7.2   &  27.9   &   99.9 & 3.4 & 5.1 & 13.5  &  30.5 \\
    & MV-Net (ours-baseline) &  2.4*  & 99.8 & 0.1 & 1.1 & 6.2  &  26.8  & 99.0 & 0.0 & 7.3 & 24.8 &  28.5 \\
    & MVA-Net (a) (ours) & 3.6* & 99.0 & 0.0 & 5.9 & 7.5 & 28.1 & 99.5 & 0.0 & 11.1 & 14.0 & 31.1 \\
    & MVA-Net (b) (ours) & 4.8* & 99.8 & 6.6 & 7.4 & \textbf{\textcolor{red}{32.9}} & \underline{\textcolor{blue}{36.7}} & 99.9 & 12.5 & 13.8 & \textbf{\textcolor{red}{49.5}} & \underline{\textcolor{blue}{43.9}} \\

& TMVA-Net (ours) &  5.6*  & 99.8 & \textbf{\textcolor{red}{26.0}} & \underline{\textcolor{blue}{8.6}} & \underline{\textcolor{blue}{30.7}} & \textbf{\textcolor{red}{41.3}} & 99.9 & \textbf{\textcolor{red}{41.3}} & \underline{\textcolor{blue}{15.9}} & \underline{\textcolor{blue}{47.0}} & \textbf{\textcolor{red}{51.0}} \\
\bottomrule
\end{tabularx}
\vspace{2pt}
 \caption[Semantic segmentation performance on the CARRADA-Test dataset for Range-Doppler (RD) and Range-Angle (RA) views)]{\textbf{Semantic segmentation performance on the CARRADA-Test dataset for \acl{RD} and \acl{RA} views}.
 The number of trainable parameters (in millions) for each method corresponds to a single view-segmentation model; Two such models, one for each view, are required for all methods but ours. In contrast, the number of parameters reported for our methods (`*') corresponds to a single model that segments both \ac{RD} and \ac{RA} views.
 The RSS-Net and RAMP-CNN methods have been modified to be trained on both tasks (see Section \ref{sec:mvrss_method_modif}). Performances are evaluated with the \acl{IoU} and the Dice score per class, and their averages, \ac{mIoU} and \ac{mDice}, over the four classes.
 The best scores are in red and bold type, the second best in blue and underlined.}
 \label{table:mvrss_main_quanti_results}
\end{table}

\smallskip\noindent\textbf{Training procedures.~}
Methods presented in Section \ref{sec:mvrss_methods_and_archis} are trained using CARRADA-Train and CARRADA-Val splits and tested on the CARRADA-Test. At each timestamp of a \ac{RADAR} sequence, the provided \ac{RAD} tensor is processed according to the method presented in Section \ref{sec:background_signal_process}.

For each frame, $q$ past frames are also considered for both training and testing phases: The views from $t-q$ to $t$ are stacked into the time-$t$ input (see Figure \ref{fig:mvrss_teaser}).
For the methods that do not explicitly process the time dimension, $q=2$ for a total input sequence length of 3. Time-based methods using 3D convolutions have specific sequence lengths: $q=8$ for RAMP-CNN and $q=4$ for TMVA-Net.

The competing architectures have been trained with the \ac{CE} loss, except for the RSS-Net, which is trained with a \ac{wCE} using the formulation in \cite{kaul_rss-net_2020}. Our proposed methods are trained using the combination of loss terms detailed in Section \ref{sec:mvrss_combi_loss}. Our proposed formulation of the \ac{wCE} loss has been used (Section \ref{sec:mvrss_wce}) with weights computed on CARRADA-Train.

All the training procedures use the Adam optimiser \cite{kingma_adam_2015} with the recommended parameters ($\beta_1 = 0.9$, $\beta_2 = 0.999$ and $\varepsilon=10^{-8}$). Since each method has its own set of hyper-parameters, further details are provided in the Appendix \ref{sec_app:mvrss_preproc_procedures}, namely batch sizes, learning rates, learning rate decays, numbers of epochs and corresponding pre-processing steps for each one of them. Training was performed using the PyTorch framework with a single GeForce RTX 2080 Ti graphic card.

\begin{figure}[t]
\centering
\includegraphics[width=0.8\linewidth]{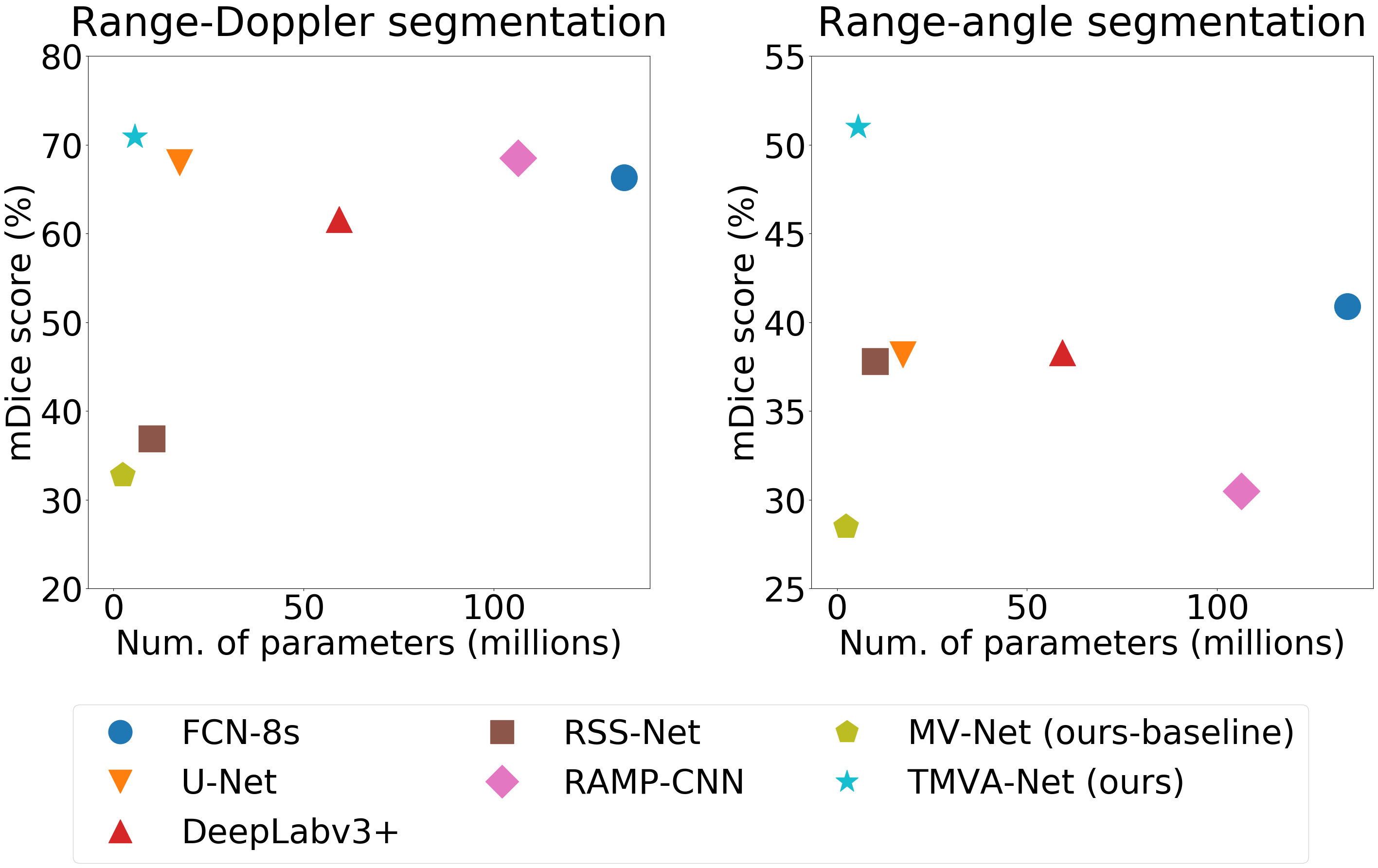}
    \caption[Performance-\textit{vs}.-complexity plots for all methods in Range-Doppler (RD) and Range-Angle (RA) tasks]{\textbf{Performance-\textit{vs}.-complexity plots for all methods in \acl{RD} and \acl{RA} tasks}. Performance is assessed by \acl{mDice} (\%) and complexity by the number of parameters (in millions) \textit{for a single task}. 
    Top-left models are the best performing and the lightest. 
    Only our proposed models, MV-Net and TMVA-Net, are able to segment both views simultaneously. For all the other methods, two distinct models must be trained to address both tasks, which doubles the number of actual parameters.}
\label{fig:mvrss_perf_params}
\end{figure}

\smallskip\noindent\textbf{Quantitative results.~}
The performance for both \ac{RD} and \ac{RA} semantic segmentation tasks on CARRADA-Test are shown in Table \ref{table:mvrss_main_quanti_results}. Our proposed TMVA-Net
achieves the best scores for both \ac{mDice} and \ac{mIoU} metrics and for both segmentation tasks.
Moreover, the proposed methods are the only ones to perform both tasks simultaneously. 
TMVA-Net also provides the best trade-off between 
performance and number of parameters for both tasks, as illustrated in Figure \ref{fig:mvrss_perf_params} with \ac{mDice} metric (similar plots with \ac{mIoU} metric are presented in Appendix \ref{sec_app:mvrss_quanti_results}). 
A study of performance variability considering four trained models for each method leads to the same conclusion as detailed in Appendix \ref{sec_app:mvrss_variability_method}.
Note that the number of parameters reported for each method in Table \ref{table:mvrss_main_quanti_results}, in Figure \ref{fig:mvrss_perf_params} and in Appendix \ref{sec_app:mvrss_quanti_results} corresponds to a single trained model, while competing methods require two independent models (hence twice more parameters) to perform both \ac{RD} and \ac{RA} segmentation tasks.

\smallskip\noindent\textbf{Ablation Studies.~} 
Table \ref{table:mvrss_ablation_architectures} reports the performance of the four architectures which have been introduced in Section \ref{sec:mvrss_archi_ours}. It shows that the additional \ac{ASPP} modules in MVA-Net(a) boosts the performance relative to MV-Net for both \ac{RD} and \ac{RA} segmentation. 
The performance is further improved with MVA-Net(b) by the  additional encoder that extracts features from the \ac{AD} view and provides relevant information to separate object signatures. 
Finally, TMVA-Net is the most effective regardless of the metric, thanks to its ability to learn spatio-temporal features with 3D convolutions. The temporal dimension indeed helps distinguish between objects and the speckle noise, and to categorise them according to the shape variations.

Table \ref{table:mvrss_ablation_losses} accesses the importance of the different losses on the performance of TMVA-Net.
The best combination of two loss terms is wCE$+$SDice for both tasks. The performance is further improved on the \ac{RA} segmentation task by adding our proposed \ac{CoL} term, while slightly reduced on \ac{RD} views. This loss improves the coherence between the tasks by better detecting objects in the \ac{RA} views as discussed in the following section.
A loss ablation study considering the performance variability is proposed in Appendix \ref{sec_app:mvrss_variability_losses}. Four models have been trained for each combination of loss functions leading to the same conclusion.

\begin{table}
\scriptsize
\begin{minipage}[b]{.5\linewidth}
\begin{center}
\resizebox{.99\textwidth}{!}{
\begin{tabular}{c l r c c }
\toprule
View & Method & \# Param. & \ac{mIoU} & \ac{mDice} \\
\midrule
\multirow{4}{*}{\textbf{\ac{RD}}} 
& MV-Net (baseline) & 2.4M & 29.0 & 32.8 \\
& MVA-Net (a) & 3.6M & 48.9 & 60.4 \\
& MVA-Net (b) & 4.8M & 52.9 & 64.3\\
& TMVA-Net & 5.6M & \textbf{59.3} & \textbf{71.5}\\
\midrule
\multirow{4}{*}{\textbf{\ac{RA}}} 
& MV-Net (baseline) & 2.4M & 26.8 & 28.5 \\
& MVA-Net (a) & 3.6M & 28.1 & 31.1 \\
& MVA-Net (b) & 4.8M & 36.7 & 43.9\\
& TMVA-Net & 5.6M & \textbf{40.1} & \textbf{49.3}\\
\bottomrule
\end{tabular}}
\end{center}
\caption[Ablation study of our proposed architectures]{\textbf{Ablation study of our proposed architectures}. Each architecture has been trained using 
the wCE$+$SDice combination loss.
TMVA-Net delivers the best performances under both \ac{mIoU} and \ac{mDice} metrics and for both \ac{RD} and \ac{RA} views.} 
\label{table:mvrss_ablation_architectures}
\end{minipage}
\qquad
\begin{minipage}[b]{.5\linewidth}
\begin{center}
\resizebox{.98\textwidth}{!}{
\scriptsize
\begin{tabular}{lcccc}
\toprule
     & \multicolumn{2}{c}{\ac{RD} view} & \multicolumn{2}{c}{\ac{RA} view} \\
     \cmidrule(lr){2-3} \cmidrule(lr){4-5}
Loss & \ac{mIoU} & \ac{mDice} & \ac{mIoU} & \ac{mDice} \\
\midrule
CE & 56.1 & 67.8 &  39.1 & 48.3 \\
SDice & 58.5 & 70.3 & 37.1 & 44.8\\
wCE & 51.1 & 62.8 & 34.3 & 41.1  \\
CE$+$SDice & 45.2 & 54.0 & 38.8 & 46.9 \\
wCE$+$SDice & \textbf{59.3} & \textbf{71.5} & \underline{40.1} & \underline{49.3} \\
wCE$+$SDice$+$CoL & \underline{58.7} & \underline{70.9} &  \textbf{41.3} & \textbf{51.0}\\
\bottomrule
\end{tabular}}
\end{center}
\caption[Ablation study of the combination of losses]{\textbf{Ablation study of the combination of losses}. Each individual or combination of loss(es) is used to train a TMVA-Net model. Our proposed combination (wCE$+$SDice$+$CoL) reaches the best \ac{mIoU} and \ac{mDice} for the \ac{RA} view and the second best scores for the \ac{RD} view.}
\label{table:mvrss_ablation_losses}
\end{minipage}
\end{table}

\begin{figure}[t]
\begin{center}
\includegraphics[width=1\linewidth]{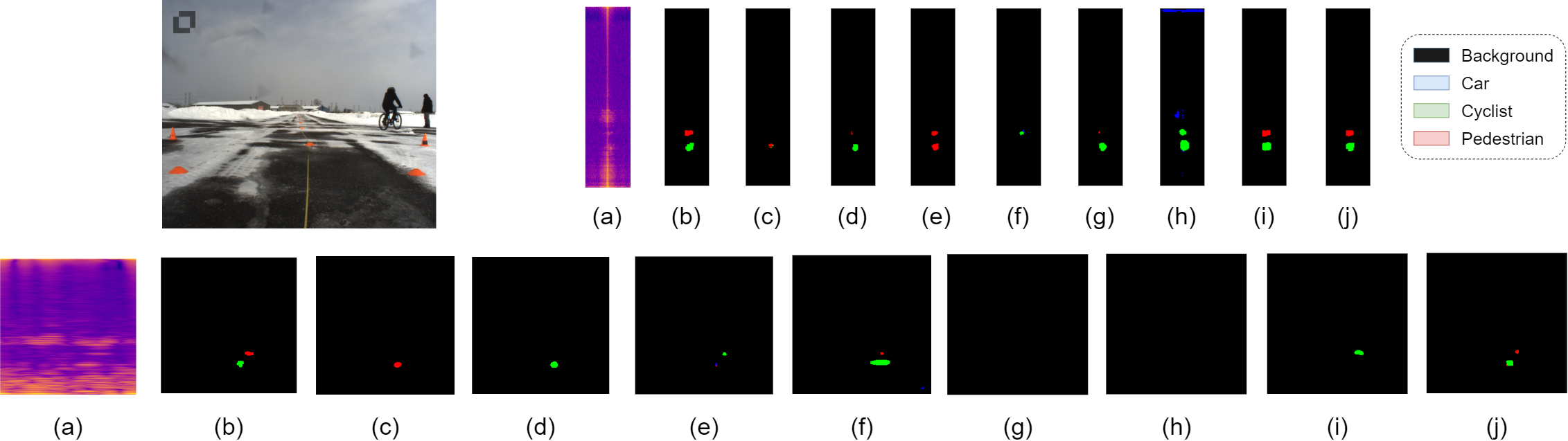}
\end{center}
   \caption[Qualitative results on a test scene of CARRADA]{\textbf{Qualitative results on a test scene of CARRADA}. (\textit{Top}) camera image of the scene and results of the \acl{RD} segmentation; (\textit{Bottom}) Results of the \acl{RA} Segmentation. (a) \ac{RADAR} view signal, (b) ground-truth mask, (c) FCN8s, (d) U-Net, (e) DeepLabv3+, (f) RSS-Net, (g) RAMP-CNN, (h) MV-Net (our baseline w/ wCE$+$SDice loss), (i) TMVA-Net (ours, w/ wCE$+$SDice loss), (j) TMVA-Net (ours, w/ wCE$+$SDice$+$CoL loss).}
\label{fig:mvrss_quali_results}
\end{figure}

\smallskip\noindent\textbf{Qualitative results.~} Figure \ref{fig:mvrss_quali_results} shows qualitative results of each method on a scene from CARRADA-Test. The results of TMVA-Net (i-j) display well segmented \ac{RD} views in terms of localisation and classification. Only TMVA-Net with \ac{CoL} (j) is able to localise and classify both objects in the \ac{RD} and \ac{RA} views. 
The enforcement of the coherence of predictions across views succeeds in correctly classifying the same objects in the two views. This is not the case for TMVA-Net without \ac{CoL}, as illustrated in the example (i), where the model predicts a cyclist instead of a pedestrian in the \ac{RA} view. Moreover, the coherence loss also helps to discover new objects: In (i), TMVA-Net predicts a single object in the \ac{RA} view, while in (j),  it localises and classifies both objects well with the help of \ac{CoL}.
Additional qualitative results leading to the same conclusions are proposed in Appendix \ref{sec_app:mvrss_quali_results_carrada}.

\subsection{Experiments on complex urban scenes datasets}
\label{sec:mvrss_expe_urban}

This section presents additional experiments in complex urban scenes highlighting the suitability of our proposed TMVA-Net architecture, trained with its corresponding loss function. Quantitative and qualitative results are presented using the RADDet dataset. A qualitative evaluation is also proposed using an in-house dataset without annotations.

\subsubsection{RADDet dataset}
\label{sec:mvrss_raddet_dataset}

In their work, \cite{zhang_raddet_2021} have proposed a dataset with synchronised \ac{RADAR} and stereo cameras, a method to annotate the \ac{RADAR} data and a neural network architecture for \ac{RADAR} object detection. To the best for our knowledge, it is the only automotive \ac{RADAR} dataset providing annotated views of the \ac{RAD} tensor in complex urban scenes.

\smallskip\noindent\textbf{Dataset.~}
The dataset consists in 10 158 frames of synchronised \ac{RADAR} and stereo cameras mounted on a stationary car. The scenarios consist in complex urban scenes (crossroads, sidewalks, crowded roads) recorded in Canada. 
The authors have used a \textit{Texas Instruments} AWR1843-BOOST \footnote{\url{www.ti.com}} \ac{RADAR} sensor as in the CARRADA dataset (see Section \ref{sec:carrada_dataset}). The authors have publicly released \ac{RAD} tensors of size $\BinR {\times} \BinA {\times} \BinD = 256 {\times} 256 {\times} 64$ for each frame.
They have proposed an annotation method to detect dynamic road users on \ac{RADAR} representations similarly to the method presented in Section \ref{fig:carrada_annot_methods}. First, a Mask R-CNN \cite{he_mask_2017} segments objects in a single camera image. A Semi-Global Block Matching method \cite{hirschmuller_stereo_2008} estimates the image depth with stereo cameras. The Cartesian coordinates of each object are then deduced using the calibration matrix of the cameras.

A CFAR algorithm \cite{rohling_radar_1983} is applied on the \ac{RD} and \ac{RA} views of the \ac{RAD} tensor to extract a point cloud of high intensity values. The point cloud of each \ac{RADAR} view is is then clustered with a DB-SCAN algorithm \cite{ester_density-based_1996}. The object instances from the stereo cameras are projected to a \ac{RADAR} \ac{BEV} representation using a projection matrix. The objects are associated to the corresponding clustered point cloud. Finally, the bounding box annotations are deduced from the \ac{RADAR} point cloud associated to each object.

The annotation pipeline proposed by \cite{zhang_raddet_2021} extends our proposed method presented in Section \ref{sec:carrada_annot_pipeline} by using the depth estimated by stereo cameras. It helps to reduce the projection error from the camera to the Cartesian coordinates.
However, their method has a major limitation: it relies on the camera images at each timestamp. It is therefore affected by any ambiguity of camera projection or occlusion problems.
A future work could combine both methods to better instantiate the annotation from the camera images and track it in the \ac{RADAR} sequences.
In the RADDet dataset, the selected $10,158$ frames correspond to robust annotations.

\begin{figure}[t!]
\centering
\includegraphics[width=1\textwidth]{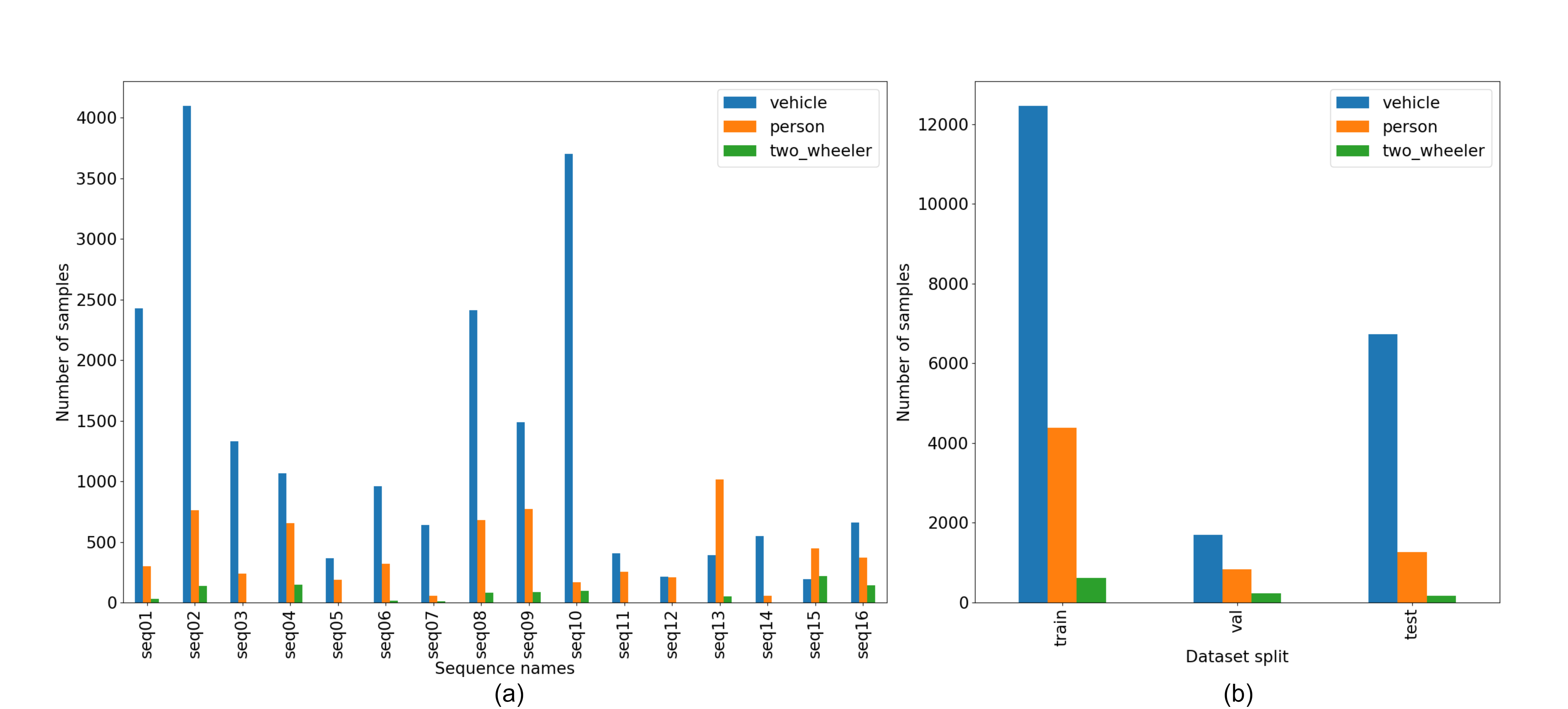}
\caption[RADDet dataset distributions]{\textbf{RADDet dataset distributions}. 
    (a) Distribution of each category by sequence; (b) Distribution of each category by proposed dataset split (training, validation and test).}
\label{fig:mvrss_raddet_distribs}
\end{figure}

The dataset is composed of six unbalanced categories: person, bicycle, car, motorcycle, bus and truck. 
In the following experiments, classes are grouped to compensate the balance between the classes.
Using only three classes as in the CARRADA dataset helps to train and evaluate the architectures proposed in Section \ref{sec:mvrss_archi_ours}; and competing methods in Sections \ref{sec:mvrss_img_based_methods} and \ref{sec:mvrss_radar_based_methods}. The bicycle and motorcycle categories have been merged to create the two-wheeler class, while the car, bus and truck categories have been reunited in the vehicle class.
In their work, \cite{zhang_raddet_2021} neither considered the temporal dimension to build the dataset nor to train their proposed object detection method. The RADDet dataset has been manually rearranged in sequences to suit the training settings detailed in Section \ref{sec:mvrss_carrada_train_and_res}. The distributions of the grouped categories considering the proposed sequentially splitted RADDet dataset are illustrated in Figure \ref{fig:mvrss_raddet_distribs} (a). The distributions of the proposed training, validation and test splits of the sequential RADDet dataset are illustrated in Figure \ref{fig:mvrss_raddet_distribs} (b).

\begin{figure}[t]
\begin{center}
\includegraphics[width=1\linewidth]{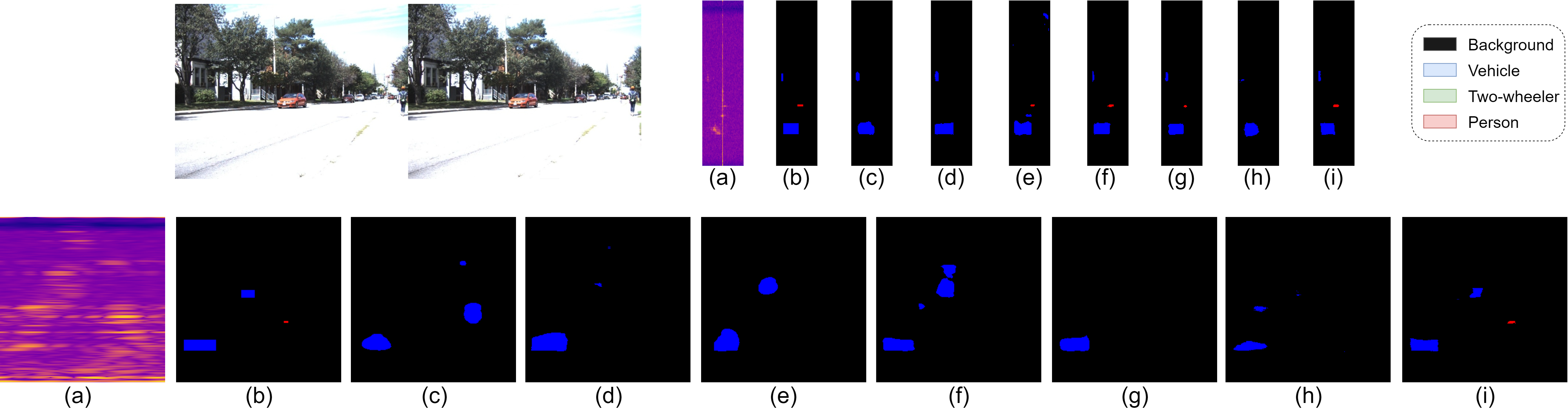}
\end{center}
   \caption[Qualitative results on a test scene of RADDet]{\textbf{Qualitative results on a test scene of RADDet}. (\textit{Top}) camera image of the scene and results of the \acl{RD} segmentation; (\textit{Bottom}) Results of the \acl{RA} Segmentation. (a) \ac{RADAR} view signal, (b) ground-truth mask, (c) FCN8s, (d) U-Net, (e) DeepLabv3+, (f) RSS-Net, (g) RAMP-CNN, (h) MV-Net (our baseline w/ wCE$+$SDice$+$CoL loss), (i) TMVA-Net (ours, w/ wCE$+$SDice$+$CoL loss).}
\label{fig:mvrss_quali1_raddet}
\end{figure}

The RADDet dataset contains bounding box annotations for each object in the RAD tensor and thus, in the \ac{RD} and \ac{RA} views. In order to perform \ac{RADAR} semantic segmentation, the entire bounding box is considered as a dense segmentation annotation as illustrated in Figure \ref{fig:mvrss_quali1_raddet}. There are multiple drawbacks due to this hypothesis: the annotations do not correspond to the object signature in the \ac{RADAR} representation (it is not rectangular), the annotation contains speckle noise and the model will be penalized if it does not succeed to estimate a rectangle even if the mask covers the signature.
However, the CARRADA and the RADDet datasets are the only publicly avaiable datasets providing \ac{RAD} tensors with object-wise annotations \footnote{Note that the methods presented in Sections \ref{sec:mvrss_img_based_methods}, \ref{sec:mvrss_radar_based_methods} and \ref{sec:mvrss_archi_ours} are not able to scale with \ac{HD} \ac{RADAR} regarding the size of the \ac{RAD} tensor and the computational cost to estimate it. The following experiments can not be applied to the RADIal dataset presented in Chapter \ref{chap:hd_radar}.}. 
\ac{RADAR} semantic segmentation experiments applied to the RADDet dataset are therefore considered as relevant to evaluate our proposed methods in complex urban scenes.

\smallskip\noindent\textbf{Experiments.~}
The same training procedures as detailed in Section \ref{sec:mvrss_carrada_train_and_res} have been followed. The \ac{RAD} tensors of the RADDet dataset have been aggregated in \ac{RADAR} views according to Equation \ref{eq:agg_method_average}. The methods presented in Sections \ref{sec:mvrss_img_based_methods}, \ref{sec:mvrss_radar_based_methods} and \ref{sec:mvrss_archi_ours} have been trained according to their corresponding loss function and set of hyper-parameters detailed in Appendix \ref{sec_app:mvrss_preproc_procedures}, namely batch sizes, learning rates, learning rate decays, numbers of epochs and corresponding pre-processing steps.


\begin{table}
\scriptsize
\begin{center}
\scriptsize
\begin{tabular}{lrcccc}
\toprule
     \multirow{2}{*}{Method} & \multirow{2}{*}{\# Params. (M)} & \multicolumn{2}{c}{\ac{RD} view} & \multicolumn{2}{c}{\ac{RA} view} \\
     \cmidrule(lr){3-4} \cmidrule(lr){5-6}
 &  & mIoU & mDice & mIoU & mDice \\
\midrule
FCN-8s \cite{long_fully_2015} & 134.3 & 51.3 & 58.5 & 31.4 & 36.0 \\
U-Net \cite{ronneberger_u-net_2015} & 7.3& 49.6 & 57.1 & \underline{41.8} & \underline{49.3}\\
DeepLabv3+ \cite{chen_deeplab_2018} & 59.3& 49.0 & 55.8 & 37.9 & 43.6  \\
RSS-Net & 10.1 & \textbf{54.1} & \underline{60.9} & 39.1 & 43.8 \\
RAMP-CNN & 106.4& 44.8 & 50.6 & 39.2 & 45.2 \\
MV-Net & 2.4*& 41.3 & 47.8 & 32.6 & 37.8\\
TMVA-Net & 5.6*& \underline{54.0} & \textbf{61.8} &  \textbf{44.2} & \textbf{51.4}\\
\bottomrule
\end{tabular}
\end{center}
\caption[Semantic segmentation performance on the RADDet-Test dataset for Range-Doppler and Range-Angle views]{\textbf{Semantic segmentation performance on the RADDet-Test dataset for \acl{RD} and \acl{RA} views}. 
The number of trainable parameters (in millions) for each method corresponds to a single view-segmentation model; Two such models, one for each view, are required for all methods but ours. In contrast, the number of parameters reported for our methods (`*') corresponds to a single model that segments both \ac{RD} and \ac{RA} views.
 The RSS-Net and RAMP-CNN methods have been modified to be trained on both tasks (see Section \ref{sec:mvrss_method_modif}). Performances are evaluated with the \ac{IoU} and the Dice score per class, and their averages, \ac{mIoU} and \ac{mDice}, over the four classes.
 The best scores are bold type, the second best are underlined.}
\label{table:mvrss_raddet_quanti}
\end{table}

\smallskip\noindent\textbf{Results.~}
Quantitative results on the RADDet-Test dataset are presented in Table \ref{table:mvrss_raddet_quanti}. Our TMVA-Net architecture, trained with our proposed combination of loss functions (wCE$+$SDice$+$CoL), performs the best according to both \ac{mIoU} and \ac{mDice} metrics on the \ac{RA} view segmentation, and to the \ac{mDice} on the \ac{RD} view segmentation. Performances according to the \ac{mIoU} metric on the \ac{RD} view segmentation are similar than the U-Net method \cite{ronneberger_u-net_2015}. 
However, our proposed method are the only one to perform both task simultaneously.
As mentioned in Section \ref{sec:mvrss_carrada_train_and_res}, TMVA-Net still provides the best trade-off between performance on the RADDet dataset and number of parameters for both tasks.

Figure \ref{fig:mvrss_quali1_raddet} shows qualitative results of each method trained on the RADDet-Train and RADDet-Validation datasets, and tested on a scene from RADDet-Test.
Once again, the TMVA-Net trained with \ac{CoL} (i) is the only method to correctly segment two vehicles and a pedestrian in both \ac{RD} and \ac{RA} views. The spatial coherence is enforced between the two views enabling the detection of the pedestrian in the \ac{RA} view.
Qualitative results on additional urban scenes are provided in Appendix \ref{sec_app:mvrss_quali_results_raddet}. These results leads to the same conclusions, our proposed method succeeds to segment multiple objects with a spatial coherence consistency outperforming competing methods while performing both tasks simultaneously.

\subsubsection{In-house dataset}

\smallskip\noindent\textbf{Dataset.~}
For qualitative evaluation only, an in-house dataset has been employed. A few sequences have been recorded by Valeo team members in cities in Canada with a stationary car. The dataset is composed of complex urban scenes with synchronised camera and \ac{RADAR} data. It has not been released publicly.
For these sequences, the \ac{RAD} tensors have the same dimensions as in CARRADA while the resolution in range is divided by two. The \ac{RADAR} views are not annotated, which does not allow quantitative evaluation.

\smallskip\noindent\textbf{Experiments.~} The \ac{RADAR} views of this dataset are not annotated and can not be used for training. The methods presented in Sections \ref{sec:mvrss_img_based_methods}, \ref{sec:mvrss_radar_based_methods} and \ref{sec:mvrss_archi_ours} have been trained on the CARRADA-Train dataset and tested on the unannotated views of the in-house dataset. The objective is to qualitatively assess the generalisation capacity in complex urban scenes of our proposed methods trained on a simple and controlled environment.

\smallskip\noindent\textbf{Qualitative results.~}
Qualitative results on these additional urban scenes are provided in Appendix \ref{sec_app:mvrss_quali_results_sc}.
They show that our proposed methods unlike others can generalise well on \ac{RD} and \ac{RA} views. 
Indeed, TMVA-Net succeeds in learning object signatures on the CARRADA dataset and recognizing them in a different environment. It also succeeds in detecting and classifying several objects in the scenes, although it has been trained to segment a maximum of two objects at a time.

\subsection{Conclusions and perspectives}
\label{sec:mvrss_conclusions}

In this section, lightweight architectures are proposed for multi-view \ac{RADAR} semantic segmentation and a combination of loss terms to train them. Our proposed methods localise and delineate objects in the \ac{RADAR} scene while simultaneously determining their relative velocity. Experiments show that both the information from the \ac{RAD} \ac{RADAR} tensor and from its temporal evolution are important for these tasks.
The proposed methods significantly outperform competing architectures specialised either in natural image semantic segmentation or in \ac{RADAR} scene understanding using the CARRADA dataset. 
The experiments conducted on the RADDet dataset support these conclusions in complex urban scenes with multiple objects to detect.
Preliminary experiments on the in-house dataset also show qualitatively that our method, trained on CARRADA only, generalizes better to new complex urban scenes without fine-tuning.

Our future investigations will focus on improving the segmentation of cyclists and pedestrians, which remain difficult to distinguish. 
Exploiting \ac{RADAR} properties could be interesting to improve both \ac{RAD} tensor aggregation and class-specific data augmentation methods for the benefit of learning algorithms.
We note that experiments considering the entire \ac{RAD} tensor (with or without the temporal information) as input of a neural network architecture were unsuccessful due to the large amount of noise.
As a first step, we are exploring aggregation methods applied to the \ac{RAD} to better separate the object signature distributions while reducing the variance of the speckle noise in the \ac{RADAR} data. The selection of specific slices per view could also be considered with a measure of signal disparity. 
We are also trying to reformulate the \acl{CoL} by integrating the ground truth, or by back-propagating the information in specific branches of the network. 

The \ac{RADAR} sensor has benefits, \textit{e.g.} it provides the Doppler information and it is robust to adverse weather conditions. It has also limitations, \textit{e.g.} low angular resolution and ghost reflections. 
An ideal framework for scene understanding is to use multiple sensors to accumulate their benefits while compensating for their limitations.
The next section will present preliminary works exploring senor fusion between \ac{LiDAR} and \ac{RADAR} to benefit from both sensor properties.

\section{Sensor fusion}
\label{sec:sensor_fusion}

\begin{figure}[t]
   \includegraphics[width=1.0\linewidth]{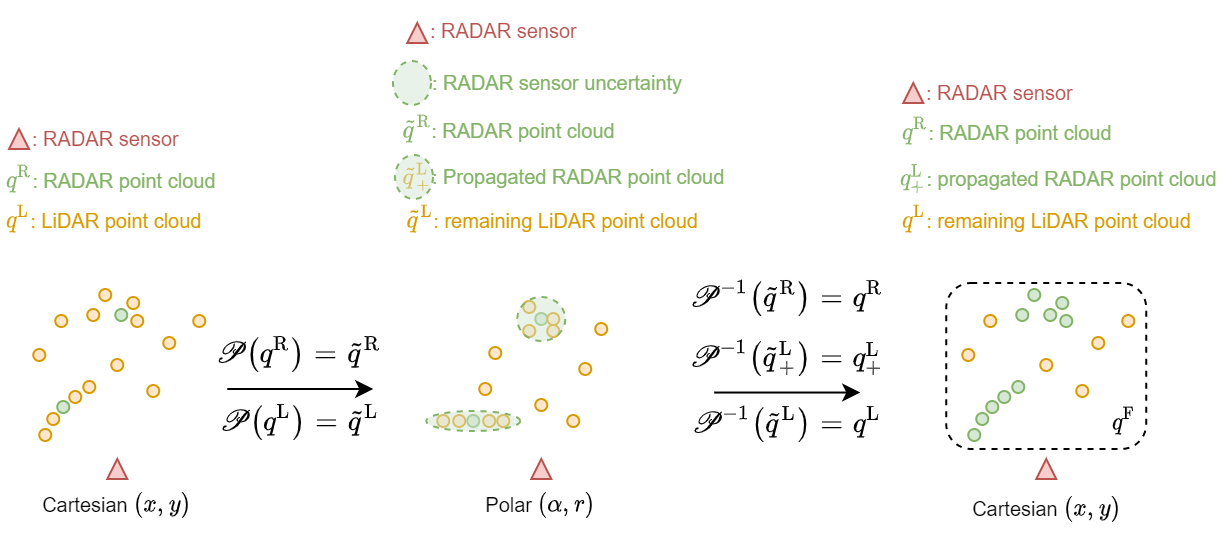}
   \caption[Overview of our propagation and fusion approach for RADAR and LiDAR point clouds]{\textbf{Overview of our propagation and fusion approach for RADAR and LiDAR point clouds}. The \ac{RADAR} ($\vect{q}^\text{R}$) and \ac{LiDAR} ($\vect{q}^\text{L}$) point clouds in \acl{BEV} Cartesian coordinates are projected in polar coordinates (respectively $\tilde{\vect{q}}^\text{R}$ and $\tilde{\vect{q}}^\text{L}$) with $\mathcal{P}(.)$ (see Eq. \ref{eq:polar_projection}). \ac{LiDAR} points are associated to each \ac{RADAR} point according to the sensor uncertainties illustrated with green ellipses ($\vect{q}_{+}^\text{L}$). The Doppler and reflectivities information are shared between the two point clouds in the uncertainty areas. They are projected back in Cartesian coordinates and combined to create the fused point cloud ($\vect{q}^\text{F}$).
   }
\label{fig:fusion_teaser}
\end{figure}

\subsection{Introduction}

A \ac{LiDAR} is an active sensor transmitting laser beams in the environment, so it is not affected by lighting conditions as during night.
The sensor measures the time delay of the reflected light to be received back, recording the distance and the light, or intensity, of a reflection.
Multiple laser beams (usually 16, 32 or 64 depending on the sensor) scan the car's surroundings, generally by considering a $360^{\circ}$ \ac{FoV}.
The \ac{LiDAR} sensor provides a dense 3D point cloud measuring the geometry of a scene with a resolution below the degree for both azimuth and elevation angles.
These advantages have brought \ac{LiDAR} to the forefront in understanding 3D scenes.

The \ac{RADAR} sensor emits electromagnetic waves, which are not impacted by adverse weather conditions, and records the location, the Doppler and the reflectivity of the objects in the scenes using the received signals (see Sections \ref{sec:background_radar_theory} and \ref{sec:background_signal_process}).
The \ac{RADAR} point cloud is a lightweight representation that is easy to manipulate in Cartesian coordinates. It is obtained after a pre-processing pipeline degrading the objects' reflections. The \ac{RADAR} point cloud is a sparse representation composed of about ten points considering a \ac{LD} \ac{RADAR} and cannot be used as-is for scene understanding.

\sloppy
The \ac{LiDAR} sensor is perturbed by adverse weather conditions (fog, rain, snow) because the light is reflected by droplets, creating artifacts \cite{bijelic_benchmark_2018, guan_through_2020, karlsson_probabilistic_2021}. 
Moreover, it only provides a dense point cloud at short range and cannot reach objects as far as \ac{RADAR}.
Consequently, fusing \ac{RADAR} and \ac{LiDAR} point clouds should lead to benefit from their respective advantages while compensating for their limitations. As they are both represented in Cartesian coordinates, an early fusion module is easily feasible.

As detailed in Section \ref{sec:related_fusion} and to the best of our knowledge, there is no related work on \ac{RADAR} and \ac{LiDAR} point clouds early fusion to exploit both representations as a single enriched point cloud. 
In this section, we propose an early fusion module to propagate \ac{RADAR} information through the \ac{LiDAR} point cloud by considering the resolution and accuracy of the \ac{RADAR} sensor as a quantification of its uncertainty.

Section \ref{sec:fusion_method} will describe our propagation and fusion method which will be simulated in Section \ref{sec:fusion_simulation}. Section \ref{sec:fusion_applications} will present an application to the nuScenes dataset \cite{caesar_nuscenes_2020}. Finally, Section \ref{sec:fusion_discussions} will discuss our proposed method and our future work.


\subsection{Method}
\label{sec:fusion_method}

The proposed method consists in projecting the \ac{RADAR} and \ac{LiDAR} point clouds in a 2D \ac{BEV} coordinates plane\footnote{The elevation angle of the \ac{LD} RADAR is not considered as relevant since the position of the antennas in the sensor does not allow for multiple elevation angles to be recorded.}.
Then it defines an uncertainty area of each \ac{RADAR} points w.r.t. the sensor specificity, \textit{i.e.} a zone in which the points are not perfectly positioned.
Finally, it propagates the \ac{RADAR} point properties through the LiDAR points belonging to its uncertainty area and fusing all the points in a unified point cloud.

Let $\vect{q}^\text{R} \in \mathbb{R}^{m {\times} 5}$ be a \ac{RADAR} point cloud of $m$ points, where the $i$-th point is written $(x^\text{R}_i, y^\text{R}_i, v_{x, i}, v_{y, i}, \sigma^\text{R}_i)$, with $(x^\text{R}_i, y^\text{R}_i)$ its Cartesian coordinates, $\sigma^\text{R}_i$ the \ac{RCS} of the reflected signal and $(v_{x, i}, v_{y, i})$ the Doppler vector in Cartesian coordinates, compensated by the ego vehicle velocity.

Let $\vect{q}^\text{L} \in \mathbb{R}^{n {\times} 3}$ be a \ac{LiDAR} point cloud of $n$ points, where the $j$-th point is written $(x^\text{L}_j, y^\text{L}_j, \sigma^\text{L}_j)$, with $(x^\text{L}_j, y^\text{L}_j)$ its Cartesian coordinates and $\sigma^\text{L}_j$ the intensity of the reflected light.

The propagation and fusion module creates a single point cloud $\vect{q}^\text{F} \in \mathbb{R}^{(m+n) {\times} 6}$, where the $k$-th point is written $(x^\text{F}_k, y^\text{F}_k, \sigma^\text{L}_k, v_{x, k}, v_{y, k}, \sigma^\text{R}_k)$ including both $\vect{q}^\text{L}$ and $\vect{q}^\text{R}$ information, \textit{i.e.} carrying its 2D Cartesian coordinates, the reflectivity of the \ac{LiDAR} point, the \ac{RCS} and the two components of the Doppler vector.

In this section, we introduce the \ac{RADAR} sensor uncertainty, \textit{i.e.} the quantified uncertainty related to the accuracy and resolution of the sensor measurements. Depending on the \ac{RADAR} sensor, it will have several modes of transmission and reception of the signals. The modes will define the maximum distance and the \ac{FoV} that the signal can reach. 
An example of the \ac{RADAR} specificity is presented in Table \ref{tab:nuscenes_radar_specs}; it corresponds to the sensor used in the nuScenes dataset. 
This \ac{RADAR} sensor has three modes: a far-range mode and two short-range modes (either with $\pm 45^\circ$ or $\pm 60^\circ$ \ac{FoV}). Each one of these modes has its own specificity in resolution, accuracy and maximum measurements in both distance and azimuth angle.
Additional details on the application of the presented method are provided in Section \ref{sec:fusion_applications}.
The distance resolution is assumed to be linearly increasing and its accuracy is fixed for each mode. The azimuth resolution and accuracy are fixed for each mode.

By knowing the location of the \ac{RADAR} point, we can associate a resolution and an accuracy in both distance and azimuth angle.
These sensor uncertainty measurements in range and azimuth angle, respectively written $\delta r$ and $\delta \alpha$, correspond to the minor and major axis of an ellipse in polar coordinates.

Let us consider a \ac{RADAR} and a \ac{LiDAR} point cloud. Each point cloud is projected in polar coordinates to propagate the \ac{RADAR} information in the \ac{LiDAR} points belonging to each uncertainty ellipse.
Let $\mathcal{P}$ be the map defined as
\begin{equation}
    \label{eq:polar_projection}
    \begin{array}[t]{lrcl}
\mathcal{P} : & \mathbb{R}^{k \times 6} & \longrightarrow & \mathbb{R}^{k \times 6} \\
    & \vect{q} = 
        \left[
        \vect{q}_{0},
        \vect{q}_{1},
        \vect{q}_{2},
        \hdots,
        \vect{q}_{5}
        \right] 
         & \longmapsto & \tilde{\vect{q}} =
        \left[
        \tan^{-1} (\sfrac{\vect{q}_{1}}{\vect{q}_{0}}),
        \sqrt{\vect{q}_{0}^2 + \vect{q}_{1}^2},
        \vect{q}_{2},
        \hdots,
        \vect{q}_{5}
        \right]
    \end{array}
\end{equation}
transforming an arbitrary point cloud $\vect{q}$ of $k$ points in Cartesian coordinates into a point cloud $\tilde{\vect{q}}$ in polar coordinates w.r.t. the sensor coordinates.
We will note $(\vect{x}^\text{R}, \vect{y}^\text{R})$ the 2D Cartesian coordinates of the \ac{RADAR} point cloud $\vect{q}^\text{R}$, and $(\bm{\alpha}^\text{R}, \vect{r}^\text{R})$ their corresponding 2D polar coordinates.

In practice, we initialise the \ac{LiDAR} and \ac{RADAR} point cloud with constant values for their respective missing features to use $\mathcal{P}$. \textit{E.g} the \ac{RADAR} point cloud will have an additional dimension with constant values noted $\sigma^\text{cst}$ filling the missing information of a \ac{LiDAR} point. The \ac{LiDAR} point cloud will have three additional dimensions written $\sigma^\text{cst}, (v_x^\text{cst}, v_y^\text{cst})$ filling the missing information of a \ac{RADAR} point cloud, reflectively the \ac{RCS} and the Doppler components.
This way, the two points clouds $\vect{q}^\text{L} \in \mathbb{R}^{n {\times} 6}$ and $\vect{q}^\text{R} \in \mathbb{R}^{m {\times} 6}$ are projected in polar coordinates, respectively written $\mathcal{P}(\vect{q}^\text{L}) = \tilde{\vect{q}}^\text{L}$ and $\mathcal{P}(\vect{q}^\text{R}) = \tilde{\vect{q}}^\text{R}$.
%

Considering the $i$-th \ac{RADAR} point $(\alpha^\text{R}_i, r^\text{R}_i)$ in polar coordinates, its uncertainty ellipse is defined as $\mathcal{B}_{(\delta \alpha, \delta r)} (\alpha^\text{R}_i, r^\text{R}_i)$, a ball centered on $(\alpha^\text{R}_i, r^\text{R}_i)$, where $\delta \alpha, \delta r$ are respectively its azimuth major axis and its range minor axis.
On the one hand, the \ac{RADAR} properties of the point $(\alpha^\text{R}_i, r^\text{R}_i)$ are propagated to the \ac{LiDAR} points belonging to this ellipse.
On the other hand, the properties of the \ac{LiDAR} points belonging to the ellipse will be averaged to be propagated to the \ac{RADAR} point.
In other words, we propose to propagate the \ac{RCS} and the Doppler components, respectively noted $\sigma^\text{R}_i$ and $(v_{x,i}, v_{y,i})$, corresponding to the \ac{RADAR} point $(\alpha^\text{R}_i, r^\text{R}_i)$,  to the \ac{LiDAR} points in its ellipse.
The reflected intensity vector of the \ac{LiDAR} points in the ellipse, noted $\bm{\sigma}^\text{L}$, is averaged and associated to the RADAR point at the center of the ellipse.
Note that the proposed method is asymmetric to adapt the difference in sparsity between the \ac{RADAR} and \ac{LiDAR} point clouds.

Considering the $j$-th \ac{LiDAR} point $(\alpha^\text{L}_j, r^\text{L}_j)$ in polar coordinates, it belongs to the \ac{RADAR} uncertainty ellipse $\mathcal{B}_{(\delta \alpha, \delta r)} (\alpha^\text{R}_i, r^\text{R}_i)$ if
\begin{equation}
    \frac{(\alpha^\text{L}_j - \alpha^\text{R}_i)^2}{\delta \alpha} + \frac{(r^\text{L}_j - r^\text{R}_i)^2}{\delta r} \leq 1
\end{equation}
is verified.
The \ac{LiDAR} points belonging to the ellipse are grouped in a sub point cloud written $\tilde{\vect{q}}^\text{L}_{+}$ and removed from the initial \ac{LiDAR} point cloud $\tilde{\vect{q}}^\text{L}$.

\raggedbottom
After the propagation step, three distinct point clouds are obtained: $\tilde{\vect{q}}^\text{F}_{+}$ the sub-\ac{LiDAR} point cloud with propagated \ac{RADAR} information, $\tilde{\vect{q}}^\text{F}$ the sub-\ac{LiDAR} point cloud containing the remaining points and $\tilde{\vect{q}}^\text{R}$ the \ac{RADAR} point cloud.
A single point cloud $\tilde{\vect{q}}^\text{F} = [ (\tilde{\vect{q}}^\text{L}_{+})^\top, (\tilde{\vect{q}}^\text{L})^\top, (\tilde{\vect{q}}^\text{R})^\top]^\top \in \mathbb{R}^{(m+n) {\times} 6}$ is created by stacking them, containing both the \ac{RADAR} and \ac{LiDAR} point clouds with the propagated information or constant values where appropriate.
The final point cloud is projected back in Cartesian coordinates: $\vect{q}^\text{F} = \mathcal{P}^{-1} (\tilde{\vect{q}}^\text{F})$.
The entire method is detailed in Algorithm \ref{alg:fusion_propag}. A simulation of this method will be presented in the following section.

\begin{algorithm}[H]
\setstretch{1.35}
\caption{\textbf{Proposed propagation and fusion module}. We note $\vect{A}_{i,j}$ the element of the matrix $\vect{A}$ at the $i$-th row and $j$-th column, $\vect{A}_{.,j}$ the elements of the matrix $\vect{A}$ corresponding to the all rows of the $j$-th column, and $\vect{A}_{i,.}$ the elements of the matrix $\vect{A}$ corresponding to all the columns of the $i$-th row.}
\label{alg:fusion_propag}
\begin{algorithmic}[1]
\REQUIRE $\vect{q}^\text{R} \in \mathbb{R}^{m {\times} 5}, \vect{q}^\text{L} \in \mathbb{R}^{n {\times} 3}$ \\ \COMMENT{$m$ the number of RADAR points, $n$ the number of LiDAR points.}
\STATE $[\bm{\sigma}^\text{cst}, \vect{v}_x^\text{cst}, \vect{v}_y^\text{cst}] \leftarrow [\vect{c}_0, \vect{c}_1, \vect{c}_2]$ \COMMENT{Set constant values.}

\STATE $\vect{q}^\text{R} \leftarrow [\vect{q}^\text{R}_{\cdot, 0}, \vect{q}^\text{R}_{\cdot, 1}, \bm{\sigma}^\text{cst}, \vect{q}^\text{R}_{\cdot, 2}, \vect{q}^\text{R}_{\cdot, 3}, \vect{q}^\text{R}_{\cdot, 4}]$ \COMMENT{Set $\vect{q}^\text{R} \in \mathbb{R}^{m {\times} 6}$ with cst: $\bm{\sigma}^\text{cst}$.}

\STATE $\vect{q}^\text{L} \leftarrow [\vect{q}^\text{L}_{\cdot, 0}, \vect{q}^\text{L}_{\cdot, 1}, \vect{q}^\text{L}_{\cdot, 2}, \vect{v}_x^\text{cst}, \vect{v}_y^\text{cst}, \bm{\sigma}^\text{cst}]$ \COMMENT{Set $\vect{q}^\text{L} \in \mathbb{R}^{n {\times} 6}$ with cst: $\bm{\sigma}^\text{cst}$, $\vect{v}_x^\text{cst}$, $\vect{v}_y^\text{cst}$.}
\STATE $\tilde{\vect{q}}^\text{R}, \tilde{\vect{q}}^\text{L} \leftarrow \mathcal{P}(\vect{q}^\text{R}), \mathcal{P}(\vect{q}^\text{L})$ \COMMENT{Project $\tilde{\vect{q}}^\text{R}$ and $\tilde{\vect{q}}^\text{L}$ in polar coordinates.}
\STATE $h \leftarrow 0$ \COMMENT{Set counter to 0.}

\FOR[$m$ RADAR points.]{$i=0$ \TO $m-1$}
    \STATE $[\alpha^\text{R}_i, r^\text{R}_i, \sigma^\text{cst}_i, v_{x, i}, v_{y, i}, \sigma^\text{R}_i] \leftarrow \tilde{\vect{q}}^\text{R}_{i, \cdot}$ \COMMENT{Set the RADAR scalar values.}
    \STATE $(\delta \alpha, \delta r) \leftarrow$ get\_uncertainty$(\alpha^\text{R}_i, r^\text{R}_i)$ \COMMENT{Get the ellipse parameters.}
    \STATE $\vect{A}:$ empty matrix with 6 columns \COMMENT{Buffer of enriched LiDAR points.}
    \STATE $k \leftarrow 0$ \COMMENT{Set counter to 0.}
    
    \FOR[Number of LiDAR points in $\tilde{\vect{q}}^\text{L}$.]{$j=0$ \TO $|\tilde{\vect{q}}^\text{L}|-1$}
        \STATE $[\alpha^\text{L}_j, r^\text{L}_j, \sigma^\text{L}_j, v_{x, i}^\text{cst}, v_{y, i}^\text{cst}, \sigma^\text{cst}] \leftarrow \tilde{\vect{q}}^\text{L}_{j,\cdot}$ \COMMENT{Set the LiDAR scalar values.}
        \IF[Check if the LiDAR point is in the ellipse]{$\frac{(\alpha^\text{L}_j - \alpha^\text{R}_i)^2}{\delta \alpha} + \frac{(r^\text{L}_j - r^\text{R}_i)^2}{\delta r} \leq 1$} 
        \STATE $\vect{A}_{k, \cdot} \leftarrow [\alpha^\text{L}_j, r^\text{L}_j, \sigma^\text{L}_j, v_{x, i}, v_{y, i}, \sigma^\text{R}_i]$ \COMMENT{Add the point to the enriched buffer.}
        \STATE $k \leftarrow k + 1$ \COMMENT{Update the counter.}
        \ENDIF
    \ENDFOR
    
    \IF[Check if LiDAR points have been added to the buffer.]{$k \neq 0$}
    \STATE $\overline{\sigma} = \text{mean}(\vect{A}_{\cdot, 2})$ \COMMENT{Compute the mean of LiDAR's $\sigma$ in the buffer.}
    \STATE $\tilde{\vect{q}}^\text{R}_{i, \cdot} \leftarrow [\alpha^\text{R}_i, r^\text{R}_i, \overline{\sigma}, v_{x, i}, v_{y, i}, \sigma^\text{R}_i]$ \\ \COMMENT{Update the RADAR point with the averaged LiDAR's $\sigma$.}
    \ENDIF
    \STATE $\tilde{\vect{q}}^\text{L}_{+, h:h+k, \cdot} \leftarrow \vect{A}$ 
    \COMMENT{Add the buffer to the enriched LiDAR point cloud.}
    \STATE $\tilde{\vect{q}}^\text{L} \leftarrow \tilde{\vect{q}}^\text{L}$.delete$(\vect{A})$ \COMMENT{Remove the buffer from initial LiDAR point cloud.}
    \STATE $h \leftarrow h + k$ \COMMENT{Update the counter.}
\ENDFOR

\STATE $\tilde{\vect{q}}^\text{F} \leftarrow 
        \begin{bmatrix}
        \tilde{\vect{q}}^\text{R} \\
        \tilde{\vect{q}}^\text{L}_{+} \\
        \tilde{\vect{q}}^\text{L}
    \end{bmatrix}$ \COMMENT{Fuse the point clouds.}
\STATE $\vect{q}^\text{F} \leftarrow \mathcal{P}^{-1}(\tilde{\vect{q}}^\text{F})$ \COMMENT{Map the fused point cloud in Cartesian coordinates.}
\end{algorithmic}
\end{algorithm}

\subsection{Simulation}
\label{sec:fusion_simulation}

The proposed method detailed in Algorithm \ref{alg:fusion_propag} has been simulated with randomly generated point clouds.
Let $n=2000$ and $m=20$ be the number of points in the \ac{LiDAR} and \ac{RADAR} point clouds respectively. 
For both point clouds, random Cartesian coordinates are drawn as $\vect{x} \sim \mathcal{U}([-50, 50])$ and $\vect{y} \sim \mathcal{U}([0, 250])$, the \ac{RADAR} sensor position has been fixed to $(0, 0)$.
The order of magnitude of the size of each point cloud has been chosen according to real data recorded from a 32-beam \ac{LiDAR} and \ac{LD} \ac{RADAR}.

\begin{figure}[!t]
\begin{center}
\includegraphics[width=1\linewidth]{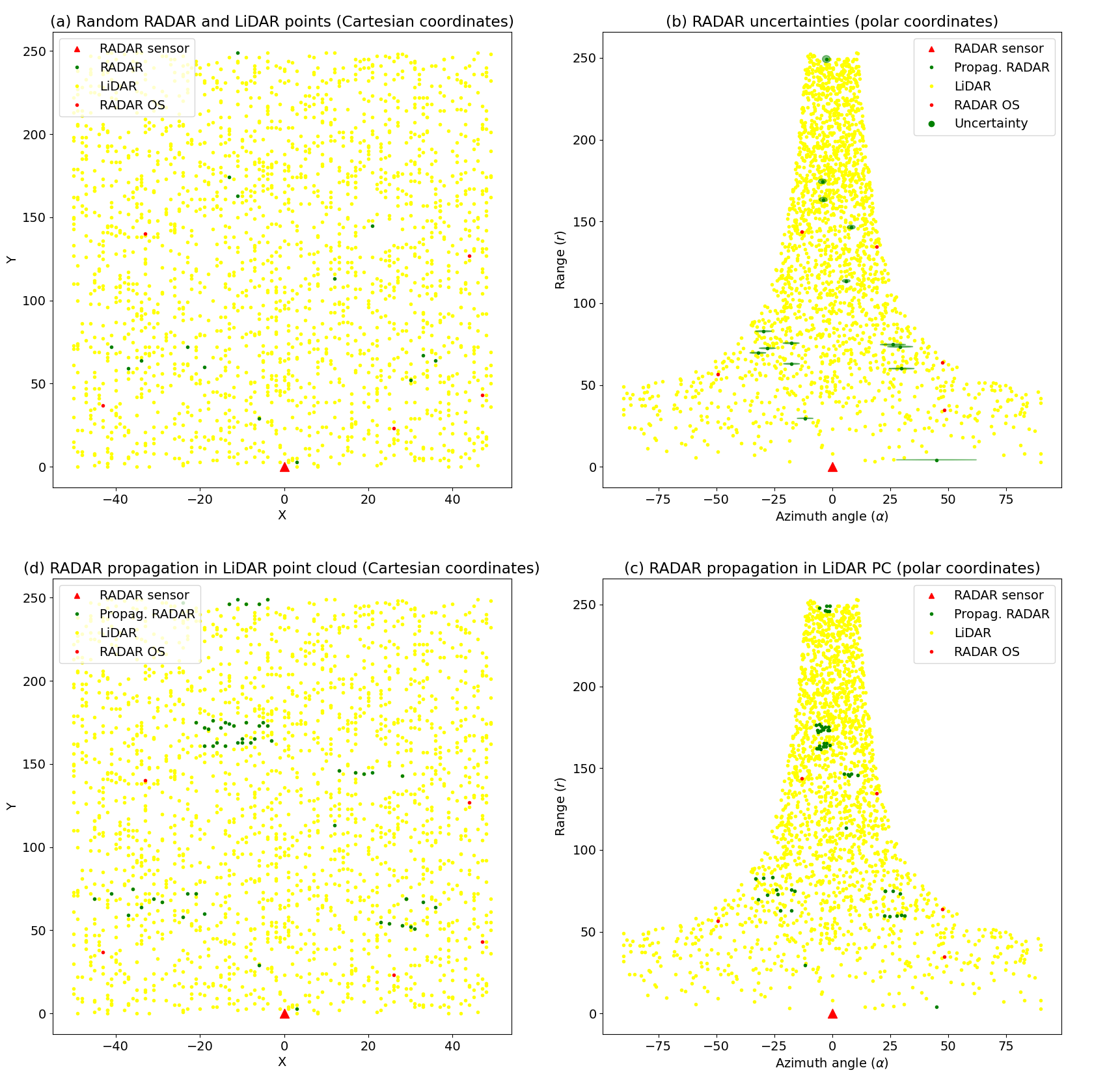}
\end{center}
  \caption[Simulation of the LiDAR and RADAR point cloud propagation and fusion method]{\textbf{Simulation of the LiDAR and RADAR point cloud propagation and fusion method}. (a) The \ac{LiDAR} and \ac{RADAR} point clouds are randomly drawn in Cartesian coordinates, the \ac{RADAR} points are filtered according to the sensor specificity (Out-of-Scope (OS) points). (b) The point clouds are transformed in polar coordinates with the \ac{RADAR} uncertainty ellipses illustrated with a scaling factor of $2.0$. (c) The \ac{RADAR} point information is propagated to the \ac{LiDAR} points and (d) transformed back in Cartesian coordinates. Note that the ``Propag. RADAR'' point cloud groups the \ac{RADAR} points and \ac{LiDAR} points in the uncertainty areas.}
\label{fig:fusion_simulation}
\end{figure}

The simulation considers the specificity of the \ac{RADAR} sensor used in the nuScenes dataset \cite{caesar_nuscenes_2020} described in Table \ref{tab:nuscenes_radar_specs}.
Figure \ref{fig:fusion_simulation}(a) illustrates the randomly drawn \ac{LiDAR} and \ac{RADAR} point clouds, respectively in yellow and green. The Out-of-Scope (OS) \ac{RADAR} points are not belonging to any \ac{RADAR} mode considering the geometric priors of the sensor specificity; they are excluded from the simulation.
Both point clouds are then mapped in polar coordinates and the uncertainty ellipse of each \ac{RADAR} point is computed in polar coordinates as depicted in Figure \ref{fig:fusion_simulation}(b)\footnote{Note that a scaling factor of $2.0$ has been applied to the ellipses for visualization purpose. The qualitative results presented in the next section are obtained with a scaling factor fixed to $1.0$.}.
The resolution and accuracy in distance and azimuth angle are defined according the method presented in Section \ref{sec:fusion_method} and with the sensor specificity described in Table \ref{tab:nuscenes_radar_specs}.
As detailed in Algorithm \ref{alg:fusion_propag}, the \ac{LiDAR} points belonging to an ellipse obtain the propagated information of the \ac{RADAR} point at the center of the ellipse. These points are denoted as ``Propag. RADAR'' in Figure \ref{fig:fusion_simulation}(c).
Finally, Figure \ref{fig:fusion_simulation}(d) illustrates in green both the \ac{RADAR} points and the \ac{LiDAR} points which have benefited from the propagation method. The aim of the fusion is to consider both green and yellow points as a single point cloud.
The simulation shows that the proposed method succeeds to locally propagate the \ac{RADAR} information in a denser point cloud by only considering the resolution and the accuracy of the sensor.
The following section will present the application of this method to a real dataset.

\subsection{Application to the nuScenes dataset}
\label{sec:fusion_applications}

The nuScenes dataset \cite{caesar_nuscenes_2020} is considered as one of the largest automotive dataset publicly available containing radar data. It has been briefly described in Section \ref{sec:related_datasets} and Table \ref{tab:related_public_dataset}.
It is composed of 5.5 hours of recorded sequences in two countries including night and rain weather conditions. 
The car is mounted with a 32-beam \ac{LiDAR}, 6 cameras and 5 \ac{LD} ARS 408-21 \acp{RADAR}\footnote{\url{https://www.continental-automotive.com/}} recording simultaneously with a $360^\circ$ \ac{FoV}.
The parameters and settings of each RADAR sensor are described in Table \ref{tab:nuscenes_radar_specs}.
The authors proposed 3D bounding boxes annotations with tracking metrics and a recent extension containing semantic segmentation annotations while considering 23 classes and 8 attributes.
The dataset is composed of 1000 scenes divided in three sub-sets: ``mini'', ``trainval'' and ``test''.
In our preliminary work, only the ``mini'' dataset containing 10 scenes has been explored yet.

\begin{table}[!ht]
\centering
\scriptsize
\begin{tabular}{c c c c} \toprule
\multirow{2}{*}{Specificity} & \multicolumn{3}{c}{RADAR mode}\\ 
\cmidrule(lr){2-4}
& Far range & Near range ($45^\circ$) & Near range ($60^\circ$) \\
\midrule
Azimuth angle \ac{FoV} (deg) & $\pm 9^\circ$ & $\pm 45^\circ$ & $\pm 60^\circ$ \\
Azimuth angle resolution (deg) & $1.6^\circ$ &  $4.5^\circ$ & $12.3^\circ$ \\
Accuracy azimuth angle (deg) & $\pm 0.1^\circ$ & $\pm 1.0^\circ$ & $\pm 5.0^\circ$ \\
Distance range (m) & $[0.2, 250]$ & $[0.2, 100]$ & $[0.2, 20]$ \\
Distance range resolution (m) & $[0, 1.79]$ & $[0, 0.39]$ & $[0, 0.39]$ \\
Accuracy distance (m) & $\pm 0.40$ & $\pm 0.10$ & $\pm 0.10$ \\
\bottomrule
\end{tabular}
\caption[Parameters and settings of the RADAR sensors used in the nuScenes dataset]{\textbf{Parameters and settings of the RADAR sensors used in the nuScenes dataset}.}
\label{tab:nuscenes_radar_specs}
\end{table}

Each sensor used to record the data has its own orientation axis as described in Figure \ref{tab:nuscenes_radar_specs} of Appendix \ref{sec_app:nuscenes_sensor_settings}.
Therefore our method has been applied iteratively on each \ac{RADAR} orientation axis by projecting the \ac{LiDAR} point cloud in the same coordinates system. By doing so, the considered \ac{RADAR} sensor is at the origin of its coordinates system and the polar coordinates of each point cloud will be computed w.r.t. its position. The procedure described in Section \ref{sec:fusion_method} and Algorithm \ref{alg:fusion_propag} is applied independently for each \ac{RADAR} sensor while considering the same $360^\circ$ \ac{LiDAR} point cloud. 

\begin{figure}[!t]
\begin{center}
\includegraphics[width=1.1\linewidth,center]{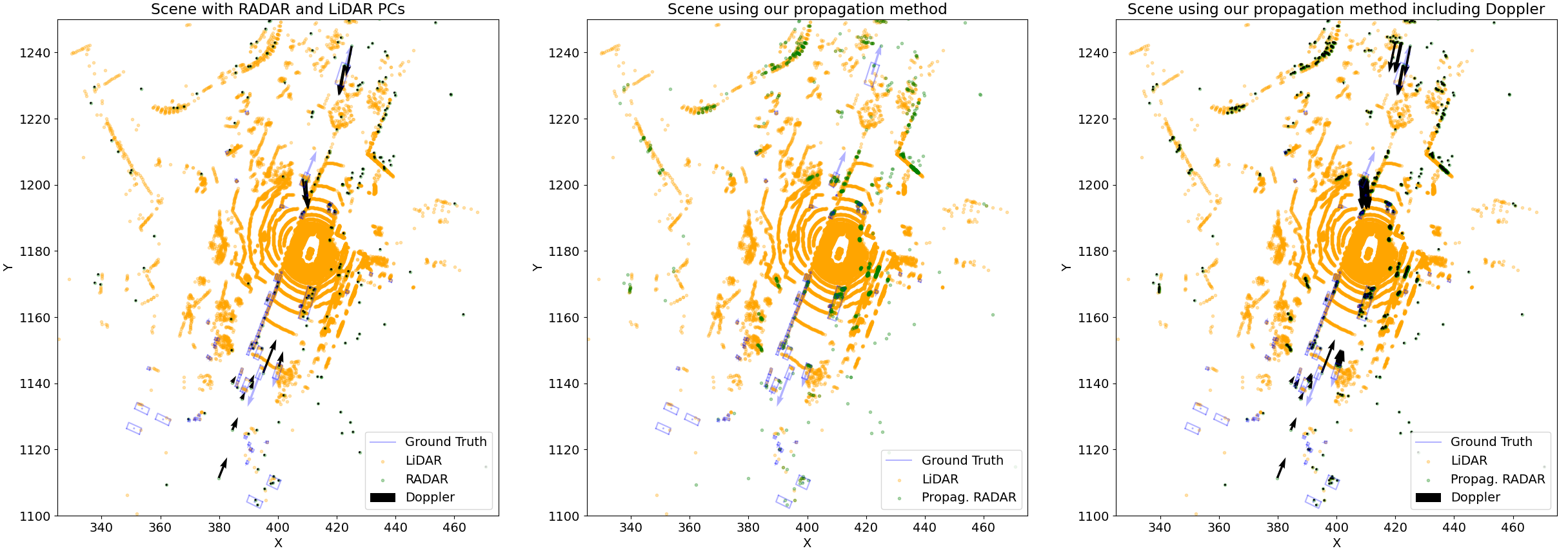}
\end{center}
  \caption[Qualitative results on the nuScenes dataset of our propose propagation and fusion module]{\textbf{Qualitative results on the nuScenes dataset of our propose propagation and fusion module}. (Left) Scene in \acl{BEV} representation with \ac{LiDAR} and \ac{RADAR} point clouds. (Middle) The point cloud illustrated in green groups the \ac{RADAR} and propagated \ac{RADAR} points with Doppler and reflectivities. (Right) The propagated Doppler information is illustrated with black arrows to distinguish moving objects.}
\label{fig:fusion_quali_results}
\end{figure}

Figure \ref{fig:fusion_quali_results} (left) illustrates a scene with a \ac{LiDAR} point cloud in orange and the \ac{RADAR} point cloud in green. Each \ac{RADAR} point is characterized by a black arrow corresponding to its Doppler compensated by the ego-vehicle velocity. The light-blue boxes and their corresponding arrows are the annotated objects in the scenes with their velocity. 
By applying our method, we propagate the \ac{RADAR} information to the \ac{LiDAR} points according to the sensor uncertainty defined by its specificity (Table \ref{tab:nuscenes_radar_specs}). The propagated points combined with the \ac{RADAR} point clouds are depicted in green in Figure \ref{fig:fusion_quali_results} (middle).
By propagating the \ac{RADAR} information in the \ac{LiDAR} point cloud, we show qualitatively in Figure \ref{fig:fusion_quali_results} (right) that our method helps to obtain a denser Doppler information in the combined point cloud.
Additional qualitative results are presented in Figure \ref{fig:fusion_app_quali_nuscenes} of Appendix \ref{sec_app:fusion_quali_results} leading to the same conclusions.
Knowing that these points also carry the \ac{RCS} and the \ac{LiDAR} reflection intensity, we hope that our proposed propagation and fusion module will help to improve detection and classification of objects considering point cloud representations.
The following section will discuss our results and future work.

\subsection{Discussions and future work}
\label{sec:fusion_discussions}

In this section, we presented our propagation and fusion module quantifying the measurement uncertainty of the \ac{RADAR} sensor. By considering both \ac{RADAR} and \ac{LiDAR} point clouds, we propose to propagate the \ac{RADAR} point properties to the \ac{LiDAR} points belonging to its uncertainty area. The other way around, the \ac{RADAR} point will also carry the average reflection of all the \ac{LiDAR} points in its uncertainty area.
Our module proposes to fuse both point clouds to benefit from the density of the \ac{LiDAR} data and the Doppler information of the \ac{RADAR} data while containing the reflectivity information of both sensors.

However, we notice that our method is impacted by ghost and multi-path \ac{RADAR} reflections. In these cases, the propagation may assign a Doppler or \ac{RCS} to an object which it does not belong to. Inspired by the work of \cite{kopp_fast_2021}, rule-based method could be integrated to filter the \ac{RADAR} point cloud before the propagation step. 
Another limitation of our method is to consider that each \ac{LiDAR} point has an equivalent probability to belong to the uncertainty ellipse of a \ac{RADAR} point. We will tackle this issue by considering this ellipse has a multi-variate Gaussian probability distribution, centered on the \ac{RADAR} point and with a variance-covariance matrix defined by its distance and azimuth angle axis (previously noted $\delta r$ and $\delta \alpha$). This way, each \ac{LiDAR} point will be associated to a probability to belong to this ellipse.

The aim of this work is to benefit from both \ac{LiDAR} and \ac{RADAR} point clouds to improve scene understanding. In our future work, we will compare a point cloud based deep learning algorithm (\textit{e.g.} PointNet \cite{qi_pointnet_2017} or PointNet++ \cite{qi_pointnet_2017-1}) while being trained with either a single sensor point cloud or with our enhanced point cloud.
In a second approach, we will explore self-supervised learning by training a model to predict a modality of our enhanced point cloud and fine-tune it on a downstream task. \textit{E.g.} a model trying to predict the Doppler information will learn to discriminate moving and static objects while using a single timestamp point cloud. The neural network will be then fine-tuned to improve an object detection or a segmentation task.

\section{Conclusions}

This chapter has described innovative approaches for \ac{RADAR} scene understanding using deep learning algorithms.
In Section \ref{sec:mvrss}, we proposed a method for multi-view \ac{RADAR} semantic segmentation based on the exploitation of the CARRADA dataset presented in Section \ref{sec:carrada}.
We have detailed several deep learning architectures, with their associated loss functions, outperforming competing methods with significantly fewer parameters. Our method is the only capable to simultaneously perform \ac{RD} and \ac{RA} semantic segmentation. We have introduced an unsupervised Coherence loss to enforce the spatial coherence between \ac{RADAR} views during training. Our proposed method has achieved the state of the art performances on the CARRADA dataset and on the recently published RADDet dataset. Conducted experiments have shown that our method is able to segment objects well in multi-views of the \ac{RADAR} tensor in both simple and complex urban scenes.
This second main contribution of the thesis was presented at the International Conference of Computer Vision (ICCV).
In our future work, we will improve the aggregation of the \ac{RAD} tensor to avoid object signature attenuation. We will also explore an extended Coherence loss formulation to introduce the ground truth and better quantify the spatial errors.  

In Section \ref{sec:sensor_fusion}, we presented a preliminary work on \ac{LiDAR} and \ac{RADAR} point cloud fusion. Our fusion and propagation method quantifies the uncertainty of the polar coordinate point location by considering the specificity of the \ac{RADAR} sensor.
On one hand, our method propagates the Doppler and \ac{RCS} information of a \ac{RADAR} point to the \ac{LiDAR} points contained in its uncertainty area.
On the other hand, it propagates the average \ac{LiDAR} reflections contained in this area to the \ac{RADAR} point.
This way, an enriched point cloud is proposed, containing the velocity and reflection information of the \ac{RADAR} point cloud while benefiting from the density of the \ac{LiDAR} point cloud.
In our future work, we will train deep neural network architectures specialized in point cloud representation with the proposed enriched point cloud to improve scene understanding tasks, \textit{e.g.} object detection and semantic segmentation.
In addition, we will explore self-supervised learning schemes applied to multi-sensor fusion. In particular, we will exploit the enriched point cloud to learn a neural network for Doppler prediction. We hope that this method will learn representations to distinguish moving objects with any additional annotations. This model will be then fine-tuned on well-known scene understanding tasks to show quantitative improvement from this process.

Further details on our future work will be provided in Section \ref{sec:conclusion_future}.
In the following chapter, we will present a collaborative project proposing RADIal, a recent dataset with synchronized camera, \ac{LiDAR} and raw \ac{HD} \ac{RADAR} data with annotations for 2D object detection and free driving space segmentation. We will also present a deep learning architecture for multi-task learning while estimating a part of the costly \ac{RADAR} pre-processing.

\chapter{High-definition RADAR}
\label{chap:hd_radar}
\minitoc

\ac{HD} \ac{RADAR} reaches a high angular resolution thanks to the large number of virtual antennas that it is composed of. However, it generates a large volume of data, directly impacting real-time applications.
This chapter details RADIal, a unique dataset with raw \ac{HD} \ac{RADAR} data synchronized with \ac{LiDAR} and camera.
Annotations for 2D object detection and free driving space segmentation, \textit{i.e.} classify each pixel of a representation as free to be driven or not, are generated with a semi-automatic pipeline. 
It also presents, FFT-RadNet, a multi-task deep neural network architecture processing raw \ac{RADAR} signals and succeeding to estimate the \ac{RADAR} processing steps avoiding large computational cost.

This chapter is organized as follows: Section \ref{sec:radial_dataset} introduces the RADIal dataset, Section \ref{sec:fft_radnet_method} details the proposed method, Section \ref{sec:fft_radnet_expe_radial} presents the experimental results, finally Section \ref{sec:hd_radar_discussions} concludes.

This chapter presents a work carried out in collaboration with Julien Rebut and mainly inspired from our article published at the Conference on Computer Vision and Pattern Recognition (CVPR) \cite{rebut_raw_2021}. 

\section{Motivations}
\label{sec:hd_radar_motivations}

Recent progress towards \ac{HD} Imaging \ac{RADAR} has driven the angular resolution below the degree, thus approaching \ac{LiDAR} performance. 
By using dense virtual antenna arrays, these sensors achieve high angular resolution both in azimuth and elevation (horizontal and vertical angular positions, respectively) and produce denser \ac{RADAR} point clouds.
The previous chapters focused on \ac{LD} \acp{RADAR} with a single elevation pitch due to the geometry of their antennas. As a consequence, the previously noted \ac{RA} view represented the Azimuth angle. 
In this chapter, we distinguish the Elevation angle deduced from adjacent pairs of antennas in the vertical axis so the \ac{RA} representation includes both Azimuth and Elevation angles. 
As a consequence, \ac{HD} \ac{RADAR} representations are cumbersome, with an order of magnitude larger than the data recorded by a \ac{LD} \ac{RADAR}.

\begin{figure}[!t]
\centering
\includegraphics[width=1\textwidth]{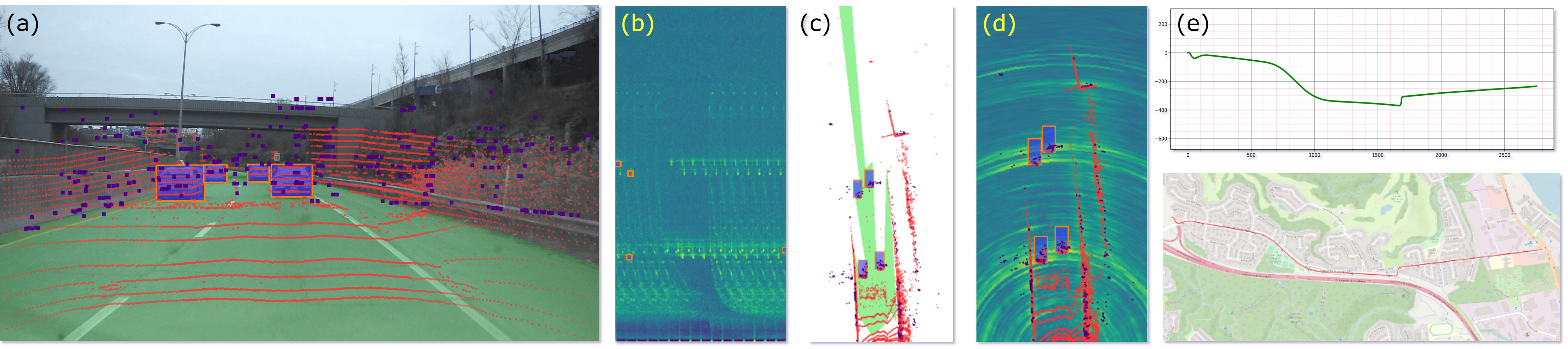}
\caption[Overview of our RADIal dataset]{\textbf{Overview of our RADIal dataset}. RADIal includes a set of 3 sensors (camera, \ac{LiDAR}, high-definition radar) and comes with GPS and vehicle's CAN traces; 25,000 synchronized samples are recorded in raw format. (a) Camera image with projected \ac{LiDAR} point cloud in red and \ac{RADAR} point cloud in indigo, vehicle annotation in orange and free driving space annotation in green; (b) \ac{RADAR} power spectrum with bounding box annotations; (c) Free driving space annotation in bird-eye view, with annotated vehicle bounding boxes in orange, \ac{RADAR} point cloud in indigo and \ac{LiDAR} point cloud in red; (d) Range-Azimuth map in Cartesian coordinates overlayed with \ac{RADAR} point cloud and \ac{LiDAR} point cloud; (e) GPS trace in red and odometry trajectory reconstruction in green.
}
\label{fig:radial_teaser}
\end{figure}

As seen in Chapter \ref{chap:related_work}, most of the recent works exploit the Range-Azimuth representation of the \ac{RADAR} data (either in polar or cartesian coordinates).
Similar to a \ac{BEV} (see Figure \ref{fig:radial_teaser}), this representation is easy to interpret and allows simple data augmentation with translations and rotations. However, one barely-mentioned drawback is that the generation of the \ac{RA} \ac{RADAR} view incurs significant processing costs (tens of GOPS, see Section \ref{sec:fft_radnet_complexity}), which compromises its viability on embedded hardware. While novel \ac{HD} \acp{RADAR} offer better resolution, they make this computational complexity issue even more acute.

As detailed in Section \ref{sec:background_signal_process},
the \ac{AoA} is deduced by applying an inverse \ac{FFT} on the pairs of antennas axis of the recorded tensor in the frequency domain (see Figure \ref{fig:schema_fft}).
An alternative is to correlate the \ac{RD} in the complex domain with a calibration matrix to estimate both the azimuth and the elevation angles.
The complexity of this operation for a single point of the RD tensor is $\mathcal{O}(\NTx\NRx\BinA\BinE)$, where $\NTx$ is the number of transmitting antennas, $\NRx$ the number of receiving antennas, $\BinA$ and $\BinE$ are respectively the number of discretization bins for azimuth and elevation angles in the calibration matrix. 
For a 4D representation in \ac{RAED}, this operation would need to be performed for each point of the RD tensor\footnote{Considering a \ac{HD} \ac{RADAR} with $0.2^{\circ}$ of azimuth resolution over $180^{\circ}$ of horizontal \ac{FoV} and 11 elevations, it would require 498 \ac{GFLOPS} to be computed.}.
Considering an embedded \ac{HD} \ac{RADAR}, traditional signal processing cannot be applied as it is too resource greedy in terms of both computation requirements and memory footprint.
For driving assistance systems, there is therefore the challenge of increasing radar's angular accuracy while keeping the processing costs under control.

In this chapter, we propose a unique dataset, nick-named RADIal\footnote{RADIal is available online at \url{https://github.com/valeoai/RADIal}.} for ``RADAR, LiDAR \textit{et al}.'', with the first raw \ac{HD} \ac{RADAR} dataset including several other automotive-grade sensors, as described in Table \ref{tab:related_public_dataset}. It is composed of 2 hours of raw data from synchronized automotive-grade sensors (camera, \ac{LiDAR}, \ac{HD} radar), annotated for object detection and free space segmentation, and collected in various environments (city street, highway, countryside road).
We also propose a novel \ac{HD} \ac{RADAR} sensing model, FFT-RadNet, that eliminates the overhead of computing the \ac{RAD} 3D tensor, learning instead to recover angles from an \ac{RD} view. FFT-RadNet is trained both to detect vehicles and to segment free driving space. On both tasks, it competes with recent scene understanding models while requiring less computations and memory.

\section{RADIal dataset}
\label{sec:radial_dataset}

As depicted in Table \ref{tab:related_public_dataset}, publicly-available datasets do not provide raw \ac{RADAR} signal, either for \ac{LD} \ac{RADAR} nor for \ac{HD} \ac{RADAR}. Therefore, we built RADIal, a new dataset to allow research on automotive \ac{HD} \ac{RADAR}.

As RADIal includes 3 sensor modalities --camera, \ac{RADAR} and \ac{LiDAR}--, it should also permit one to investigate the fusion of \ac{HD} \ac{RADAR} with other common sensors. The specifications of the sensor suite are detailed in Table \ref{tab:radial_sensors_specs}.  
Except for the camera, all sensors are automotive-grade qualified. 
On top of that, the \ac{GPS} position and full \ac{CAN bus} of the vehicle (including odometry) are also provided.
Sensor signals were recorded simultaneously in a raw format, without any signal pre-processing. 
In the case of the \ac{HD} \ac{RADAR}, the raw signal is the \ac{ADC}. From this \ac{ADC} data, all conventional \ac{RADAR} representations can be generated: \ac{RAD} tensor, \ac{RA} and \ac{RD} views or point cloud.

\begin{table}[!t]
\begin{minipage}[b]{.48\linewidth}

\centering
\tiny
\begin{tabular}{clccc}
\toprule
 & & \textbf{HD Radar} &  \textbf{LiDAR} & \textbf{Camera}\\
\midrule
\multirow{3}{*}{\rotatebox{90}{FOV}} & Range   & 103\,m & 150\,m & --\\
& Azimuth   & $180^{\circ}$ & $133^{\circ}$ & $100^{\circ}$\\
& Elevation  & $12^{\circ}$ & $10^{\circ}$ & $75^{\circ}$\\
\midrule
\multirow{4}{*}{\rotatebox{90}{resolution}} & Range & 0.2\,m & 0.1\,m & \\
& Azimuth & $0.1^{\circ}$& $0.125$-$0.25^{\circ}$ & $2592$\,px\\
& Elevation & $1^{\circ}$ & $0.6^{\circ}$ & $1944$\,px \\
& Velocity & $0.1\,\text{m}{\cdot}\text{s}^{-1}$ & -- & -- \\
\midrule
\multicolumn{2}{l}{Frame rate} & 5fps & 25fps & 30fps \\
\multicolumn{2}{l}{Height above ground} &  80\,cm & 42\,cm & 145\,cm \\
\bottomrule
\end{tabular}
\label{tab:radial_sensors_specs}
\caption[Specifications of the RADIal's sensor suite]{\textbf{Specifications of the RADIal's sensor suite}. The main characteristics of the \ac{HD} \ac{RADAR}, the \ac{LiDAR} and the camera are reported. Their synchronized signals are complemented by \ac{GPS} and \ac{CAN bus} information.}
\end{minipage}
\quad
\begin{minipage}[b]{.48\linewidth}

\centering
\includegraphics[width=0.7\columnwidth]{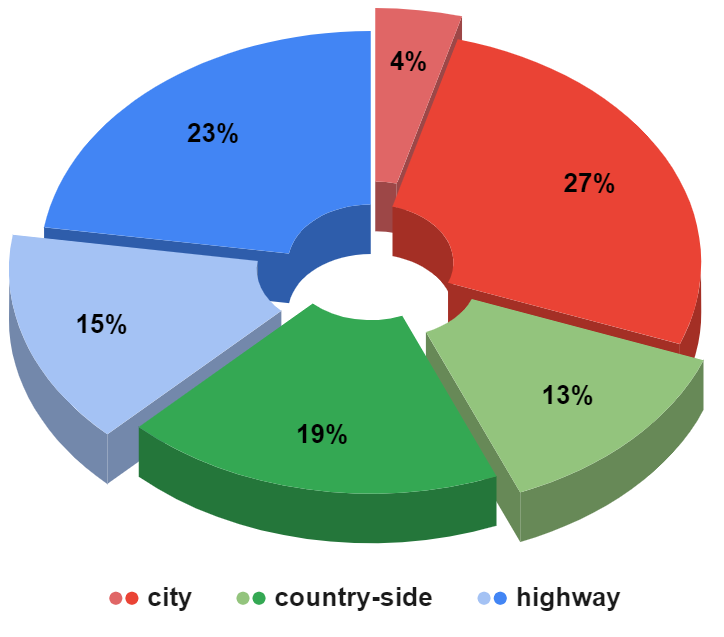}
\label{fig:radial_dataset}
\caption[Scene-type proportions in RADIal]{\textbf{Scene-type proportions in RADIal}. The dataset contains 91 sequences in total, captured on city streets, highway or country-side roads, for a total of 25k synchronized frames (dark colors), out of which 8,252 are labelled (light colors).}
\end{minipage}
\end{table}

Central to the proposed RADIal dataset, our \ac{HD} \ac{RADAR} is composed of $\NRx = 16$ receiving antennas and $\NTx = 12$ transmitting antennas, leading to $\NRx \cdot \NTx=192$ virtual antennas. This virtual-antenna array enables reaching a high azimuth angular resolution while estimating objects' elevation angles as well. 
As the \ac{RADAR} signal is difficult to interpret by annotators and practitioners alike, a 16-layer automotive-grade \ac{LiDAR} and a 5 Mpix RGB camera are also provided. The camera is placed below the interior mirror behind the windshield while the \ac{RADAR} and the \ac{LiDAR} are installed in the middle of the front ventilation grid, one above the other. The three sensors have parallel horizontal lines of sight, pointing in the driving direction. Their extrinsic parameters are provided together with the dataset.
RADIal also offers synchronized \ac{GPS} and \ac{CAN bus} traces which give access to the geo-referenced position of the vehicle as well as its driving information such as speed, steering wheel angle and yaw rate.
The sensors' specifications are detailed in Table \ref{tab:radial_sensors_specs}.

RADIal contains 91 sequences of about 1-4 minutes, for a total of 2 hours. This amounts to approximately 25,000 synchronized frames in total, out of which 8,252 are annotated with 9,550 vehicles.
These sequences are categorized in highway, country-side and city driving. The distribution of the sequences is indicated in Figure \ref{fig:radial_dataset}.

The annotation of the \ac{RADAR} signal is hard to achieve as the \ac{RD} representation is not meaningful for the human eye. 
Vehicle detection labels were first generated automatically using supervision from the camera and \ac{LiDAR}.
A RetinaNet model \cite{lin_focal_2017} was used to extract object proposals from the camera. Then, these proposals were validated when both \ac{RADAR} and \ac{LiDAR} agree on the object position from their respective point cloud. Finally, manual verification was conducted to reject or validate the labels. 
The free-space annotation was done fully automatically on the camera images. A DeepLabV3+ \cite{chen_encoder-decoder_2018}, pre-trained on Cityscape, has been fine-tuned with 2 classes (\textit{free} or \textit{occupied}) on a small manually-annotated part of our dataset.

This model segmented each video frame and the obtained segmentation mask was projected from the camera's coordinate system to the radar's one thanks to known calibration. 
Finally, already available vehicle bounding boxes were subtracted from the free-space mask. 
The quality of the segmentation mask is limited due to the automatic method we employed and to the projection inaccuracy from camera to real world.
In the next section, the proposed FFT-RadNet architecture and its associated multi-task loss will be presented.

\section{Proposed method}
\label{sec:fft_radnet_method}

\begin{figure}[!t]
\centering
\includegraphics[width=1\linewidth]{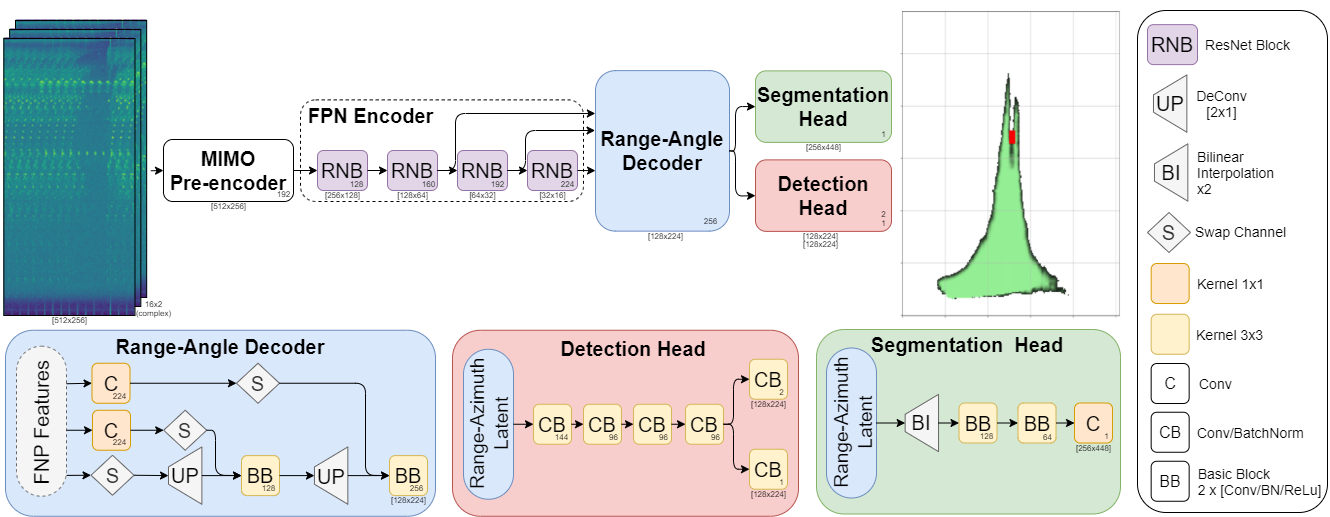}
  \caption[Overview of FFT-RadNet]{\textbf{Overview of FFT-RadNet}. FFT-RadNet is a lightweight multi-task architecture. It does not use any \acl{RA} maps or \acl{RAD} tensor which would require costly pre-processing. Instead, it leverages complex \acl{RD} containing all the range, azimuth and elevation information. This data is de-interleaved and compressed by the \ac{MIMO} pre-encoder. A \acl{FPN} encoder extracts a pyramid of features which the \acl{RA} decoder converts into a latent Range-Azimuth representation. Based on this representation, multi-task heads finally detect vehicles (red rectangle) and predict the free driving space (green shape) in a \acl{BEV} map illustrated in the image on the right.
  } 
  \label{fig:fft_radnet_archi}
\end{figure}

Our approach has been motivated by automotive constraints: automotive-grade sensors must be used and only limited processing/memory resources are available on the embedded hardware.
In this context, the \ac{RD} is the only representation that is practical for \ac{HD} \ac{RADAR}. Based on it, we propose a multi-task architecture, compatible with above requirements, which is composed of five blocks (see Figure \ref{fig:fft_radnet_archi}):
\begin{itemize}
  \item A pre-encoder reorganizing and compressing the \ac{RD} tensor into a meaningful and compact representation;
  \item A shared \ac{FPN} encoder combining low-resolution semantic information with high-resolution details;
  \item A \ac{RA} decoder building a Range-Azimuth latent representation from the feature pyramid;
  \item A detection head localizing vehicles in Range-Azimuth coordinates;
  \item A segmentation head predicting the free driving space.
\end{itemize}

\subsection{MIMO pre-encoder}
\label{sec:fft_radnet_mimo_layer}

As explained in Section \ref{sec:background_radar_theory}, the \ac{MIMO} configuration implies one complex \ac{RD} representation per receiver after the Range-\ac{FFT} and the Doppler-\ac{FFT}.
This results in a complex 3D tensor of dimension $(\BinR,\BinD,\NRx)$, where $\BinR$ and $\BinD$ are the number of discretization bins for range and Doppler respectively. 
It is important to understand how a given reflecting object, say a car in front, appears in this data. Denote $R$ the actual radial distance of this object to the \ac{RADAR} and $D$ its relative radial velocity expressed in Doppler effect. For each receiver, its signature will be visible $\NTx$ times, one per transmitter. More specifically, it will be measured at \ac{RD} positions $(R,(D+k\Delta)[D_\text{max}])_{k=1\cdots\NTx}$, where $\Delta$ is the Doppler shift (induced by the phase shift $\Delta_{\phi}= \phi(t)$ in the transmitted signal, see Section \ref{sec:background_radar_theory}) and $D_{\text{max}}$ is the largest Doppler that can be measured. The measured Doppler values are modulo this maximum. This phenomena is illustrated in Figure \ref{fig:radial_teaser}(b).

\begin{figure}[t]
    \centering
    \includegraphics[width=0.5\columnwidth]{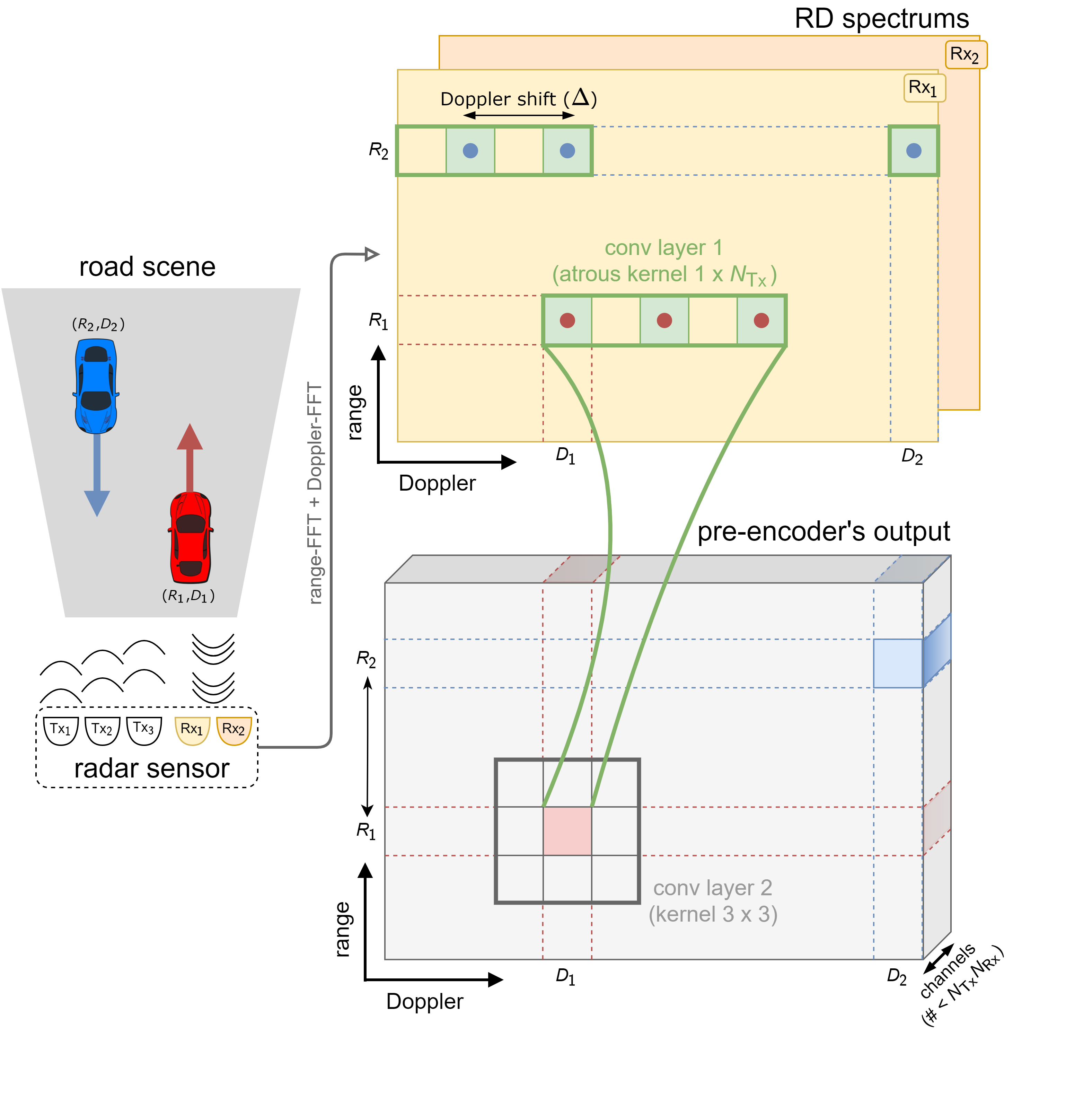}
    \caption[Trainable MIMO pre-encoder]{\textbf{Trainable MIMO pre-encoder}. Considering three transmitters ($\NTx{=}3$) and two receivers ($\NRx{=}2$), an object's signature is visible $\NTx$ times in the \acl{RD} representation. The pre-encoder uses atrous convolutions to organize and compress signatures in fewer than $\NTx \cdot \NRx$ output channels.
    }
    \label{fig:mimo_preencoder}
\end{figure}

This signal intricacy calls for a rearrangement of the \ac{RD} tensor that will facilitate a subsequent exploitation of the \ac{MIMO} information (to recover angles) while keeping data volume under control.
To this end, we propose a new trainable pre-encoder illustrated in Figure \ref{fig:mimo_preencoder} that performs a compact reorganization of the input tensor. 
In order to handle its specific structure along the Doppler axis, we use first a suitably-defined atrous convolution layer that gathers Tx and Rx information at the right positions. 
The size of its kernel for one input channel is $1{\times}\NTx$, hence defined by the number of Tx antennas, and its dilation amounts to $\delta = \frac{\Delta \BinD}{D_{\text{max}}}$, the number of Doppler bins corresponding to the Doppler shift $\Delta$. The number of input channels is the number $\NRx$ of Rx antennas.
A second convolution layer, with a $3 {\times} 3$ kernel, learns how to combine these channels and compresses the signal.
The two-layer pre-encoder is trained end-to-end with the rest of the proposed architecture.

\subsection{FPN encoder}
Using a pyramidal structure to learn multi-scale features is a common practice in object detection and semantic segmentation as detailed in Sections \ref{sec:background_detection} and \ref{sec:background_segmentation}.
Our \ac{FPN} architecture uses 4 blocks composed of 3, 6, 6, 3 residual layers \cite{he_deep_2016} respectively.
The feature maps of these residual blocks form the feature pyramid. This encoder has been optimized considering the nature of the data while controlling its complexity. 
The channel dimensions are chosen to encode at best the azimuth angle over the entire distance range (\textit{i.e.}, high resolution and narrow field of view at far range, low resolution and wider field of view at near range). 

To prevent losing the signature of small objects (typically few pixels in the \ac{RD} representation), the \ac{FPN} encoder performs a $2{\times}2$ down-sampling per block, leading to a total reduction of the tensor size by a factor of 16 in height and width.
For similar reasons and to avoid overlaps between adjacent Tx's, it uses $3{\times}3$ convolution kernels.

\subsection{Range-Angle decoder}
The \ac{RA} decoder aims to expand the input feature maps to higher resolution representations.
This up-scaling is usually achieved through multiple deconvolution layers whose output is combined with previous feature maps to preserve spatial details. 
In our case, the representation is unusual due to the physical nature of the axes: The dimensions of the input tensor correspond respectively to range, Doppler and azimuth angle, whereas the feature maps that will be sent to the subsequent task heads should correspond to a Range-Azimuth representation. 
Consequently, we swap the Doppler and azimuth axes to match the final axis ordering and then upscale the feature maps. 

However, the range axis has a lower size compared to the azimuth one, since it was decimated by a factor of 2 after each residual block, while the azimuth axis (formerly the channel axis) was increasing. 
Prior to these operations, we apply a $1{\times}1$ convolution to the feature maps from the encoder to the decoder. It adjusts the dimension of the azimuth channel to its final size, right before swapping the axes. The deconvolution layers upscale only the range axis, producing feature maps that are concatenated with those from the previous pyramid level. 
A final block of two Conv-BatchNorm-ReLU layers is applied, generating the final Range-Azimuth latent representation. 
The proposed \ac{RA} decoder is illustrated in Figure \ref{fig:fft_radnet_archi}.

\subsection{Multi-task learning}
\label{sec:fft_radnet_multi-task}

\paragraph{Detection task.}
The detection head is inspired from PIXOR \cite{yang_pixor_2018}, an efficient and scalable single-stage model. Further details on the PIXOR architecture and method are provided in Section \ref{sec:background_3d}. 
It takes the \ac{RA} latent representation as input and processes it using a first common sequence of four Conv-BatchNorm layers with 144, 96, 96 and 96 filters respectively.
The branch is then divided in a classification and a regression heads.
The classification part is a convolution layer with sigmoid activation that predicts a probability map.
This output corresponds to a binary classification of each ``pixel'' as occupied or not by a vehicle.
In order to reduce computational complexity, it predicts a coarse \ac{RA} map, where each cell has a resolution of $0.8$m in range and $0.8^{\circ}$ in azimuth (\textit{i.e.}, $\sfrac{1}{4}$ and $\sfrac{1}{8}$ of native resolutions resp. in range and azimuth). This cell size is enough to dissociate two close objects. Then, the regression part finely predicts the range and azimuth values corresponding to the detected object. To do so, a unique $3{\times}3$ convolution layer outputs two feature maps corresponding to the final range and azimuth values. 

This two-fold detection head is trained with a multi-task loss composed of a focal loss, specialized in unbalanced classification, applied to all the locations and of a ``smooth L1'' loss for the regression applied only on positive detection as detailed in the work of \cite{yang_pixor_2018}.
Let $\vect{x}$ be a training example, $\vect{y}_{\text{clas}} \in \{ 0, 1\}^{\sfrac{\BinR}{4}{\times}\sfrac{\BinA}{8}}$ its classification ground truth and $\vect{y}_{\text{reg}} \in \mathbb{R}^{2{\times}\sfrac{\BinR}{4}{\times}\sfrac{\BinA}{8}}$ the associated regression ground truth. 
The detection head of FFT-RadNet predicts a detection map $\hat{\vect{y}}_{\text{clas}} \in [0, 1]^{\sfrac{\BinR}{4}{\times}\sfrac{\BinA}{8}}$ and associated regression map $\hat{\vect{y}}_{\text{reg}}\in \mathbb{R}^{2{\times}\sfrac{\BinR}{4} \times \sfrac{\BinA}{8}}$. Its training loss is written:
\begin{equation}
    \mathcal{L}_{\text{det}}(\vect{x},\vect{y}_{\text{clas}},\vect{y}_{\text{reg}}) = \text{focal}(\vect{y}_{\text{clas}}, \hat{\vect{y}}_{\text{clas}}) + \beta\,\text{smooth-L1}(\vect{y}_{\text{reg}} - \hat{\vect{y}}_{\text{reg}}),
\end{equation}
where $\beta>0$ is a balancing hyper-parameter. The $\text{focal}$ and $\text{smooth-L1}$ are respectively defined as:
\begin{equation}
    \text{focal}(\vect{y}, \mathbf{p})= 
\begin{cases}
    -(1- \mathbf{p})^{ \gamma }log(\mathbf{p}) & \text{if } \vect{y}=1,\\
    -\mathbf{p}^{ \gamma }log(1-\mathbf{p})   & \text{otherwise},
\end{cases}
\end{equation}
where $\gamma$ is a down-weighting factor and
\begin{equation}
        \text{smooth-L1}(\vect{x})= 
\begin{cases}
    0.5\vect{x}^{2}  & \text{if } \abs{\vect{x}}<1,\\
    |\vect{x}|-0.5    & \text{otherwise},
\end{cases}
\end{equation}
where $|.|$ denotes the absolute value.

\paragraph{Segmentation task.}
The free driving space segmentation task is formulated as a pixel-level binary classification. The segmentation mask has a resolution of $0.4$m in range and $0.2^{\circ}$ in azimuth. It corresponds to half of the native range and azimuth resolutions  while considering only half of the entire azimuth \ac{FoV} (within $[-45^\circ, 45^\circ]$). The \ac{RA} latent representation is processed by two consecutive groups of two Conv-BatchNorm-ReLu blocks, producing respectively 128 and 64 feature maps. A final $1{\times}1$ convolution outputs a 2D feature map followed by a sigmoid activation to estimate the probability of each location to be drivable. 
Let $\vect{x}$ be a training example, $\vect{y}_{\text{seg}} \in \{ 0, 1\}^{\sfrac{\BinR}{2}{\times}\sfrac{\BinA}{4}}$ its one-hot ground truth and $\hat{\vect{y}}_{\text{seg}} \in \left[ 0, 1\right]^{\sfrac{\BinR}{2}{\times}\sfrac{\BinA}{4}}$ the predicted soft detection map.
The segmentation task is learnt using a \ac{BCE} loss:
\begin{equation}
    \mathcal{L}_{\text{free}}(\vect{x},\vect{y}_{\text{seg}}) = \sum_{(r, a) \in \Omega} \text{BCE}(\vect{y}_{\text{seg}}(r,a),\hat{\vect{y}}_{\text{seg}}(r,a)),
\end{equation}
where $\Omega = \llbracket 1,\frac{\BinR}{2}\rrbracket \times \llbracket 1,\frac{\BinA}{4}\rrbracket$.

\paragraph{End-to-end multi-task training.}
The whole FFT-RadNet model is trained by minimizing a combination of the previous detection and segmentation losses:
\begin{equation}
    \mathcal{L}_{\text{MTL}} = \sum_{\vect{x}} \mathcal{L}_{\text{det}}(\vect{x},\vect{y}_{\text{clas}},\vect{y}_{\text{reg}}) + \lambda \mathcal{L}_{\text{free}}(\vect{x},\vect{y}_{\text{seg}}),
\end{equation}
w.r.t. the parameters of the \ac{MIMO} pre-encoder, of the \ac{FPN} encoder, of the \ac{RA} decoder and of the two heads, where $\lambda$ is a positive hyper-parameter that balances the two tasks.
The following section will present the experiments and results of the FFT-RadNet with the RADIal dataset.

\section{Experiments and Results}
\label{sec:fft_radnet_expe_radial}

\begin{table}[!t]
\tiny
\centering
\setlength{\tabcolsep}{3pt}
\begin{tabular}{l | c| c c c c| c c c c |c c c c}
\toprule
\multirow{2}{*}{Model} & RADAR &\multicolumn{4}{c|}{Overall} & \multicolumn{4}{c|}{Easy} & \multicolumn{4}{c}{Hard}\\
 & Input & AP($\%$) $\uparrow$ & AR($\%$) $\uparrow$ & R(cm) $\downarrow$ &  A($^{\circ}$) $\downarrow$ & AP($\%$) $\uparrow$ & AR($\%$) $\uparrow$ & R(cm) $\downarrow$ &  A($^{\circ}$) $\downarrow$ & AP($\%$) $\uparrow$ & AR($\%$) $\uparrow$ & R(cm) $\downarrow$ &  A($^{\circ}$) $\downarrow$\\
\midrule
PIXOR & PC & 96.46 & 32.32 & 0.17 & 0.25 & $\textbf{99.02}$ & 28.83 & 0.15 & 0.19 & 93.28 & 38.69 & 0.19 & 0.33\\
PIXOR & RA & 96.56 & 81.68 & $\textbf{0.10}$ & 0.20 &
               96.86 & 88.02 & $\textbf{0.09}$ & 0.16 &
               $\textbf{95.88}$ & $\textbf{70.10}$ & $\textbf{0.12}$ & 0.27\\
FFT-RadNet (ours) & RD & $\textbf{96.84}$ & $\textbf{82.18}$ & 0.11 & $\textbf{0.17}$	&
                            98.49 & $\textbf{91.69}$ & 0.10 & $\textbf{0.13}$ & 
                            92.93 & 64.82 & 0.13 & $\textbf{0.26}$\\ 
\bottomrule
\end{tabular}
\label{tab:fft_raddet_perf_detection}
\caption[Object detection performances on RADIal Test split]{\textbf{Object detection performances on RADIal Test split}.
Comparison between PIXOR \cite{yang_pixor_2018} trained with Point Cloud (PC) or Range-Azimuth (RA) representations, and the proposed FFT-RadNet requiring only \ac{RD} as input. Our method obtains similar or better overall performances than baselines in both Average Precision (AP) and Average Recall (AP) for a 50\% IoU threshold. It also reaches similar or better Range (R) and Angle (A) accuracy, showing it successfully learns a signal processing pipeline that estimates the \ac{AoA} with significantly fewer operations, as detailed in Table \ref{tab:fft_radnet_complexity}.
}
\end{table}

\subsection{Training details}
The proposed architecture has been trained on the RADIal dataset using exclusively the \ac{RD} representations as input. The \ac{RD} being composed of complex numbers, their real and imaginary parts are stacked along the channel axis and used as input of the \ac{MIMO} pre-encoder.
The dataset has been split into Training, Validation and Test sets (approximately 70\%, 15\% and 15\% of the dataset, respectively) in such a way that frames from a same sequence can not appear in different sets.
We manually split the Test dataset into ``hard'' and ``easy'' cases. Hard cases are mostly situations where the \ac{RADAR} signal is perturbed, \textit{e.g.}, by interference with other \acp{RADAR}, important side-lobes effects or significant reflections on metallic surfaces. 

The FFT-RadNet architecture is trained using the multi-task loss detailed in Section \ref{sec:fft_radnet_multi-task} with the following hyper-parameters set-up empirically: $\lambda=100$, $\beta=100$ and $\gamma=2$.
The training process uses the Adam optimizer \cite{kingma_adam_2015} during 100 epochs, with an initial learning rate of $10^{-4}$ and a decay of 0.9 every 10 epochs.

\subsection{Baselines}

The proposed architecture has been compared to recent contributions in deep learning and in \ac{RADAR} scene understanding.
Most of the competing methods presented in Chapter \ref{chap:related_work} have been designed for \ac{LD} \ac{RADAR} and can not scale with \ac{HD} \ac{RADAR} data due to memory limitation.
Instead, baselines with similar complexity have been selected regarding their input representation (Range-Azimuth or point cloud) for a fair comparison.
Input representations (\ac{RD}, \ac{RA} or point cloud) are generated for the entire Training, Validation and Test sets using a conventional signal processing pipeline.

\paragraph{Object detection with point cloud.} 
\sloppy
The PIXOR \cite{yang_pixor_2018} method has been adapted to detect vehicles after voxelization of the \ac{RADAR} point cloud into a 3D volume of 
$[0\,\text{m}, 103\,\text{m}] {\times} [−40 \,\text{m}, 40 \,\text{m}] {\times}[-2.5 \,\text{m}, 2.0 \,\text{m}]$ around the \ac{RADAR} (longitudinal, lateral and vertical ranges), sampled at $0.1$m in each direction. 
The size for this input 3D grid is thus $800 {\times} 1030{\times} 45$.
PIXOR is a lightweight architecture intended to be real-time. However, its input representation generates 96MB of data, which becomes a challenge for embedded devices.

\paragraph{Object detection with RA tensor.}
As detailed in Chapter \ref{chap:related_work} and Section \ref{sec:mvrss}, several methods \cite{major_vehicle_2019, gao_ramp-cnn_2020} including ours used views of the \ac{RAD} tensor as input. However, the memory usage would be too extensive for \ac{HD} \ac{RADAR} data. As \cite{major_vehicle_2019} showed that using only the \ac{RA} view leads to better performance for object detection, we compared our method to a PIXOR architecture without the voxelization module. It takes as input the \ac{RA} representation in RADIal, of size $512{\times}896$ with range values in $[0 \text{m}, 103 \text{m}]$ and azimuth in $[-90^\circ, 90^\circ]$.

\paragraph{Freespace segmentation.} 
We selected PolarNet \cite{nowruzi_deep_2020} to evaluate against our approach. It is a lightweight architecture designed to process \ac{RA} maps and predict free space. We re-implemented the model to the best of our ability.

\subsection{Evaluation metric}
For object detection, the \ac{AP} and \ac{AR} are used considering an \ac{IoU} threshold of 50\%. 
For semantic segmentation, the \ac{mIoU} metric is used on a binary classification task (\textit{free} or \textit{occupied}). The metric is computed on a reduced $[0 \text{m},50 \text{m}]$ range as the boundaries of the road surface are hardly visible beyond this distance. The evaluation metrics are detailed in Section \ref{sec:background_detection}.

\begin{figure}[!t]
\centering
\includegraphics[width=1\textwidth]{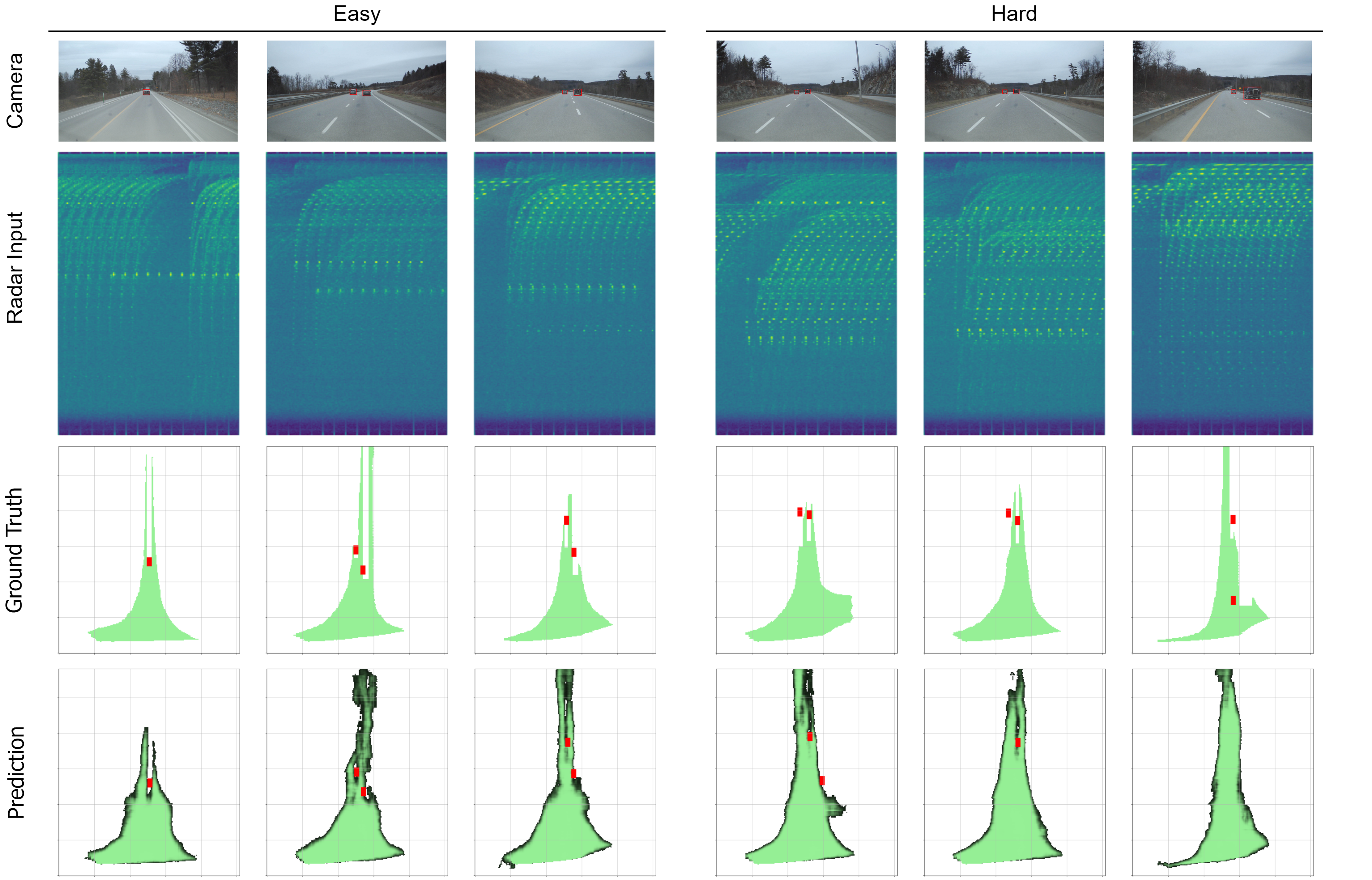}
\caption[Qualitative results for object detection and free space segmentation on Easy and Hard samples]{\textbf{Qualitative results for object detection and free space segmentation on Easy and Hard samples}. Camera views (1st row) are displayed for visual reference only; \acl{RD} representation (2nd row) are the only inputs to the model; Ground truths (3rd row) and predictions (4th row) are shown for both tasks.
Ground truths and predictions are represented by \acl{BEV} maps with 2D red boxes for object detection and green shapes for free driving space segmentation.
Note that there could be a projection error of the free driving space from camera to real world due to vehicle pitch variations.}
\label{fig:fft_radnet_qualitative}
\end{figure}

\subsection{Performance analysis}

\paragraph{Object detection.} 
Performances for object detection are reported in Table \ref{tab:fft_raddet_perf_detection}. We observe that FFT-RadNet using \ac{RD} as input outperforms all baselines overall. 
The position accuracy, both in range and azimuth angle, is similar, and even better in angle, compared to PIXOR using \ac{RA} as input (PIXOR-RA).
These results show that our approach successfully learns the azimuth angle from the data. From a manufacturing viewpoint, note that this opens cost saving opportunities as the end-of-line calibration of the sensor is no longer required in the proposed framework. 
In the Easy Test set, FFT-RadNet delivers +1.6\% AP and +3.6\% AR compared to PIXOR-RA. However, on the Hard test set, PIXOR-RA performs the best.  
The \ac{RA} approach works well with the hard samples because the data is pre-processed by a signal processing pipeline that already solves some of these cases. 
In contrast, the performance with point-cloud input is much lower than all others. Indeed, the recall is low due to the limited number of points at far range. 
Qualitative results of FFT-RadNet on the Easy and Hard Test sets are illustrated in Figure \ref{fig:fft_radnet_qualitative}.

\paragraph{Free driving space segmentation.} 
The performance for the free driving space segmentation is provided in Table \ref{tab:fft_radnet_perf_freespace}. We observe that FFT-RadNet significantly outperforms PolarNet by 13.4\% \ac{IoU} on average. This is partly explained by the lack of elevation information in the \ac{RA} representation, an information that is present in the \ac{RD}. 
Segmentation results on the Easy and Hard Test sets are also presented in Figure \ref{fig:fft_radnet_qualitative}.

\subsection{Complexity analysis}
\label{sec:fft_radnet_complexity}

\begin{table}[!t]
\begin{minipage}[b]{.48\linewidth}
\centering
\scriptsize
\setlength{\tabcolsep}{5pt}

\begin{tabular}{l  c  c  c  c}
\toprule
\multirow{2}{*}{Model} & \multirow{2}{*}{RADAR input} & \multicolumn{3}{c}{mIoU (\%) $\uparrow$}\\ \cmidrule(l){3-5}
&  & Overall & Easy & Hard\\
\midrule
PolarNet & RA & 60.6 & 61.9 & 57.4 \\
FFT-RadNet & RD & \textbf{74.0} & \textbf{74.6} & \textbf{72.3} \\
\bottomrule
\end{tabular}
\label{tab:fft_radnet_perf_freespace}
\caption[Free driving space segmentation performances]{\textbf{Free driving space segmentation performances}. FFT-RadNet successfully approximates the angle information in the \ac{RADAR} data while reaching better performance than PolarNet. Note that this performance is achieved by FFT-RadNet while simultaneously performing object detection, as our model is multi-task.}
\end{minipage}
\quad
\begin{minipage}[b]{.48\linewidth}
\centering
\scriptsize
\setlength{\tabcolsep}{2pt}

\begin{tabular}{l c c c c}
\toprule
\multirow{2}{*}{Method} & \multirow{2}{*}{\shortstack{Input size \\ (MB)  $\downarrow$}} & \# \multirow{2}{*}{\shortstack{Params. \\ ($10^{6}$) $\downarrow$}} & \multicolumn{2}{c}{Complexity (GFLOPS) $\downarrow$}\\
& & & AoA processing & Model\\
\midrule
PCL PIXOR  & 98.30 & 6.93 & 8 & 741\\
RA PIXOR   & $~~\textbf{1.75}$ & 6.92 & 45* & 761\\
FFT-RadNet  & 16.00 & $\textbf{3.79}$ & $\textbf{0}$ & $\textbf{584}$\\ 
\bottomrule
\end{tabular}%
\caption[Complexity analysis of FFT-RadNet]{\textbf{Complexity analysis of FFT-RadNet}. The proposed method reaches the best trade-off between the size of the input, the number of parameters of the model and the computational complexity. Note that the \acl{AoA} processing of the \ac{RA} PIXOR method (*) considers only a single elevation, otherwise it is up to 496 \ac{GFLOPS} for the whole set of $\BinE\! =\! 11$ elevations.}
\label{tab:fft_radnet_complexity}
\end{minipage}
\end{table}

FFT-RadNet has been designed to compress of the signal processing chains that transform the \ac{ADC} data into either a sparse point cloud or denser representations (\ac{RA} or \ac{RAD}), without compromising the richness of the signal.
Because the input data remains quite large, we designed a compact model to bound the complexity in terms of number of operations, as a trade-off between performance and range/angle accuracy. Moreover, the pre-encoder layer compresses significantly the input data. An ablation study has been performed to define the best trade-off between the size of the feature maps and the model's performance (details in Appendix \ref{sec_app:fft_radnet_ablation}).

As shown in Table \ref{tab:fft_radnet_complexity}, FFT-RadNet is the only method that does not require the \ac{AoA} estimation. As explained in Section \ref{sec:fft_radnet_mimo_layer}, the pre-encoder layer compresses the \ac{MIMO} signal containing all the information to recover both the azimuth and elevation angles. The \ac{AoA} for the point cloud approach generates 3D coordinates for a sparse cloud of around 1000 points on average, leading to 8 \ac{GFLOPS} worth of computing, prior to applying PIXOR for object detection. 

To produce the \ac{RA} or \ac{RAD} tensor, the \ac{AoA} runs for each single bin of the \ac{RD} representation, but only considering one elevation. Such a model is thus unable to estimate the elevation of objects such as bridges or lost cargo (low object). For one elevation, the complexity is about 45 \ac{GFLOPS}, but it would increase up to 495 \ac{GFLOPS} for all the 11 elevations. We have demonstrated that FFT-RadNet can cut these processing costs without compromising the quality of the estimation.

\section{Conclusions and discussions}
\label{sec:hd_radar_discussions}

This chapter introduced RADIal, a new dataset containing sequences of automotive-grade sensor signals (\ac{HD} \ac{RADAR}, camera and \ac{LiDAR}).
Synchronized sensor data are available in a raw format so that various representations can be evaluated and further research can be conducted, possibly with fusion-based approaches.
The FFT-RadNet is also presented as a novel trainable architecture to process and analyse \ac{HD} \ac{RADAR} signals. We demonstrated that it effectively alleviates the need for costly pre-processing to estimate \ac{RA} or \ac{RAD} representations. Instead, it detects and estimates objects position while segmenting free driving space from the \ac{RD} representations directly. Experiments on the RADIal dataset shown that FFT-RadNet slightly outperforms \ac{RA}-based and point cloud-based approaches while reducing processing requirements. 

The annotations of the RADIal dataset have been performed semi-automatically and thus could be improved. Figure \ref{fig:fft_radnet_qualitative} illustrates misalignment of the camera's semantic segmentation mask projections to the Cartesian coordinates in the ground truths where the object boxes do not perfectly match with the shape of the free space segmentation.
An additional correction could be integrated by quantifying the projection error due to the calibration of the camera.
%
The azimuth angle estimation could be improved by defining additional pyramid features in the \ac{MIMO} pre-encoder layer of the FFT-RadNet architecture. The number of feature maps defined at each level of the pyramid will correspond to a smoother angle resolution at each scale of the input.
Finally, an extensive ablation study of the \ac{MIMO} pre-encoder could also be explored to better understand the contribution of the virtual pairs of antennas regarding their position in the array of antennas in the sensor. 
Our experiments showed that best results are reached by considering a subset of the virtual antennas. A deeper analysis on which antennas contribute the most could be interesting to selectively build the array of antennas of a \ac{HD} \ac{RADAR} and possibly reducing its cost.

\ac{HD} \ac{RADAR} is expected to be increasingly used in the near feature for scene understanding due to its improved angular resolution. It is complementary to the \ac{LiDAR} and camera sensor thanks to its robustness to adverse weather conditions and the recorded Doppler information. It could even replace the \ac{LiDAR} sensor thanks to the azimuth and elevation resolutions achieved with hundreds of virtual antennas. Sensor fusion considering \ac{HD} \ac{RADAR}, \ac{LiDAR} and camera will be explored to further improve scene understanding tasks in the context of autonomous driving.

\chapter{Conclusion}
\label{chap:conclusion}
\minitoc

\section{Contributions}

Advanced driving assistance systems and autonomous driving require a detailed perception around the ego vehicle. Robust scene understanding is key to ensuring safety of road users. Thus, complementary sensors recording redundant information are required to compensate for their limitations while taking advantage of their own physical properties.
Since the recent golden age of deep learning, the performance of algorithms for scene understanding has increased dramatically and autonomous driving has become within reach.
The \ac{RADAR} sensor has been left behind due to its cumbersome and noisy representations which are complex to understand for human eyes.
However, it is the only sensor that is robust to adverse weather conditions, while proving information on the velocity and position of the surrounding objects.

This thesis explored \ac{RADAR} scene understanding using deep learning algorithms. We have addressed the fundamental problem of the lack of annotated datasets while proposing appropriate algorithms to explore \ac{RADAR} representations.
First, we created a simple simulation of \ac{RD} representation by taking into account the properties of moving objects while including \ac{RADAR} sensor noise (Section \ref{sec:rd_simulation}). We experimented with generative approaches to reconstruct \ac{RD} representations from camera images with the underlying idea to automatically transfer annotations between the two domains (Section \ref{sec:radar_generation}). Both methods aimed to build annotated \ac{RADAR} dataset without the costly intervention of human experts. These methods have not been able to produce sufficiently realistic \ac{RADAR} data to investigate these directions further.
Therefore we proposed the CARRADA dataset (Section \ref{sec:carrada}), composed of synchronised camera and annotated \ac{RADAR} data for scene understanding. 
As detailed in Table \ref{tab:related_public_dataset}, it is still the only dataset providing raw and processed \ac{RADAR} data with both object detection and semantic segmentation annotations.
A semi-automatic annotation pipeline has also been proposed, taking advantage of camera images to reduce the cost of manual annotation.
Our annotation method has been slightly improved in the work of \cite{zhang_raddet_2021}, as detailed in Section \ref{sec:mvrss_raddet_dataset}, demonstrating the relevance of its application to complex urban scenes.
However, the efficiency of the camera projections in the \ac{RADAR} data is based on strong assumptions, \textit{e.g.} non-occluded objects or objects at close range, and leads to inaccuracies in the annotations. Our proposed propagation of the annotations in the \ac{RADAR} sequence alleviates the dependence on the camera image, but it is also limited regarding the object signature tracking which remains difficult when small objects are close together. 
Using increased human supervision or a higher resolution sensor could solve these problems.
Finally, our method is limited in the generation of the \ac{RA} annotations because it does not consider the angular resolution of the sensor. Therefore the generated annotations do not cover entirely the objects' signature on the \ac{RA} view. This could be mitigated by incorporating the angular resolution of the sensor into the shape extension method presented in Section \ref{sec:proj_and_annot}.

In the past two years, the release of annotated datasets has opened up research for \ac{RADAR} scene understanding using deep learning algorithms. As detailed in Chapter \ref{chap:related_work}, mostly object detection has been explored, however it is not well suited to the shapes of objects' signatures. 
We proposed a novel approach for multi-view \ac{RADAR} semantic segmentation exploiting the raw \ac{RAD} tensor of \ac{RADAR} data while reducing its cumbersome representation and its noise (Section \ref{sec:mvrss}). Our proposed approach reached the state of the art performances of semantic segmentation considering several competing methods while requiring significantly fewer parameters. 
These results were confirmed after various ablation studies and evaluations on several datasets, including simple and complex urban scenes.
Nonetheless, these small datasets are unbalanced and it is still difficult to provide a fair and precise evaluation on vulnerable road users (pedestrians and cyclists), which are usually under represented.
Another limitation of our work is the aggregation method we employed for the \ac{RAD} tensor. It helps to reduce the noise, as detailed in Section \ref{sec:background_signal_process}, but it also attenuates the high reflections of the objects making difficult to separate them.
Moreover, the proposed \acl{CoL} in Section \ref{sec:mvrss_col} showed limits in improving radar semantic segmentation performances only on a single view. The spatial coherence could integrate the ground truth segmentation to penalize the network prediction in a robust manner.

Each sensor used for automotive scene understanding has its strengths and weaknesses. Sensor fusion aims to benefit from their advantages while compensating their limitations. In this thesis, we proposed a preliminary work to exploit jointly \ac{RADAR} and \ac{LiDAR} point clouds with an early fusion method (Section \ref{sec:sensor_fusion}). It aims to quantify the resolution and the accuracy of a low-definition \ac{RADAR} to construct a sensor uncertainty area for each measurement. The \ac{RADAR} information of a point is then propagated to the \ac{LiDAR} points belonging to its uncertainty area and vice-versa. 
The presented qualitative results showed that the \ac{RADAR} information (Doppler and \ac{RCS}) are well-propagated locally. The final point cloud benefits from the density of the \ac{LiDAR} point cloud while carrying the Doppler and \ac{RCS} information in dense local groups of points.
The presented method is limited by the ghost and multi-path reflections of the \ac{RADAR} point cloud propagating information in the wrong local space. However it can be tackled by filtering the upstream points.
The \ac{RADAR} and \ac{LiDAR} point cloud fusion aims to improve scene understanding in general by exploiting both sensors at the same time. Various opportunities are discussed in the following section.

Our recent collaborative project (Chapter \ref{chap:hd_radar}) explored the \ac{HD} \ac{RADAR} sensor by proposing the RADIal dataset, the first dataset composed of camera, \ac{LiDAR} and raw \ac{HD} \ac{RADAR} data annotated for object detection and free driving space segmentation. 
A deep neural network architecture is also proposed to directly learn from the raw  \ac{RADAR} and replace the costly pre-processing steps for angle estimation. It succeeds in attaining better performances than competing methods while requiring fewer parameters and operations. It also operates directly on the raw data, which is an important advantage for real-time applications using \ac{HD} \ac{RADAR}.
The proposed dataset is an important step for \ac{RADAR} understanding, but it has only a single class for object detection and the annotations are performed semi-automatically based on camera calibration, which is inaccurate and could be manually corrected.
The proposed pre-encoder layer could also be improved by considering a smoother estimation of the angular resolution of the sensor which increases with the distance. 
Finally, a deeper study on the impact of the selected pairs of virtual antennas should be realized to better characterize the impact of each one of them. Since we reached the best performances by estimating fewer pairs of virtual antennas than existing in the sensor, such analysis should help to improve the geometric positioning of each antenna in an array of a \ac{RADAR}.

\section{Future work}
\label{sec:conclusion_future}

Multiple datasets with \ac{RADAR} data have been created in the past two years (see Table \ref{tab:related_public_dataset}) opening up research in object detection and semantic segmentation for scene understanding.
However, the community is still missing a large scale dataset containing camera, \ac{LiDAR} and raw \ac{RADAR} data with various scenarios (lighting and weather conditions) and diverse annotations. It is important to obtain such datasets to further explore fully supervised learning using \ac{RADAR} data that the car will rely on in certain scenarios.
It is also important to provide the raw data to identify the most appropriate representation in different driving scenarios. \textit{E.g.}, considering \ac{LD} \ac{RADAR}, a point cloud representation is enough to detect vehicles but it will miss pedestrians at middle and long range.

We discussed in Chapter \ref{chap:radar_scene_understanding} the importance of raw data, in particular the \ac{RAD} tensor, with semantic segmentation annotation for scene understanding using \ac{LD} \ac{RADAR} sensor. 
A simple aggregation method (see Equation \ref{eq:agg_method_average}) is then applied to reduce the noise and the size of the representation. The downside of this method is that high reflections are attenuated. 
In a current work, we are exploring diverse aggregation methods to further reduce the noise while respecting the high reflection distributions and separating them from the different classes.
One way to carry out this idea is to use a weighted average depending on the bin intensity values.
We note that the selection of specific slices per view could also be considered with a measure of signal disparity.
We are also formulating the \acl{CoL} differently to better balance the performance in multi-view \ac{RADAR} semantic segmentation (see Section \ref{sec:mvrss_col}). In particular, we are trying to integrate the ground truth in the loss to better impose the spatial coherence between the prediction in the two branches of the neural network architecture. Since this loss aims to improve the performances in the \ac{RA} view, we are also experimenting with a local back-propagation of the Coherence loss in specific branches of the network to avoid penalization in the \ac{RD} branches.

In our opinion, raw data is the most relevant for scene understanding, but most of the available datasets and integrated algorithms use lightweight \ac{RADAR} point clouds. This representation is sparse and misses numerous objects, but it carries the Doppler and the signal reflection (\ac{RCS}). We proposed a method which propagates these information through the dense \ac{LiDAR} point cloud while fusing them to benefit from their respective properties (Section \ref{sec:sensor_fusion}). 
Our next work will consist in training deep neural network architectures specialized in point clouds (\textit{e.g.} PointNet \cite{qi_pointnet_2017} or PointNet++ \cite{qi_pointnet_2017-1}) using our enriched point cloud to improve scene understanding tasks. 
Motivated by the lack of annotations and the difficulty of creating them even for a human, we will explore self-supervised learning consisting in training a model with a pretext task and to fine-tune it on a downstream application. As a first experiment, we will used our enriched point cloud to learn a model to predict the propagated Doppler components. This way, the model will be able to learn geometric features while distinguishing moving and static objects without any annotation cost.
Our propagation and fusion module could be useful to consider \ac{RADAR} data as prior weak annotations in a self-supervised setting. The main advantage is that it could be applied to any dataset with a \ac{LiDAR} and at least one \ac{RADAR}, either \ac{LD} or \ac{HD}.

The recent \ac{HD} \ac{RADAR} sensors attain a resolution similar to a \ac{LiDAR} in azimuth and elevation angles but this requires costly pre-processing steps. This issue is barely discussed in the related work on \ac{HD} \ac{RADAR} but it is essential for real time predictions. 
In our collaborative work (Chapter \ref{chap:hd_radar}), we showed that a neural network architecture is able to estimate the \ac{RADAR} pre-processing while decreasing its computational complexity. This multi-task architecture also increases performance compared to recent competing methods, as each method is specialised in its own task.
Real time predictions and high performances are reached with a \ac{HD} \ac{RADAR}. Moreover it provides representations as dense as \ac{LiDAR} does with a comparable resolution. Additional experiments must be conducted to compare \ac{HD} \ac{RADAR} and \ac{LiDAR} sensors performance individually on automotive scene understanding tasks. The \ac{HD} \ac{RADAR} suffers from ghost and multi-path reflections but if a neural network can distinguish the noise, as it does in our experiments, one can argue that this sensor may replace \ac{LiDAR} in autonomous cars.

With respect to the limitations of the camera and \ac{LiDAR} sensors, it is clear that the \ac{RADAR} must be integrated into autonomous vehicles in order to obtain sufficient performance in any scenario to ensure the safety of drivers.
In the near future, supervised learning using a \ac{RADAR} sensor will be explored in depth to be able to rely solely on this sensor in several scenarios.
Methods for sensor fusion using \ac{RADAR} are at their early stage but they will be explored with any sensor mounted on a car. In particular, \ac{LD} \ac{RADAR} and \ac{LiDAR} are particularly well suited to be fused and benefit from each of them at lower cost. Future research will also focus on \ac{HD} \ac{RADAR} for fusion, either with camera or \ac{LiDAR}, to provide a dense representation of the scene surrounding the vehicle while benefiting from the properties of each one of them.

Scene understanding for autonomous driving is increasingly being mastered by exploiting multiple sensors in various driving scenarios. It goes hand in hand with decision making that uses the observation data to drive the car. 
The automotive industry estimates that vehicles with a level 3 of driving automation, \textit{i.e.} full capacity to perform driving tasks but requiring human intervention, will be sold to the public in 2024 in European countries. 
If research in this area continues to progress so rapidly, we can estimate that a level of automation that no longer requires a human driver could be available to the public within one or two decades.

\bibliographystyle{ThesisStyleWithEtAl}
\bibliography{manuscript}

\begin{thebibliography}{xxxx}

\bibitem[Abramowitz \& Stegun~1965]{abramowitz_handbook_1965}
Milton Abramowitz and Irene~A. Stegun, editors.
\newblock Handbook of mathematical functions: with formulas, graphs, and
  mathematical tables.
\newblock Dover books on mathematics, 1965.

\bibitem[Aydogdu \textit{et~al.}~2020]{aydogdu_multi-modal_2020}
Cem~Yusuf Aydogdu, Souvik Hazra, Avik Santra and Robert Weigel.
\newblock {\em Multi-{Modal} {Cross} {Learning} for {Improved} {People}
  {Counting} using {Short}-{Range} {FMCW} {Radar}}.
\newblock In {IEEE} {International} {Radar} {Conference} ({RADAR}), 2020.

\bibitem[Azam \textit{et~al.}~2021]{azam_channel_2021}
Shoaib Azam, Farzeen Munir and Moongu Jeon.
\newblock {\em Channel {Boosting} {Feature} {Ensemble} for {Radar}-based
  {Object} {Detection}}.
\newblock In {IEEE} {Intelligent} {Vehicles} {Symposium} ({IV}), 2021.

\bibitem[Badrinarayanan \textit{et~al.}~2017]{badrinarayanan_segnet_2017}
Vijay Badrinarayanan, Alex Kendall and Roberto Cipolla.
\newblock {\em {SegNet}: {A} {Deep} {Convolutional} {Encoder}-{Decoder}
  {Architecture} for {Image} {Segmentation}}.
\newblock In {IEEE} {Transactions} on {Pattern} {Analysis} and {Machine}
  {Intelligence} ({TPAMI}), 2017.

\bibitem[Bai \textit{et~al.}~2020]{bai_vehicle_2020}
Jie Bai, Yifan Zhang, Libo Huang and Sen Li.
\newblock {\em Vehicle {Detection} {Based} on {Deep} {Neural} {Network}
  {Combined} with {Radar} {Attention} {Mechanism}}.
\newblock In Automotive {Technical} {Papers}, 2020.

\bibitem[Barnes \textit{et~al.}~2020]{barnes_oxford_2020}
Dan Barnes, Matthew Gadd, Paul Murcutt, Paul Newman and Ingmar Posner.
\newblock {\em The {Oxford} {Radar} {RobotCar} {Dataset}: {A} {Radar}
  {Extension} to the {Oxford} {RobotCar} {Dataset}}.
\newblock In {IEEE} {International} {Conference} on {Robotics} and {Automation}
  ({ICRA}), 2020.

\bibitem[Bewley \textit{et~al.}~2016]{bewley_simple_2016}
Alex Bewley, Zongyuan Ge, Lionel Ott, Fabio Ramos and Ben Upcroft.
\newblock {\em Simple {Online} and {Realtime} {Tracking}}.
\newblock In {IEEE} {International} {Conference} on {Image} {Processing}
  ({ICIP}), 2016.

\bibitem[Bijelic \textit{et~al.}~2018]{bijelic_benchmark_2018}
Mario Bijelic, Tobias Gruber and Werner Ritter.
\newblock {\em A {Benchmark} for {Lidar} {Sensors} in {Fog}: {Is} {Detection}
  {Breaking} {Down}?}
\newblock In {IEEE} {Intelligent} {Vehicles} {Symposium} ({IV}), 2018.

\bibitem[Bijelic \textit{et~al.}~2020]{bijelic_seeing_2020}
Mario Bijelic, Tobias Gruber, Fahim Mannan, Florian Kraus, Werner Ritter, Klaus
  Dietmayer and Felix Heide.
\newblock {\em Seeing {Through} {Fog} {Without} {Seeing} {Fog}: {Deep}
  {Multimodal} {Sensor} {Fusion} in {Unseen} {Adverse} {Weather}}.
\newblock In {IEEE} {Conference} on {Computer} {Vision} and {Pattern}
  {Recognition} ({CVPR}), 2020.

\bibitem[Brodeski \textit{et~al.}~2019]{brodeski_deep_2019}
Daniel Brodeski, Igal Bilik and Raja Giryes.
\newblock {\em Deep {Radar} {Detector}}.
\newblock In {IEEE} {Radar} {Conference} ({RadarConf}), 2019.

\bibitem[Brooker~2005]{brooker_understanding_2005}
Graham Brooker.
\newblock {\em Understanding millimetre wave {FMCW} radars}.
\newblock In {IEEE} {International} {Conference} on {Software} {Testing}
  ({ICST}), 2005.

\bibitem[Brooks \textit{et~al.}~2018]{brooks_temporal_2018}
Daniel~A. Brooks, Olivier Schwander, Frederic Barbaresco, Jean-Yves Schneider
  and Matthieu Cord.
\newblock {\em Temporal {Deep} {Learning} for {Drone} {Micro}-{Doppler}
  {Classification}}.
\newblock In International {Radiation} {Symposium} ({IRS}), 2018.

\bibitem[Bugeau \& Pérez~2007]{bugeau_bandwidth_2007}
Aurélie Bugeau and Patrick Pérez.
\newblock {\em Bandwidth selection for kernel estimation in mixed
  multi-dimensional spaces}.
\newblock Technical report RR-6286, INRIA, 2007.

\bibitem[Caesar \textit{et~al.}~2020]{caesar_nuscenes_2020}
Holger Caesar, Varun Bankiti, Alex~H. Lang, Sourabh Vora, Venice~Erin Liong,
  Qiang Xu, Anush Krishnan, Yu~Pan, Giancarlo Baldan and Oscar Beijbom.
\newblock {\em {nuScenes}: {A} {Multimodal} {Dataset} for {Autonomous}
  {Driving}}.
\newblock In {IEEE} {Conference} on {Computer} {Vision} and {Pattern}
  {Recognition} ({CVPR}), 2020.

\bibitem[Capobianco \textit{et~al.}~2017]{capobianco_vehicle_2017}
Samuele Capobianco, Luca Facheris, Fabrizio Cuccoli and Simone Marinai.
\newblock {\em Vehicle classification based on convolutional networks applied
  to {FM}-{CW} radar signals}.
\newblock In European {Conference} on {Traffic} {Mining} {Applied} to {Police}
  {Activities} ({TRAP}), 2017.

\bibitem[Cennamo \textit{et~al.}~2020]{cennamo_leveraging_2020}
Alessandro Cennamo, Florian Kaestner and Anton Kummert.
\newblock {\em Leveraging {Radar} {Features} to {Improve} {Point} {Clouds}
  {Segmentation} with {Neural} {Networks}}.
\newblock In International {Conference} on {Engineering} {Applications} of
  {Neural} {Networks} ({EANN}), 2020.

\bibitem[Chadwick \textit{et~al.}~2019]{chadwick_distant_2019}
Simon Chadwick, Will Maddern and Paul Newman.
\newblock {\em Distant {Vehicle} {Detection} {Using} {Radar} and {Vision}}.
\newblock In {IEEE} {International} {Conference} on {Robotics} and {Automation}
  ({ICRA}), 2019.

\bibitem[Chen \textit{et~al.}~2017]{chen_rethinking_2017}
Liang-Chieh Chen, George Papandreou, Florian Schroff and Hartwig Adam.
\newblock {\em Rethinking {Atrous} {Convolution} for {Semantic} {Image}
  {Segmentation}}.
\newblock In {ArXiv}, 2017.

\bibitem[Chen \textit{et~al.}~2018a]{chen_deeplab_2018}
Liang-Chieh Chen, George Papandreou, Iasonas Kokkinos, Kevin Murphy and Alan~L.
  Yuille.
\newblock {\em {DeepLab}: {Semantic} {Image} {Segmentation} with {Deep}
  {Convolutional} {Nets}, {Atrous} {Convolution}, and {Fully} {Connected}
  {CRFs}}.
\newblock In {IEEE} {Transactions} on {Pattern} {Analysis} and {Machine}
  {Intelligence} ({TPAMI}), 2018.

\bibitem[Chen \textit{et~al.}~2018b]{chen_encoder-decoder_2018}
Liang-Chieh Chen, Yukun Zhu, George Papandreou, Florian Schroff and Hartwig
  Adam.
\newblock {\em Encoder-{Decoder} with {Atrous} {Separable} {Convolution} for
  {Semantic} {Image} {Segmentation}}.
\newblock In European {Conference} on {Computer} {Vision} ({ECCV}), 2018.

\bibitem[Chen \textit{et~al.}~2021a]{chen_contactless_2021}
Jinbo Chen, Dongheng Zhang, Zhi Wu, Fang Zhou, Qibin Sun and Yan Chen.
\newblock {\em Contactless {Electrocardiogram} {Monitoring} with {Millimeter}
  {Wave} {Radar}}.
\newblock In {ArXiv}, 2021.

\bibitem[Chen \textit{et~al.}~2021b]{chen_attention-based_2021}
Shiliang Chen, Wentao He, Jianfeng Ren and Xudong Jiang.
\newblock {\em Attention-based {Dual}-stream {Vision} {Transformer} for {Radar}
  {Gait} {Recognition}}.
\newblock In {ArXiv}, 2021.

\bibitem[Cheng \& Liu~2021]{cheng_person_2021}
Yuwei Cheng and Yimin Liu.
\newblock {\em Person {Reidentification} {Based} on {Automotive} {Radar}
  {Point} {Clouds}}.
\newblock In {IEEE} {Transactions} on {Geoscience} and {Remote} {Sensing},
  2021.

\bibitem[Cheng \textit{et~al.}~2021a]{cheng_new_2021}
Yuwei Cheng, Jingran Su, Hongyu Chen and Yimin Liu.
\newblock {\em A {New} {Automotive} {Radar} {4D} {Point} {Clouds} {Detector} by
  {Using} {Deep} {Learning}}.
\newblock In {IEEE} {International} {Conference} on {Acoustics}, {Speech}, \&
  {Signal} {Processing} ({ICASSP}), 2021.

\bibitem[Cheng \textit{et~al.}~2021b]{cheng_robust_2021}
Yuwei Cheng, Hu~Xu and Yimin Liu.
\newblock {\em Robust {Small} {Object} {Detection} on the {Water} {Surface}
  {Through} {Fusion} of {Camera} and {Millimeter} {Wave} {Radar}}.
\newblock In {IEEE} {International} {Conference} on {Computer} {Vision}
  ({ICCV}), 2021.

\bibitem[Chollet~2017]{chollet_xception_2017}
François Chollet.
\newblock {\em Xception: {Deep} {Learning} with {Depthwise} {Separable}
  {Convolutions}}.
\newblock In {IEEE} {Conference} on {Computer} {Vision} and {Pattern}
  {Recognition} ({CVPR}), 2017.

\bibitem[Comaniciu \& Meer~2002]{comaniciu_mean_2002}
D.~Comaniciu and P.~Meer.
\newblock {\em Mean shift: a robust approach toward feature space analysis}.
\newblock In {IEEE} {Transactions} on {Pattern} {Analysis} and {Machine}
  {Intelligence} ({TPAMI}), 2002.

\bibitem[Cordts \textit{et~al.}~2016]{cordts_cityscapes_2016}
Marius Cordts, Mohamed Omran, Sebastian Ramos, Timo Rehfeld, Markus Enzweiler,
  Rodrigo Benenson, Uwe Franke, Stefan Roth and Bernt Schiele.
\newblock {\em The {Cityscapes} {Dataset} for {Semantic} {Urban} {Scene}
  {Understanding}}.
\newblock In {IEEE} {Conference} on {Computer} {Vision} and {Pattern}
  {Recognition} ({CVPR}), 2016.

\bibitem[Cui \textit{et~al.}~2021]{cui_3d_2021}
Hang Cui, Junzhe Wu, Jiaming Zhang, Girish Chowdhary and William~R. Norris.
\newblock {\em {3D} {Detection} and {Tracking} for {On}-road {Vehicles} with a
  {Monovision} {Camera} and {Dual} {Low}-cost {4D} {mmWave} {Radars}}.
\newblock In International {Conference} on {Intelligent} {Transportation}
  {Systems} ({ITSC}), 2021.

\bibitem[Dai \textit{et~al.}~2016]{dai_r-fcn_2016}
Jifeng Dai, Yi~Li, Kaiming He and Jian Sun.
\newblock {\em R-{FCN}: {Object} {Detection} via {Region}-{Based} {Fully}
  {Convolutional} {Networks}}.
\newblock In Conference on {Neural} {Information} {Processing} {Systems}
  ({NeurIPS}), 2016.

\bibitem[Dai \textit{et~al.}~2017]{dai_deformable_2017}
Jifeng Dai, Haozhi Qi, Yuwen Xiong, Yi~Li, Guodong Zhang, Han Hu and Yichen
  Wei.
\newblock {\em Deformable {Convolutional} {Networks}}.
\newblock In {IEEE} {International} {Conference} on {Computer} {Vision}
  ({ICCV}), 2017.

\bibitem[Dalsasso \textit{et~al.}~2021]{dalsasso_sar2sar_2021}
Emanuele Dalsasso, Loïc Denis and Florence Tupin.
\newblock {\em {SAR2SAR}: a semi-supervised despeckling algorithm for {SAR}
  images}.
\newblock In Journal of {Selected} {Topics} in {Applied} {Earth} {Observations}
  and {Remote} {Sensing}, 2021.

\bibitem[Danzer \textit{et~al.}~2019]{danzer_2d_2019}
Andreas Danzer, Thomas Griebel, Martin Bach and Klaus Dietmayer.
\newblock {\em {2D} {Car} {Detection} in {Radar} {Data} with {PointNets}}.
\newblock In International {Conference} on {Intelligent} {Transportation}
  {Systems} ({ITSC}), 2019.

\bibitem[de~Oliveira \& Bekooij~2020]{de_oliveira_deep_2020}
Marcio L.~Lima de~Oliveira and Marco J.~G. Bekooij.
\newblock {\em Deep {Convolutional} {Autoencoder} {Applied} for {Noise}
  {Reduction} in {Range}-{Doppler} {Maps} of {FMCW} {Radars}}.
\newblock In {IEEE} {International} {Radar} {Conference} ({RADAR}), 2020.

\bibitem[Dekker \textit{et~al.}~2017]{dekker_gesture_2017}
B.~Dekker, S.~Jacobs, A.~S. Kossen, M.~C. Kruithof, A.~G. Huizing and
  M.~Geurts.
\newblock {\em Gesture recognition with a low power {FMCW} radar and a deep
  convolutional neural network}.
\newblock In {IEEE} {European} {Radar} {Conference} ({EuRAD}), 2017.

\bibitem[Deledalle \textit{et~al.}~2017]{deledalle_mulog_2017}
Charles-Alban Deledalle, Loic Denis, Sonia Tabti and Florence Tupin.
\newblock {\em {MuLoG}, or {How} to {Apply} {Gaussian} {Denoisers} to
  {Multi}-{Channel} {SAR} {Speckle} {Reduction}?}
\newblock In {IEEE} {Transactions} on {Image} {Processing}, 2017.

\bibitem[Deng \textit{et~al.}~2009]{deng_imagenet_2009}
Jia Deng, Wei Dong, Richard Socher, Li-Jia Li, Kai Li and Li~Fei-Fei.
\newblock {\em Imagenet: {A} large-scale hierarchical image database}.
\newblock In {IEEE} {Conference} on {Computer} {Vision} and {Pattern}
  {Recognition} ({CVPR}), 2009.

\bibitem[Djuric \textit{et~al.}~2021]{djuric_multixnet_2021}
Nemanja Djuric, Henggang Cui, Zhaoen Su, Shangxuan Wu, Huahua Wang, Fang-Chieh
  Chou, Luisa~San Martin, Song Feng, Rui Hu, Yang Xu, Alyssa Dayan, Sidney
  Zhang, Brian~C. Becker, Gregory~P. Meyer, Carlos Vallespi-Gonzalez and
  Carl~K. Wellington.
\newblock {\em {MultiXNet}: {Multiclass} {Multistage} {Multimodal} {Motion}
  {Prediction}}.
\newblock In {IEEE} {Intelligent} {Vehicles} {Symposium} ({IV}), 2021.

\bibitem[Dong \textit{et~al.}~2020]{dong_probabilistic_2020}
Xu~Dong, Pengluo Wang, Pengyue Zhang and Langechuan Liu.
\newblock {\em Probabilistic {Oriented} {Object} {Detection} in {Automotive}
  {Radar}}.
\newblock In {IEEE} {Conference} on {Computer} {Vision} and {Pattern}
  {Recognition} {Workshop} ({CVPRW}), 2020.

\bibitem[Dong \textit{et~al.}~2021]{dong_radar_2021}
Xu~Dong, Binnan Zhuang, Yunxiang Mao and Langechuan Liu.
\newblock {\em Radar {Camera} {Fusion} via {Representation} {Learning} in
  {Autonomous} {Driving}}.
\newblock In {IEEE} {Conference} on {Computer} {Vision} and {Pattern}
  {Recognition} ({CVPR}), 2021.

\bibitem[Donnet \& Longstaff~2006]{donnet_mimo_2006}
B~Donnet and I~Longstaff.
\newblock {\em {MIMO} {Radar}, {Techniques} and {Opportunities}}.
\newblock In {IEEE} {European} {Radar} {Conference} ({EuRAD}), 2006.

\bibitem[Dosovitskiy \textit{et~al.}~2021]{dosovitskiy_image_2021}
Alexey Dosovitskiy, Lucas Beyer, Alexander Kolesnikov, Dirk Weissenborn,
  Xiaohua Zhai, Thomas Unterthiner, Mostafa Dehghani, Matthias Minderer, Georg
  Heigold, Sylvain Gelly, Jakob Uszkoreit and Neil Houlsby.
\newblock {\em An {Image} is {Worth} 16x16 {Words}: {Transformers} for {Image}
  {Recognition} at {Scale}}.
\newblock In International {Conference} on {Learning} {Representations}
  ({ICLR}), 2021.

\bibitem[Endres \& Schindelin~2003]{endres_new_2003}
D.M. Endres and J.E. Schindelin.
\newblock {\em A new metric for probability distributions}.
\newblock In {IEEE} {Transactions} on {Information} {Theory}, 2003.

\bibitem[Ester \textit{et~al.}~1996]{ester_density-based_1996}
Martin Ester, Hans-Peter Kriegel, Jörg Sander and Xiaowei Xu.
\newblock {\em A {Density}-{Based} {Algorithm} for {Discovering} {Clusters} in
  {Large} {Spatial} {Databases} with {Noise}}.
\newblock In Knowledge {Discovery} and {Data} {Mining} ({KDD}), 1996.

\bibitem[Everingham \textit{et~al.}~2015]{everingham_pascal_2015}
M.~Everingham, S.~M.~A. Eslami, L.~Van~Gool, C.~K.~I. Williams, J.~Winn and
  A.~Zisserman.
\newblock {\em The {Pascal} {Visual} {Object} {Classes} {Challenge}: {A}
  {Retrospective}}.
\newblock In International {Journal} of {Computer} {Vision} ({IJCV}), 2015.

\bibitem[Farag~2021]{farag_real-time_2021}
Wael Farag.
\newblock {\em Real-time lidar and radar fusion for road-objects detection and
  tracking}.
\newblock In International {Journal} of {Computational} {Science} and
  {Engineering} ({IJCSE}), 2021.

\bibitem[Feng \textit{et~al.}~2019]{feng_point_2019}
Zhaofei Feng, Shuo Zhang, Martin Kunert and Werner Wiesbeck.
\newblock {\em Point {Cloud} {Segmentation} with a {High}-{Resolution}
  {Automotive} {Radar}}.
\newblock In {IEEE} {Automotive} meets {Electronics} ({AmE}), 2019.

\bibitem[Fu \textit{et~al.}~2018]{fu_deep_2018}
Huan Fu, Mingming Gong, Chaohui Wang, Kayhan Batmanghelich and Dacheng Tao.
\newblock {\em Deep {Ordinal} {Regression} {Network} for {Monocular} {Depth}
  {Estimation}}.
\newblock In {IEEE} {Conference} on {Computer} {Vision} and {Pattern}
  {Recognition} ({CVPR}), 2018.

\bibitem[Gadd \textit{et~al.}~2021]{gadd_unsupervised_2021}
Matthew Gadd, Daniele De~Martini and Paul Newman.
\newblock {\em Unsupervised {Place} {Recognition} with {Deep} {Embedding}
  {Learning} over {Radar} {Videos}}.
\newblock In {IEEE} {International} {Conference} on {Robotics} and {Automation}
  {Workshop} ({ICRAW}), 2021.

\bibitem[Gao \textit{et~al.}~2019]{gao_experiments_2019}
Xiangyu Gao, Guanbin Xing, Sumit Roy and Hui Liu.
\newblock {\em Experiments with {mmWave} {Automotive} {Radar} {Test}-bed}.
\newblock In Asilomar {Conference}, 2019.

\bibitem[Gao \textit{et~al.}~2020]{gao_ramp-cnn_2020}
Xiangyu Gao, Guanbin Xing, Sumit Roy and Hui Liu.
\newblock {\em {RAMP}-{CNN}: {A} {Novel} {Neural} {Network} for {Enhanced}
  {Automotive} {Radar} {Object} {Recognition}}.
\newblock In {IEEE} {Sensors} {Journal}, 2020.

\bibitem[Garcia \textit{et~al.}~2012]{garcia_data_2012}
Fernando Garcia, Pietro Cerri, Alberto Broggi, Arturo de~la Escalera and
  Jose~Maria Armingol.
\newblock {\em Data fusion for overtaking vehicle detection based on radar and
  optical flow}.
\newblock In {IEEE} {Intelligent} {Vehicles} {Symposium} ({IV}), 2012.

\bibitem[Garnot \textit{et~al.}~2021]{garnot_multi-modal_2021}
Vivien Sainte~Fare Garnot, Loic Landrieu and Nesrine Chehata.
\newblock {\em Multi-{Modal} {Temporal} {Attention} {Models} for {Crop}
  {Mapping} from {Satellite} {Time} {Series}}.
\newblock In {ArXiv}, 2021.

\bibitem[Gasperini \textit{et~al.}~2021]{gasperini_r4dyn_2021}
Stefano Gasperini, Patrick Koch, Vinzenz Dallabetta, Nassir Navab, Benjamin
  Busam and Federico Tombari.
\newblock {\em {R4Dyn}: {Exploring} {Radar} for {Self}-{Supervised} {Monocular}
  {Depth} {Estimation} of {Dynamic} {Scenes}}.
\newblock In {IEEE} {International} {Conference} on {3D} {Vision} ({3DV}),
  2021.

\bibitem[Geiger \textit{et~al.}~2013]{geiger_vision_2013}
A~Geiger, P~Lenz, C~Stiller and R~Urtasun.
\newblock {\em Vision meets robotics: {The} {KITTI} dataset}.
\newblock In International {Journal} of {Robotics} {Research}, 2013.

\bibitem[Ghaleb~2009]{ghaleb_micro-doppler_2009}
Antoine Ghaleb.
\newblock {\em Micro-{Doppler} analysis of non-stationary moving targets in
  radar imaging}.
\newblock PhD thesis, Telecom Paris, France, 2009.

\bibitem[Girshick \textit{et~al.}~2016]{girshick_region-based_2016}
Ross Girshick, Jeff Donahue, Trevor Darrell and Jitendra Malik.
\newblock {\em Region-{Based} {Convolutional} {Networks} for {Accurate}
  {Object} {Detection} and {Segmentation}}.
\newblock In {IEEE} {Transactions} on {Pattern} {Analysis} and {Machine}
  {Intelligence} ({TPAMI}), 2016.

\bibitem[Girshick~2015]{girshick_fast_2015}
Ross Girshick.
\newblock {\em Fast {R}-{CNN}}.
\newblock In {IEEE} {International} {Conference} on {Computer} {Vision}
  ({ICCV}), September 2015.

\bibitem[Glorot \& Bengio~2010]{glorot_understanding_2010}
Xavier Glorot and Yoshua Bengio.
\newblock {\em Understanding the difficulty of training deep feedforward neural
  networks}.
\newblock In International {Conference} on {Artificial} {Intelligence} and
  {Statistics} ({AISTATS}), 2010.

\bibitem[Goodfellow \textit{et~al.}~2014]{goodfellow_generative_2014}
Ian~J. Goodfellow, Jean Pouget-Abadie, Mehdi Mirza, Bing Xu, David
  Warde-Farley, Sherjil Ozair, Aaron Courville and Yoshua Bengio.
\newblock {\em Generative {Adversarial} {Networks}}.
\newblock In Conference on {Neural} {Information} {Processing} {Systems}
  ({NeurIPS}), 2014.

\bibitem[Goodfellow \textit{et~al.}~2016]{goodfellow_deep_2016}
Ian~J. Goodfellow, Yoshua Bengio and Aaron Courville.
\newblock Deep {Learning}.
\newblock MIT Press, 2016.

\bibitem[Goodman~1976]{goodman_fundamental_1976}
J.~W. Goodman.
\newblock {\em Some fundamental properties of speckle}.
\newblock In Journal of the {Optical} {Society} of {America}, 1976.

\bibitem[Goodman~2007]{goodman_speckle_2007}
Joseph~W Goodman.
\newblock Speckle phenomena in optics: theory and applications.
\newblock Roberts and Company Publishers, 2007.

\bibitem[Graham \textit{et~al.}~2018]{graham_3d_2018}
Benjamin Graham, Martin Engelcke and Laurens van~der Maaten.
\newblock {\em {3D} {Semantic} {Segmentation} with {Submanifold} {Sparse}
  {Convolutional} {Networks}}.
\newblock In {IEEE} {Conference} on {Computer} {Vision} and {Pattern}
  {Recognition} ({CVPR}), 2018.

\bibitem[Graves \textit{et~al.}~2006]{graves_connectionist_2006}
Alex Graves, Santiago Fernández, Faustino Gomez and Jürgen Schmidhuber.
\newblock {\em Connectionist temporal classification: labelling unsegmented
  sequence data with recurrent neural networks}.
\newblock In International {Conference} on {Machine} {Learning} ({ICML}), 2006.

\bibitem[Griebel \textit{et~al.}~2021]{griebel_anomaly_2021}
Thomas Griebel, Dominik Authaler, Markus Horn, Matti Henning, Michael Buchholz
  and Klaus Dietmayer.
\newblock {\em Anomaly {Detection} in {Radar} {Data} {Using} {PointNets}}.
\newblock In {IEEE} {Intelligent} {Transportation} {Systems} {Conference}
  ({ITSC}), 2021.

\bibitem[Grimm \textit{et~al.}~2020]{grimm_warping_2020}
Christopher Grimm, Tai Fei, Ernst Warsitz, Ridha Farhoud, Tobias Breddermann
  and Reinhold Haeb-Umbach.
\newblock {\em Warping of {Radar} {Data} into {Camera} {Image} for
  {Cross}-{Modal} {Supervision} in {Automotive} {Applications}}.
\newblock In {ArXiv}, 2020.

\bibitem[Guan \textit{et~al.}~2020]{guan_through_2020}
Junfeng Guan, Sohrab Madani, Suraj Jog, Saurabh Gupta and Haitham Hassanieh.
\newblock {\em Through {Fog} {High}-{Resolution} {Imaging} {Using} {Millimeter}
  {Wave} {Radar}}.
\newblock In {IEEE} {Conference} on {Computer} {Vision} and {Pattern}
  {Recognition} ({CVPR}), 2020.

\bibitem[Guo \textit{et~al.}~2020]{guo_deep_2020}
Yulan Guo, Hanyun Wang, Qingyong Hu, Hao Liu, Li~Liu and Mohammed Bennamoun.
\newblock {\em Deep {Learning} for {3D} {Point} {Clouds}: {A} {Survey}}.
\newblock In {IEEE} {Transactions} on {Pattern} {Analysis} and {Machine}
  {Intelligence} ({TPAMI}), 2020.

\bibitem[Hastie \textit{et~al.}~2001]{hastie_elements_2001}
Trevor Hastie, Robert Tibshirani and Jerome Friedman.
\newblock The {Elements} of {Statistical} {Learning}.
\newblock Springer Series in Statistics, 2001.

\bibitem[Hazra \& Santra~2019]{hazra_radar_2019}
Souvik Hazra and Avik Santra.
\newblock {\em Radar {Gesture} {Recognition} {System} in {Presence} of
  {Interference} using {Self}-{Attention} {Neural} {Network}}.
\newblock In {IEEE} {International} {Conference} on {Machine} {Learning} and
  {Applications} ({ICMLA}), 2019.

\bibitem[He \textit{et~al.}~2016]{he_deep_2016}
Kaiming He, Xiangyu Zhang, Shaoqing Ren and Jian Sun.
\newblock {\em Deep residual learning for image recognition}.
\newblock In {IEEE} {Conference} on {Computer} {Vision} and {Pattern}
  {Recognition} ({CVPR}), 2016.

\bibitem[He \textit{et~al.}~2017]{he_mask_2017}
Kaiming He, Georgia Gkioxari, Piotr Dollar and Ross Girshick.
\newblock {\em Mask {R}-{CNN}}.
\newblock In {IEEE} {International} {Conference} on {Computer} {Vision}
  ({ICCV}), 2017.

\bibitem[Hearst \textit{et~al.}~1998]{hearst_support_1998}
M.A. Hearst, S.T. Dumais, E.~Osuna, J.~Platt and B.~Scholkopf.
\newblock {\em Support vector machines}.
\newblock In {IEEE} {Intelligent} {Systems} and their {Applications}, 1998.

\bibitem[Hinton \textit{et~al.}~2012]{hinton_improving_2012}
Geoffrey~E. Hinton, Nitish Srivastava, Alex Krizhevsky, Ilya Sutskever and
  Ruslan~R. Salakhutdinov.
\newblock {\em Improving neural networks by preventing co-adaptation of feature
  detectors}.
\newblock In {ArXiv}, 2012.

\bibitem[Hirschmuller~2008]{hirschmuller_stereo_2008}
H.~Hirschmuller.
\newblock {\em Stereo {Processing} by {Semiglobal} {Matching} and {Mutual}
  {Information}}.
\newblock In {IEEE} {Transactions} on {Pattern} {Analysis} and {Machine}
  {Intelligence} ({TPAMI}), 2008.

\bibitem[Hochreiter \& Schmidhuber~1997]{hochreiter_long_1997}
Sepp Hochreiter and Jürgen Schmidhuber.
\newblock {\em Long short-term memory}.
\newblock In Neural {Computation}, 1997.

\bibitem[Hoermann \textit{et~al.}~2018]{hoermann_dynamic_2018}
Stefan Hoermann, Martin Bach and Klaus Dietmayer.
\newblock {\em Dynamic {Occupancy} {Grid} {Prediction} for {Urban} {Autonomous}
  {Driving}: {A} {Deep} {Learning} {Approach} with {Fully} {Automatic}
  {Labeling}}.
\newblock In {IEEE} {International} {Conference} on {Robotics} and {Automation}
  ({ICRA}), 2018.

\bibitem[Hsu \textit{et~al.}~2021]{hsu_efficient-rod_2021}
Chih-Chung Hsu, Chieh Lee, Lin Chen, Min-Kai Hung, Yu-Lun Lin and Xian-Yu Wang.
\newblock {\em Efficient-{ROD}: {Efficient} {Radar} {Object} {Detection} based
  on {Densely} {Connected} {Residual} {Network}}.
\newblock In {ACM} {International} {Conference} on {Multimedia} {Retrieval}
  ({ICMR}), 2021.

\bibitem[Hu \textit{et~al.}~2018]{hu_squeeze-and-excitation_2018}
Jie Hu, Li~Shen and Gang Sun.
\newblock {\em Squeeze-and-{Excitation} {Networks}}.
\newblock In {IEEE} {Conference} on {Computer} {Vision} and {Pattern}
  {Recognition} ({CVPR}), 2018.

\bibitem[Huang \textit{et~al.}~2019]{huang_apolloscape_2019}
Xinyu Huang, Peng Wang, Xinjing Cheng, Dingfu Zhou, Qichuan Geng and Ruigang
  Yang.
\newblock {\em The {ApolloScape} {Open} {Dataset} for {Autonomous} {Driving}
  and its {Application}}.
\newblock In {IEEE} {Transactions} on {Pattern} {Analysis} and {Machine}
  {Intelligence} ({TPAMI}), 2019.

\bibitem[Hussain \textit{et~al.}~2021]{hussain_rvmde_2021}
Muhamamd~Ishfaq Hussain, Muhammad~Aasim Rafique and Moongu Jeon.
\newblock {\em {RVMDE}: {Radar} {Validated} {Monocular} {Depth} {Estimation}
  for {Robotics}}.
\newblock In {ArXiv}, 2021.

\bibitem[Ioffe \& Szegedy~2015]{ioffe_batch_2015}
Sergey Ioffe and Christian Szegedy.
\newblock {\em Batch {Normalization}: {Accelerating} {Deep} {Network}
  {Training} by {Reducing} {Internal} {Covariate} {Shift}}.
\newblock In International {Conference} on {Machine} {Learning} ({ICML}), 2015.

\bibitem[Iovescu \& Rao~2017]{iovescu_fundamentals_2017}
Cesar Iovescu and Sandeep Rao.
\newblock {\em The fundamentals of millimeter wave radar sensors}.
\newblock Technical report, Texas Instruments, 2017.

\bibitem[Ishak \textit{et~al.}~2018]{ishak_human_2018}
K.~Ishak, N.~Appenrodt, J.~Dickmann and C.~Waldschmidt.
\newblock {\em Human {Motion} {Training} {Data} {Generation} for {Radar}
  {Based} {Deep} {Learning} {Applications}}.
\newblock In {IEEE} {International} {Conference} on {Microwaves} for
  {Intelligent} {Mobility}, 2018.

\bibitem[Isola \textit{et~al.}~2017]{isola_image--image_2017}
Phillip Isola, Jun-Yan Zhu, Tinghui Zhou and Alexei~A. Efros.
\newblock {\em Image-to-{Image} {Translation} with {Conditional} {Adversarial}
  {Networks}}.
\newblock In {IEEE} {Conference} on {Computer} {Vision} and {Pattern}
  {Recognition} ({CVPR}), 2017.

\bibitem[Janai \textit{et~al.}~2020]{janai_computer_2020}
Joel Janai, Fatma Güney, Aseem Behl and Andreas Geiger.
\newblock {\em Computer {Vision} for {Autonomous} {Vehicles}: {Problems},
  {Datasets} and {State} of the {Art}}.
\newblock In {FNT} in {Computer} {Graphics} and {Vision}, 2020.

\bibitem[Ji \textit{et~al.}~2012]{ji_3d_2012}
Shuiwang Ji, Wei Xu, Ming Yang and Kai Yu.
\newblock {\em {3D} convolutional neural networks for human action
  recognition}.
\newblock In {IEEE} {Transactions} on {Pattern} {Analysis} and {Machine}
  {Intelligence} ({TPAMI}), 2012.

\bibitem[Jin \textit{et~al.}~2021]{jin_traffic_2021}
Shaojie Jin, Ying Gao, Shoucai Jing, Fei Hui, Xiangmo Zhao and Jianzhen Liu.
\newblock {\em Traffic {Flow} {Parameters} {Collection} under {Variable}
  {Illumination} {Based} on {Data} {Fusion}}.
\newblock In Journal of {Advanced} {Transportation}, 2021.

\bibitem[Ju \textit{et~al.}~2021]{ju_danet_2021}
Bo~Ju, Wei Yang, Jinrang Jia, Xiaoqing Ye, Qu~Chen, Xiao Tan, Hao Sun, Yifeng
  Shi and Errui Ding.
\newblock {\em {DANet}: {Dimension} {Apart} {Network} for {Radar} {Object}
  {Detection}}.
\newblock In {ACM} {International} {Conference} on {Multimedia} {Retrieval}
  ({ICMR}), 2021.

\bibitem[Karlsson \textit{et~al.}~2021]{karlsson_probabilistic_2021}
Robin Karlsson, David~Robert Wong, Kazunari Kawabata, Simon Thompson and Naoki
  Sakai.
\newblock {\em Probabilistic {Rainfall} {Estimation} from {Automotive}
  {Lidar}}.
\newblock In {ArXiv}, 2021.

\bibitem[Karpathy~2021]{karpathy_cs231n_2021}
Andrej Karpathy.
\newblock {\em {CS231n} {Convolutional} {Neural} {Networks} for {Visual}
  {Recognition}.}, 2021.

\bibitem[Kaul \textit{et~al.}~2020]{kaul_rss-net_2020}
Prannay Kaul, Daniele De~Martini, Matthew Gadd and Paul Newman.
\newblock {\em {RSS}-{Net}: {Weakly}-{Supervised} {Multi}-{Class} {Semantic}
  {Segmentation} with {FMCW} {Radar}}.
\newblock In {IEEE} {Intelligent} {Vehicles} {Symposium} ({IV}), 2020.

\bibitem[Kiefer \& Wolfowitz~1952]{kiefer_stochastic_1952}
J.~Kiefer and J.~Wolfowitz.
\newblock {\em Stochastic {Estimation} of the {Maximum} of a {Regression}
  {Function}}.
\newblock In The {Annals} of {Mathematical} {Statistics}, 1952.

\bibitem[Kim \& Moon~2016]{kim_human_2016}
Youngwook Kim and Taesup Moon.
\newblock {\em Human {Detection} and {Activity} {Classification} {Based} on
  {Micro}-{Doppler} {Signatures} {Using} {Deep} {Convolutional} {Neural}
  {Networks}}.
\newblock In Geoscience and {Remote} {Sensing} {Letters}, 2016.

\bibitem[Kim \& Toomajian~2016]{kim_hand_2016}
Y.~Kim and B.~Toomajian.
\newblock {\em Hand {Gesture} {Recognition} {Using} {Micro}-{Doppler}
  {Signatures} {With} {Convolutional} {Neural} {Network}}.
\newblock In {IEEE} {Access}, 2016.

\bibitem[Kim \textit{et~al.}~2020a]{kim_mulran_2020}
Giseop Kim, Yeong~Sang Park, Younghun Cho, Jinyong Jeong and Ayoung Kim.
\newblock {\em {MulRan}: {Multimodal} {Range} {Dataset} for {Urban} {Place}
  {Recognition}}.
\newblock In {IEEE} {International} {Conference} on {Robotics} and {Automation}
  ({ICRA}), 2020.

\bibitem[Kim \textit{et~al.}~2020b]{kim_low-level_2020}
Jinhyeong Kim, Youngseok Kim and Dongsuk Kum.
\newblock {\em Low-level {Sensor} {Fusion} {Network} for {3D} {Vehicle}
  {Detection} using {Radar} {Range}-{Azimuth} {Heatmap} and {Monocular}
  {Image}}.
\newblock In Asian {Conference} on {Computer} {Vision} ({ACCV}), 2020.

\bibitem[Kim \textit{et~al.}~2020c]{kim_grif_2020}
Youngseok Kim, Jun~Won Choi and Dongsuk Kum.
\newblock {\em {GRIF} {Net}: {Gated} {Region} of {Interest} {Fusion} {Network}
  for {Robust} {3D} {Object} {Detection} from {Radar} {Point} {Cloud} and
  {Monocular} {Image}}.
\newblock In {IEEE} {International} {Conference} on {Intelligent} {Robots} and
  {Systems} ({IROS}), 2020.

\bibitem[Kingma~2015]{kingma_adam_2015}
Diederik~P. Kingma.
\newblock {\em Adam: a method for stochastic optimization}.
\newblock In International {Conference} on {Learning} {Representations}
  ({ICLR}), 2015.

\bibitem[Klarenbeek \textit{et~al.}~2017]{klarenbeek_multi-target_2017}
G.~Klarenbeek, R.~I.~A. Harmanny and L.~Cifola.
\newblock {\em Multi-target human gait classification using {LSTM} recurrent
  neural networks applied to micro-{Doppler}}.
\newblock In {IEEE} {European} {Radar} {Conference} ({EuRAD}), 2017.

\bibitem[Kopp \textit{et~al.}~2021]{kopp_fast_2021}
Johannes Kopp, Dominik Kellner, Aldi Piroli and Klaus Dietmayer.
\newblock {\em Fast {Rule}-{Based} {Clutter} {Detection} in {Automotive}
  {Radar} {Data}}.
\newblock In International {Conference} on {Intelligent} {Transportation}
  {Systems} ({ITSC}), 2021.

\bibitem[Kowol \textit{et~al.}~2021]{kowol_yodar_2021}
Kamil Kowol, Matthias Rottmann, Stefan Bracke and Hanno Gottschalk.
\newblock {\em {YOdar}: {Uncertainty}-based {Sensor} {Fusion} for {Vehicle}
  {Detection} with {Camera} and {Radar} {Sensors}}.
\newblock In International {Conference} on {Agents} and {Artificial}
  {Intelligence} ({ICAART}), 2021.

\bibitem[Krizhevsky \textit{et~al.}~2012]{krizhevsky_imagenet_2012}
Alex Krizhevsky, Ilya Sutskever and Geoffrey~E Hinton.
\newblock {\em {ImageNet} {Classification} with {Deep} {Convolutional} {Neural}
  {Networks}}.
\newblock In Conference on {Neural} {Information} {Processing} {Systems}
  ({NeurIPS}), 2012.

\bibitem[Krähenbühl \& Koltun~2011]{krahenbuhl_efficient_2011}
Philipp Krähenbühl and Vladlen Koltun.
\newblock {\em Efficient {Inference} in {Fully} {Connected} {CRFs} with
  {Gaussian} {Edge} {Potentials}}.
\newblock In Conference on {Neural} {Information} {Processing} {Systems}
  ({NeurIPS}), 2011.

\bibitem[Ku \textit{et~al.}~2018]{ku_joint_2018}
Jason Ku, Melissa Mozifian, Jungwook Lee, Ali Harakeh and Steven Waslander.
\newblock {\em Joint {3D} {Proposal} {Generation} and {Object} {Detection} from
  {View} {Aggregation}}.
\newblock In {IEEE} {International} {Conference} on {Intelligent} {Robots} and
  {Systems} ({IROS}), 2018.

\bibitem[Kuang \textit{et~al.}~2020]{kuang_multi-modality_2020}
Hongwu Kuang, Xiaodong Liu, Jingwei Zhang and Zicheng Fang.
\newblock {\em Multi-{Modality} {Cascaded} {Fusion} {Technology} for
  {Autonomous} {Driving}}.
\newblock In {IEEE} {International} {Conference} on {Robotics} and {Automation}
  {Sciences} ({ICRAS}), 2020.

\bibitem[Kullback \& Leibler~1951]{kullback_information_1951}
S.~Kullback and R.~A. Leibler.
\newblock {\em On {Information} and {Sufficiency}}.
\newblock In The {Annals} of {Mathematical} {Statistics}, 1951.

\bibitem[Kung \textit{et~al.}~2021]{kung_radar_2021}
Pou-Chun Kung, Chieh-Chih Wang and Wen-Chieh Lin.
\newblock {\em Radar {Occupancy} {Prediction} with {Lidar} {Supervision} while
  {Preserving} {Long}-{Range} {Sensing} and {Penetrating} {Capabilities}}.
\newblock In {ArXiv}, 2021.

\bibitem[Kurup \& Bos~2021]{kurup_dsor_2021}
Akhil Kurup and Jeremy Bos.
\newblock {\em {DSOR}: {A} {Scalable} {Statistical} {Filter} for {Removing}
  {Falling} {Snow} from {LiDAR} {Point} {Clouds} in {Severe} {Winter}
  {Weather}}.
\newblock In {ArXiv}, 2021.

\bibitem[LeCun \textit{et~al.}~1989]{lecun_backpropagation_1989}
Y.~LeCun, B.~Boser, J.~S. Denker, D.~Henderson, R.~E. Howard, W.~Hubbard and
  L.~D. Jackel.
\newblock {\em Backpropagation {Applied} to {Handwritten} {Zip} {Code}
  {Recognition}}.
\newblock In Neural {Computation}, 1989.

\bibitem[LeCun \textit{et~al.}~2012]{lecun_efficient_2012}
Yann LeCun, Léon Bottou, Genevieve~B. Orr and Klaus-Robert Müller.
\newblock {\em Efficient {BackProp}.}
\newblock In Neural {Networks}: {Tricks} of the {Trade}. Springer, 2012.

\bibitem[{Lee, Wei-Yu} \textit{et~al.}~2021]{lee_wei-yu_spatio-temporal_2021}
{Lee, Wei-Yu}, Martin Dimitrievski, Ljubomir Jovanov and Wilfried Philips.
\newblock {\em Spatio-temporal consistency for semi-supervised learning using
  {3D} radar cubes}.
\newblock In {IEEE} {Intelligent} {Vehicles} {Symposium} ({IV}), 2021.

\bibitem[Lei \textit{et~al.}~2020]{lei_continuous_2020}
Wentai Lei, Xinyue Jiang, Long Xu, Jiabin Luo, Mengdi Xu and Feifei Hou.
\newblock {\em Continuous {Gesture} {Recognition} {Based} on {Time} {Sequence}
  {Fusion} {Using} {MIMO} {Radar} {Sensor} and {Deep} {Learning}}.
\newblock In Electronics, 2020.

\bibitem[Lekic \& Babic~2019]{lekic_automotive_2019}
Vladimir Lekic and Zdenka Babic.
\newblock {\em Automotive radar and camera fusion using {Generative}
  {Adversarial} {Networks}}.
\newblock In Computer {Vision} and {Image} {Understanding} ({CVIU}), 2019.

\bibitem[Li \& Xie~2020]{li_feature_2020}
Liang-qun Li and Yuan-liang Xie.
\newblock {\em A {Feature} {Pyramid} {Fusion} {Detection} {Algorithm} {Based}
  on {Radar} and {Camera} {Sensor}}.
\newblock In {IEEE} {International} {Conference} on {Intelligent} {Computing}
  and {Signal} {Processing} ({ICSP}), 2020.

\bibitem[Liang \textit{et~al.}~2020]{liang_pnpnet_2020}
Ming Liang, Bin Yang, Wenyuan Zeng, Yun Chen, Rui Hu, Sergio Casas and Raquel
  Urtasun.
\newblock {\em {PnPNet}: {End}-to-{End} {Perception} and {Prediction} with
  {Tracking} in the {Loop}}.
\newblock In {IEEE} {Conference} on {Computer} {Vision} and {Pattern}
  {Recognition} ({CVPR}), 2020.

\bibitem[Lin \textit{et~al.}~2014a]{lin_network_2014}
Min Lin, Qiang Chen and Shuicheng Yan.
\newblock {\em Network {In} {Network}}.
\newblock In International {Conference} on {Learning} {Representations}
  ({ICLR}), 2014.

\bibitem[Lin \textit{et~al.}~2014b]{lin_microsoft_2014}
Tsung-Yi Lin, Michael Maire, Serge Belongie, James Hays, Pietro Perona, Deva
  Ramanan, Piotr Dollár and C.~Lawrence Zitnick.
\newblock {\em Microsoft {COCO}: {Common} {Objects} in {Context}}.
\newblock In European {Conference} on {Computer} {Vision} ({ECCV}), 2014.

\bibitem[Lin \textit{et~al.}~2017a]{lin_feature_2017}
Tsung-Yi Lin, Piotr Dollar, Ross Girshick, Kaiming He, Bharath Hariharan and
  Serge Belongie.
\newblock {\em Feature {Pyramid} {Networks} for {Object} {Detection}}.
\newblock In {IEEE} {Conference} on {Computer} {Vision} and {Pattern}
  {Recognition} ({CVPR}), 2017.

\bibitem[Lin \textit{et~al.}~2017b]{lin_focal_2017}
Tsung-Yi Lin, Priya Goyal, Ross Girshick, Kaiming He and Piotr Dollár.
\newblock {\em Focal {Loss} for {Dense} {Object} {Detection}}.
\newblock In {IEEE} {International} {Conference} on {Computer} {Vision}
  ({ICCV}), 2017.

\bibitem[Lin \textit{et~al.}~2020]{lin_depth_2020}
Juan-Ting Lin, Dengxin Dai and Luc Van~Gool.
\newblock {\em Depth {Estimation} from {Monocular} {Images} and {Sparse}
  {Radar} {Data}}.
\newblock In {IEEE} {International} {Conference} on {Intelligent} {Robots} and
  {Systems} ({IROS}), 2020.

\bibitem[Liu \textit{et~al.}~2015]{liu_parsenet_2015}
Wei Liu, Andrew Rabinovich and Alexander~C. Berg.
\newblock {\em {ParseNet}: {Looking} {Wider} to {See} {Better}}.
\newblock In {ArXiv}, 2015.

\bibitem[Liu \textit{et~al.}~2016]{liu_ssd_2016}
Wei Liu, Dragomir Anguelov, Dumitru Erhan, Christian Szegedy, Scott~E. Reed,
  Cheng-Yang Fu and Alexander~C. Berg.
\newblock {\em {SSD}: {Single} {Shot} {MultiBox} {Detector}}.
\newblock In European {Conference} on {Computer} {Vision} ({ECCV}), 2016.

\bibitem[Liu \textit{et~al.}~2018]{liu_path_2018}
Shu Liu, Lu~Qi, Haifang Qin, Jianping Shi and Jiaya Jia.
\newblock {\em Path {Aggregation} {Network} for {Instance} {Segmentation}}.
\newblock In {IEEE} {Conference} on {Computer} {Vision} and {Pattern}
  {Recognition} ({CVPR}), 2018.

\bibitem[Liu \textit{et~al.}~2021a]{liu_deep_2021}
Jianan Liu, Weiyi Xiong, Liping Bai, Yuxuan Xia and Bing Zhu.
\newblock {\em Deep {Instance} {Segmentation} with {High}-{Resolution}
  {Automotive} {Radar}}.
\newblock In {ArXiv}, 2021.

\bibitem[Liu \textit{et~al.}~2021b]{liu_surrounding_2021}
Ze~Liu, Yingfeng Cai, Hai Wang and Long Chen.
\newblock {\em Surrounding {Objects} {Detection} and {Tracking} for
  {Autonomous} {Driving} {Using} {LiDAR} and {Radar} {Fusion}}.
\newblock In Chinese {Journal} of {Mechanical} {Engineering} ({CJME}), 2021.

\bibitem[Liu \textit{et~al.}~2021c]{liu_robust_2021}
Ze~Liu, Yingfeng Cai, Hai Wang, Long Chen, Hongbo Gao, Yunyi Jia and Yicheng
  Li.
\newblock {\em Robust {Target} {Recognition} and {Tracking} of {Self}-{Driving}
  {Cars} {With} {Radar} and {Camera} {Information} {Fusion} {Under} {Severe}
  {Weather} {Conditions}}.
\newblock In {IEEE} {Transactions} on {Intelligent} {Transportation} {Systems},
  2021.

\bibitem[Liu \textit{et~al.}~2021d]{liu_swin_2021}
Ze~Liu, Yutong Lin, Yue Cao, Han Hu, Yixuan Wei, Zheng Zhang, Stephen Lin and
  Baining Guo.
\newblock {\em Swin {Transformer}: {Hierarchical} {Vision} {Transformer} using
  {Shifted} {Windows}}.
\newblock In {IEEE} {International} {Conference} on {Computer} {Vision}
  ({ICCV}), 2021.

\bibitem[Lo \& Vandewalle~2021]{lo_depth_2021}
Chen-Chou Lo and Patrick Vandewalle.
\newblock {\em Depth {Estimation} from {Monocular} {Images} and {Sparse} radar
  using {Deep} {Ordinal} {Regression} {Network}}.
\newblock In {IEEE} {International} {Conference} on {Image} {Processing}
  ({ICIP}), 2021.

\bibitem[Lombacher \textit{et~al.}~2017]{lombacher_semantic_2017}
Jakob Lombacher, Kilian Laudt, Markus Hahn, Jurgen Dickmann and Christian
  Wohler.
\newblock {\em Semantic radar grids}.
\newblock In {IEEE} {Intelligent} {Vehicles} {Symposium} ({IV}), 2017.

\bibitem[Long \textit{et~al.}~2015]{long_fully_2015}
Jonathan Long, Evan Shelhamer and Trevor Darrell.
\newblock {\em Fully convolutional networks for semantic segmentation}.
\newblock In {IEEE} {Conference} on {Computer} {Vision} and {Pattern}
  {Recognition} ({CVPR}), 2015.

\bibitem[Long \textit{et~al.}~2021a]{long_full-velocity_2021}
Yunfei Long, Daniel Morris, Xiaoming Liu, Marcos Castro, Punarjay Chakravarty
  and Praveen Narayanan.
\newblock {\em Full-{Velocity} {Radar} {Returns} by {Radar}-{Camera} {Fusion}}.
\newblock In {IEEE} {International} {Conference} on {Computer} {Vision}
  ({ICCV}), 2021.

\bibitem[Long \textit{et~al.}~2021b]{long_radar-camera_2021}
Yunfei Long, Daniel Morris, Xiaoming Liu, Marcos Castro, Punarjay Chakravarty
  and Praveen Narayanan.
\newblock {\em Radar-{Camera} {Pixel} {Depth} {Association} for {Depth}
  {Completion}}.
\newblock In {IEEE} {Conference} on {Computer} {Vision} and {Pattern}
  {Recognition} ({CVPR}), 2021.

\bibitem[Lowe~2004]{lowe_distinctive_2004}
David~G. Lowe.
\newblock {\em Distinctive {Image} {Features} from {Scale}-{Invariant}
  {Keypoints}}.
\newblock In International {Journal} of {Computer} {Vision} ({IJCV}), 2004.

\bibitem[Major \textit{et~al.}~2019]{major_vehicle_2019}
Bence Major, Daniel Fontijne, Amin Ansari, Ravi~Teja Sukhavasi, Radhika
  Gowaikar, Michael Hamilton, Sean Lee, Slawomir Grzechnik and Sundar
  Subramanian.
\newblock {\em Vehicle {Detection} {With} {Automotive} {Radar} {Using} {Deep}
  {Learning} on {Range}-{Azimuth}-{Doppler} {Tensors}}.
\newblock In {IEEE} {International} {Conference} on {Computer} {Vision}
  {Workshop} ({ICCVW}), 2019.

\bibitem[Masci \textit{et~al.}~2011]{masci_stacked_2011}
Jonathan Masci, Ueli Meier, Dan Cireşan and Jürgen Schmidhuber.
\newblock {\em Stacked {Convolutional} {Auto}-{Encoders} for {Hierarchical}
  {Feature} {Extraction}}.
\newblock In International {Conference} on {Artificial} {Neural} {Networks}
  ({ICANN}), 2011.

\bibitem[McCulloch \& Pitts~1943]{mcculloch_logical_1943}
Warren~S. McCulloch and Walter Pitts.
\newblock {\em A logical calculus of the ideas immanent in nervous activity}.
\newblock In Bulletin of {Mathematical} {Biophysics}, 1943.

\bibitem[Meyer \& Kuschk~2019a]{meyer_deep_2019}
M.~Meyer and G.~Kuschk.
\newblock {\em Deep {Learning} {Based} {3D} {Object} {Detection} for
  {Automotive} {Radar} and {Camera}}.
\newblock In {IEEE} {European} {Radar} {Conference} ({EuRAD}), 2019.

\bibitem[Meyer \& Kuschk~2019b]{meyer_automotive_2019}
Michael Meyer and Georg Kuschk.
\newblock {\em Automotive {Radar} {Dataset} for {Deep} {Learning} {Based} {3D}
  {Object} {Detection}}.
\newblock In {IEEE} {European} {Radar} {Conference} ({EuRAD}), 2019.

\bibitem[Meyer \textit{et~al.}~2021]{meyer_graph_2021}
Michael Meyer, Georg Kuschk and Sven Tomforde.
\newblock {\em Graph {Convolutional} {Networks} for {3D} {Object} {Detection}
  on {Radar} {Data}}.
\newblock In {IEEE} {International} {Conference} on {Computer} {Vision}
  {Workshop} ({ICCVW}), 2021.

\bibitem[Milletari \textit{et~al.}~2016]{milletari_v-net_2016}
Fausto Milletari, Nassir Navab and Seyed-Ahmad Ahmadi.
\newblock {\em V-{Net}: {Fully} {Convolutional} {Neural} {Networks} for
  {Volumetric} {Medical} {Image} {Segmentation}}.
\newblock In {IEEE} {International} {Conference} on {3D} {Vision} ({3DV}),
  2016.

\bibitem[Minsky \& Papert~1969]{minsky_perceptrons_1969}
Marvin Minsky and Seymour Papert.
\newblock Perceptrons: {An} {Introduction} to {Computational} {Geometry}.
\newblock MIT Press, 1969.

\bibitem[Mohta \textit{et~al.}~2020]{mohta_investigating_2020}
Abhishek Mohta, Fang-Chieh Chou, Brian~C. Becker, Carlos Vallespi-Gonzalez and
  Nemanja Djuric.
\newblock {\em Investigating the {Effect} of {Sensor} {Modalities} in
  {Multi}-{Sensor} {Detection}-{Prediction} {Models}}.
\newblock In Conference on {Neural} {Information} {Processing} {Systems}
  ({NeurIPS}), 2020.

\bibitem[Molchanov \textit{et~al.}~2015]{molchanov_multi-sensor_2015}
Pavlo Molchanov, Shalini Gupta, Kihwan Kim and Kari Pulli.
\newblock {\em Multi-sensor system for driver's hand-gesture recognition}.
\newblock In {IEEE} {International} {Conference} and {Workshops} on {Automatic}
  {Face} and {Gesture} {Recognition} ({FG}), 2015.

\bibitem[Molchanov~2014]{molchanov_radar_2014}
P.~Molchanov.
\newblock {\em Radar {Target} {Classification} by {Micro}-{Doppler}
  {Contributions}}.
\newblock PhD thesis, Tampere University, Finland, 2014.

\bibitem[Mostajabi \textit{et~al.}~2020]{mostajabi_high-resolution_2020}
Mohammadreza Mostajabi, Ching~Ming Wang, Darsh Ranjan and Gilbert Hsyu.
\newblock {\em High-{Resolution} {Radar} {Dataset} for {Semi}-{Supervised}
  {Learning} of {Dynamic} {Objects}}.
\newblock In {IEEE} {Conference} on {Computer} {Vision} and {Pattern}
  {Recognition} {Workshop} ({CVPRW}), 2020.

\bibitem[Mottaghi \textit{et~al.}~2014]{mottaghi_role_2014}
Roozbeh Mottaghi, Xianjie Chen, Xiaobai Liu, Nam-Gyu Cho, Seong-Whan Lee, Sanja
  Fidler, Raquel Urtasun and Alan Yuille.
\newblock {\em The {Role} of {Context} for {Object} {Detection} and {Semantic}
  {Segmentation} in the {Wild}}.
\newblock In {IEEE} {Conference} on {Computer} {Vision} and {Pattern}
  {Recognition} ({CVPR}), 2014.

\bibitem[Nabati \& Qi~2020]{nabati_radar-camera_2020}
Ramin Nabati and Hairong Qi.
\newblock {\em Radar-{Camera} {Sensor} {Fusion} for {Joint} {Object}
  {Detection} and {Distance} {Estimation} in {Autonomous} {Vehicles}}.
\newblock In {IEEE} {International} {Conference} on {Intelligent} {Robots} and
  {Systems} ({IROS}), 2020.

\bibitem[Nabati \& Qi~2021]{nabati_centerfusion_2021}
Ramin Nabati and Hairong Qi.
\newblock {\em {CenterFusion}: {Center}-based {Radar} and {Camera} {Fusion} for
  {3D} {Object} {Detection}}.
\newblock In {IEEE} {Workshop} on {Applications} of {Computer} {Vision}
  ({WACV}), 2021.

\bibitem[Ng \textit{et~al.}~2020]{ng_range-doppler_2020}
Weichong Ng, Guohua Wang, {Siddhartha}, Zhiping Lin and Bhaskar~Jyoti Dutta.
\newblock {\em Range-{Doppler} {Detection} in {Automotive} {Radar} with {Deep}
  {Learning}}.
\newblock In {IEEE} {International} {Joint} {Conference} on {Neural} {Networks}
  ({IJCNN}), 2020.

\bibitem[Niesen \& Unnikrishnan~2020]{niesen_camera-radar_2020}
Urs Niesen and Jayakrishnan Unnikrishnan.
\newblock {\em Camera-{Radar} {Fusion} for 3-{D} {Depth} {Reconstruction}}.
\newblock In {IEEE} {Intelligent} {Vehicles} {Symposium} ({IV}), 2020.

\bibitem[Nobis \textit{et~al.}~2019]{nobis_deep_2019}
Felix Nobis, Maximilian Geisslinger, Markus Weber, Johannes Betz and Markus
  Lienkamp.
\newblock {\em A {Deep} {Learning}-based {Radar} and {Camera} {Sensor} {Fusion}
  {Architecture} for {Object} {Detection}}.
\newblock In {IEEE} {Sensor} {Data} {Fusion}: {Trends}, {Solutions},
  {Applications} ({SDF}), 2019.

\bibitem[Nobis \textit{et~al.}~2021a]{nobis_kernel_2021}
Felix Nobis, Felix Fent, Johannes Betz and Markus Lienkamp.
\newblock {\em Kernel {Point} {Convolution} {LSTM} {Networks} for {Radar}
  {Point} {Cloud} {Segmentation}}.
\newblock In Applied {Sciences}, 2021.

\bibitem[Nobis \textit{et~al.}~2021b]{nobis_radar_2021}
Felix Nobis, Ehsan Shafiei, Phillip Karle, Johannes Betz and Markus Lienkamp.
\newblock {\em Radar {Voxel} {Fusion} for {3D} {Object} {Detection}}.
\newblock In Applied {Sciences}, 2021.

\bibitem[Noh \textit{et~al.}~2015]{noh_learning_2015}
Hyeonwoo Noh, Seunghoon Hong and Bohyung Han.
\newblock {\em Learning {Deconvolution} {Network} for {Semantic}
  {Segmentation}}.
\newblock In {IEEE} {International} {Conference} on {Computer} {Vision}
  ({ICCV}), 2015.

\bibitem[Nowruzi \textit{et~al.}~2020]{nowruzi_deep_2020}
F.~E. Nowruzi, D.~Kolhatkar, Prince Kapoor, E.~J. Heravi, R.~Laganiere, Julien
  Rebut and Waqas Malik.
\newblock {\em Deep open space segmentation using automotive radar}.
\newblock In {IEEE} {International} {Conference} on {Microwaves} for
  {Intelligent} {Mobility} ({ICMIM}), 2020.

\bibitem[Ouaknine \textit{et~al.}~2020]{ouaknine_carrada_2020}
A.~Ouaknine, A.~Newson, J.~Rebut, F.~Tupin and P.~Pérez.
\newblock {\em {CARRADA} {Dataset}: {Camera} and {Automotive} {Radar} with
  {Range}-{Angle}-{Doppler} {Annotations}}.
\newblock In {IEEE} {International} {Conference} on {Pattern} {Recognition}
  ({ICPR}), 2020.

\bibitem[Ouaknine \textit{et~al.}~2021]{ouaknine_multi-view_2021}
Arthur Ouaknine, Alasdair Newson, Patrick Pérez, Florence Tupin and Julien
  Rebut.
\newblock {\em Multi-{View} {Radar} {Semantic} {Segmentation}}.
\newblock In {IEEE} {International} {Conference} on {Computer} {Vision}
  ({ICCV}), 2021.

\bibitem[Palffy \textit{et~al.}~2020]{palffy_cnn_2020}
Andras Palffy, Jiaao Dong, Julian F.~P. Kooij and Dariu~M. Gavrila.
\newblock {\em {CNN} based {Road} {User} {Detection} using the {3D} {Radar}
  {Cube}}.
\newblock In {IEEE} {Robotics} and {Automation} {Letter} ({RA}-{L}), 2020.

\bibitem[Pegoraro \& Rossi~2021]{pegoraro_real-time_2021}
Jacopo Pegoraro and Michele Rossi.
\newblock {\em Real-time {People} {Tracking} and {Identification} from {Sparse}
  mm-{Wave} {Radar} {Point}-clouds}.
\newblock In {IEEE} {Access}, 2021.

\bibitem[Pham \& Lefevre~2021]{pham_very_2021}
Minh-Tan Pham and Sebastien Lefevre.
\newblock {\em Very high resolution {Airborne} {PolSAR} {Image}
  {Classification} using {Convolutional} {Neural} {Networks}}.
\newblock In {IEEE} {European} {Conference} on {Synthetic} {Aperture}
  ({EUSAR}), 2021.

\bibitem[Pinheiro \textit{et~al.}~2015]{pinheiro_learning_2015}
Pedro~O. Pinheiro, Ronan Collobert and Piotr Dollár.
\newblock {\em Learning to {Segment} {Object} {Candidates}}.
\newblock In Conference on {Neural} {Information} {Processing} {Systems}
  ({NeurIPS}), 2015.

\bibitem[Pinheiro \textit{et~al.}~2016]{pinheiro_learning_2016}
Pedro~O. Pinheiro, Tsung-Yi Lin, Ronan Collobert and Piotr Dollàr.
\newblock {\em Learning to {Refine} {Object} {Segments}}.
\newblock In European {Conference} on {Computer} {Vision} ({ECCV}), 2016.

\bibitem[Prophet \textit{et~al.}~2019]{prophet_semantic_2019}
Robert Prophet, Gang Li, Christian Sturm and Martin Vossiek.
\newblock {\em Semantic {Segmentation} on {Automotive} {Radar} {Maps}}.
\newblock In {IEEE} {Intelligent} {Vehicles} {Symposium} ({IV}), 2019.

\bibitem[Prophet \textit{et~al.}~2020]{prophet_semantic_2020}
Robert Prophet, Anastasios Deligiannis, Juan-Carlos Fuentes-Michel, Ingo Weber
  and Martin Vossiek.
\newblock {\em Semantic {Segmentation} on {3D} {Occupancy} {Grids} for
  {Automotive} {Radar}}.
\newblock In {IEEE} {Access}, 2020.

\bibitem[Pérez \textit{et~al.}~2019]{perez_deep_2019}
Rodrigo Pérez, Falk Schubert, Ralph Rasshofer and Erwin Biebl.
\newblock {\em Deep {Learning} {Radar} {Object} {Detection} and
  {Classification} for {Urban} {Automotive} {Scenarios}}.
\newblock In Kleinheubach {Conference}, 2019.

\bibitem[Qi \textit{et~al.}~2017a]{qi_pointnet_2017}
Charles~R. Qi, Hao Su, Mo~Kaichun and Leonidas~J. Guibas.
\newblock {\em {PointNet}: {Deep} {Learning} on {Point} {Sets} for {3D}
  {Classification} and {Segmentation}}.
\newblock In {IEEE} {Conference} on {Computer} {Vision} and {Pattern}
  {Recognition} ({CVPR}), 2017.

\bibitem[Qi \textit{et~al.}~2017b]{qi_pointnet_2017-1}
Charles~R. Qi, Li~Yi, Hao Su and Leonidas~J. Guibas.
\newblock {\em {PointNet}++: {Deep} {Hierarchical} {Feature} {Learning} on
  {Point} {Sets} in a {Metric} {Space}}.
\newblock In Conference on {Neural} {Information} {Processing} {Systems}
  ({NeurIPS}), 2017.

\bibitem[Qi \textit{et~al.}~2018]{qi_frustum_2018}
Charles~R. Qi, Wei Liu, Chenxia Wu, Hao Su and Leonidas~J. Guibas.
\newblock {\em Frustum {PointNets} for {3D} {Object} {Detection} from {RGB}-{D}
  {Data}}.
\newblock In {IEEE} {Conference} on {Computer} {Vision} and {Pattern}
  {Recognition} ({CVPR}), 2018.

\bibitem[Qian \textit{et~al.}~2021]{qian_robust_2021}
Kun Qian, Shilin Zhu, Xinyu Zhang and Li~Erran Li.
\newblock {\em Robust {Multimodal} {Vehicle} {Detection} in {Foggy} {Weather}
  {Using} {Complementary} {Lidar} and {Radar} {Signals}}.
\newblock In {IEEE} {Conference} on {Computer} {Vision} and {Pattern}
  {Recognition} ({CVPR}), 2021.

\bibitem[Rahnemoonfar \textit{et~al.}~2020]{rahnemoonfar_radar_2020}
Maryam Rahnemoonfar, Masoud Yari and John Paden.
\newblock {\em Radar {Sensor} {Simulation} with {Generative} {Adversarial}
  {Network}}.
\newblock In {IEEE} {International} {Geoscience} and {Remote} {Sensing}
  {Symposium} ({IGARSS}), 2020.

\bibitem[Rebut \textit{et~al.}~2022]{rebut_raw_2021}
Julien Rebut, Arthur Ouaknine, Waqas Malik and Patrick Pérez.
\newblock {\em Raw {High}-{Definition} {Radar} for {Multi}-{Task} {Learning}}.
\newblock In {IEEE} {Conference} on {Computer} {Vision} and {Pattern}
  {Recognition} ({CVPR}), 2022.

\bibitem[Redmon \& Farhadi~2017]{redmon_yolo9000_2017}
Joseph Redmon and Ali Farhadi.
\newblock {\em {YOLO9000}: {Better}, {Faster}, {Stronger}}.
\newblock In {IEEE} {Conference} on {Computer} {Vision} and {Pattern}
  {Recognition} ({CVPR}), 2017.

\bibitem[Redmon \& Farhadi~2018]{redmon_yolov3_2018}
Joseph Redmon and Ali Farhadi.
\newblock {\em {YOLOv3}: {An} {Incremental} {Improvement}}.
\newblock In {ArXiv}, 2018.

\bibitem[Redmon \textit{et~al.}~2016]{redmon_you_2016}
Joseph Redmon, Santosh Divvala, Ross Girshick and Ali Farhadi.
\newblock {\em You {Only} {Look} {Once}: {Unified}, {Real}-{Time} {Object}
  {Detection}}.
\newblock In {IEEE} {Conference} on {Computer} {Vision} and {Pattern}
  {Recognition} ({CVPR}), 2016.

\bibitem[Ren \textit{et~al.}~2015]{ren_faster_2015}
Shaoqing Ren, Kaiming He, Ross Girshick and Jian Sun.
\newblock {\em Faster {R}-{CNN}: {Towards} {Real}-{Time} {Object} {Detection}
  with {Region} {Proposal} {Networks}}.
\newblock In Conference on {Neural} {Information} {Processing} {Systems}
  ({NeurIPS}), 2015.

\bibitem[Robbins \& Monro~1951]{robbins_stochastic_1951}
Herbert Robbins and Sutton Monro.
\newblock {\em A {Stochastic} {Approximation} {Method}}.
\newblock In The {Annals} of {Mathematical} {Statistics}, 1951.

\bibitem[Rohling~1983]{rohling_radar_1983}
Hermann Rohling.
\newblock {\em Radar {CFAR} {Thresholding} in {Clutter} and {Multiple} {Target}
  {Situations}}.
\newblock In Transactions on {Aerospace} and {Electronic} {Systems}, 1983.

\bibitem[Ronneberger \textit{et~al.}~2015]{ronneberger_u-net_2015}
Olaf Ronneberger, Philipp Fischer and Thomas Brox.
\newblock {\em U-{Net}: {Convolutional} {Networks} for {Biomedical} {Image}
  {Segmentation}}.
\newblock In International {Conference} on {Medical} {Image} {Computing} and
  {Computer} {Assisted} {Intervention} ({MICCAI}), 2015.

\bibitem[Rosenblatt~1958]{rosenblatt_perceptron_1958}
F.~Rosenblatt.
\newblock {\em The perceptron: {A} probabilistic model for information storage
  and organization in the brain.}
\newblock In Psychological {Review}, 1958.

\bibitem[Russakovsky \textit{et~al.}~2015]{russakovsky_imagenet_2015}
Olga Russakovsky, Jia Deng, Hao Su, Jonathan Krause, Sanjeev Satheesh, Sean Ma,
  Zhiheng Huang, Andrej Karpathy, Aditya Khosla, Michael Bernstein,
  Alexander~C. Berg and Li~Fei-Fei.
\newblock {\em {ImageNet} {Large} {Scale} {Visual} {Recognition} {Challenge}}.
\newblock In International {Journal} of {Computer} {Vision} ({IJCV}), 2015.

\bibitem[Scheiner \textit{et~al.}~2020]{scheiner_seeing_2020}
Nicolas Scheiner, Florian Kraus, Fangyin Wei, Buu Phan, Fahim Mannan, Nils
  Appenrodt, Werner Ritter, Jürgen Dickmann, Klaus Dietmayer, Bernhard Sick
  and Felix Heide.
\newblock {\em Seeing {Around} {Street} {Corners}: {Non}-{Line}-of-{Sight}
  {Detection} and {Tracking} {In}-the-{Wild} {Using} {Doppler} {Radar}}.
\newblock In {IEEE} {Conference} on {Computer} {Vision} and {Pattern}
  {Recognition} ({CVPR}), 2020.

\bibitem[Scherer \textit{et~al.}~2021]{scherer_tinyradarnn_2021}
Moritz Scherer, Michele Magno, Jonas Erb, Philipp Mayer, Manuel Eggimann and
  Luca Benini.
\newblock {\em {TinyRadarNN}: {Combining} {Spatial} and {Temporal}
  {Convolutional} {Neural} {Networks} for {Embedded} {Gesture} {Recognition}
  with {Short} {Range} {Radars}}.
\newblock In {IEEE} {Internet} of {Things} {Journal} ({IoT}-{J}), 2021.

\bibitem[Schumann \textit{et~al.}~2018]{schumann_semantic_2018}
Ole Schumann, Markus Hahn, Jurgen Dickmann and Christian Wohler.
\newblock {\em Semantic {Segmentation} on {Radar} {Point} {Clouds}}.
\newblock In {IEEE} {International} {Conference} on {Information} {Fusion}
  ({FUSION}), 2018.

\bibitem[Schumann \textit{et~al.}~2021]{schumann_radarscenes_2021}
Ole Schumann, Markus Hahn, Nicolas Scheiner, Fabio Weishaupt, Julius~F. Tilly,
  Jürgen Dickmann and Christian Wöhler.
\newblock {\em {RadarScenes}: {A} {Real}-{World} {Radar} {Point} {Cloud} {Data}
  {Set} for {Automotive} {Applications}}.
\newblock In {ArXiv}, 2021.

\bibitem[Shah \textit{et~al.}~2020]{shah_liranet_2020}
Meet Shah, Zhiling Huang, Ankit Laddha, Matthew Langford, Blake Barber, Sidney
  Zhang, Carlos Vallespi-Gonzalez and Raquel Urtasun.
\newblock {\em {LiRaNet}: {End}-to-{End} {Trajectory} {Prediction} using
  {Spatio}-{Temporal} {Radar} {Fusion}}.
\newblock In Conference on {Robot} {Learning} ({CoRL}), 2020.

\bibitem[Sheeny \textit{et~al.}~2020]{sheeny_300_2020}
Marcel Sheeny, Andrew Wallace and Sen Wang.
\newblock {\em 300 {GHz} {Radar} {Object} {Recognition} based on {Deep}
  {Neural} {Networks} and {Transfer} {Learning}}.
\newblock In {IET} {Radar}, {Sonar} and {Navigation}, 2020.

\bibitem[Sheeny \textit{et~al.}~2021]{sheeny_radiate_2021}
Marcel Sheeny, Emanuele De~Pellegrin, Saptarshi Mukherjee, Alireza Ahrabian,
  Sen Wang and Andrew Wallace.
\newblock {\em {RADIATE}: {A} {Radar} {Dataset} for {Automotive} {Perception}}.
\newblock In {IEEE} {International} {Conference} on {Robotics} and {Automation}
  ({ICRA}), 2021.

\bibitem[Shuai \textit{et~al.}~2021]{shuai_millieye_2021}
Xian Shuai, Yulin Shen, Yi~Tang, Shuyao Shi, Luping Ji and Guoliang Xing.
\newblock {\em {milliEye}: {A} {Lightweight} {mmWave} {Radar} and {Camera}
  {Fusion} {System} for {Robust} {Object} {Detection}}.
\newblock In {IEEE} {International} {Conference} on {Internet} of {Things}
  {Design} and {Implementation} ({IoTDI}), 2021.

\bibitem[Silver \textit{et~al.}~2016]{silver_mastering_2016}
David Silver, Aja Huang, Chris~J. Maddison, Arthur Guez, Laurent Sifre, George
  van~den Driessche, Julian Schrittwieser, Ioannis Antonoglou, Veda
  Panneershelvam, Marc Lanctot, Sander Dieleman, Dominik Grewe, John Nham, Nal
  Kalchbrenner, Ilya Sutskever, Timothy Lillicrap, Madeleine Leach, Koray
  Kavukcuoglu, Thore Graepel and Demis Hassabis.
\newblock {\em Mastering the game of {Go} with deep neural networks and tree
  search}.
\newblock In Nature, 2016.

\bibitem[Simonyan \& Zisserman~2015]{simonyan_very_2015}
Karen Simonyan and Andrew Zisserman.
\newblock {\em Very {Deep} {Convolutional} {Networks} for {Large}-{Scale}
  {Image} {Recognition}}.
\newblock In International {Conference} on {Learning} {Representations}
  ({ICLR}), 2015.

\bibitem[Sless \textit{et~al.}~2019]{sless_road_2019}
Liat Sless, Gilad Cohen, Bat~El Shlomo and Shaul Oron.
\newblock {\em Road {Scene} {Understanding} by {Occupancy} {Grid} {Learning}
  from {Sparse} {Radar} {Clusters} using {Semantic} {Segmentation}}.
\newblock In {IEEE} {International} {Conference} on {Computer} {Vision}
  {Workshop} ({ICCVW}), 2019.

\bibitem[Srivastava \textit{et~al.}~2014]{srivastava_dropout_2014}
Nitish Srivastava, Geoffrey Hinton, Alex Krizhevsky, Ilya Sutskever and Ruslan
  Salakhutdinov.
\newblock {\em Dropout: {A} {Simple} {Way} to {Prevent} {Neural} {Networks}
  from {Overfitting}}.
\newblock In Journal of {Machine} {Learning} {Research}, 2014.

\bibitem[Stephan \textit{et~al.}~2021]{stephan_human_2021}
Michael Stephan, Avik Santra and Georg Fischer.
\newblock {\em Human {Target} {Detection} and {Localization} with {Radars}
  {Using} {Deep} {Learning}}.
\newblock In Deep {Learning} {Applications}, 2021.

\bibitem[Stroescu \textit{et~al.}~2020]{stroescu_object_2020}
Ana Stroescu, Liam Daniel, Dominic Phippen, Mikhail Cherniakov and Marina
  Gashinova.
\newblock {\em Object {Detection} on {Radar} {Imagery} for {Autonomous}
  {Driving} {Using} {Deep} {Neural} {Networks}}.
\newblock In {IEEE} {European} {Radar} {Conference} ({EuRAD}), 2020.

\bibitem[Sun \textit{et~al.}~2017]{sun_revisiting_2017}
Chen Sun, Abhinav Shrivastava, Saurabh Singh and Abhinav Gupta.
\newblock {\em Revisiting {Unreasonable} {Effectiveness} of {Data} in {Deep}
  {Learning} {Era}}.
\newblock In {IEEE} {International} {Conference} on {Computer} {Vision}
  ({ICCV}), 2017.

\bibitem[Sun \textit{et~al.}~2019]{sun_automatic_2019}
Yuliang Sun, Tai Fei, Shangyin Gao and Nils Pohl.
\newblock {\em Automatic {Radar}-based {Gesture} {Detection} and
  {Classification} via a {Region}-based {Deep} {Convolutional} {Neural}
  {Network}}.
\newblock In {IEEE} {International} {Conference} on {Acoustics}, {Speech}, \&
  {Signal} {Processing} ({ICASSP}), 2019.

\bibitem[Sun \textit{et~al.}~2020]{sun_scalability_2020}
Pei Sun, Henrik Kretzschmar, Xerxes Dotiwalla, Aurelien Chouard, Vijaysai
  Patnaik, Paul Tsui, James Guo, Yin Zhou, Yuning Chai, Benjamin Caine, Vijay
  Vasudevan, Wei Han, Jiquan Ngiam, Hang Zhao, Aleksei Timofeev, Scott
  Ettinger, Maxim Krivokon, Amy Gao, Aditya Joshi, Yu~Zhang, Jonathon Shlens,
  Zhifeng Chen and Dragomir Anguelov.
\newblock {\em Scalability in {Perception} for {Autonomous} {Driving}: {Waymo}
  {Open} {Dataset}}.
\newblock In {IEEE} {Conference} on {Computer} {Vision} and {Pattern}
  {Recognition} ({CVPR}), 2020.

\bibitem[Sun \textit{et~al.}~2021]{sun_squeeze-and-excitation_2021}
Pengliang Sun, Xuetong Niu, Pengfei Sun and Kele Xu.
\newblock {\em Squeeze-and-{Excitation} network-{Based} {Radar} {Object}
  {Detection} {With} {Weighted} {Location} {Fusion}}.
\newblock In {ACM} {International} {Conference} on {Multimedia} {Retrieval}
  ({ICMR}), 2021.

\bibitem[Svenningsson \textit{et~al.}~2021]{svenningsson_radar-pointgnn_2021}
Peter Svenningsson, Francesco Fioranelli and Alexander Yarovoy.
\newblock {\em Radar-{PointGNN}: {Graph} {Based} {Object} {Recognition} for
  {Unstructured} {Radar} {Point}-cloud {Data}}.
\newblock In {IEEE} {Radar} {Conference} ({RadarConf}), 2021.

\bibitem[Szegedy \textit{et~al.}~2015]{szegedy_going_2015}
Christian Szegedy, {Wei Liu}, {Yangqing Jia}, Pierre Sermanet, Scott Reed,
  Dragomir Anguelov, Dumitru Erhan, Vincent Vanhoucke and Andrew Rabinovich.
\newblock {\em Going deeper with convolutions}.
\newblock In {IEEE} {Conference} on {Computer} {Vision} and {Pattern}
  {Recognition} ({CVPR}), 2015.

\bibitem[Szegedy \textit{et~al.}~2016]{szegedy_rethinking_2016}
C.~Szegedy, V.~Vanhoucke, S.~Ioffe, J.~Shlens and Z.~Wojna.
\newblock {\em Rethinking the {Inception} {Architecture} for {Computer}
  {Vision}}.
\newblock In {IEEE} {Conference} on {Computer} {Vision} and {Pattern}
  {Recognition} ({CVPR}), 2016.

\bibitem[Szegedy \textit{et~al.}~2017]{szegedy_inception-v4_2017}
Christian Szegedy, Sergey Ioffe, Vincent Vanhoucke and Alexander~A. Alemi.
\newblock {\em Inception-v4, {Inception}-{ResNet} and the {Impact} of
  {Residual} {Connections} on {Learning}}.
\newblock In {AAAI} {Conference} on {Artificial} {Intelligence}, 2017.

\bibitem[Taigman \textit{et~al.}~2014]{taigman_deepface_2014}
Yaniv Taigman, Ming Yang, Marc'Aurelio Ranzato and Lior Wolf.
\newblock {\em {DeepFace}: {Closing} the {Gap} to {Human}-{Level} {Performance}
  in {Face} {Verification}}.
\newblock In {IEEE} {Conference} on {Computer} {Vision} and {Pattern}
  {Recognition} ({CVPR}), 2014.

\bibitem[Touvron \textit{et~al.}~2021]{touvron_training_2021}
Hugo Touvron, Matthieu Cord, Matthijs Douze, Francisco Massa, Alexandre
  Sablayrolles and Hervé Jégou.
\newblock {\em Training data-efficient image transformers \& distillation
  through attention}.
\newblock In International {Conference} on {Machine} {Learning} ({ICML}), 2021.

\bibitem[Truong \& Yanushkevich~2019]{truong_generative_2019}
Thomas Truong and Svetlana Yanushkevich.
\newblock {\em Generative {Adversarial} {Network} for {Radar} {Signal}
  {Synthesis}}.
\newblock In {IEEE} {International} {Joint} {Conference} on {Neural} {Networks}
  ({IJCNN}), 2019.

\bibitem[Uijlings \textit{et~al.}~2013]{uijlings_selective_2013}
J.~R.~R. Uijlings, K.~E. A. van~de Sande, T.~Gevers and {A.W.M. Smeulders}.
\newblock {\em Selective {Search} for {Object} {Recognition}}.
\newblock In International {Journal} of {Computer} {Vision} ({IJCV}), 2013.

\bibitem[Wallace \textit{et~al.}~2021]{wallace_combining_2021}
Andrew~M. Wallace, Saptarshi Mukherjee, Bemsibom Toh and Alireza Ahrabian.
\newblock {\em Combining automotive radar and {LiDAR} for surface detection in
  adverse conditions}.
\newblock In {IET} {Radar}, {Sonar} \& {Navigation}, 2021.

\bibitem[Wang \textit{et~al.}~2016]{wang_interacting_2016}
Saiwen Wang, Jie Song, Jaime Lien, Ivan Poupyrev and Otmar Hilliges.
\newblock {\em Interacting with {Soli}: {Exploring} {Fine}-{Grained} {Dynamic}
  {Gesture} {Recognition} in the {Radio}-{Frequency} {Spectrum}}.
\newblock In {ACM} {User} {Interface} {Software} and {Technology} ({UIST}),
  2016.

\bibitem[Wang \textit{et~al.}~2019]{wang_rammar_2019}
Yong Wang, Xiuqian Jia, Mu~Zhou, Xiaolong Yang and Zengshan Tian.
\newblock {\em Rammar: {RAM} {Assisted} {Mask} {R}-{CNN} for {FMCW} {Sensor}
  {Based} {HGD} {System}}.
\newblock In {IEEE} {International} {Conference} on {Communications} ({ICC}),
  2019.

\bibitem[Wang \textit{et~al.}~2020a]{wang_high_2020}
Leichen Wang, Tianbai Chen, Carsten Anklam and Bastian Goldluecke.
\newblock {\em High {Dimensional} {Frustum} {PointNet} for {3D} {Object}
  {Detection} from {Camera}, {LiDAR}, and {Radar}}.
\newblock In {IEEE} {Intelligent} {Vehicles} {Symposium} ({IV}), 2020.

\bibitem[Wang \textit{et~al.}~2020b]{wang_deep_2020}
Yuchen Wang, Mingze Xu, John Paden, Lora Koenig, Geoffrey Fox and David
  Crandall.
\newblock {\em Deep {Tiered} {Image} {Segmentation} {forDetecting} {Internal}
  {Ice} {Layers} in {Radar} {Imagery}}.
\newblock In {IEEE} {International} {Conference} on {Multimedia} \& {Expo}
  ({ICME}), 2020.

\bibitem[Wang \textit{et~al.}~2021a]{wang_rodnet_2021}
Yizhou Wang, Zhongyu Jiang, Yudong Li, Jenq-Neng Hwang, Guanbin Xing and Hui
  Liu.
\newblock {\em {RODNet}: {A} {Real}-{Time} {Radar} {Object} {Detection}
  {Network} {Cross}-{Supervised} by {Camera}-{Radar} {Fused} {Object} {3D}
  {Localization}}.
\newblock In Journal of {Selected} {Topics} in {Signal} {Processing}, 2021.

\bibitem[Wang \textit{et~al.}~2021b]{wang_rethinking_2021}
Yizhou Wang, Gaoang Wang, Hung-Min Hsu, Hui Liu and Jenq-Neng Hwang.
\newblock {\em Rethinking of {Radar}'s {Role}: {A} {Camera}-{Radar} {Dataset}
  and {Systematic} {Annotator} via {Coordinate} {Alignment}}.
\newblock In {IEEE} {Conference} on {Computer} {Vision} and {Pattern}
  {Recognition} {Workshop} ({CVPRW}), 2021.

\bibitem[Wang \textit{et~al.}~2021c]{wang_human_2021}
Zeyu Wang, Chenglin Yao, Jianfeng Ren and Xudong Jiang.
\newblock {\em Human {Activity} {Recognition} {Using} {3D}
  {Orthogonally}-projected {EfficientNet} on {Radar} {Time}-{Range}-{Doppler}
  {Signature}}.
\newblock In {ArXiv}, 2021.

\bibitem[Weston \textit{et~al.}~2019]{weston_probably_2019}
Rob Weston, Sarah Cen, Paul Newman and Ingmar Posner.
\newblock {\em Probably {Unknown}: {Deep} {Inverse} {Sensor} {Modelling}
  {Radar}}.
\newblock In {IEEE} {International} {Conference} on {Robotics} and {Automation}
  ({ICRA}), 2019.

\bibitem[Weston \textit{et~al.}~2021]{weston_there_2021}
Rob Weston, Oiwi~Parker Jones and Ingmar Posner.
\newblock {\em There and {Back} {Again}: {Learning} to {Simulate} {Radar}
  {Data} for {Real}-{World} {Applications}}.
\newblock In {IEEE} {International} {Conference} on {Robotics} and {Automation}
  ({ICRA}), 2021.

\bibitem[Xie \textit{et~al.}~2017]{xie_aggregated_2017}
Saining Xie, Ross Girshick, Piotr Dollár, Zhuowen Tu and Kaiming He.
\newblock {\em Aggregated {Residual} {Transformations} for {Deep} {Neural}
  {Networks}}.
\newblock In {IEEE} {Conference} on {Computer} {Vision} and {Pattern}
  {Recognition} ({CVPR}), 2017.

\bibitem[Xu \textit{et~al.}~2021]{xu_rpfa-net_2021}
Baowei Xu, Xinyu Zhang, Li~Wang, Xiaomei Hu, Zhiwei Li, Shuyue Pan, Jun Li and
  Yongqiang Deng.
\newblock {\em {RPFA}-{Net}: a {4D} {RaDAR} {Pillar} {Feature} {Attention}
  {Network} for {3D} {Object} {Detection}}.
\newblock In {IEEE} {International} {Conference} on {Intelligent}
  {Transportation} {Systems} ({ITSC}), 2021.

\bibitem[Yadav \textit{et~al.}~2020]{yadav_radarrgb_2020}
Ritu Yadav, Axel Vierling and Karsten Berns.
\newblock {\em Radar+{RGB} {Attentive} {Fusion} for {Robust} {Object}
  {Detection} in {Autonomous} {Vehicles}}.
\newblock In {IEEE} {International} {Conference} on {Image} {Processing}
  ({ICIP}), 2020.

\bibitem[Yang \textit{et~al.}~2018]{yang_pixor_2018}
Bin Yang, Wenjie Luo and Raquel Urtasun.
\newblock {\em {PIXOR}: {Real}-time {3D} {Object} {Detection} from {Point}
  {Clouds}}.
\newblock In {IEEE} {Conference} on {Computer} {Vision} and {Pattern}
  {Recognition} ({CVPR}), 2018.

\bibitem[Yang \textit{et~al.}~2020]{yang_radarnet_2020}
Bin Yang, Runsheng Guo, Ming Liang, Sergio Casas and Raquel Urtasun.
\newblock {\em {RadarNet}: {Exploiting} {Radar} for {Robust} {Perception} of
  {Dynamic} {Objects}}.
\newblock In European {Conference} on {Computer} {Vision} ({ECCV}), 2020.

\bibitem[Yu \& Koltun~2016]{yu_multi-scale_2016}
Fisher Yu and Vladlen Koltun.
\newblock {\em Multi-{Scale} {Context} {Aggregation} by {Dilated}
  {Convolutions}}.
\newblock In International {Conference} on {Learning} {Representations}
  ({ICLR}), 2016.

\bibitem[Yu \textit{et~al.}~2020]{yu_bdd100k_2020}
Fisher Yu, Wenqi Xian, Yingying Chen, Fangchen Liu, Mike Liao, Vashisht
  Madhavan and Trevor Darrell.
\newblock {\em {BDD100K}: {A} {Diverse} {Driving} {Video} {Database} with
  {Scalable} {Annotation} {Tooling}}.
\newblock In {IEEE} {Conference} on {Computer} {Vision} and {Pattern}
  {Recognition} ({CVPR}), 2020.

\bibitem[Zeiler \& Fergus~2014]{zeiler_visualizing_2014}
Matthew~D. Zeiler and Rob Fergus.
\newblock {\em Visualizing and {Understanding} {Convolutional} {Networks}}.
\newblock In European {Conference} on {Computer} {Vision} ({ECCV}), 2014.

\bibitem[Zhang \textit{et~al.}~2017a]{zhang_deep_2017}
Hang Zhang, Jia Xue and Kristin Dana.
\newblock {\em Deep {TEN}: {Texture} {Encoding} {Network}}.
\newblock In {IEEE} {Conference} on {Computer} {Vision} and {Pattern}
  {Recognition} ({CVPR}), 2017.

\bibitem[Zhang \textit{et~al.}~2017b]{zhang_complex-valued_2017}
Zhimian Zhang, Haipeng Wang, Feng Xu and Ya-Qiu Jin.
\newblock {\em Complex-{Valued} {Convolutional} {Neural} {Network} and {Its}
  {Application} in {Polarimetric} {SAR} {Image} {Classification}}.
\newblock In Transactions on {Geoscience} and {Remote} {Sensing}, 2017.

\bibitem[Zhang \textit{et~al.}~2018a]{zhang_context_2018}
Hang Zhang, Kristin Dana, Jianping Shi, Zhongyue Zhang, Xiaogang Wang, Ambrish
  Tyagi and Amit Agrawal.
\newblock {\em Context {Encoding} for {Semantic} {Segmentation}}.
\newblock In {IEEE} {Conference} on {Computer} {Vision} and {Pattern}
  {Recognition} ({CVPR}), 2018.

\bibitem[Zhang \textit{et~al.}~2018b]{zhang_latern_2018}
Zhenyuan Zhang, Zengshan Tian and Mu~Zhou.
\newblock {\em Latern: {Dynamic} {Continuous} {Hand} {Gesture} {Recognition}
  {Using} {FMCW} {Radar} {Sensor}}.
\newblock In Sensors {Journal}, 2018.

\bibitem[Zhang \textit{et~al.}~2019]{zhang_u-deephand_2019}
Zhenyuan Zhang, Zengshan Tian, Ying Zhang, Mu~Zhou and Bang Wang.
\newblock {\em u-{DeepHand}: {FMCW} {Radar}-{Based} {Unsupervised} {Hand}
  {Gesture} {Feature} {Learning} {Using} {Deep} {Convolutional}
  {Auto}-{Encoder} {Network}}.
\newblock In Sensors {Journal}, 2019.

\bibitem[Zhang \textit{et~al.}~2020]{zhang_polsar_2020}
Lamei Zhang, Siyu Zhang, Hongwei Dong and Da~Lu.
\newblock {\em Polsar {Image} {Classification} via {Complex}-{Valued}
  {Multi}-{Scale} {Convolutional} {Neural} {Network}}.
\newblock In {IEEE} {International} {Geoscience} and {Remote} {Sensing}
  {Symposium} ({IGARSS}), 2020.

\bibitem[Zhang \textit{et~al.}~2021a]{zhang_raddet_2021}
Ao~Zhang, Farzan~Erlik Nowruzi and Robert Laganiere.
\newblock {\em {RADDet}: {Range}-{Azimuth}-{Doppler} based {Radar} {Object}
  {Detection} for {Dynamic} {Road} {Users}}.
\newblock In Computer {Vision} and {Robotics} ({CVR}), 2021.

\bibitem[Zhang \textit{et~al.}~2021b]{zhang_rvdet_2021}
Jingwei Zhang, Ming Zhang, Zicheng Fang, Yulong Wang, Xian Zhao and Shiliang
  Pu.
\newblock {\em {RVDet}: {Feature}-level {Fusion} of {Radar} and {Camera} for
  {Object} {Detection}}.
\newblock In {IEEE} {International} {Conference} on {Intelligent}
  {Transportation} {Systems} ({ITSC}), 2021.

\bibitem[Zhao \textit{et~al.}~2017]{zhao_pyramid_2017}
Hengshuang Zhao, Jianping Shi, Xiaojuan Qi, Xiaogang Wang and Jiaya Jia.
\newblock {\em Pyramid {Scene} {Parsing} {Network}}.
\newblock In {IEEE} {Conference} on {Computer} {Vision} and {Pattern}
  {Recognition} ({CVPR}), 2017.

\bibitem[Zheng \textit{et~al.}~2021]{zheng_scene-aware_2021}
Zangwei Zheng, Xiangyu Yue, Kurt Keutzer and Alberto~Sangiovanni Vincentelli.
\newblock {\em Scene-aware {Learning} {Network} for {Radar} {Object}
  {Detection}}.
\newblock In {ACM} {International} {Conference} on {Multimedia} {Retrieval}
  ({ICMR}), 2021.

\bibitem[Zhou \& Tuzel~2018]{zhou_voxelnet_2018}
Yin Zhou and Oncel Tuzel.
\newblock {\em {VoxelNet}: {End}-to-{End} {Learning} for {Point} {Cloud}
  {Based} {3D} {Object} {Detection}}.
\newblock In {IEEE} {Conference} on {Computer} {Vision} and {Pattern}
  {Recognition} ({CVPR}), 2018.

\bibitem[Zhu \textit{et~al.}~2017]{zhu_unpaired_2017}
Jun-Yan Zhu, Taesung Park, Phillip Isola and Alexei~A. Efros.
\newblock {\em Unpaired {Image}-to-{Image} {Translation} using
  {Cycle}-{Consistent} {Adversarial} {Networks}}.
\newblock In {IEEE} {International} {Conference} on {Computer} {Vision}
  ({ICCV}), 2017.

\bibitem[Zhu \textit{et~al.}~2020]{zhu_classification_2020}
JianPing Zhu, HaiQuan Chen and WenBin Ye.
\newblock {\em Classification of {Human} {Activities} {Based} on {Radar}
  {Signals} using {1D}-{CNN} and {LSTM}}.
\newblock In {IEEE} {International} {Symposium} on {Circuits} and {Systems}
  ({ISCAS}), 2020.

\end{thebibliography}

\appendix

\cleardoublepage
\mtcaddpart[Appendices]
\part*{Appendices}

\chapter{Background}
\label{chap:appendix2}

\section{Deep learning}

\subsection{Object detection}
\label{sec_app:background_detection}

This section details deep learning methods and architectures for object detection in natural images. It provides additional details to those presented in Section \ref{sec:background_detection}.

\paragraph{Region-based Fully Convolutional Network (R-FCN).}
Fast and Faster R-CNN requires multiple sub-networks to process the region proposals produced by the \ac{RPN}.
The R-FCN proposed by \cite{dai_r-fcn_2016} uses a single network, shared on the entire image, processing the region proposals.
This module is fully convolutional except the last fully connected layer used for classification.
The architecture starts with a ResNet-101 \cite{he_deep_2016} processing the input and generating feature maps corresponding to the number of classes.
These feature maps are called position-sensitive score maps because they take into account the
spatial localisation of a particular object. 
There are $s {\times} s {\times} (K+1)$ score maps forming a ``score bank'', where $s$ is the size of the score map and $K$ the number of classes. The idea of these maps is to create patches recognizing a single part of an object, each patch being specialized in a category.
A \ac{RPN} generates \ac{RoI} in parallel. The \ac{RoI} extracted in the feature maps are placed into bins which are compared with the score bank. A vote is performed, if enough bins are activated (regarding the corresponding score), the patch localizes and classifies the object.
The best R-FCN have reached \ac{mAP} scores of 83.6\% on the 2007 PASCAL VOC challenge while bein trained using the 2007, 2012 PASCAL VOC datasets and the COCO dataset. 
On 2015 COCO challenge, they performed a \ac{mAP} of 53.2\% for an IoU = 0.5 and a
score of 31.5\% for the official \ac{mAP} metric. The authors noticed that the R-FCN
is 2.5 to 20 times faster than the Faster R-CNN counterpart.

\paragraph{YOLO extensions.}
In their work, \cite{redmon_yolo9000_2017} improved the YOLO presented in a previous paragraph in YOLOv2 and proposed a new model, YOLO9000, capable of detecting more than 9000 categories while running in almost real time (around 10 \ac{FPS}). 

The YOLOv2 focuses on improving the accuracy while still being a fast detector. Batch normalization is added to prevent over-fitting without using dropout. 
The model is also able to process higher resolution images to potentially detect smaller objects: $608 {\times} 608$ instead of $448 {\times} 448$ in the initial version.
The final fully connected layer of the YOLO model predicting the bounding box coordinates has been removed to use anchor boxes in the same way as Faster R-CNN. 
The input image is divided into a grid of cells, each one containing 5 anchor boxes. YOLOv2 uses $19 {\times} 19 {\times} 5$ = 1805 anchor boxes by image instead of 98 boxes for the YOLO model. YOLOv2 predicts the correction of each anchor box relative to the location of the grid cell (ranging between 0 and 1) and selects the boxes according to their confidence as the SSD model.
The dimensions of the anchor boxes have been fixed using a k-means clustering on the training set of bounding boxes.
It uses a ResNet architecture to stack high and low resolution feature
maps to detect smaller objects. Their proposed backbone called Darknet-19 is composed of 19 convolutional layers with $3 {\times} 3$ and $1 {\times} 1$ kernels,  followed by max-pooling layers to reduce the output dimension. A final $1 {\times} 1$ convolutional layer outputs 5 boxes per cell of the grid with 5 coordinates and a probability for each class (20 in the case of the PASCAL VOC dataset).
The YOLOv2 model trained with the 2007 and 2012 PASCAL VOC datasets obtained a 78.6\% \ac{mAP} score on the 2007 PASCAL VOC challenge at 40 \ac{FPS} during inference. The model trained with the 2015 COCO dataset got \ac{mAP} scores on the corresponding challenge of 44.0\% for an \ac{IoU} threshold of 0.5, 19.2\% for an \ac{IoU} threshold of 0.75 and 21.6\% for the official \ac{mAP} metric.

To train their second proposed architecture, the authors have combined the ImageNet dataset with the COCO dataset. The ImageNet dataset for classification contains 1000 categories and the 2015 COCO dataset only 80 categories. The ImageNet classes are based on the WordNet\footnote{\url{https://wordnet.princeton.edu/}} lexicon developed by the Princeton University which is composed of more than 20,000 words. \cite{redmon_yolo9000_2017} detail a method to construct a tree version of the WordNet. 
The model predicting on an image will use a softmax applied on a group of labels with the same hyponym\footnote{An hyponym is a word of more specific meaning than a general or superordinate term applicable to. Source: Oxford Languages.}. 
The final probability associated to a label is computed with posterior probabilities in the tree. When the authors extend the concept to the entire WordNet lexicon excluding under-represented categories, they obtain more than 9,000 categories. 
The ImageNet and COCO dataset combination is used to train a YOLOv2 architecture with 3 prior convolution layers instead of 5 to limit the output size. This model called YOLO9000 is theoretically able to learn categories of the presented lexicon.
It is evaluated on the ImageNet dataset for the detection task with around 200 labels. Only 44 labels are shared between the training and the testing dataset so the results are not significant. It gets a 19.7\% \ac{mAP} score overall the test dataset.

\subsection{Segmentation}
\label{sec_app:background_segmentation}

This section details deep learning methods and architectures for semantic segmentation of natural images. It provides additional details to those presented in Section \ref{sec:background_segmentation}.

\paragraph{ParseNet.}
In their work, \cite{liu_parsenet_2015} present improvements on the \ac{FCN} method \cite{long_fully_2015}. They claimed that the \ac{FCN} model looses the global context of the image in its deep layers by specializing the generated feature maps. 
The ParseNet is an end-to-end convolutional network 
with ``contexture'' modules.
This module takes feature maps as input and processes them with two pathways. The first one aggregates the feature maps with a global pooling to obtain global features normalised with an L2-norm which are finally up-pooled (replicated) to recover the original dimensions.
The second pathway simply normalises the input feature maps with a L2-norm. The outputs of both pathways are stacked and transmitted to the next layer.
The normalization is helpful to scale the concatenated feature maps values and it leads to better performances. The ParseNet is a \ac{FCN} with contexture modules instead of simple convolutional layers. It obtained a 69.8\% \ac{mIoU} score on the 2012 PASCAL VOC segmentation challenge and a 40.4\% \ac{mIoU} score on the PASCAL-Context challenge.

\paragraph{Convolution and deconvolutional networks.}
\cite{noh_learning_2015} proposed an end-to-end architecture composed of two linked parts.
The first one is a convolutional network with a VGG16 architecture with only two fully connected layers.
It takes as input a proposal, \textit{e.g.} a bounding box generated by an object detection model. The proposal is processed and transformed by the convolutional network to generate a vector of features. The deconvolutional network takes the vector of features as input and aims to classify pixel-wise the entire image. 
The deconvolutional network is composed of successive layers of un-pooling and deconvolutions.
The un-pooling targets the location of the maximum value in the mirrored feature map reduced by the max-pooling in the first convolutional network. It expands the feature map with sparse values, the value taken as input is placed at the recorded position while the others are filled with zeros.
The deconvolution takes a single value as input and generates a $k {\times} k$ output depending of the kernel size. It aims in learning the expansion of the feature maps while recovering the dimension of the input and keeping the information dense.
The authors analyzed deconvolution feature maps and noticed that low-level ones are specific to the shape while the higher-level ones help to classify the proposal. Finally, when all the proposals of an image are processed by the entire network, the maps are concatenated to obtain the fully segmented image. This network obtained a 72.5\% \ac{mIoU} on the 2012 PASCAL VOC segmentation challenge.

\paragraph{Pyramid scene parsing network (PSPNet).}
\cite{zhao_pyramid_2017} developed the PSPNet to better learn the global context representation of a scene.
Patterns are extracted from the input image using a ResNet \cite{he_deep_2016} as feature extractor with a dilated network strategy\footnote{The dilated convolutional layer has been proposed by \cite{yu_multi-scale_2016}. It consists in a convolutional layer with an expanded kernel (the neurons of the kernel are no more side-by-side). A dilation rate fixes the gap between two neurons in term of pixel. More details are provided in the DeepLab paragraph.}. 
The feature maps are used as input of a pyramid pooling module to distinguish patterns with different scales. 
This pooling layer consists in four parallel convolutions with different kernel sizes. Their respective outputs are processed by $1 {\times} 1$ convolutions to reduce the number of feature maps and up-sampled using bilinear interpolation to match their dimensions.
Each pyramid level analyses sub-regions of the image at different locations.
The outputs of the pyramid levels are concatenated to the initial feature maps. The output thus contains the local and the global context information. They are finally processed by a convolutional layer to generate the pixel-wise predictions. The best PSPNet with a pre-trained ResNet (using the COCO dataset) reached a 85.4\% \ac{mIoU} score on the 2012 PASCAL VOC semantic segmentation challenge.

\paragraph{Path aggregation network (PANet).}

In their work, \cite{liu_path_2018} proposed the PANet, an improvement of the information propagation in the Mask R-CNN and FPN frameworks.
The feature extractor of the network uses a FPN architecture with an additional augmented bottom-up pathway of convolutions improving the propagation of low-layer features. 
Each stage of this third pathway takes as input the feature maps of the previous stage and processes them with a $3 {\times} 3$ convolutional layer. The output is summed with the same stage feature maps of the top-down pathway using lateral connections. These maps are used as input to the next stage.

The feature maps of the augmented bottom-up pathway are pooled with a RoIAlign layer to extract proposals from all level features. An adaptive feature pooling layer processes the maps of each stage with a fully connected layer and concatenate all the outputs.
The output of the adaptive feature pooling layer is taken as input of three branches similarly to the Mask R-CNN. The two first branches uses a fully connected layer to generate the predictions of the regression for the location and the classification. 
The third branch processes the \ac{RoI} with a \ac{FCN} to predict a binary pixel-wise mask for the detected object. 
An additional path processes the output of a convolutional layer of the \ac{FCN} with a fully connected layer improving the location of the predicted pixels. Finally the output of the parallel path is reshaped and concatenated to the output of the \ac{FCN} generating the binary mask.

The PANet achieved 42.0\% \ac{AP} score on the 2016 COCO segmentation challenge using a ResNeXt as feature extractor. They also performed the 2017 COCO segmentation challenge with an 46.7\% \ac{AP} score using a ensemble of seven feature extractors: ResNet \cite{he_deep_2016}, ResNeXt \cite{xie_aggregated_2017} and SENet \cite{hu_squeeze-and-excitation_2018}.

\paragraph{Context encoding network (EncNet).}
\cite{zhang_context_2018} created the EncNet architecture capturing global information in an image to improve semantic segmentation. 
The model starts by using a ResNet \cite{he_deep_2016} backbone followed by a ``context encoding'' module inspired from the encoding layer of \cite{zhang_deep_2017}. 
It learns visual centers and smoothing factors creating an embedding of contextual information while specializing feature maps by class. 
The final layer of the module learns scaling factors of contextual information by class with an attention layer (fully connected layer). Theses factors are applied to the output feature maps of the backbone network via a skip connection.
In parallel, an additional fully connected layer classifies the global context of the module with a softmax activation. This branch is trained with a  \acf{BCE} loss, named semantic encoding loss (SE-Loss), regularizing the training of the module with the entire context (at the contrary to pixel-wise loss). The outputs of the context encoding module are reshaped and processed by a dilated convolution while minimizing two SE-losses and a final pixel-wise loss. The best EncNet reached 52.6\% \ac{mIoU} on the PASCAL-Context challenge. It also achieved a 85.9\% \ac{mIoU} score on the 2012 PASCAL VOC segmentation challenge.


\chapter{RADAR scene understanding}
\label{chap:appendix5}

\section{Multi-view RADAR semantic segmentation}

\subsection{RAD tensor visualisation}
\label{sec_app:mvrss_rad_tensor_vis}

\begin{figure}[h]
\begin{center}
\includegraphics[width=1\linewidth]{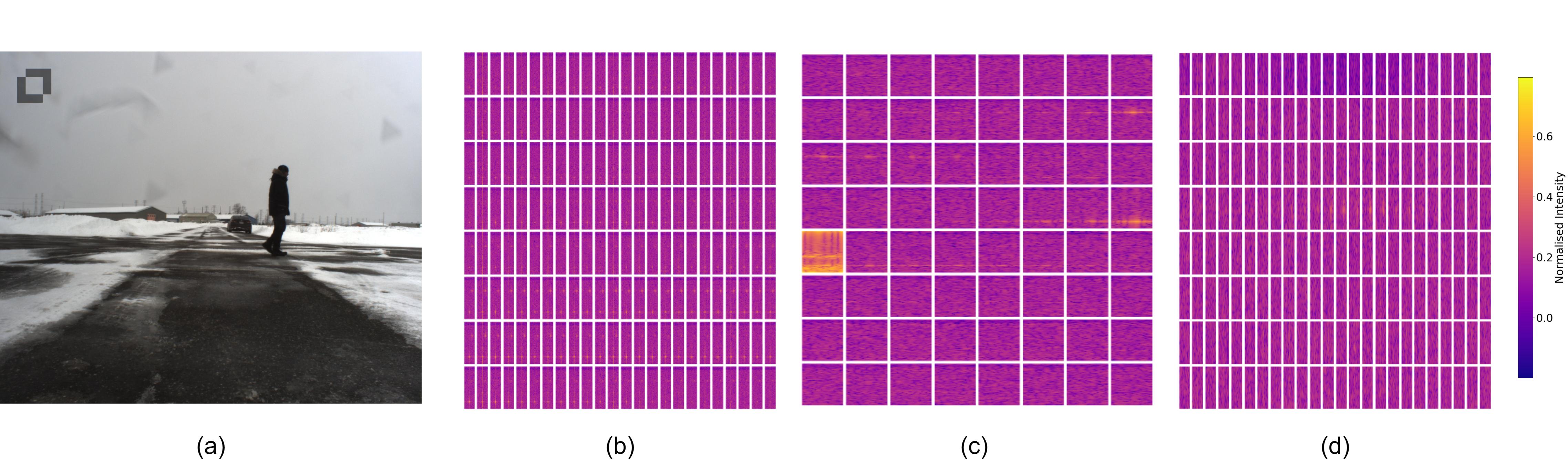}
\end{center}
   \caption[Visualisation of the Range-Angle-Doppler (RAD) tensor]{\textbf{Visualisation of the \acf{RAD} tensor}. (a) Camera image of the scene. The corresponding \ac{RAD} tensor is visualised by slices of (b) range-Doppler, (c) range-angle or (d) angle-Doppler views w.r.t. their discretized third axis.}
\label{fig:mvrss_rad_tensor}
\end{figure}

An illustration of the \ac{RAD} tensor is proposed in Figure \ref{fig:mvrss_rad_tensor}. Each slice of 2D views corresponds to a discretized bin of the third axis. In Figure \ref{fig:mvrss_rad_tensor}(b) for instance, the $256$ range-Doppler slices correspond to the view of each discretized value of the angle axis. One can observe redundant signal information and a significant level of noise for each group of 2D-view slices.

\subsection{Detailed multi-view architectures}
\label{sec_app:mvrss_archi_details}

The architecture details of the proposed multi-view network (MV-Net) and temporal multi-view network with ASPP modules (TMVA-Net) are respectively provided in Tables \ref{table:mvrss_mvnet_archi} and \ref{table:mvrss_tmvanet_archi}.
For each layer, the parameters of the operations used are specified in the following manner:
\begin{itemize}
    \item $n$-dim convolution: Conv$n$D\,(input\_channels, output\_channels, kernel\_size, stride, padding, dilation\_rate);
    \item \mbox{$n$-dim up-convolution: ConvTranspose$n$D\,(input\_chan-}\\nels, output\_channels, kernel\_size, stride, padding, dilation\_rate);
    \item maximum pooling: MaxPool2D\,(kernel\_size, stride);
    \item atrous spatial pyramid pooling: ASPP\,(input\_channels, output\_channels);
    \item $n$-D batch normalisation: BN$n$D\,(input\_channels);
    \item Leaky ReLU activation:  LeakyReLU\,(negative\_slope);
\end{itemize}
Where $n\!\in\!\{1,2,3\}$ is the dimension of the associated operation.

The ASPP module \cite{chen_deeplab_2018} has the same architecture as the one introduced by Kaul \etal \cite{kaul_rss-net_2020} for range-angle semantic segmentation. We note that the `output\_channels' parameter for the ASPP module corresponds to the number of output channels for each parallel convolution. We also note that the `stride' parameter can be either a scalar or a tuple of scalars depending on the axis on which it is applied.


\begin{table*}
\begin{center}
\tiny
\renewcommand{\arraystretch}{1.8}
\begin{tabular}{c c c c l}
\toprule
 & Layer & Inputs & \makecell[c]{Output resolution \\ ($C \times H \times W$)} & Operations \\
\midrule
\multirow{5}{*}{\textbf{RD Encoder}} 
& rd\_layer1 & RD view & $128 \times 256 \times 64$ & \makecell[l]{Conv2D(3, 128, $3 \times 3$, 1, 1, 1) $+$ BN2D(128) $+$ LeakyReLU(0.01) \\ Conv2D(128, 128, $3 \times 3$, 1, 1, 1) $+$ BN2D(128) $+$ LeakyReLU(0.01)}\\

& rd\_layer2 & rd\_layer1 & $128 \times 128 \times 64$ & MaxPool2D(2, (2, 1)) \\

& rd\_layer3 & rd\_layer2 & $128 \times 128 \times 64$ & \makecell[l]{Conv2D(128, 128, $3 \times 3$, 1, 1, 1) $+$ BN2D(128) $+$ LeakyReLU(0.01) \\ Conv2D(128, 128, $3 \times 3$, 1, 1, 1) $+$ BN2D(128) $+$ LeakyReLU(0.01)}\\

& rd\_layer4 & rd\_layer3 & $128 \times 64 \times 64$ & MaxPool2D(2, (2, 1)) \\

& rd\_layer5 & rd\_layer4 & $128 \times 64 \times 64$ & Conv1D(128, 128, $1 \times 1$, 1, 0, 1)\\

\midrule
\multirow{5}{*}{\textbf{RA Encoder}} 
& ra\_layer1  & RA view & $128 \times 256 \times 256$ & \makecell[l]{Conv2D(3, 128, $3 \times 3$, 1, 1, 1) $+$ BN2D(128) $+$ LeakyReLU(0.01) \\ Conv2D(128, 128, $3 \times 3$, 1, 1, 1) $+$ BN2D(128) $+$ LeakyReLU(0.01)} \\

& ra\_layer2 & ra\_layer1 & $128 \times 128 \times 128$ & MaxPool2D(2, 2)\\

& ra\_layer3 & ra\_layer2 & $128 \times 128 \times 128$ & \makecell[l]{Conv2D(128, 128, $3 \times 3$, 1, 1, 1) $+$ BN2D(128) $+$ LeakyReLU(0.01) \\ Conv2D(128, 128, $3 \times 3$, 1, 1, 1) $+$ BN2D(128) $+$ LeakyReLU(0.01)} \\

& ra\_layer4 & ra\_layer3 & $128 \times 64 \times 64$ & MaxPool2D(2, 2)\\

& ra\_layer5 & ra\_layer4  & $128 \times 64 \times 64$ & Conv1D(128, 128, $1 \times 1$, 1, 0, 1)\\

\midrule
\textbf{Latent space} & layer6 & rd\_layer5, ra\_layer5 & $256 \times 64 \times 64$ &	concatenate(rd\_layer5, ra\_layer5) \\

\midrule
\multirow{6}{*}{\textbf{RD Decoder}} 
& rd\_layer7 & layer6 & $128 \times 64 \times 64$ & Conv1D(256, 128, $1 \times 1$, 1, 0, 1) \\

& rd\_layer8 & rd\_layer7 & $128 \times 128 \times 64$ & ConvTranspose2D(128, 128, $2 \times 1$, (2, 1), 1, 1) \\

& rd\_layer9 & rd\_layer8 & $128 \times 128 \times 64$ & \makecell[l]{Conv2D(128, 128, $3 \times 3$, 1, 1, 1) $+$ BN2D(128) $+$ LeakyReLU(0.01) \\ Conv2D(128, 128, $3 \times 3$, 1, 1, 1) $+$ BN2D(128) $+$ LeakyReLU(0.01)} \\

& rd\_layer10 & rd\_layer9 & $128 \times 256 \times 64$	& ConvTranspose2D(128, 128, $2 \times 1$, (2, 1), 1, 1) \\

& rd\_layer11 & rd\_layer10 & $128 \times 256 \times 64$ & \makecell[l]{Conv2D(128, 128, $3 \times 3$, 1, 1, 1) $+$ BN2D(128) $+$ LeakyReLU(0.01) \\ Conv2D(128, 128, $3 \times 3$, 1, 1, 1) $+$ BN2D(128) $+$ LeakyReLU(0.01)} \\

& rd\_layer12 & rd\_layer11 & $K \times 256 \times 64$ & Conv1D(128, $K$, $1 \times 1$, 1, 0, 1) \\

\midrule
\multirow{6}{*}{\textbf{RA Decoder}} 
& ra\_layer7 & layer6 &	$128 \times 64 \times 64$ & Conv1D(256, 128, $1 \times 1$, 1, 0, 1) \\

& ra\_layer8 & ra\_layer7 & $128 \times 128 \times 128$ & ConvTranspose2D(128, 128, $2 \times 2$, 2, 1, 1) \\

& ra\_layer9 & ra\_layer8 & $128 \times 128 \times 128$ & \makecell[l]{Conv2D(128, 128, $3 \times 3$, 1, 1, 1) $+$ BN2D(128) $+$ LeakyReLU(0.01) \\ Conv2D(128, 128, $3 \times 3$, 1, 1, 1) $+$ BN2D(128) $+$ LeakyReLU(0.01)} \\

& ra\_layer10 & ra\_layer9 & $128 \times 256 \times 256$ & ConvTranspose2D(128, 128, $2 \times 2$, 2, 1, 1) \\

& ra\_layer11 & ra\_layer10 & $128 \times 256 \times 256$ & \makecell[l]{Conv2D(128, 128, $3 \times 3$, 1, 1, 1) $+$ BN2D(128) $+$ LeakyReLU(0.01) \\ Conv2D(128, 128, $3 \times 3$, 1, 1, 1) $+$ BN2D(128) $+$ LeakyReLU(0.01)} \\

& ra\_layer12 & ra\_layer11 & $K \times 256 \times 256$ & Conv1D(128, $K$, $1 \times 1$, 1, 0, 1) \\

\bottomrule
\end{tabular}
\end{center}
\caption[Multi-view network (MV-Net) architecture]{\textbf{Multi-view network (MV-Net) architecture}. This table lists all the layers contained in the model taking as input multi-view \ac{RADAR} representations (\ac{RD} and \ac{RA} views) to predict segmentation maps for each multi-view output. 
Details about the parameters of each operation are provided in Section \ref{sec_app:mvrss_archi_details}.
Let $K$ be the number of classes.
The number of input channels in the first layer corresponds to the consecutive frames of each view stacked in temporal dimension, here $q=2$ and thus the number of channels is $3$.}
\label{table:mvrss_mvnet_archi}
\end{table*}


\begin{table*}
\begin{center}
\tiny
\renewcommand{\arraystretch}{1}
\begin{tabular}{c c c c l}
\toprule
 & Layer & Inputs & \makecell[c]{Output resolution \\ ($C \times H \times W$)} & Operations \\
\midrule
\multirow{9}{*}{\textbf{RD Encoder}} 
& rd\_layer1 & RD view & $128 \times 256 \times 64$ & \makecell[l]{Conv3D(1, 128, $3 \times 3 \times 3$, 1, (0, 1, 1), 1) \\
$+$ BN3D(128) $+$ LeakyReLU(0.01) \\ 
Conv3D(128, 128, $3 \times 3 \times 3$, 1, (0, 1, 1), 1) \\
$+$ BN3D(128) $+$ LeakyReLU(0.01)} \\

& rd\_layer2 & rd\_layer1 & $128 \times 128 \times 64$ & MaxPool2D(2, (2, 1)) \\

& rd\_layer3 & rd\_layer2 & $128 \times 128 \times 64$ & \makecell[l]{Conv2D(128, 128, $3 \times 3$, 1, 1, 1) $+$ BN2D(128) \\
$+$ LeakyReLU(0.01) \\ 
Conv2D(128, 128, $3 \times 3$, 1, 1, 1) $+$ BN2D(128) \\
$+$ LeakyReLU(0.01)}\\

& rd\_layer4 & rd\_layer3 & $128 \times 64 \times 64$ & MaxPool2D(2, (2, 1)) \\

& rd\_layer5 & rd\_layer4 & $128 \times 64 \times 64$ & Conv1D(128, 128, $1 \times 1$, 1, 0, 1)\\

& rd\_layer6 & rd\_layer5 & $640 \times 64 \times 64$ & ASPP(128, 128) \\

& rd\_layer7 & rd\_layer6 & $128 \times 64 \times 64$ & Conv1D(640, 128,  $1 \times 1$, 1, 0, 1) \\

\midrule
\multirow{9}{*}{\textbf{AD Encoder}} 
& ad\_layer1 & AD view & $128 \times 256 \times 64$ & \makecell[l]{Conv3D(1, 128, $3 \times 3 \times 3$, 1, (0, 1, 1), 1) \\
$+$ BN3D(128) $+$ LeakyReLU(0.01) \\
Conv3D(128, 128, $3 \times 3 \times 3$, 1, (0, 1, 1), 1) \\
$+$ BN3D(128) $+$ LeakyReLU(0.01)} \\

& ad\_layer2 & ad\_layer1 & $128 \times 128 \times 64$ & MaxPool2D(2, (2, 1)) \\

& ad\_layer3 & ad\_layer2 & $128 \times 128 \times 64$ & \makecell[l]{Conv2D(128, 128, $3 \times 3$, 1, 1, 1) $+$ BN2D(128) \\
$+$ LeakyReLU(0.01) \\ 
Conv2D(128, 128, $3 \times 3$, 1, 1, 1) $+$ BN2D(128) \\
$+$ LeakyReLU(0.01)}\\

& ad\_layer4 & ad\_layer3 & $128 \times 64 \times 64$ & MaxPool2D(2, (2, 1)) \\

& ad\_layer5 & ad\_layer4 & $128 \times 64 \times 64$ & Conv1D(128, 128, $1 \times 1$, 1, 0, 1)\\

& ad\_layer6 & ad\_layer5 & $640 \times 64 \times 64$ & ASPP(128, 128) \\

& ad\_layer7 & ad\_layer6 & $128 \times 64 \times 64$ & Conv1D(640, 128,  $1 \times 1$, 1, 0, 1) \\

\midrule
\multirow{9}{*}{\textbf{RA Encoder}} 

& ra\_layer1 & RA view & $128 \times 256 \times 256$ & \makecell[l]{Conv3D(1, 128, $3 \times 3 \times 3$, 1, (0, 1, 1), 1) \\
$+$ BN3D(128) $+$ LeakyReLU(0.01) \\ 
Conv3D(128, 128, $3 \times 3 \times 3$, 1, (0, 1, 1), 1) \\
$+$ BN3D(128) $+$ LeakyReLU(0.01)} \\

& ra\_layer2 & ra\_layer1 & $128 \times 128 \times 128$ & MaxPool2D(2, 2)\\

& ra\_layer3 & ra\_layer2 & $128 \times 128 \times 128$ & \makecell[l]{Conv2D(128, 128, $3 \times 3$, 1, 1, 1) $+$ BN2D(128) \\
$+$ LeakyReLU(0.01) \\ 
Conv2D(128, 128, $3 \times 3$, 1, 1, 1) $+$ BN2D(128) \\
$+$ LeakyReLU(0.01)} \\

& ra\_layer4 & ra\_layer3 & $128 \times 64 \times 64$ & MaxPool2D(2, 2)\\

& ra\_layer5 & ra\_layer4  & $128 \times 64 \times 64$ & Conv1D(128, 128, $1 \times 1$, 1, 0, 1)\\

& ra\_layer6 & ra\_layer5 & $640 \times 64 \times 64$ & ASPP(128, 128) \\

& ra\_layer7 & ra\_layer6 & $128 \times 64 \times 64$ & Conv1D(640, 128,  $1 \times 1$, 1, 0, 1) \\

\midrule
\textbf{Latent space} & layer8 & rd\_layer5, ra\_layer5, ad\_layer5 & $384 \times 64 \times 64$ & concatenate(rd\_layer5, ra\_layer5, ad\_layer5) \\

\midrule
\multirow{12}{*}{\textbf{RD Decoder}} 
& rd\_layer9 & layer8 & $128 \times 64 \times 64$ & Conv1D(384, 128, $1 \times 1$, 1, 0, 1) \\

& rd\_layer10 & rd\_layer7, rd\_layer9, ad\_layer7 & $384 \times 64 \times 64$ & concatenate(rd\_layer7, rd\_layer9, ad\_layer7) \\

& rd\_layer11 & rd\_layer10 & $128 \times 128 \times 64$ & ConvTranspose2D(384, 128, $2 \times 1$, (2, 1), 1, 1) \\

& rd\_layer12 & rd\_layer11 & $128 \times 128 \times 64$ & \makecell[l]{Conv2D(128, 128, $3 \times 3$, 1, 1, 1) $+$ BN2D(128) \\
$+$ LeakyReLU(0.01) \\ 
Conv2D(128, 128, $3 \times 3$, 1, 1, 1) $+$ BN2D(128) \\
$+$ LeakyReLU(0.01)} \\

& rd\_layer13 & rd\_layer12 & $128 \times 256 \times 64$	& ConvTranspose2D(128, 128, $2 \times 1$, (2, 1), 1, 1) \\

& rd\_layer14 & rd\_layer13 & $128 \times 256 \times 64$ & \makecell[l]{Conv2D(128, 128, $3 \times 3$, 1, 1, 1) $+$ BN2D(128) \\
$+$ LeakyReLU(0.01) \\ 
Conv2D(128, 128, $3 \times 3$, 1, 1, 1) \\
$+$ BN2D(128) $+$ LeakyReLU(0.01)} \\

& rd\_layer15 & rd\_layer14 & $K \times 256 \times 64$ & Conv1D(128, $K$, $1 \times 1$, 1, 0, 1) \\

\midrule
\multirow{12}{*}{\textbf{RA Decoder}} 
& ra\_layer9 & layer8 & $128 \times 64 \times 64$ & Conv1D(384, 128, $1 \times 1$, 1, 0, 1) \\

& ra\_layer10 & ra\_layer7, ra\_layer9, ad\_layer7 & $384 \times 64 \times 64$ & concatenate(ra\_layer7, ra\_layer9, ad\_layer7) \\

& ra\_layer11 & ra\_layer10 & $384 \times 128 \times 128$ & ConvTranspose2D(128, 128, $2 \times 2$, 2, 1, 1) \\

& ra\_layer12 & ra\_layer11 & $128 \times 128 \times 128$ & \makecell[l]{Conv2D(128, 128, $3 \times 3$, 1, 1, 1) $+$ BN2D(128) \\
$+$ LeakyReLU(0.01) \\ 
Conv2D(128, 128, $3 \times 3$, 1, 1, 1) $+$ BN2D(128) \\
$+$ LeakyReLU(0.01)} \\

& ra\_layer13 & ra\_layer12 & $128 \times 256 \times 256$ & ConvTranspose2D(128, 128, $2 \times 2$, 2, 1, 1) \\

& ra\_layer14 & ra\_layer13 & $128 \times 256 \times 256$ & \makecell[l]{Conv2D(128, 128, $3 \times 3$, 1, 1, 1) $+$ BN2D(128) \\
$+$ LeakyReLU(0.01) \\ 
Conv2D(128, 128, $3 \times 3$, 1, 1, 1) $+$ BN2D(128) \\
$+$ LeakyReLU(0.01)} \\

& ra\_layer15 & ra\_layer14 & $K \times 256 \times 256$ & Conv1D(128, $K$, $1 \times 1$, 1, 0, 1) \\

\bottomrule
\end{tabular}
\end{center}
\caption[Temporal multi-view network with ASPP modules (TMVA-Net) architecture]{\textbf{Temporal multi-view network with ASPP modules (TMVA-Net) architecture}. 
Details about the parameters of each operation are provided in Sec.\,\ref{sec_app:mvrss_archi_details}. 
Let $K$ be the number of classes.
The number of input channels in the first layer is fixed to $1$ (frames are considered as a sequence), here $q=4$, thus the number of channels is $5$.
}

\label{table:mvrss_tmvanet_archi}
\end{table*}





\subsection{Pre-processing and training procedures}
\label{sec_app:mvrss_preproc_procedures}

The experiments in the main paper have been conducted using the parameters detailed in Table \ref{table:mvrss_details_training}. 
An exponential decay with $\gamma = 0.9$ has been applied to each learning rate with an epoch step specific to each model (see Table \ref{table:mvrss_details_training}).
The competing methods have been trained using the Cross Entropy (CE) loss, except for the RSS-Net, which is trained with a
weighted Cross Entropy (wCE) using the formulation in \cite{kaul_rss-net_2020}. Our methods have been trained with the proposed combination of losses using the following parameters set up empirically: $\lambda_{\text{wCE}}= 1$, $\lambda_{\text{SDice}} = 10$ and $\lambda_{\text{CoL}}= 5$.

The architectures with which we compare our work have been designed to process inputs of size $256 \times 256$. Since the size of the range-Doppler view is $256 \times 64$ in the CARRADA dataset \cite{ouaknine_carrada_2020}, it is resized in the Doppler dimension to train these competing models. 
On the other hand, the proposed architectures are composed of down-sampling layers adapted to the size of the Doppler dimension, thus they do not require this pre-processing step.
The range-angle view has a size of $256 \times 256$ and does not require a resizing in both cases. 
For all methods, we used vertical and horizontal flip as data augmentation to reduce over-fitting.

Each view is normalised between 0 and 1 using local batch statistics for the competing methods. Our normalisation strategy consists in using the global statistics of the entire CARRADA dataset to normalise the input views.

\begin{table}
\begin{center}
\scriptsize
\begin{tabular}{c l r c c c c c }
\toprule
View & Method & Param. \# & $q$ & Batch size & LR & LR step & Epoch \#  \\
\midrule
\multirow{7}{*}{\textbf{RD}} 
& FCN-8s \cite{long_fully_2015} & 134.3~~ & 0 & 20 & $ 10^{-4}$ & 10 & 100 \\
& U-Net \cite{ronneberger_u-net_2015} & 17.3~~ & 3 & 6 & $ 10^{-4}$ & 20 & 150 \\
& DeepLabv3+ \cite{chen_encoder-decoder_2018} & 59.3~~ & 3 & 20 & $ 10^{-4}$ & 20 & 150 \\
& RSS-Net & 10.1~~ & 3 & 6 & $ 10^{-3}$ & 10 & 100 \\
& RAMP-CNN & 106.4~~ & 9 & 2 & $ 10^{-5}$ & 20 & 150 \\
& MV-Net (ours-baseline) & 2.4* & 3 & 13 & $ 10^{-4}$ & 20 & 300 \\
& TMVA-Net (ours) & 5.6* & 5 & 6 & $ 10^{-4}$ & 20 & 300 \\
\midrule
\multirow{7}{*}{\textbf{RA}}
& FCN-8s \cite{long_fully_2015} & 134.3~~ & 0 & 10 & $ 10^{-4}$ & 10 & 100 \\
& U-Net \cite{ronneberger_u-net_2015} & 17.3~~ & 3 & 6 & $ 10^{-4}$ & 20 & 150 \\
& DeepLabv3+ \cite{chen_encoder-decoder_2018} & 59.3~~ & 3 & 20 & $ 10^{-4}$ & 20 & 150 \\
& RSS-Net & 10.1~~ & 3 & 6 & $ 10^{-4}$ & 10 & 100 \\
& RAMP-CNN & 106.4~~ & 9 & 2 & $ 10^{-5}$ & 20 & 150 \\
& MV-Net (ours-baseline) & 2.4* & 3 & 13 & $ 10^{-4}$ & 20 & 300 \\
& TMVA-Net (ours) & 5.6* & 5 & 6 & $ 10^{-4}$ & 20 & 300 \\
\bottomrule
\end{tabular}
\end{center}
\caption[Hyper-parameters used for training]{\textbf{Hyper-parameters used for training}. The number of trainable parameters (in millions) for each method corresponds to a single view-segmentation model; Two such models, one for each view, are required for all methods but ours. In contrast, the number of parameters reported for our methods (`*') corresponds to a single model that segments both \ac{RD} and \ac{RA} views. RSS-Net and RAMP-CNN have been modified to be trained on both tasks (see Section \ref{sec:mvrss_carrada_train_and_res} of the main article). The input of a model consists in $q+1$ successive \ac{RAD} frames, where $q$ is the number of considered past frames, if any. The learning rate (`LR') step is in epochs.}
\label{table:mvrss_details_training}
\end{table}

\subsection{Quantitative results}
\label{sec_app:mvrss_quanti_results}

\begin{figure}[t]
\begin{center}
   \includegraphics[width=0.8\linewidth]{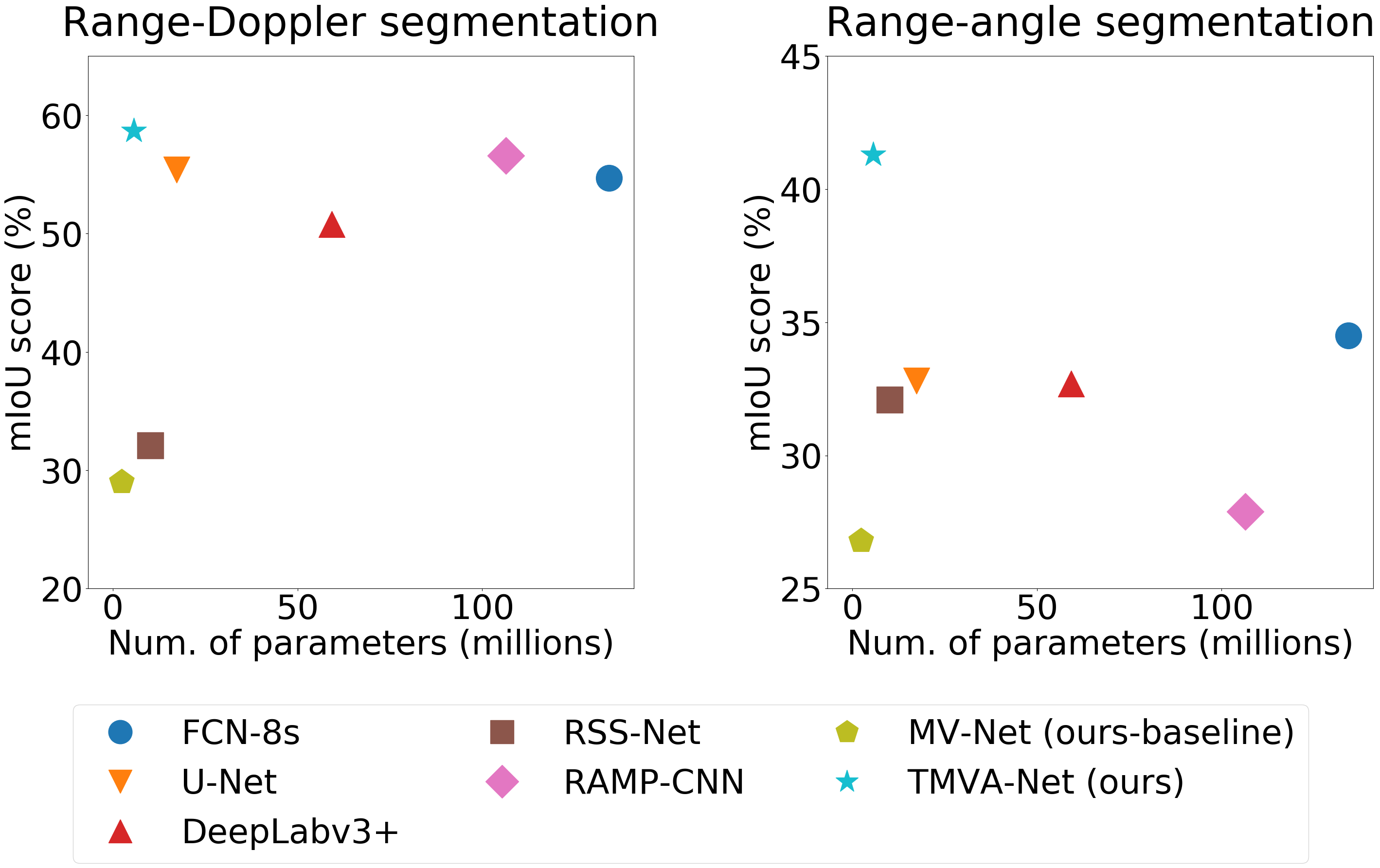}
\end{center}
   \caption[Performance-\textit{vs}.-complexity plots for all methods in RD and RA tasks]{\textbf{Performance-\textit{vs}.-complexity plots for all methods in \acf{RD} and \acf{RA} tasks}. Performance is assessed by \acf{mIoU} (\%) and complexity by the number of parameters (in millions) \textit{for a single task}.
   Top-left models correspond to the best performing and the lightest.
    Only our models, MV-Net and TMVA-Net, are able to segment both views simultaneously. For all the other methods, two distinct models must be trained to address both tasks, which doubles the number of actual parameters.}
\label{fig:mvrss_perf_params_iou}
\end{figure}

The proposed TMVA-Net architecture provides the best trade-off between performance and number of parameters for both range-Doppler and range-angle semantic segmentation tasks, as illustrated in Figure \ref{fig:mvrss_perf_params_iou} with mIoU metric.

\subsection{Variability of the quantitative results by method}
\label{sec_app:mvrss_variability_method}

\begin{figure}[t]
\begin{center}
   \includegraphics[width=0.8\linewidth]{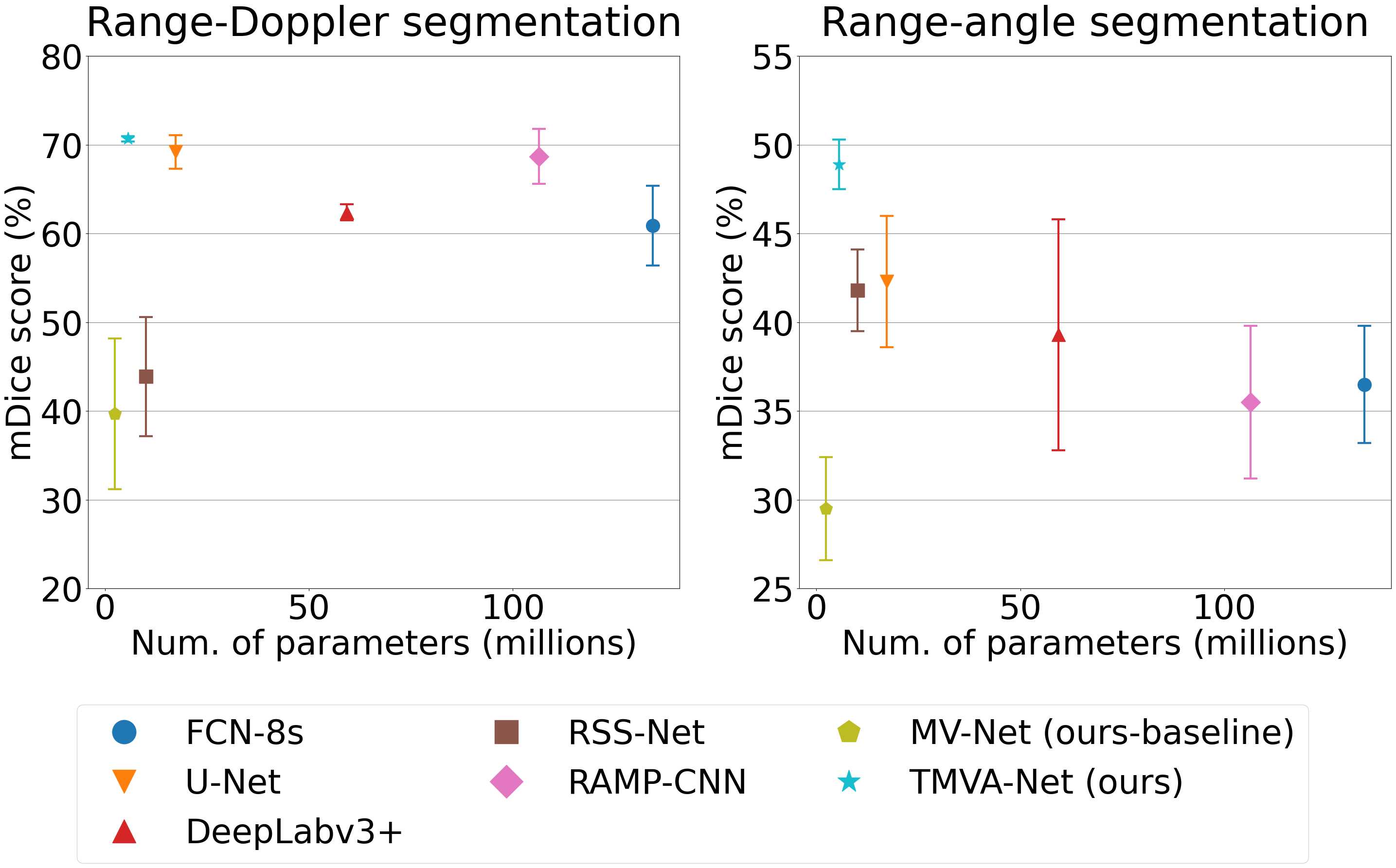}
\end{center}
   \caption[Variability of the performances-\textit{vs}.-complexity plots for all methods in Range-Doppler and Range-Angle tasks regarding the mDice metric.]{\textbf{Variability of the performances-\textit{vs}.-complexity plots for all methods in \acf{RD} and \acf{RA} tasks regarding the mDice metric.}. Performance is assessed by \ac{mDice} (\%) and complexity by the number of parameters (in millions) \textit{for a single task}.
   Top-left models correspond to the best performing and the lightest.
   The variability of the performances consists in evaluating the average (point) and standard deviation (vertical line) of four trained models.
    Only the proposed models, MV-Net and TMVA-Net, are able to segment both views simultaneously. For all the other methods, two distinct models must be trained to address both tasks, which doubles the number of actual parameters.}
\label{fig:mvrss_variability_methods_dice}
\end{figure}

\begin{figure}[t]
\begin{center}
   \includegraphics[width=0.8\linewidth]{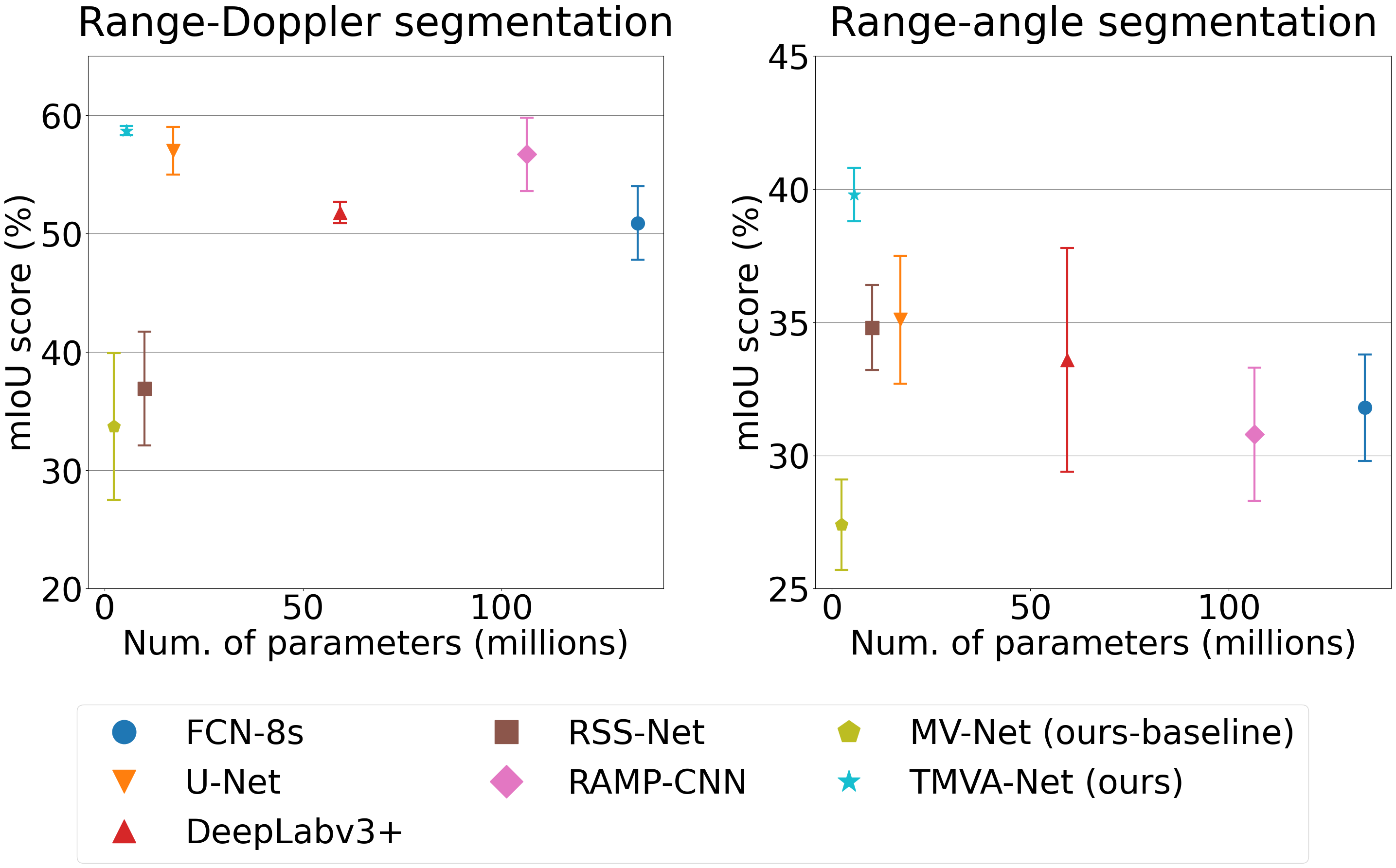}
\end{center}
   \caption[Variability of the performances-\textit{vs}.-complexity plots for all methods in Range-Doppler and Range-Angle tasks regarding the mIoU metric.]{\textbf{Variability of the performances-\textit{vs}.-complexity plots for all methods in \acf{RD} and \acf{RA} tasks regarding the \ac{mIoU} metric.}. Performance is assessed by \acl{mIoU} (\%) and complexity by the number of parameters (in millions) \textit{for a single task}.
   Top-left models correspond to the best performing and the lightest.
   he variability of the performances consists in evaluating the average (point) and standard deviation (vertical line) of four trained models.
    Only our models, MV-Net and TMVA-Net, are able to segment both views simultaneously. For all the other methods, two distinct models must be trained to address both tasks, which doubles the number of actual parameters.}
\label{fig:mvrss_variability_methods_iou}
\end{figure}

This section details a study of performance variability of the proposed and competing methods. Each model has been trained four times using the CARRADA-Train and CARRADA-Val datasets following the procedures explained in Section \ref{sec:mvrss_expe_carrada}. Three of the four models have been trained with different weight initialisation. The training process of the fourth starts from an already explored weight initialisation to take into account the stochasticity of the optimization. 
The variability is quantified by considering the average and standard deviation of the performances over the CARRADA-Test dataset regarding the four models for each method and each \ac{RADAR} view. 
The variability of the quantitative results by method is illustrated in Figures \ref{fig:mvrss_variability_methods_dice} and \ref{fig:mvrss_variability_methods_iou} for the \ac{mDice} and \ac{mIoU} evaluation metrics respectively. The TMVA-Net architecture trained with the proposed combination of losses (\ac{wCE}$+$\ac{SDice}$+$\ac{CoL}) reached the best average performances with the lower standard deviation, and thus the most stable results, for both \ac{RD} and \ac{RA} views, and both \ac{mDice} and \ac{mIoU} metrics. 

\subsection{Variability of the loss ablation study}
\label{sec_app:mvrss_variability_losses}

\begin{figure}[t]
\begin{center}
   \includegraphics[width=1\linewidth]{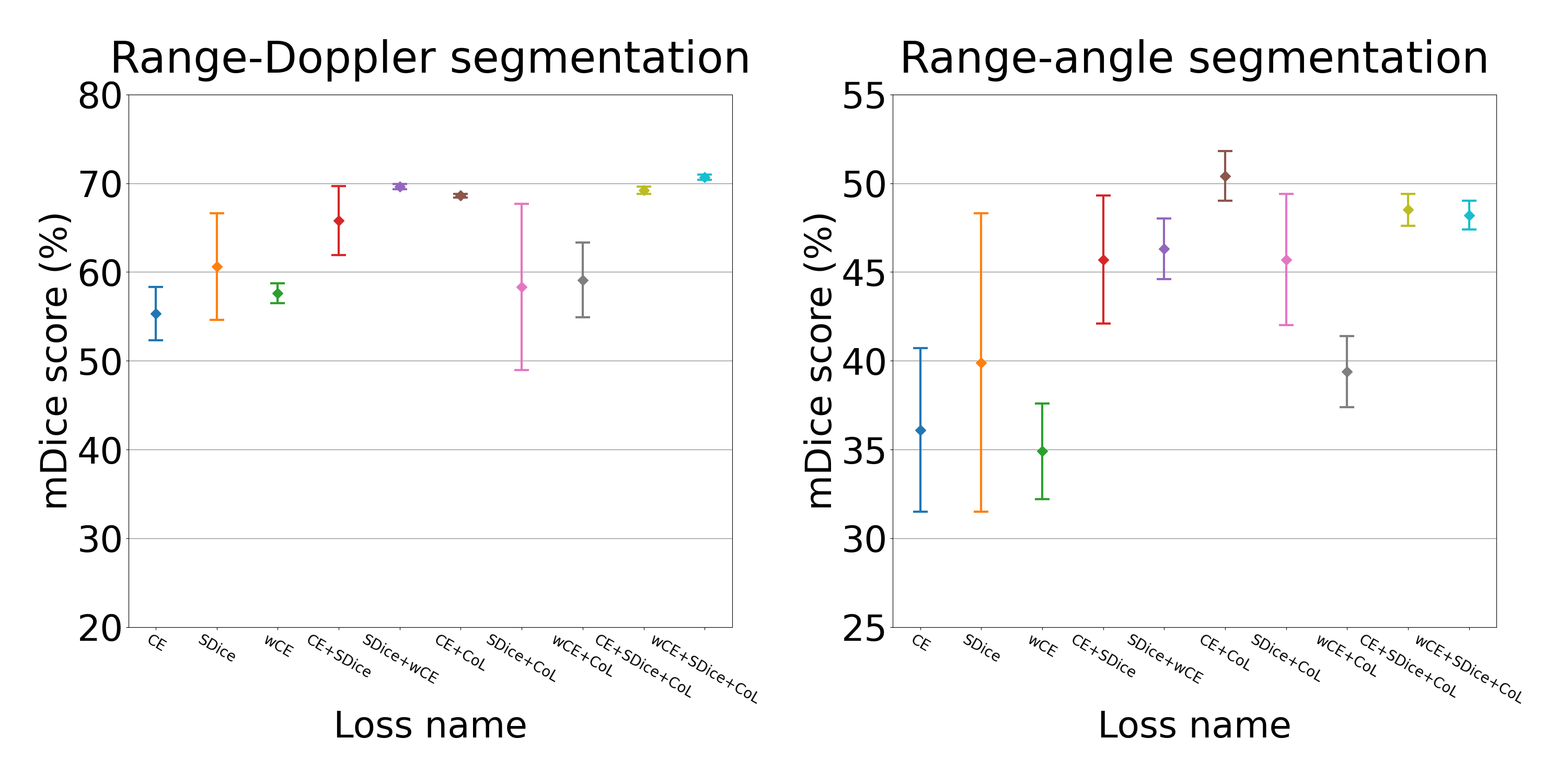}
\end{center}
   \caption[Variability of the performances plots for all combination of losses in Range-Doppler and Range-Angle tasks regarding the mDice metric.]{\textbf{Variability of the performances plots for all combination of losses in \acf{RD} and \acf{RA} tasks regarding the \ac{mDice} metric.}. Performance is assessed by \ac{mDice} (\%). All the possible combination of losses are considered, either with the \acf{CE} or \acf{wCE} loss, mixed with or without the \acf{SDice} and \acf{CoL} losses.
   The performance variability consists in evaluating the mean (point) and standard deviation (vertical line) of four trained models.
   The proposed combination (wCE$+$SDice$+$CoL) is the best trade-off between performance and stability on both \ac{RD} and \ac{RA}.
   }
\label{fig:mvrss_variability_losses}
\end{figure}

This section analyses the variability of the ablation study comparing the combination of loss functions to train the proposed multi-view architectures. The TMVA-Net architecture has been optimised considering either the \ac{CE} or the \ac{wCE} loss function, combined with the \ac{SDice} or \ac{CoL}, or both. Details on the loss functions are provided in Section \ref{sec:mvrss_losses}. Each model has been trained four times using the CARRADA-Train and CARRADA-Val datasets following the procedures explained in Section \ref{sec:mvrss_expe_carrada}. Three of the four models have been trained with different weight initialisation. 
The training process of the fourth starts from an already explored weight initialisation to take into account the stochasticity of the optimization. 
The variability is quantified by considering the average and standard deviation of the performances over the CARRADA-Test dataset regarding the four models for each method and each \ac{RADAR} view. 
The variability of the loss ablation study usng the TMVA-Net architecture is illustrated in Figure \ref{fig:mvrss_variability_losses} for the \ac{mDice} evaluation metric. 
The average performances increased for both \ac{RD} and \ac{RA} segmentation by integrating the \ac{CoL} term to each individual loss term of \ac{CE}, \ac{wCE} and \ac{SDice}\footnote{Except for the combination \ac{SDice}$+$\ac{CoL} for the \ac{RD} view.}. It indicates that the \ac{CoL} term has a significant impact on the training process.
The combination \ac{wCE}$+$\ac{SDice}$+$\ac{CoL} reached the best trade-off between performances and stability of the results for both \ac{RD} and \ac{RA}.

\subsection{Qualitative results on CARRADA}
\label{sec_app:mvrss_quali_results_carrada}

\begin{figure}
\begin{center}
\includegraphics[width=1\linewidth]{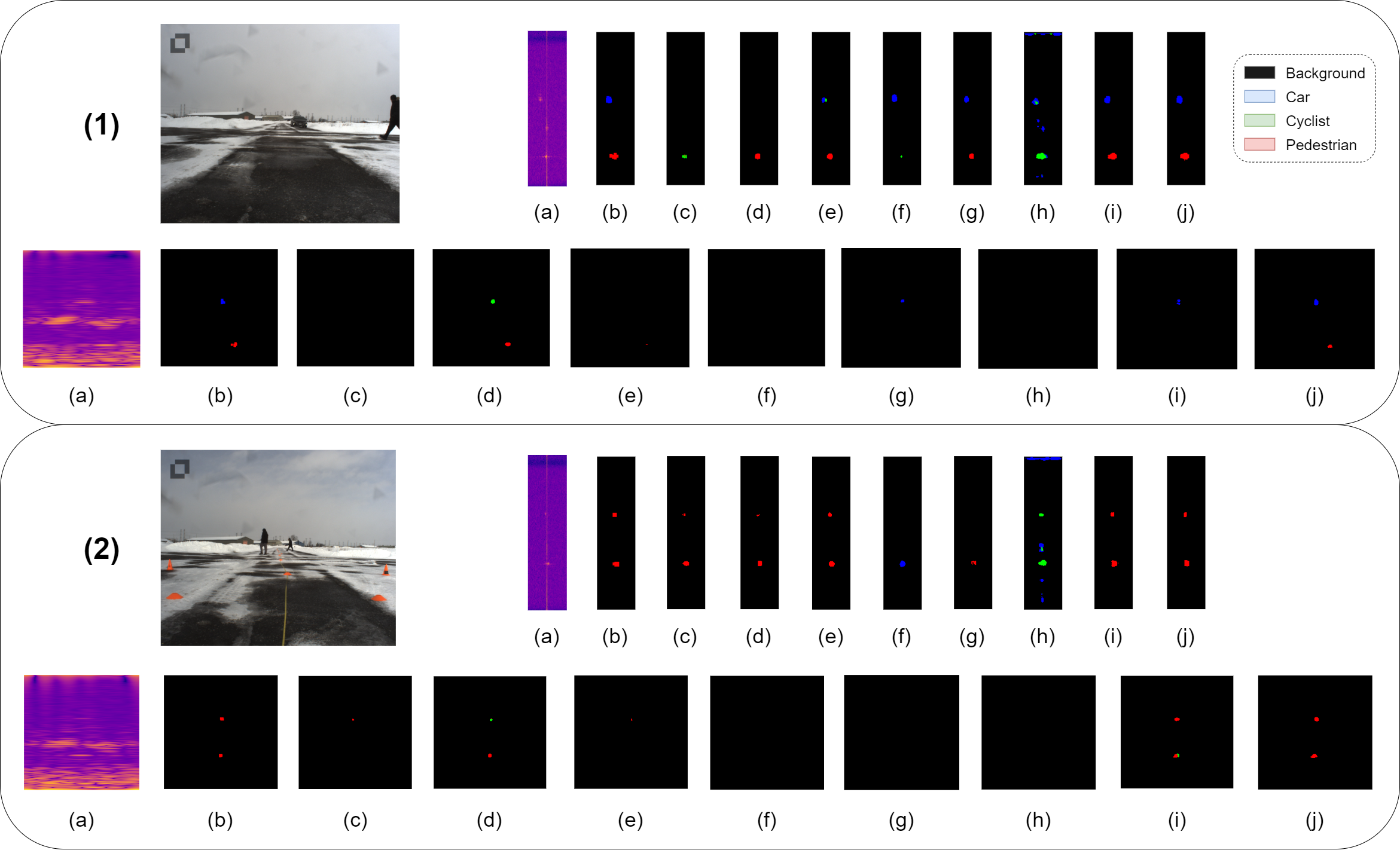}
\end{center}
   \caption[Qualitative results on two test scenes of CARRADA-Test]{\textbf{Qualitative results on two test scenes of CARRADA-Test}. (1) and (2) are two independent examples. (\textit{Top}) camera image of the scene and results of the \acl{RD} segmentation; (\textit{Bottom}) Results of the \acl{RA} Segmentation. (a) \ac{RADAR} view signal, (b) ground-truth mask, (c) FCN8s, (d) U-Net, (e) DeepLabv3+, (f) RSS-Net, (g) RAMP-CNN, (h) MV-Net (our baseline w/ wCE$+$SDice loss), (i) TMVA-Net (ours, w/ wCE$+$SDice loss), (j) TMVA-Net (ours, w/ wCE$+$SDice$+$CoL loss).}
\label{fig:mvrss_app_quali_results_carrada}
\end{figure}

Additional qualitative results are shown in Figure \ref{fig:mvrss_app_quali_results_carrada} for each method on scenes (1-2) from the CARRADA-Test. For the scene (1), RAMP-CNN (g) and TMVA-Net (i-j) display well segmented \ac{RD} views. However, only TMVA-Net with CoL (j) is able to localise and classify both objects in the \ac{RD} and \ac{RA} views of the first example. 
In scene (2), four methods (d-e-i-j) are able to well localise objects in the \ac{RD} view. Once again, only TVMA-Net with CoL (j) is able to well segment objects in both \ac{RD} and \ac{RA} views while our method without CoL (i) predicts pedestrian and cyclist categories for pixels of the same object.

\subsection{Qualitative results on RADDet}
\label{sec_app:mvrss_quali_results_raddet}

\begin{figure}[t]
\begin{center}
\includegraphics[width=1\linewidth]{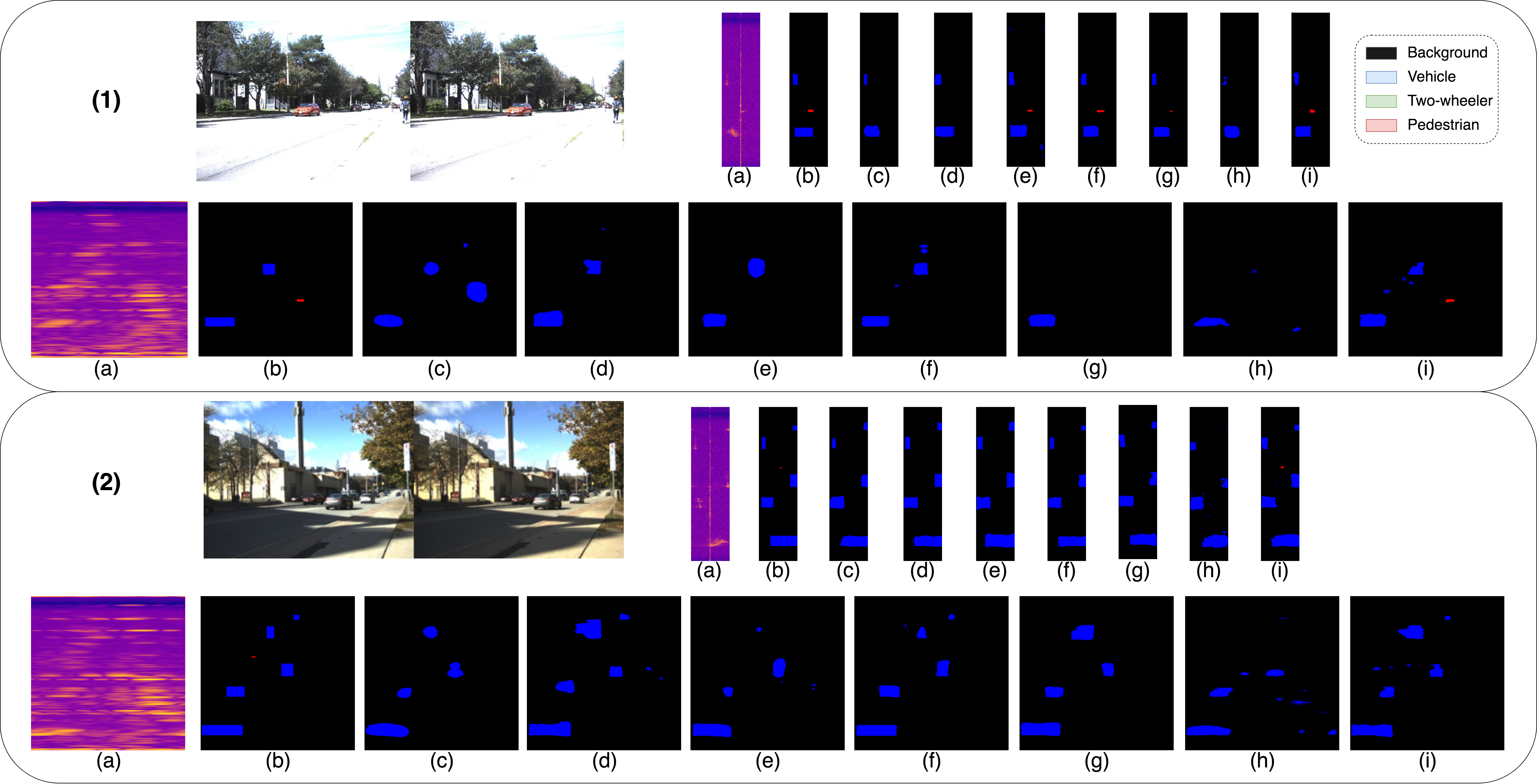}
\end{center}
   \caption[Qualitative results on a test scene of RADDet]{\textbf{Qualitative results on a test scene of RADDet}. (\textit{Top}) camera image of the scene and results of the \acl{RD} segmentation; (\textit{Bottom}) Results of the \acl{RA} Segmentation. (a) Radar view signal, (b) ground-truth mask, (c) FCN8s, (d) U-Net, (e) DeepLabv3+, (f) RSS-Net, (g) RAMP-CNN, (h) MV-Net (our baseline w/ wCE$+$SDice$+$CoL loss), (i) TMVA-Net (ours, w/ wCE$+$SDice$+$CoL loss).}
\label{fig:mvrss_app_quali_results_raddet}
\end{figure}

Figure \ref{fig:mvrss_app_quali_results_raddet} shows qualitative results of two complex urban scenes (1-2) for each method trained on RADDet-Train and RADDet-Val, and tested on RADDet-Test.
For the scene (1), DeepLabv3+ (e), RSS-Net (f) and RMVA-Net (i) display well segmented \ac{RD} views. However, only the TMVA-Net trained with the proposed loss combination (wCE$+$SDice$+$CoL) is able to localise and classify two cars and a pedestrian in the \ac{RD} and \ac{RA} views of the first example.
In scene (2), most of the methods successfully segment four cars in both \ac{RD} and \ac{RA} views. However, only the TMVA-Net is able to localise and classify the pedestrian in the \ac{RD} view. It is well localised on the \ac{RA} but misclassified as a car instead of pedestrian. The annotated shape of the pedestrian is small and thus difficult to recognize. A reformulation of \ac{CoL} could be explored to better enforce the spatial coherence.

\subsection{Qualitative results on in-house dataset}
\label{sec_app:mvrss_quali_results_sc}

\begin{figure}
\begin{center}
\includegraphics[width=1\linewidth]{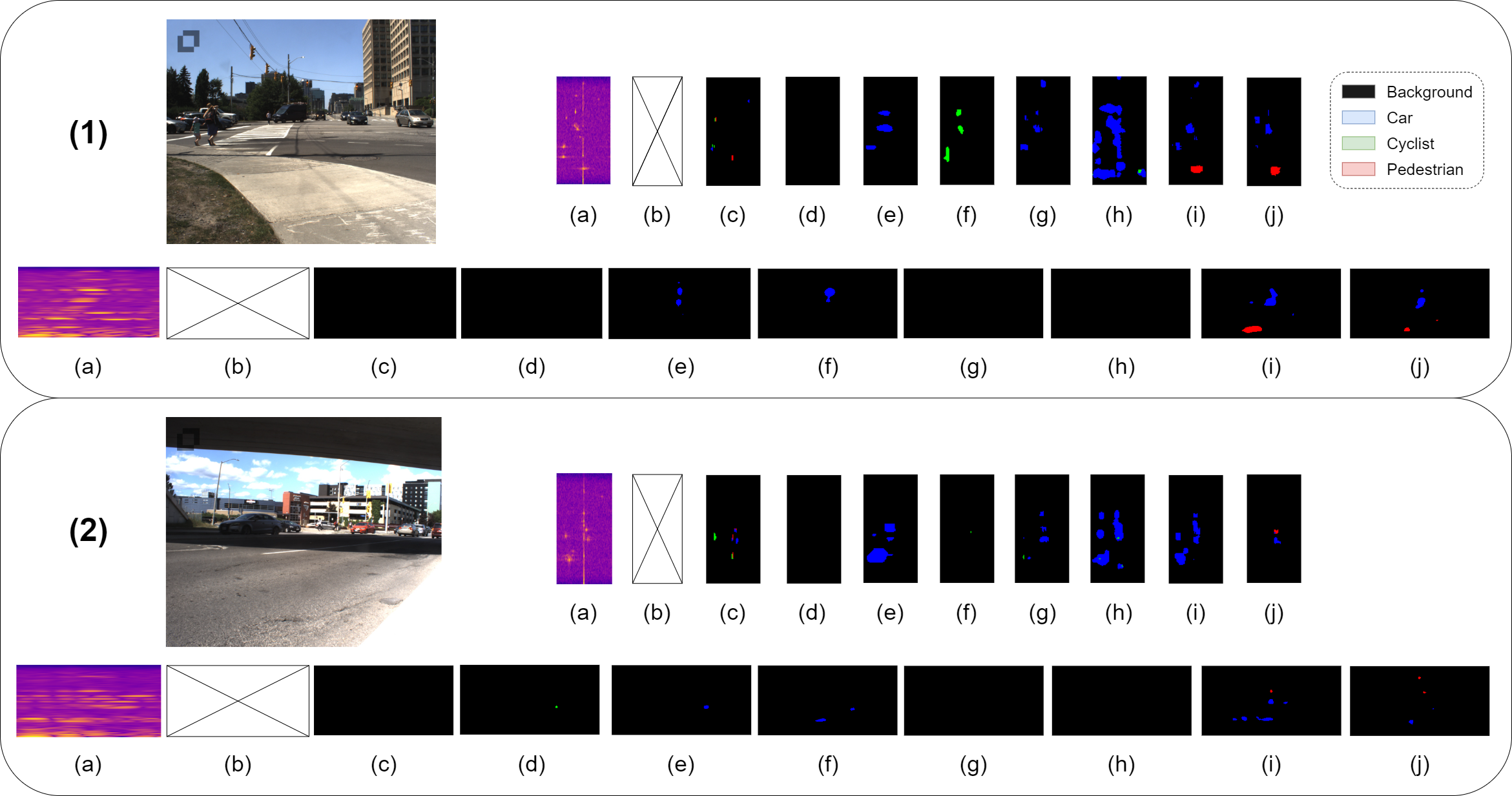}
\end{center}
   \caption[Qualitative results on two test scenes of in-house sequences]{\textbf{Qualitative results on two test scenes of in-house sequences}. (1) and (2) are two independent examples. (\textit{Top}) camera image of the scene and results of the \acl{RD} segmentation; (\textit{Bottom}) Results of the \acl{RA} Segmentation. (a) \ac{RADAR} view signal, (b) ground-truth mask, (c) FCN8s, (d) U-Net, (e) DeepLabv3+, (f) RSS-Net, (g) RAMP-CNN, (h) MV-Net (our baseline w/ wCE$+$SDice loss), (i) TMVA-Net (ours, w/ wCE$+$SDice loss), (j) TMVA-Net (ours, w/ wCE$+$SDice$+$CoL loss).}
\label{fig:mvrss_app_quali_results_sc}
\end{figure}

Figure \ref{fig:mvrss_app_quali_results_sc} shows qualitative results for each method trained on CARRADA-Train and CARRADA-Val, and tested on in-house sequences of complex urban scenes (1-2) with a different range resolution.
The qualitative examples and results have been cropped with respect to the minimum and maximum range of the dataset used for training. The ground-truth masks in columns (1-b) and (2-b) are empty because the \ac{RADAR} views are not annotated.
In scene (1), only TMVA-Net models (i-j) are able to localise and classify the signals related to the pedestrians and cars in both the \ac{RD} and the \ac{RA} views. 
In scene (2), only TMVA-Net (i-j) methods succeed in localising and classifying cars and pedestrians in the \ac{RA} view. We note that TMVA-Net without CoL (i) detects more car signals while TMVA-Net with CoL (j) is the only method capable of distinguishing pedestrian signatures on both \ac{RD} and \ac{RA} views.

These two examples in complex urban scenes suggest that our method has successfully learnt object signatures in the CARRADA dataset and is able to generalise well.

\section{Sensor fusion}

\subsection{Sensor settings of the nuScenes dataset}
\label{sec_app:nuscenes_sensor_settings}

\begin{figure}
\begin{center}
\includegraphics[width=1\linewidth]{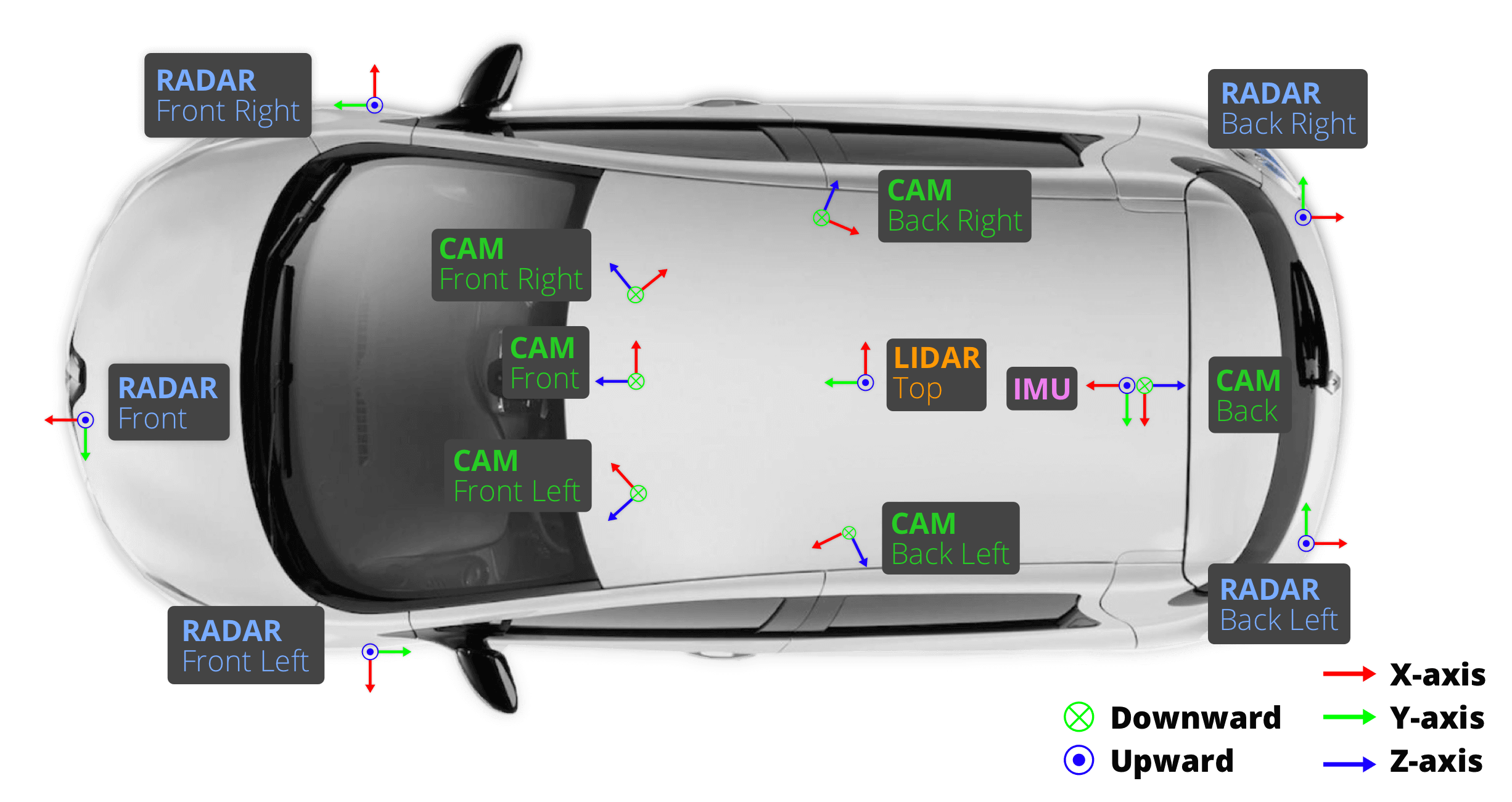}
\end{center}
   \caption[Qualitative results on two test scenes of in-house sequences]{\textbf{Qualitative results on two test scenes of in-house sequences}. Source: \cite{caesar_nuscenes_2020}.}
\label{fig:fusion_app_nuscenes_sensors}
\end{figure}

Figure \ref{fig:fusion_app_nuscenes_sensors} illustrates the sensor setup used to record the nuScenes dataset \cite{caesar_nuscenes_2020}. The orientation axis is specific to each sensor requiring transformations to manipulate both \ac{RADAR} and \ac{LiDAR} point clouds. 

\subsection{Qualitative results of the propagation and fusion module}
\label{sec_app:fusion_quali_results}

\begin{figure}
\begin{center}
\includegraphics[width=1.1\linewidth,center]{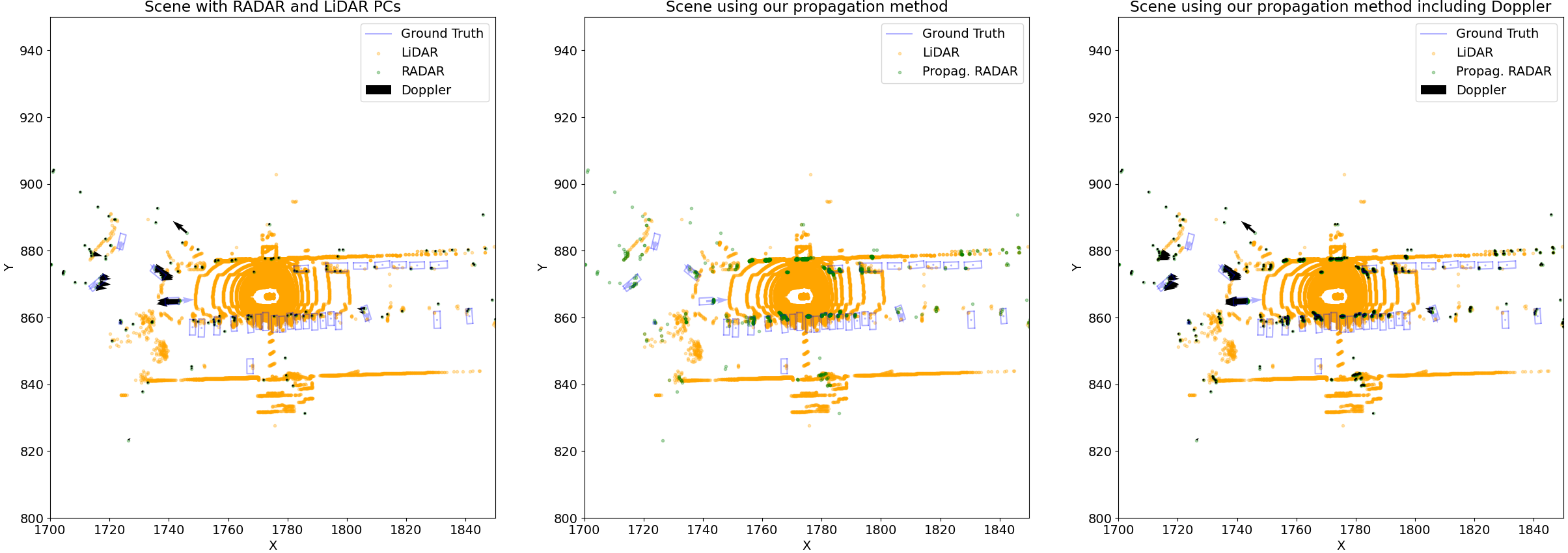}
\end{center}
   \caption[Qualitative results on the nuScenes dataset of our propose propagation and fusion module]{\textbf{Qualitative results on the nuScenes dataset of our propose propagation and fusion module}. (Left) Scene in \acl{BEV} representation with \ac{LiDAR} and \ac{RADAR} point clouds. (Middle) The point cloud illustrated in green groups the \ac{RADAR} and propagated \ac{RADAR} points with Doppler and reflectivities. (Right) The propagated Doppler information is illustrated with black arrows to distinguish moving objects.}
\label{fig:fusion_app_quali_nuscenes}
\end{figure}

Figure \ref{fig:fusion_app_quali_nuscenes} presents additional qualitative results of our proposed propagation and fusion module on a scene of the nuScenes dataset. Our proposed method detailed in Section \ref{sec:fusion_method} succeed to obtain an enriched point cloud with a denser Doppler information.

\chapter{High-definition RADAR}
\label{chap:appendix6}

\section{Ablation study of the MIMO pre-encoder}
\label{sec_app:fft_radnet_ablation}
\begin{figure}[!ht]
    \centering
    \includegraphics[width=0.6\columnwidth]{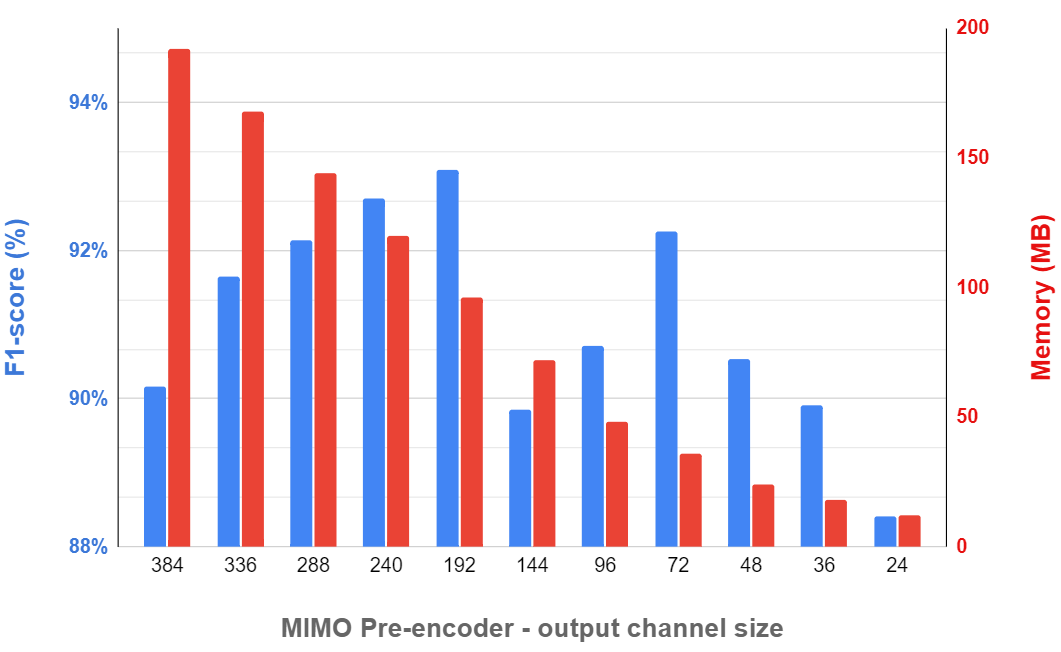}
    \label{fig:fft_radnet_ablation}
    \caption[MIMO pre-encoder ablation study]{\textbf{MIMO pre-encoder ablation study}. Influence of the number of output channels of the pre-encoder on the memory footprint and the performance of the detection head.}
\end{figure}

The role of the \ac{MIMO} pre-encoder is to de-interleave the \acf{RD} and to transform them into a representation that is compact and still allows, through learning, the prediction of azimuth angles along with other information on reflectors. 
The input of the \ac{MIMO} pre-encoder is composed of the $\NRx = 16$ \ac{RD} in complex numbers, one for each Rx. 
The real and imaginary parts are stacked, yielding an input tensor of total size 
$\BinR{\times}\BinD{\times}2\NRx$, 
\textit{i.e.}, $512{\times}256{\times}32$.   
The ablation study consists in evaluating the performance of FFT-RadNet's detection head while reducing the number of feature channels that the \ac{MIMO} pre-encoder outputs. 
The maximum number of output channels is the number of virtual antennas with a complex signal (real and imaginary parts), \textit{i.e.}, $\NTx{\cdot}2{\cdot}\NRx  = 384$. We vary the number of output channels from a minimum of 24 to this maximum value and compute the detection performance on the validation set.
The results of this ablation study are reported in Figure \ref{fig:fft_radnet_ablation}. We measure the detection performance with the F1-score, classically defined as  $\text{F1-score} = \frac{\text{AP}\cdot\text{AR}}{\text{AP} + \text{AR}}$, which aggregates in a single metric both the \acf{AP} and the \acf{AR}.
We observe that the best performance is reached with 192 output channels, hence half of the maximum output size. 
This compressed output is the one that captures at best the range and azimuth information from the \ac{RD} inputs toward the detection and segmentation tasks. 

\clearpage

\begin{minipage}{\textwidth}
\includepdf[pages=-]{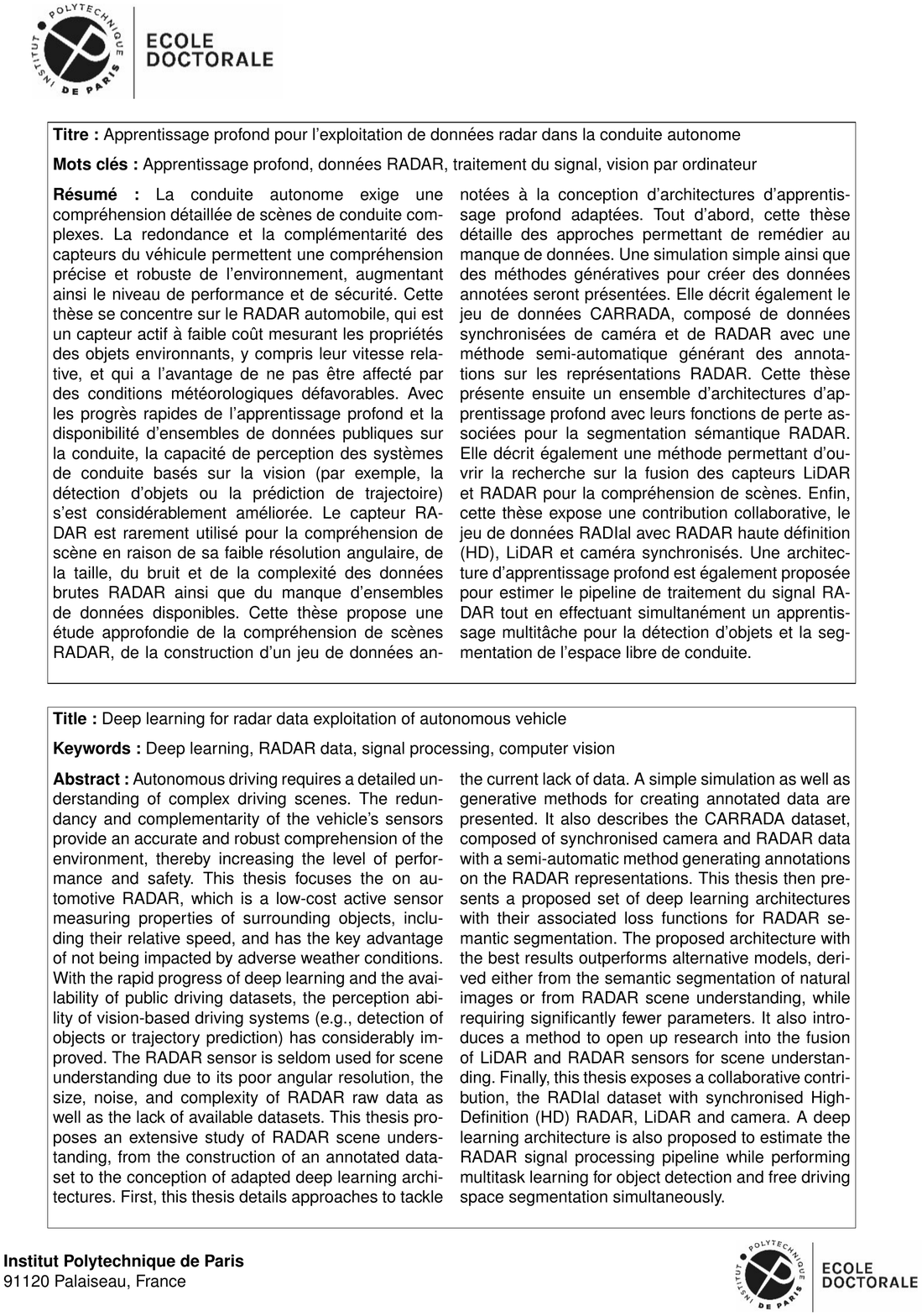}
\end{minipage}

\end{document}